\documentclass[12pt]{report}

\usepackage{tamuconfig}
\usepackage[T1]{fontenc}
\usepackage{graphicx}

\DeclareGraphicsExtensions{.png}
\DeclareGraphicsExtensions{.jpg}

\graphicspath{ {./graphic/} }

\usepackage{lscape}
\usepackage{amssymb}
\usepackage[hidelinks]{hyperref}
\usepackage{ gensymb }
\usepackage{algorithm}
\usepackage{algorithmic}
\usepackage[caption=false,font=footnotesize]{subfig}
\usepackage{multirow}
\usepackage{bm}
\usepackage{esvect}

\begin{document}

\renewcommand{\tamumanuscripttitle}{Risk-aware Path and Motion Planning for a Tethered Aerial Visual Assistant in Unstructured or Confined Environments}

\renewcommand{\tamupapertype}{Dissertation}

\renewcommand{\tamufullname}{Xuesu Xiao}

\renewcommand{\tamudegree}{Doctor of Philosophy}

\renewcommand{\tamuchairone}{Robin R. Murphy}

\renewcommand{\tamumemberone}{Dylan A. Shell}

\newcommand{\tamumembertwo}{Thomas R. Ioerger}

\newcommand{\tamumemberthree}{Suman Chakravorty}

\renewcommand{\tamudepthead}{Dilma Da Silva}

\renewcommand{\tamugradmonth}{December}

\renewcommand{\tamugradyear}{2019}

\renewcommand{\tamudepartment}{Computer Science}

%
%
%
%


\providecommand{\tabularnewline}{\\}

\begin{titlepage}
\begin{center}
\MakeUppercase{\tamumanuscripttitle}
\vspace{4em}

A \tamupapertype

by

\MakeUppercase{\tamufullname}

\vspace{4em}

\begin{singlespace}

Submitted to the Office of Graduate and Professional Studies of \\
Texas A\&M University \\

in partial fulfillment of the requirements for the degree of \\
\end{singlespace}

\MakeUppercase{\tamudegree}
\par\end{center}
\vspace{2em}
\begin{singlespace}
\begin{tabular}{ll}
 & \tabularnewline
& \cr
Chair of Committee, & \tamuchairone\tabularnewline
Committee Members, & \tamumemberone\tabularnewline
 & \tamumembertwo\tabularnewline
 & \tamumemberthree\tabularnewline
Head of Department, & \tamudepthead\tabularnewline

\end{tabular}
\end{singlespace}
\vspace{3em}

\begin{center}
\tamugradmonth \hspace{2pt} \tamugradyear

\vspace{3em}

Major Subject: \tamudepartment \par
\vspace{3em}
Copyright \tamugradyear \hspace{.5em}\tamufullname 
\par\end{center}
\end{titlepage}
\pagebreak{}

%
%
%
%

\chapter*{ABSTRACT}
\addcontentsline{toc}{chapter}{ABSTRACT} 

\pagestyle{plain} 
\pagenumbering{roman} 
\setcounter{page}{2}

\indent This research aims at developing path and motion planning algorithms for a tethered Unmanned Aerial Vehicle (UAV) to visually assist a teleoperated primary robot in unstructured or confined environments. The emerging state of the practice for nuclear operations, bomb squad, disaster robots, and other domains with novel tasks or highly occluded environments is to use two robots, a primary and a secondary that acts as a visual assistant to overcome the perceptual limitations of the sensors by providing an external viewpoint. However, the benefits of using an assistant have been limited for at least three reasons: (1) users tend to choose suboptimal viewpoints, (2) only ground robot assistants are considered, ignoring the rapid evolution of small unmanned aerial systems for indoor flying, (3) introducing a whole crew for the second teleoperated robot is not cost effective, may introduce further teamwork demands, and therefore could lead to miscommunication. This dissertation proposes to use an autonomous tethered aerial visual assistant to replace the secondary robot and its operating crew. Along with a pre-established theory of viewpoint quality based on affordances, this dissertation aims at defining and representing robot motion risk in unstructured or confined environments. Based on those theories, a novel high level path planning algorithm is developed to enable risk-aware planning, which balances the tradeoff between viewpoint quality and motion risk in order to provide safe and trustworthy visual assistance flight. The planned flight trajectory is then realized on a tethered UAV platform. The perception and actuation are tailored to fit the tethered agent in the form of a low level motion suite, including a novel tether-based localization model with negligible computational overhead, motion primitives for the tethered airframe based on position and velocity control, and two different approaches to negotiate tether with complex obstacle-occupied environments. The proposed research provides a formal reasoning of motion risk in unstructured or confined spaces, contributes to the field of risk-aware planning with a versatile planner, and opens up a new regime of indoor UAV navigation: tethered indoor flight to ensure battery duration and failsafe in case of vehicle malfunction. It is expected to increase teleoperation productivity and reduce costly errors in scenarios such as safe decommissioning and nuclear operations in the Fukushima Daiichi facility. 

\pagebreak{}

%
%
%
%

\chapter*{DEDICATION}
\addcontentsline{toc}{chapter}{DEDICATION}  

\begin{center}
\vspace*{\fill}
To my parents, Jing and Fan, whose dedications and sacrifices give me the opportunity to become the person I am today. 
\vspace*{\fill}
\end{center}

\pagebreak{}

%
%
%
%

\chapter*{ACKNOWLEDGEMENTS}
\addcontentsline{toc}{chapter}{ACKNOWLEDGEMENTS}  

I would like to first thank my advisor and committee chair, Dr. Robin R. Murphy, and my graduate committee members, Drs. Dylan A. Shell, Thomas R. Ioerger, and Suman Chakravorty for their guidance and feedback throughout the course of this research. 

Second, I wish to thank the National Science Foundation's National Robotics Initiative and the US Department of Energy for providing the funding for this research. Their support made it possible for me to pursue and finish my PhD. 

Thanks also to my labmates in the Humanitarian Robotics and AI Laboratory, Department of Computer Science and Engineering and Dwight Look College of Engineering staff for their help in the long road of my academic studies. 

Finally, thanks most of all to my parents, confidants, and dearest friends... Jing and Fan: without your support, financially and emotionally, patience, respect, and love, my academic career wouldn't even have the chance to start and I would be far away from the person who I become today. Thank you and love you!

\pagebreak{}

%
%
%
%

\chapter*{CONTRIBUTORS AND FUNDING SOURCES}
\addcontentsline{toc}{chapter}{CONTRIBUTORS AND FUNDING SOURCES}  

\subsection*{Contributors}
This work was supported by a dissertation committee consisting of Professor Robin. R. Murphy, Professor Dylan A. Shell and Professor Thomas R. Ioerge of the Department of Computer Science and Engineering, and Professor Suman Chakravorty of the Department of Aerospace Engineering.

The implementation of the low level motion planner in C++ was in collaboration with Jan Dufek. The mechanics model for tether-based UAV localization was in collaboration with Yiming Fan. The development of the server (video encoder and Raspberry Pi computer) and the user interface, both for wireless video streaming and visual assistant telemetry and command, was in collaboration with Mickie Byrd from AdventGX. 

All other work conducted for the dissertation was completed by the student independently.
\subsection*{Funding Sources}
This study was supported by NSF DOE NRI Grant DE-EM0004483 A Collaborative Visual Assistant for Robot Operations in Unstructured or Confined Environments. 
\pagebreak{}
%
%
%
%


\chapter*{NOMENCLATURE}
\addcontentsline{toc}{chapter}{NOMENCLATURE}  


\hspace*{-1.25in}
\vspace{12pt}
\begin{spacing}{1.0}
	\begin{longtable}[htbp]{@{}p{0.35\textwidth} p{0.62\textwidth}@{}}
		UAV	&	Unmanned Aerial Vehicle\\	[2ex]
		UGV		&	Unmanned Ground Vehicle\\	[2ex] 
		JAEA & Japanese Atomic Energy Agency \\ [2ex]
		DoF & Degree of Freedom \\[2ex]
		MDP & Markov Decision Process \\[2ex]
		POMDP & Partially Observable Markov Decision Process \\[2ex]
		AIS & Automated Identification System \\[2ex]
		AUV & Autonomous Underwater Vehicle \\[2ex]
		Belief Roadmap & BRM \\[2ex]
		FIRM & Feedback-based Information Roadmap \\[2ex]
		CC-RRT & Chance-Constrained Rapidly-exploring Random Tree \\[2ex]
		C-MDP & Constrained Markov Decision Process \\[2ex]
		C-POMDP & Constrained Partially Observable Markov Decision Process \\[2ex]
		CC-POMDP & Chance Constrained Partially Observable Markov Decision Process \\[2ex]
		IRA & Iterative Risk Allocation \\[2ex]
		RMPC & Robust Model Predictive Control \\[2ex]
		FAA & Federal Aviation Administration \\[2ex]
		SLAM & Simultaneous Localization and Mapping \\[2ex]
		PRM & Probabilistic Road Map \\[2ex]
		DFS & Depth First Search \\[2ex]
		
		PID & Proportional, Integral, Derivative \\[2ex]
		PoI & Point of Interest \\[2ex]
		CoM & Center of Mass \\[2ex]
		MoCap & Motion Capture \\[2ex]
		VTOL & Vertical Take-Off and Landing \\[2ex]
		MTTF & Mean Time To Failure \\[2ex]
		SDK & Software Development Kit \\[2ex]
		RMS & Root Mean Square \\[2ex]
		SD & Standard Deviation \\[2ex]
		OCU & Operator Control Unit \\[2ex]

	\end{longtable}
\end{spacing}

\pagebreak{}

%
%
%
%

\phantomsection
\addcontentsline{toc}{chapter}{TABLE OF CONTENTS}  

\begin{singlespace}
\renewcommand\contentsname{\normalfont} {\centerline{TABLE OF CONTENTS}}

\setcounter{tocdepth}{4} 

\setlength{\cftaftertoctitleskip}{1em}
\renewcommand{\cftaftertoctitle}{%
\hfill{\normalfont {Page}\par}}

\tableofcontents

\end{singlespace}

\pagebreak{}


\phantomsection
\addcontentsline{toc}{chapter}{LIST OF FIGURES}  

\renewcommand{\cftloftitlefont}{\center\normalfont\MakeUppercase}

\setlength{\cftbeforeloftitleskip}{-12pt} 
\renewcommand{\cftafterloftitleskip}{12pt}

\renewcommand{\cftafterloftitle}{%
\\[4em]\mbox{}\hspace{2pt}FIGURE\hfill{\normalfont Page}\vskip\baselineskip}

\begingroup

\begin{center}
\begin{singlespace}
\setlength{\cftbeforechapskip}{0.4cm}
\setlength{\cftbeforesecskip}{0.30cm}
\setlength{\cftbeforesubsecskip}{0.30cm}
\setlength{\cftbeforefigskip}{0.4cm}
\setlength{\cftbeforetabskip}{0.4cm}



\listoffigures

\end{singlespace}
\end{center}

\pagebreak{}

%
\phantomsection
\addcontentsline{toc}{chapter}{LIST OF TABLES}  

\renewcommand{\cftlottitlefont}{\center\normalfont\MakeUppercase}

\setlength{\cftbeforelottitleskip}{-12pt} 

\renewcommand{\cftafterlottitleskip}{1pt}

\renewcommand{\cftafterlottitle}{%
\\[4em]\mbox{}\hspace{2pt}TABLE\hfill{\normalfont Page}\vskip\baselineskip}

\begin{center}
\begin{singlespace}

\setlength{\cftbeforechapskip}{0.4cm}
\setlength{\cftbeforesecskip}{0.30cm}
\setlength{\cftbeforesubsecskip}{0.30cm}
\setlength{\cftbeforefigskip}{0.4cm}
\setlength{\cftbeforetabskip}{0.4cm}

\listoftables 

\end{singlespace}
\end{center}
\endgroup
\pagebreak{}  

%
%
%
%


\pagestyle{plain} 
\pagenumbering{arabic} 
\setcounter{page}{1}

\chapter{\uppercase {Introduction}}
A secondary assistant robot providing an external view of a task being performed by a primary robot has emerged as the state of the practice in nuclear operations, bomb squad, disaster robots, and other remote domains where situational awareness of the teleoperator is not easily achievable from the onboard camera of the primary robot. Situational awareness plays even a more vital role when the operator is performing novel and sophisticated tasks. Only using onboard camera will deteriorate perception of the remote environment, especially in highly occluded spaces, such as after-disaster scenarios. 

\begin{figure}[!ht]
\centering
	\includegraphics[scale=0.5]{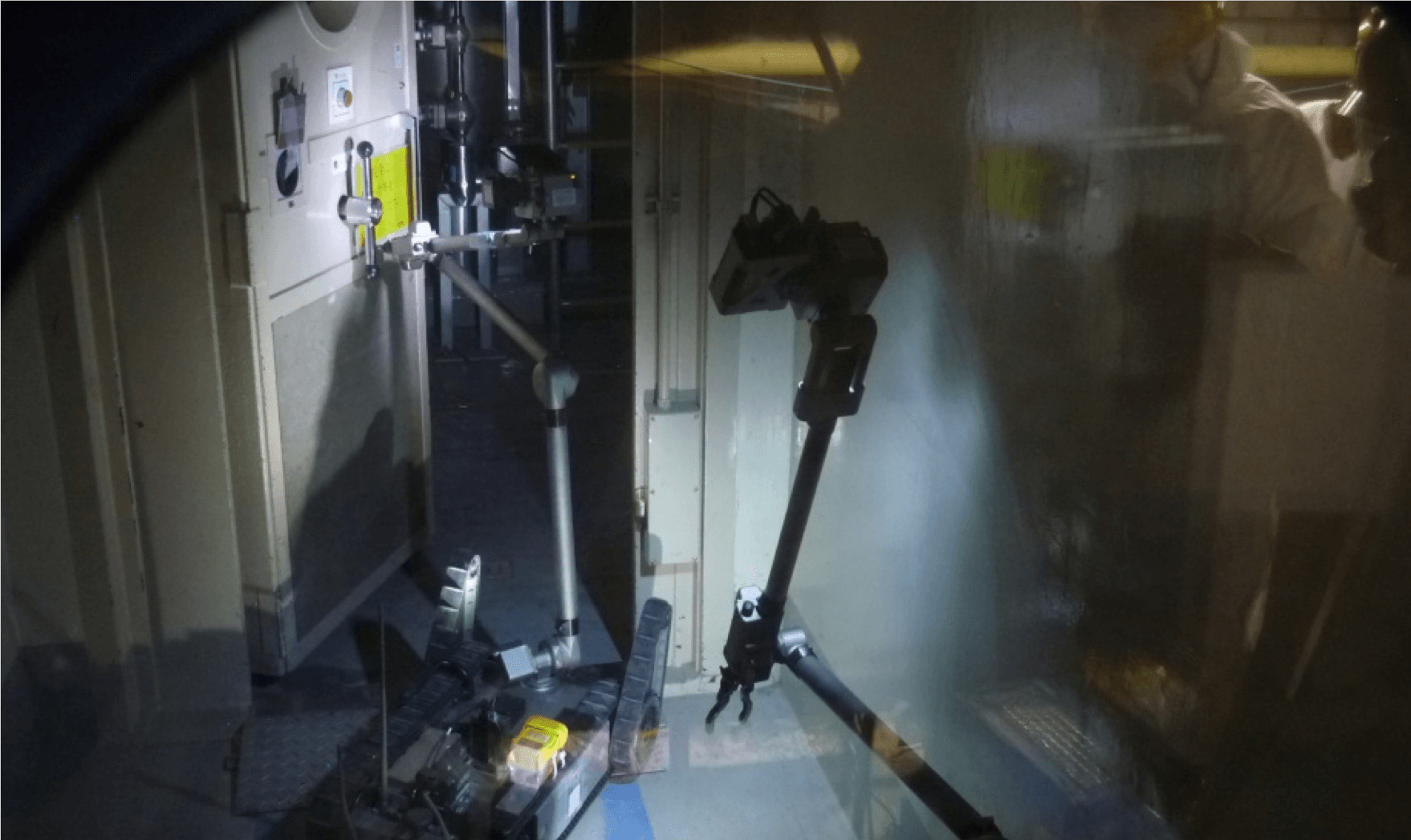}
	\caption{Two iRobot Packbots Working Together to Open a Door during the Fukushima Daiichi Nuclear Accident (Image Courtesy: JAEA)}
	\label{fig::2packbots}
\end{figure}

The Fukushima Daiichi nuclear power plant is an example of such environment. The expected duration of 30 years to conduct decommissioning makes it a living laboratory for the daily use of robots. Teleoperated robots were used in pairs to expedite mission execution. As shown in Fig. \ref{fig::2packbots}, two iRobot Packbots were used to conduct radiation surveys and read dials inside the plant facility, where the second Packbot provided camera views of the first robot in order to manipulate door handles, valves, and sensors faster. Another example is that QinetiQ Talon UGVs let operators see if their teleoperated Bobcat end loader bucket had scraped up a full load of dirt to deposit over radioactive materials. 

Since then, the use of two robots to perform a single task has been formally acknowledged as a best practice for decommissioning tasks, e.g., cutting and removing a section of irradiated pipe. However, the Japanese Atomic Energy Agency (JAEA) has reported that operators constantly try to avoid using a second robot. The two sets of robot operators find it difficult to coordinate with the other robot in order to get and maintain the desired view but a single operator becomes frustrated trying to operate both robots. However, two robots are better than one. In 2014, an iRobot Warrior costing over \$500K was damaged due to inability to see that it was about to perform an action it could not successfully complete. The experienced operator had declined to use a second robot. Not only was this a direct economic loss, the 150kg robot was too heavy to be removed without being dismantled and thus cost other robots time and increased their risk as they have to navigate around the carcass until another robot could be modified to dismantle it.

\begin{figure}[!ht]
\centering
	\includegraphics[scale=0.3]{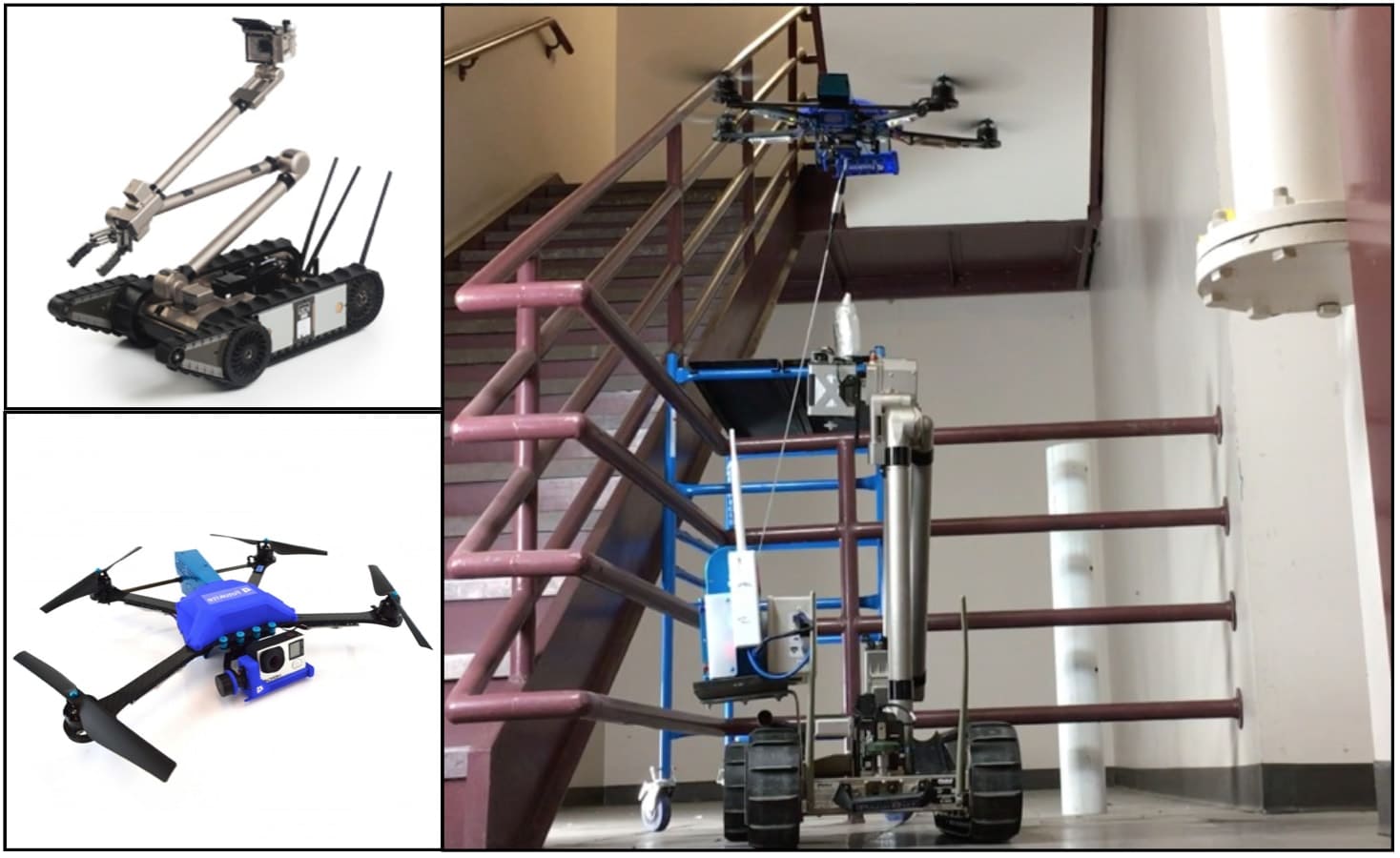}
	\caption{The Proposed Solution: An autonomous tethered UAV visually assists the teleoperation of a UGV (reprinted from \cite{xiao2019autonomous})}
	\label{fig::team}
\end{figure}

This research proposes to replace the second robot and its operating crew with a fully autonomous visual assistant robot, in particular, a tethered UAV (Fig. \ref{fig::team}). The proposed visual assistant can autonomously position itself at the cognitively best external viewpoint of the primary robot's workspace using a pre-established affordance-based viewpoint quality theory. In order for this method to be practical for unstructured or confined spaces where it offers the most benefit, the autonomous navigation has to be trustworthy. Thus, a fundamental theory is required to rate the risk associated with navigating to a viewpoint and to be incorporated into the choice of viewpoint. Considering the user's low tolerance of rapidly shifting viewpoints, the visual assistant needs to plan and execute not only safe but also visually smooth path, which is capable of maximizing the visual assistance quality not only at the goal point, but also along the entire flight. The execution of the planned safe and high-quality path is based on a novel tethered UAV, as the visual assistant platform in the marsupial robot team, paired with the teleoperated primary ground robot. 

\section{Research Question}
Based on the aforementioned motivations, the primary research question to be addressed in the scope of this dissertation is: 

\emph{How can a robotic agent reason about motion risk and balance it with mission reward during path planning and execute the path with a fully autonomous tethered UAV operating in indoor cluttered GPS-denied environments? 
}

This research first formulates a formal definition of robot motion risk and presents an explicit risk representation in unstructured or confined environments. Based on that, a risk-aware path planning algorithm is developed to balance the tradeoff between viewpoint quality reward and flight execution risk during the entire visual assistance process. In addition to the high level risk reasoning and risk-aware path planning, a low level motion suite is also developed in order to realize the high level risk-aware flight trajectory on a tethered UAV: the motion commands are tailored to fit the tethered agent by means of a set of low level motion building blocks, including a novel tether-based localization model with negligible computational overhead, motion primitives for the tethered airframe based on position and velocity controls, two different approaches to negotiate tether with complex obstacle-occupied environments, and a reactive tether-based visual servoing behavior. 

\section{Reasons to Use Tethered UAV}
Advances in small unmanned aerial vehicles, especially tethered UAVs, and personal satellite assistants such as SPHERES \cite{dorais2003personal, stoll2012spheres} suggest that flying robot helpers will soon supply the needed secondary visual perspective or even shine a spotlight on dark work areas \cite{bradshaw2000human}. The ability to hover in place and the mobility to navigate in three-dimensional spaces make UAVs a perfect fit for the visual assistance purpose. The mature controller, hardened hardware, along with the readily available onboard camera further speak for the choice of UAV as the next generation visual assistant. The live feed from the onboard camera of an autonomously flying UAV at the cognitively best external viewpoint in 3D space outperforms the video from a stationary camera or teleoperated ground robot. 

While using a tether seems of disadvantage since the tether may interfere with obstacles in the unstructured or confined environments, the tether provides two indispensable advantages: (1) UAV's onboard battery cannot provide sufficient fight time in order to match with the battery duration of a ground robot mission. A tether connecting the UGV and UAV can make the battery onboard the UGV sharable with the UAV. Therefore the mission duration for both vehicles could be matched. (2) In remote mission-critical environments, tethered operation is always mandatory due to safety concerns. In case of a UAV malfunction or crash, tether could be used for retrieval, by the UGV dragging the UAV out through the tether. Another advantage of using a tether is that it provides an alternative means of localization with negligible computational overhead in indoor GPS-denied environments, which will be discussed in detail in the following chapters. 

\section{Novelty and Contributions}
This dissertation has both intellectual merit to the field of path and motion planning and societal benefit to improve next generation safe robotic teleoperation performance. 
\subsection{Intellectual Merit}
The intellectual merit of the proposed work has three facets: a reasoning framework for robot motion risk including a formal risk definition and an explicit risk representation in unstructured or confined environments, a risk-aware path planner that balances the tradeoff between reward and risk, and a low level motion suite to accommodate and take advantage of a tethered UAV. 

\subsubsection{Motion Risk Reasoning}
This work proposes a formal robot motion risk definition and an explicit motion risk representation in unstructured or confined environments in contrast to the conventional implicit representation based on probabilistic or uncertainty models. This provides a framework to explicitly and formally reason about robot motion risk in complex spaces and a means to prioritize safer paths, and therefore improve robot motion trustworthiness in unstructured or confined environments. 

\subsubsection{High Level Risk-aware Path Planner}
Based on a pre-established affordance-based viewpoint quality (reward) theory and the explicit motion risk representation, this work proposes a high level path planner which can plan the navigational target and the path leading to it while balancing the reward and risk along the entire path. A majority of the proposed risk elements could be optimally addressed by the risk-aware planner and a new class of minimum-risk paths based on a more formal, general, and comprehensive risk representation could be computed.  

\subsubsection{Low Level Tethered Motion Suite}
This work presents a full motion suite to implement any high-level three-dimensional path on tethered UAVs in unstructured or confined environments. This motion suite can both accommodate the issues of and take advantage of the tether, to facilitate autonomous free flight in GPS-denied and obstacle-occupied indoor environments. This tethered motion suite encompasses the following building blocks from perception to actuation: 

\textbf{Tethered UAV Indoor Localization:}
This work presents a UAV localization scheme based on a taut tether in indoor GPS-denied environments. Not relying on any vision or point cloud based methods, this tether-based localizer requires negligible computational overhead. 

\textbf{Low Level Motion Primitives for Tethered UAV:}
This work presents two motion primitives for tethered UAV, with position and velocity controls. The former is based on PID controllers and is expected to be robust against singularity but less accurate when facing paths with very sparse waypoints. The latter is based on Jacobian and can generate precise flight execution even if the waypoints are sparse. However, it is sensitive to singularity. 

\textbf{Low Level Tether Handling Approaches:}
This work presents two tether planners without and with the possibility of tether contacting with environment. The tether planners maintain benefits brought by tether while allow the UAV to navigate through cluttered environments with a tether and enable free flight in Cartesian space. 

\textbf{6-DoF Visual Servoing with a Tethered UAV:}
This work presents a reactive tether-based visual servoing approach which allows the viewpoint to maintain a constant 6-Degree-of-Freedom (DoF) configuration with respect to a moving target. The usage of a UAV can increase visual servoing quality and coverage. This reactive approach is complementary to the deliberate high level risk-aware path planner. 

\subsection{Societal Benefit}
The proposed work is expected to have societal benefit in nuclear operations, bomb squad, disaster robots, and other domains with novel tasks or highly occluded environments, where two robots, a primary and a secondary that acts as a visual assistant to overcome perceptual limitations, are used. The primary robot's teleoperator would benefit from operating under a series of good viewpoints and being alleviated from team work demands since coordination between the two operating crews is no longer necessary. Due to the explicit motion risk representation, trustworthy autonomy, planner under navigational and human perceptual constraints developed by this research, it will have significant societal benefit and enable co-robots to create more resilience to disasters and public safety incidents, accelerate the safe decommissioning of nuclear operations, and even aid missions for personal assistant robots. There is potentially significant economic impact in immediate productivity gains with robots in service now and increased competitiveness with future ground and aerial robots. 

\subsection{List of Publications and Awards}
The work by the author related to the current dissertation was published in \cite{xiao2017visual, xiao2018indoor, xiao2018motion, xiao2019explicit1, xiao2019benchmarking, xiao2019explicit2, xiao2019autonomous}. The project was a Finalist for Best Student Paper Award at 2018 IEEE International Symposium on Safety, Security and Rescue Robotics (SSRR) for paper published in \cite{xiao2018indoor}.

\section{Organization}
 Chapter \ref{chapter::related_work} summarizes related work in the fields of UGV/UAV team, path planning to maximize reward under risk or constraints, and motions of tethered UAVs. Chapter \ref{chapter::risk_representation} introduces the proposed robot motion risk reasoning framework including the formal risk definition and explicit motion risk representation for unstructured or confined environments. Chapter \ref{chapter::high_level} presents the proposed high level risk-aware path planning algorithm, which plans based on the proposed risk framework in Chapter \ref{chapter::risk_representation} and balances the tradeoff between reward and risk. Chapter \ref{chapter::low_level} presents the motion suite applicable for tethered flight in indoor GPS-denied and obstacle-occupied environments. It includes low level motion preceptors, controllers, planners, and executors to take advantage of the tether and tailor the motion commands to fit on a tethered UAV to fly any path computed by the planner in Chapter \ref{chapter::high_level} in indoor cluttered environments. Chapter \ref{chapter::experiments} presents experiments for all the individual components in the low level motion suite proposed in Chapter \ref{chapter::low_level}. All experimental results are presented and discussed in detail, in order to validate the proposed tethered motion approaches. Chapter \ref{chapter::integrated_demonstration} presents an integrated demonstration conducted in a real-world physically unstructured or confined environment, which resembles the scenarios encountered by search and rescue robot and personnel in Fukushima Daiichi nuclear power plant. The macroscopic experiments focus on the validation of the proposed risk reasoning framework, risk-aware planning, and implementation of the entire motion suite in real-world physical environments. Chapter \ref{chapter::summary_and_conclusions} summarizes and concludes this dissertation. 

\chapter{RELATED WORK}
\label{chapter::related_work}
This chapter presents the related work from different aspects of this dissertation. The related work is divided into three categories: (1) UGV/UAV team, (2) path planning to maximize reward under risk or constraints, and (3) motion of tethered UAVs. 

\section{UGV/UAV Team}
UGV and UAV teams have been widely used in the robotics community, especially in field robotics. UGVs are popular in field deployment since they are stable, reliable, durable, and can effectively project human presence to avoid risking human agents. They could carry large computational devices and manipulation tools so they could be intelligent and dexterous enough to represent humans to actuate upon the real world. However, due to the direct contact of their cumbersome chassis with the terrain, relatively conservative locomotion principles such as wheels or tracks, and sophisticated and heavy payloads which are sensitive to abrupt accelerations, UGVs usually move slowly and are limited to a two dimensional work space reduced from the entire 3D work envelope. On the other hand, thanks to UAV's aerial maneuverability enabled by the light-weight airframe, their superior mobility and therefore increased coverage of the entire workspace can provide enhanced situational awareness from locations in the workspace which are inaccessible to UGVs. But they lack the capability to actuate and affect the physical world. Therefore, researchers have looked into the combination of both to utilize the advantages and avoid the disadvantages. 

\cite{cheung2008uav} used a PackBot UGV and a Raven UAV to pursue and track a dynamic target. The UAV was used to survey an area and geolocate the target. This was shared with the UGV and the UGV pursued the target. \cite{chaimowicz2005deploying} deployed air-ground multi-robot teams in order to increase situational awareness, achieve cooperative sensing, and construct radio maps to keep team connectivity. Another example that utilized the strong suits of both types of unmanned vehicles was shown in \cite{tanner2007switched}: in order to detect a target, UGVs were stabilized into a guarding formation, and the UAVs scanned the enclosed regions. \cite{phan2008cooperative} employed a hierarchical task allocation scheme to coordinate blimp, quad-rotors, and rovers for wild fire detection and fighting. The flight paths planned for the UAVs were usually mission-oriented and the UAVs were assumed to be flying in outdoor open space without the existence of any obstacles. 

The above-mentioned works mainly focused on the cooperation of UGV and UAV to improve system performance. Another body of literature used UAV to augment UGV's perception or assist UGV's task execution. \cite{frietsch2008teaming} used UAV to help geolocate UGV in order to improve navigation solution in the case of GPS loss. In \cite{chaimowicz2004experiments}, a blimp acted as ``an eye in the sky'' and determined the position of UGV. In order to address the mapping between the UGV's 3D coordinates in the world frame and 2D coordinates of its projection on the image plane, \cite{rao2004calibrating} derived a subset of the parameters of the homography from the relationship between the velocity of the UGV on the ground plane and the velocity of its projection in the image. In addition to localizing UGV in UAV's image frame, \cite{rao2003visual} used differential flatness to generate effective control strategies only based on UAV's visual feedback. In most scenarios, the UAV's enhanced perception was achieved at a stationary and elevated viewpoint. 

In this proposed research, the UGV is teleoperated by human operator, while the UAV is autonomous and flies around the UGV to visually assist the teleoperation by providing the operator with an external viewpoint. Unlike executing a flight path in outdoor open space to utilize UAV's wide coverage in the workspace or assisting the UGV in terms of localization and navigation from an elevated stationary aerial view, the UAV used in this research needs to fly through unstructured or confined environments in order to visually assist the teleoperator from an optimal external view, while balancing the viewpoint quality and flight risk along the entire path. 

\section{Path Planning to Maximize Reward under Risk or Constraints}
To understand how reward and risk were addressed in planning, it is firstly investigated what is risk for the planners and how is risk represented. Approaches of planning to mitigate risk is then researched, with a focus on not only minimizing risk but also maximizing reward at the same time. 

\subsection{Risk Definition and Representation}
In order to review how the previous path planning works dealt with risk, an understanding of risk definition and representation is a prerequisite. In the robotics literature, to the author's best knowledge, risk of robot motion is not formally defined, except being referred to as some negative impact or factor in ad hoc situations. Or it is simply treated as a numerical measure of the severity/negativity related with certain aspects of motion. Due to the lack of a formal definition of what risk is in the robotics literature, this related work review only focuses on how risk is represented. Even without a formal definition, risk is still either represented as (1) a risk function of the state or as (2) sensing and action uncertainty. 

\subsubsection{Explicit Risk as Function of State}
Explicitly representing risk in physical space is directly applicable to unstructured or confined environments. The body of literature is not large, with a majority of work focusing on risk representation as a function of state. 

\cite{soltani2004fuzzy} represented the workspace by two risk layers: hazard data layer and visibility layer. The risk of each state along the path was embedded in those layers based on fuzzy logic. The hazard data layer was a fuzzy function that mapped a state to a hazard value depending on how far is the state from a hazard source. Visibility layer mapped the state to a visibility value reflecting the obstacle density around the state. The objective function was a weighted sum of the two layers and distance and the planner used Dijkstra's search algorithm \cite{dijkstra1959note} to minimize path cost. To the author's best knowledge, this is the only work in the literature that considered more than one risk sources: risk from being close to hazard and risk from having low visibility. A similar approach was taken by \cite{de2011minimum}, where a risk map was generated based on ground orography and A* and genetic algorithm were used to minimize the risk. The ground orography is simply treated as hazard or obstacle for the robot, being close to which induces motion risk. \cite{vian1989trajectory} presented the idea of risk index for any particular location (state) and assumed risk to be a function of location only. Risk was defined in horizontal plane, vertical plane, and ad hoc risk area. Only one risk was used at a time for planning. \cite{zabarankin2002optimal} based its risk representation on the same risk index idea, whose value was proportional to risk factor and reciprocal to squared distance to threat. Risk caused by multiple threats were summed and this accumulated value was integrated along the path. \cite{gu2006comprehensive} further proposed an accumulative parametrized function based on distances to multiple threats. The set of functional parameters were set manually. \cite{feyzabadi2014risk} used a similar distance-based function to represent state-dependent risk in its experiment. 

Data-driven approaches to predict potential risk of a certain state could also be seen in prior works. In the field of Autonomous Underwater Vehicles (AUVs), \cite{pereira2011toward} defined risk as a function of state location with ship occurrences averaged over time domain since historical Automated Identification System (AIS) data was available. The planner was based on A* search. \cite{pereira2013risk} further extended this work by using expected risk in the search and using MDP. In the MDP approach, the only positive reward was assigned to the final state, resulting the robot getting to the goal. Risk was represented in the same way from AIS data and incorporated into the MDP as a negative reward. In traffic planing, \cite{krumm2017risk} utilized historical traffic data to predict crash probability to represent the risk associated with driving through each corresponding highway segment. 

\begin{table}[]
\centering
\begin{tabular}{|c|c|c|}
\hline
Reference \#       & Risk Element(s)              & Explicit Representation \\ \hline
\multirow{2}{*}{\cite{soltani2004fuzzy}} & Distance to hazard           & Fuzzy logic             \\ \cline{2-3} 
                   & visibility                   & Fuzzy logic             \\ \hline
\cite{de2011minimum}                  & Distance to local elevation  & Fuzzy logic             \\ \hline
\cite{vian1989trajectory}                  & Distance/altitude/risk area  & Risk function           \\ \hline
\cite{zabarankin2002optimal}                  & Distance to radar            & Risk function           \\ \hline
\cite{gu2006comprehensive}                   & Distance to threat center    & Risk function           \\ \hline
\cite{feyzabadi2014risk}                   & Distance to Closest obstacle & Risk function           \\ \hline
\cite{pereira2011toward} \cite{pereira2013risk}                  & Existence of another ship    & Data-driven prediction  \\ \hline
\cite{krumm2017risk}                  & Predicted traffic crash      & Data-driven prediction  \\ \hline
\end{tabular}
\caption{Explicit Risk Representation in the Literature}
\label{tab::explicit_risks}
\end{table}

The explicit risk representation in the literature is summarized in Tab. \ref{tab::explicit_risks}, in terms of what elements contribute to risk and how risk is represented. From the above-mentioned explicit risk representation, distance to closet threat is the main risk element considered, in the form of hazard, orography, radar, obstacle, etc. Although it is naturally assumed that being closer to threat brings more risk, a formal definition of what risk is is still missing. However, different ways to represent risk has been used. To represent risk as a numerical value for the planner, fuzzy logic, parameterized functions, and data-driven predictions were used. \cite{soltani2004fuzzy} is the only work that considered multiple risk elements and the approach to combine them was weighted sum based on human heuristics. All those previous works in the first category explicitly represented risk as a function of state and used search or MDP algorithms to find the minimum risk path. It also worth to note that all the above-mentioned works assumed that risk is additive. That is, the risk associated with an entire path is a simple addition of all the risks at each individual states, the minimal component of the path. However, the justification of the additivity of risk remains missing. 

Other works that did not directly formulate risk but characterized unstructured or confined environments are \cite{agarwal2014characteristics} and \cite{murphy2014disaster}. Characteristics including access elements and tortuosity were proposed, which haven't been but could be used as explicit risk elements, especially in unstructured or confined environments. 

\subsubsection{Implicit Risk as Model Uncertainty}
Another body of literature implicitly modeled risk as uncertainty in theoretical belief space. The uncertainty was either represented as partially known state, or probabilistic action model. One rationale behind modeling uncertainty is that stochastic sensor and action models may introduce risk into path execution, e.g., not knowing exactly where the robot is may lead to collision with obstacles. This is the reason why planning is conducted in belief space. Probability map of threats \cite{jun2003path}, Belief Roadmap (BRM) \cite{prentice2010belief}, Rapidly-exploring Random Belief Trees \cite{bry2011rapidly}, linear-quadratic controller based on an ensemble of paths \cite{van2011lqg}, local optimization over Partially Observable Markov Decision Process (POMDP) \cite{van2012motion}, Feedback-based Information Roadmap (FIRM) \cite{agha2014firm} were used, representing risk as model uncertainty, to plan safe path. Another approach to deal with partially-known environments is dynamic replan \cite{stentz1994optimal}. In this second category, instead of explicitly representing risk as a numerical value, the exact notion of risk does not exist, and it was embedded in probabilistic system model or uncertainty. The risk representation was implicit, i.e., risk was due to indeterministic robot sensor and action models, not explicit, i.e., state A was riskier than state B. This type of risk reasoning requires a convincing method to quantify the probabilistic model when going beyond theory and planning with real physical robots. 

\subsubsection{Proposed Approach}
In this work, the lack of formal definition of risk for robot motion is firstly mended. A formal definition for robot motion risk is proposed using propositional logic and probability theory. Based on this formal definition, the explicit risk representation approach is taken, reasoning the risk of robot motion based on path. Multiple risk elements are taken into account and the risk of path execution captures a wider variety of aspects in robotic locomotion. Due to the usage of probability theory, no assumption of additivity of risk is necessary. Instead, probability dependence and chain rule are used to depict the relationship between risk of the entire path and risk of individual states. Furthermore, propositional logic and probability theory also show that the risk robot is facing at a certain state is also dependent on history.

\subsection{Path Planning with Cost Tradeoff }
Two levels of trade-off exist in: (1) the trade-off between achieving higher reward but minimizing risk simultaneously, such as visiting good viewpoints but still guaranteeing minimal-risk motion, and (2) minimizing risk from contradicting risk elements, e.g. avoiding a long detour but still maintaining high-clearance on the entire path. The approaches to address both trade-offs remain methodologically identical. 

Although \cite{soltani2004fuzzy} is the only work that handled more than one risk elements in the planner, robots working in physical environments always face risks from multiple sources. Being closer to obstacles could cause a crash and executing a long path may cause the robot stuck in the middle as well. The robot will need to minimize the negative effect from all risk elements. 

In the presence of reward, risk is mostly treated as penalty to the reward, negative cost, safety constraint, or chance on constraint violation. The trade-off was usually handled by manually setting one and then maximize or minimize the other. 

\subsubsection{MDP: Reward with (Chance) Constraints}

A popular approach to handle reward and risk is to use (PO)MDP. As standard MDP inherently contains reward but not risk, researchers have looked into representing risk as negative reward (penalty) or constraints (C-POMDP) with unit cost for constraint violation. \cite{pereira2013risk} modeled constraints as penalties on the reward by subtracting penalty from reward function. However, as shown by \cite{undurti2010online}, the choice of penalty value that achieves the desired balance between risk and reward is not clear or even does not exist. So the planner that models both rewards and risks in the reward model can switch abruptly between being too conservative and too risky. This is why \cite{undurti2010online} proposed to separate reward and risk, using an offline constraint penalty estimate for beyond planning horizon and an online reward optimization within the planning horizon. It was further extended to continuous domain using function approximation \cite{undurtifunction} and then to address multi-agent system and dynamic constraints \cite{undurti2011decentralized}. Here, risk was treated as unit cost incurred when a hard constraint on the system would be violated. To solve C-POMDP, \cite{isom2008piecewise} used dynamic programming to find for each belief state the vector that has the best objective function valuation while still satisfying the constraint function. Approximate dynamic programming method was proposed by \cite{kim2011point} to solve POMDP using point-based value iteration, for speed and scalability. \cite{poupart2015approximate} used approximate linear programming to solve C-POMDP and outperforms the point-based value iteration in \cite{kim2011point}. Another suboptimal but efficient approximation algorithm was hierarchical Constrained MDP (C-MDP) \cite{feyzabadi2014risk}. Going beyond unit cost for constraint violation, Chance Constrained Partially Observable Markov Decision Process (CC-POMDP) was proposed by \cite{santana2016rao}, which was based on a bound on the probability (chance) of some event happening during policy execution. 

\subsubsection{(Chance) Constrained RMPC}
Besides MDP-based methods, if a system dynamic model is available, Robust Model Predictive Control (RMPC) is another alternative approach to address reward and risk at the same time, 

\cite{luders2010chance} proposed a chance-constrained rapidly-exploring random tree (CC-RRT) approach, which used chance constraints to guarantee probabilistic feasibility at each time step for linear systems subject to process noise and/or uncertain, possibly dynamic obstacles. \cite{luders2013robust} expanded this approach to consider both chance-constrained environmental boundaries and guaranteed probabilistic feasibility over entire trajectories (CC-RRT*).

Other works emphasized on risk allocation, i.e., to allocate more risk for more rewarding actions. \cite{ono2008efficient} used a two stage optimization scheme with the upper stage optimizing risk allocation and lower stage calculating optimal control sequence that maximizes reward, named Iterative Risk Allocation (IRA). \cite{ono2008iterative} further discussed IRA's optimality. \cite{vitus2011feedback} also used risk allocation and feedback controller optimization to reduce conservatism and improve performance. Risk is still represented as a probability of constraint violation (mission failure).

The majority of aforementioned works modeled risk as chance constraints, i.e., the probability of certain system constraint being violated. All the approaches in the literature, however, only focused on risk, or constraints, caused by collision with obstacles. With the system modeled within Cartesian space, the constraints of the dynamic system were formulated as no intersection between the robot trajectory and obstacles in the environment at each time step. A risk-aware plan was only a path with high collision-free probability. The path planner searched for path to reach destination, minimize cost, or finish mission, while satisfying a pre-determined bound on probability of constraint violation, i.e., probability of collision is less than or equal to certain threshold value. Being modeled only in a geometric point of view, approaches to model risk caused by any other sources than obstacles were overlooked, e.g., robot motor overheat, getting stuck in granular environments, etc. Using chance constraints, risk was only a bound or threshold of constraint violation. Furthermore, the temporal or spatial (multiple obstacles) dependencies of constraint violation probability were either assumed to be independent or relaxed using ellipsoidal relaxation technique or Boole's inequality. For example, the probability of constraint violation at this time step was only a function of $x_t$, the state at this time. Although some history information may be embedded in the system dynamics updates, these two methods, especially when residing only in Cartesian space, neglected the important dependencies on the motion history and the rough approximation introduced significant conservatism. 

\subsubsection{Multi-objective Optimization} 
Another approach to address the trade-off between reward and risk or different cost aspects is multi-objective optimization. The solution to a multi-objective optimization problem is a Pareto-optimal front, a set of solutions to each of which no changes could be made to improve one objective while not sacrificing the others. One of the most popular approaches to solve multi-objective optimization was through genetic algorithm \cite{deb2000fast, deb2002fast}. In the robotic planning community, although the two objectives were not necessarily formulated as reward and risk, multi-objective optimization was usually used to leverage path costs from different aspects. \cite{davoodi2013multi} used genetic algorithm to find a set of Pareto-optimal solutions which balance the trade-off between path length and clearance. \cite{jun2010multi} used the same idea in terms of path length, smoothness, and security. All these approaches generated a set of Pareto-optimal solutions, which could be treated as the ultimate results of the multi-objective optimization problem. However, for robot path planning, one final path still needs to be chosen from the Pareto-optimal set and then to be executed on the robot. This step remained unclear, and the researchers usually only presented a set of Pareto-optimal paths to ``provide great conveniences for a robot to choose appropriate path according to different preferences in practice'' \cite{jun2010multi}. \cite{lavin2015pareto} addressed this problem by combining the Pareto-optimality idea with traditional A*  into a A*-PO algorithm for path planning. This algorithm yielded one single optimal solution so the user or robot did not need to choose from a set of solutions. However, a further investigation into the A*-PO algorithm revealed that while operating over Pareto-optimal front during every iteration of A*, the planner was still choosing the ``best'' solution from the Pareto-optimal front based on normalization and weighted sum of different objective dimensions of the Pareto space. The planner was just transferring the multi-objective optimization problem into a single-objective optimization problem among all Pareto-optimal solutions in every iteration. Therefore in case of the reward and risk or different risk elements of this work, the trade-off is still pending ad hoc decision based on manual arbitration in practice, even though a set of Pareto-optimal solutions are available. Therefore, no matter the multi-objectives are defined as different costs (risks), or reward and risk, it is still unclear how to find the one ultimate optimal solution among the Pareto-optimal front, and then to execute on the robot. 

\subsubsection{Proposed Approach}
Both MDP and RMPC methods require an artificial definition of risk, in the form of negative reward (penalty), unit cost, or bound on probability, for constraint violation. How to choose penalty value as negative reward is not clear and a desired value may not even exist, leading to policies that are overly risk-averse or overly risk-taking. Assigning unit cost for constraint violation has incorrect probability values when constraint violation does not cause policy execution to terminate and even with this assumption belief state computations are strongly impacted \cite{santana2016rao}. 

Modeling risk as chance constraints looked like a more natural and objective way to reason about risk, but current works only addressed constraints in terms of probability of collision with obstacles and collision-free path. During planning with other (more important) objectives, risk-awareness is only achieved by maintaining a bound or threshold of probability of collision. This bound is  subjective to human choice and hard to be determined. Even risk allocation can make sure that significant risk is only taken on most valuable actions, actions that lead to large reward, what is the acceptable bound on the total risk is not clear, e.g., it is hard to define the value of scientific discovery compared to the cost of losing the robot. Being treated as (chance) constraints during planning for other objectives (such as reaching destination, minimizing cost, etc.), minimizing risk is not the objective of the planner itself, but constraint(s) to satisfy. However, in many scenarios, especially when locomoting in unstructured or confined environments, reaching the goal location in the safest manner is the only objective of interest. With the existing literature, what is a general definition and representation of risk (in addition to collision with obstacles) and how to plan with minimum risk (not only within a manually determined probability bound or threshold of collision) remain unclear. A new framework of risk is necessary to enable more general and comprehensive reasoning of risk, including risk caused by other non-geometric-related sources, finding minimum risk path in an absolute sense, not probabilistic threshold, and reasoning about risks's dependencies among individual time steps and different risk sources to tighten the conservatively relaxed risk bound.  

It is possible to model the trade-off between reward and risk or the trade-off among multiple risk sources as a multi-objective optimization problem. But choosing the ultimate optimal path from a set of Pareto-optimal solutions as the results of multi-objective optimization is still ad hoc and subject to human bias. 

In this research, the necessity of manually arbitrating either the maximum acceptable risk or minimum expected reward is avoided. For the trade-off between reward and risk, a utility function as a ratio between reward and risk along the entire path is proposed, as a measurement of how much reward is collected when taking one unit of risk, or how much risk is taken to achieve one unit of reward. Apparently, a path with higher utility value is more favorable. In this sense, the balance between reward and risk is represented as a ratio and no longer dependent on any manually chosen weights, value, or probability. In addition, the desirable goal state is not specified to the planner beforehand, but discovered by the planner based on optimal utility during the planning process. For the trade-off between contradicting risk elements, negative impacts from different sources of risk are combined using our formal risk definition with propositional logic and probability theory. It no longer depends on normalization and a set of weights from heuristics. It exceeds the scope of obstacle-related risks, i.e., constraints due to collision (intersection between motion trajectory and obstacles), and focuses on minimum motion risk in an absolute sense, not only a bound on probability of constraint violation. It formally reasons the temporal (longitudinal) and spatial (lateral) dependencies of risk among all history time steps and different risk sources. No approximation or relaxation techniques such as ellipsoidal relaxation and Boole's inequality are necessary. The risk of robot motion is therefore explicitly represented in a more general, objective, not ad hoc way. 

\section{Tethered UAVs}
Tethered UAVs have been studied in the literature. However, due to the limitations caused by the tether, they don't belong to the main stream of UAV research. Therefore, research on tether for other types of unmanned vehicles than UAVs is also included.

For many applications, especially mission-critical scenarios, tethered operation is a requirement, e.g. for power or communication. \cite{perrin2004novel} designed a self-actuated tether for rescue robots using hydraulic transients. The self-actuation mechanism made the tether capable of moving its own weight and remaining free while traversing around corners. Small robots for search and rescue could then have sufficient power and reliable communication while not worrying about the added drag of the tether. \cite{minor2002automated} designed a tether management system for astronauts during microgravity extravehicular activities. The tether system was equipped with a remotely releasable, self-locking robotic gripper and an automated tether retractor. \cite{prabhakar2005dynamics} and \cite{muttin2011umbilical} investigated dynamic modeling and control of tether for a remotely operated underwater vehicle and UAV, respectively. 

Tether can bring extended flight time to UAVs, especially due to the small battery a UAV could carry. \cite{zikou2015power} presented a tethered UAV platform with the purpose of power-over-tether considerations with a specific tether lengthening and retraction method. There was even reported field deployment of tethered UAV at Berkman Plaza II collapse by \cite{pratt2008use}. The usage of tether was due to US Federal Aviation Administration (FAA) requirements for unregulated flight below 45m. However, other than designing the mechanism to manage the tether for power considerations or using a human tether manager in the disaster field, the tethered UAV was not treated differently as a tetherless agent. 

Tether was also included into the dynamics of UAVs. For example, the reliability of tether was used for station keeping of UAVs. \cite{lupashin2013stabilization} investigated stabilization of a UAV on a taut tether using only on-board inertial sensors. \cite{nicotra2014taut} used a similar tethered UAV setup and developed a nonlinear controller with constraints to stabilize and steer the UAV to a desired set-point while maintaining a taut tether at all times. In addition to station keeping, utilizing the tension on the tether and stability of the attachment point with ground, \cite{schulz2015high} achieved high-speed steady flight on a UAV. 

The disadvantage brought by the tether was not deeply investigated, or even not looked into for UAVs at all. One important problem with tether is contact or entanglement with the environment or other agents. \cite{hert1996ties} developed a motion planning algorithm for multiple mobile tethered robots in a common planar environment. A sequential motion strategy was designed for the robots that would not entangle the tethers. 

Flights by tethered UAVs were pre-defined, either station keeping or open-loop execution, and conducted in obstacle-free environments. To the author's best knowledge, no motion planning and execution algorithms for autonomous flight of tethered UAV in indoor unstructured or confined environments exist in the current literature. 

Tether is an indispensable part of this research, not only for power extension and safety retrieval purposes. This research uses tether as an alternative means of indoor UAV localization with negligible computational overhead. The tether-based motion primitives are used to achieve planned flight trajectory or visual servoing. Low level motion planner focuses on the mitigation of the disadvantages or inconveniences caused by tether in cluttered environments, with the aim of flying tethered UAV in non-free, obstacle-occupied spaces as if it were tetherless.

\section{Summary of Related Work}
Related work of this dissertation covers a variety of topics due to the diverse components related with this work. 

This dissertation aims at developing a new paradigm of co-robots, with an autonomous marsupial visual assistant robot navigating to provide external viewpoint to the primary robot's human operator. The locomotion and navigation of the aerial visual assistant exceed the existing UGV/UAV operations in the literature, either flying in outdoor open space to utilize UAV's wide coverage in the workspace or assisting the UGV in terms of localization and navigation from an elevated stationary aerial view. The proposed aerial visual assistant needs to autonomously fly through indoor unstructured or confined environments in a risk-aware manner, and provide optimal viewpoint quality at the same time.  

To enable risk-aware behavior, a basic framework to reason about risk is a prerequisite. However, this literature review reveals the lack of formal definition of risk for robot motion. Risk was addressed in ad hoc ways, only suitable for very specific robot and application of interest. This is also the reason why related work usually only considered one element, such as maximizing distance to obstacles, ignoring other risk elements and their combined effect on robot motion. The literature also assumed risk to be additive along the execution of a path. The additivity, however, was not justified. Furthermore, existing works treated risk only as a function of a state, thus the risk a robot faces at a certain point on a path is only a function of that point alone. But apparently dependencies of risk at a certain state on the history of the path exist, e.g. the risk of battery depletion at a state is dependent on the path the robot took to come to this state. Ignoring the dependencies on history weakens our proper understanding of robot motion risk. This dissertation proposes a formal definition for robot motion using propositional logic and probability theory. Explicit approach is taken and multiple risk elements are considered. The dependencies of risk on history and non-additivity are formally justified using probability chain rule. The effect of multiple risk elements are combined with conditional independence. The proposed formal definition and explicit representation could serve as a numerical metric to quantify safety of robot motion in unstructured of confined environments.  

In the existence of reward, risk was treated as negative reward (penalty), unit cost, or bound on probability, for constraint violation. These methods only focused on obstacles, either assumed independence or introduced conservatism, and were artificial and subject to human bias. Based on the risk definition and representation proposed by this dissertation, planners are given a metric to quantify and compare absolute risk levels of different paths. Therefore the utility of taking risk could be reasoned against the reward achieved. This dissertation aims at creating a risk-aware planner that also maximizes reward and is immune to human bias.  

For tethered aerial vehicles, despite the power-over-tether advantage, how to overcome disadvantages of a moving tether was mostly avoided. Tethered UAVs mostly hovered in place or flies in wide open space. Potential interference of tether with environment has never been investigated on UAVs. No motion planning and execution algorithms for autonomous tethered flight in indoor unstructured or confined environments exist in the current literature. For this dissertation, tether is required for power extension and safety retrieval purposes. At the same time, the UAV needs to navigate through unstructured or confined spaces with that tether. How to plan and execute the motion with a tether in complex environments is the key to the implementation of tethered aerial visual assistance. This research develops a whole low level UAV motion suite revolving around the tether as a key component. From tether-based indoor localization, tether-based motion primitives, tether planning techniques along with motion executor, to tether-based visual servoing, the proposed low level motion suite takes advantage of the tether, mitigates the disadvantages or inconveniences caused by tether in cluttered environments, and therefore opens up a new regime of indoor aerial locomotion: tethered flight. It enables flying tethered UAV in non-free, obstacle-occupied spaces as if it were tetherless.

\chapter{APPROACH: RISK REASONING FRAMEWORK}
\label{chapter::risk_representation}
Autonomous robot operation inherently entails taking risk from a variety of risk sources, but a majority of risk comes from robot locomotion, which is further originated from both the robot's internal components and external interactions with the physical world. Therefore, minimizing motion risk is essential to enable safe and trustworthy robot mission, especially in unstructured or confined environments. This requires a proper framework to understand and reason about robot motion risk. 

To the author's best knowledge, despite the extensive literature on risk-aware motion planning, a formal definition of risk does not exist. Safety or risk concerns were addressed in an ad hoc fashion, depending only on the specific application of interest. This work proposes a formal definition of robot motion risk using propositional logic and probability theory. The proposed framework unifies most existing robot motion risk sources into one single metric, called \emph{risk index} (or \emph{risk}, \emph{risk value}), to explicitly represent risk levels of different motion plans (or path). This definition and representation provide an intuitive approach to compare risk of different paths to improve safe operations. As a new tool to quantify safety for robust autonomy, the risk index could be used as a new cost function for risk-aware motion planners to maximize the likelihood of motion execution success. 

Therefore, the first contribution of this dissertation is a formal definition of robot motion risk and an explicit risk representation approach. Using propositional logic and probability theory, this work reveals the dependencies of risk a robot faces at a certain point on a path on the history leading to this point. It also articulates how risks at individual steps are combined into risk of executing the entire path. It captures the combined effect from different risk sources during robot locomotion as well. This approach is formal and therefore general, comprehensive, and objective. The resulted risk index provides a formal approach to reason about motion risk with a focus on safety-oriented scenarios and gives an explicit and intuitive comparison between different motion plans, i.e. paths. It could be used for reasoning by both human and robotic agents.\footnote{A preliminary version of the risk reasoning framework was discussed and published in previous work \cite{xiao2019explicit2}.}

\section{A Motivating Example: Mine Disaster Borehole Entry}
\begin{figure}[]
\centering
	\includegraphics[width = 0.6 \columnwidth]{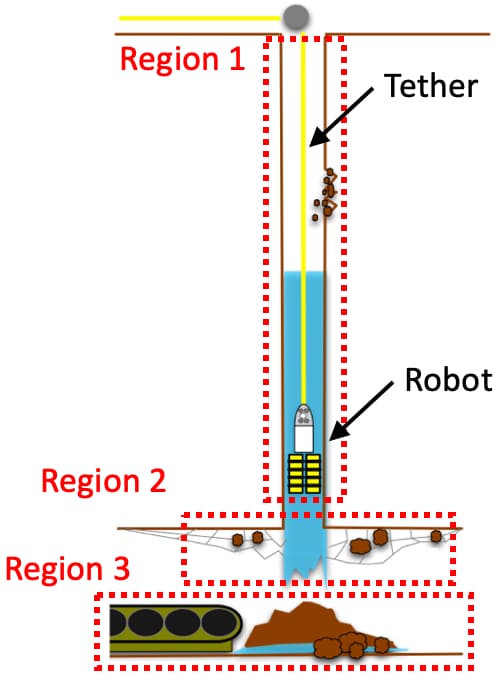}
	\caption{Mine Disaster Borehole Entry (Adapted from \cite{murphy2009mobile})}
	\label{fig::BE_3regions}
\end{figure}

When locomoting in unstructured or confined environments, robot faces risk from multiple aspects. Fig. \ref{fig::BE_3regions} shows an example of borehole entry from Crandall Canyon Mine (Utah) response in 2007 \cite{murphy2009mobile}. 

During mine disasters, robots are widely used to enter the mine, which is inaccessible or too dangers for humans to enter. Borehole entry, in contrast to surface entry and void entry, utilizes the small boreholes which are drilled into the mine into what is expected to be the affected area and the robot uses those as entry point. The idea in the borehole scenario is to insert a small robot into the boreholes, drop the robot to the floor, and explore the affected area. The advantages are that the robot that starts in the neighborhood of the presumed incident would not have to open doors and can conserve onboard power by starting close to the point of interest \cite{murphy2009mobile}. However, extra risks due to borehole entry are induced, in comparison to the risks caused by other methods of entry. 

In region 1 of Fig. \ref{fig::BE_3regions} (the borehole area), multiple risk sources exist at the same time: due the small clearance of the borehole, it is very likely that the robot will get jammed. This will not only cause losing the robot but also preventing any further use of the borehole. Due to the lack of casing of the borehole, falling rocks may damage the robot. Drilling foam, water, and debris may cause system malfunction as well. The vertically hanging robot might spin and therefore lose controllability and mobility. 

In region 2, the transition from the borehole to the mine, mesh roof exists as the existent structure of the mine. The robot faces risk due to the mesh roof interfering with hole exit and reentry. Because the robot is tethered, risk of robot tether getting tangled with the mesh roof is significant. Furthermore, the transition from vertical mobility to operating on mine floor also requires extra effort and induces risk. 

Region 3 is the inside of the mine, where extra risk sources appear after the disaster. Terrain may be unstable due to running water and mud, causing the robot getting trapped and stuck. The robot also has to traverse soft drill tailings and foam, or even equipment, before reaching the mine floor. Any of those can pose risk to the robot. Lastly, while locomoting in region 3, robot tether is still being extended or retracted, interacting with the borehole (region 1) and the mesh roof in the transition into the mine (region 2). Risk of tether entanglement still exists. 

Due to the variety of existing risk sources in borehole entry, the robot failed at Crandall Canyon Mine (Utah) in all four runs during the response in 2007. The reasons for the failure are: 

\begin{enumerate}
\item The lowering system failed.
\item The robot encountered a blockage in the borehole.
\item The robot had to be removed to clean the lens from the buildup of water, debris, and drilling foam.
\item The robot tether was entangled with the mesh and the robot was trapped on the way back. After it was freed, the robot was lost when the tether finally broke due to the actively eroding borehole's severe washout and large boulders. 
\end{enumerate}

How the variety of risk sources in the unstructured or confined environment contribute to the high failure rate (100\%) of the deployment of a sophisticated robot system designed and engineered for those purposes reveals the motivating question of this chapter: how can we formally define the risk the robot faces in unstructured or confined environments and represent the risk so that we can reason about it? 

\section{Formal Definition and Explicit Representation}
Risk is one embodiment of uncertainty. This work only considers motion risk for mobile robots executing a preplanned path. Risk in terms of a sequence of motion (path) is formally defined as \emph{the probability of the robot not being able to finish the path}. 

Before reasoning about risk of executing a path, the workspace of the robot is firstly defined based on tessellation of the Cartesian space, either in 2D or 3D, depending on where the robot resides. Each tessellation is either a viable (e.g. free) or unviable (e.g. occupied) state for the robot to locomote. A feasible path plan $P$ is defined to be an ordered sequence of viable tessellations, called \emph{states} and denoted as $s_i$:

\begin{center}
$P = \{s_0, s_1, ..., s_n\},~\lVert s_i - s_{i-1}\rVert_2 \leq r_c, \forall 1 \leq i \leq n$ 
\end{center}

where $r_c$ is the maximum distance between two consecutive states for the path to be feasible. 

A state on the path is \emph{finished} by the robot reaching the waypoint within an acceptable tolerance and ready to move on to the next waypoint. A state is \emph{not finished} due to two main reasons: the robot crashes or gets stuck. In order to finish the path of $n$ states, the robot faces $r$ different risk elements. which will possibly cause not finishing the path (crash or getting stuck). 

Three types of events are defined: 

\begin{itemize}
\item $F$ -- the event where the robot finishes path $P$
\item $F_i$ -- the event where the robot finishes state $i$
\item $F_i^k$ -- the event where risk $k$ does not cause a failure at state $i$
\end{itemize}

The reasoning about motion risk is based on three assumptions, which are expressed by propositional logic: 
\begin{enumerate}
	\item Path is finished only when all states are finished: 
	\begin{center}
	$F = F_n \cap F_{n-1} \cap ... \cap F_1 \cap F_0$
	\end{center}
	
	\item A state is finished only when all risk elements do not cause failure: 
	\begin{center}
	$F_i = F_i^1 \cap F_i^2 \cap ... \cap F_i^{r-1} \cap F_i^r$
	\end{center}
	
	\item Finish or fail a state because of one risk element is conditionally independent of finish or fail that state because of any other risk element, given the history leading to the state:  
	\begin{center}
	$(F_i^1 \vert \bigcap \limits_{j=0}^{i-1} F_j) \perp \!\!\! \perp (F_i^2 \vert \bigcap \limits_{j=0}^{i-1} F_j) \perp \!\!\! \perp ... \perp \!\!\! \perp (F_i^{r-1} \vert \bigcap \limits_{j=0}^{i-1} F_j) \perp \!\!\! \perp (F_i^r \vert \bigcap \limits_{j=0}^{i-1} F_j)$
	\end{center}
\end{enumerate}

As complement of the formal risk definition proposed by this work, the probability of the robot being able to finish the path could be written as $P(F)$. Based on assumption 1, the event $F$ is logically equivalent to $F_n \cap F_{n-1} \cap ... \cap F_1 \cap F_0$ by propositional logic: 

\begin{equation}
P(F) = P(F_n\cap F_{n-1} \cap ... \cap F_0)
\end{equation}

Using probability chain rule: 

\begin{equation}
\begin{split}
P(F_n\cap F_{n-1} \cap ... \cap F_0)
&= P(F_n \vert F_{n-1} \cap ... \cap F_0) \cdot ... \cdot P(F_1 \vert F_0) \cdot P(F_0)\\
&= \prod_{i=0}^{n} P(F_i \vert \bigcap_{j=0}^{i-1} F_j)
\end{split}
\end{equation}

Based on assumption 2, the event $F_i$ is logically equivalent to $F_i^1 \cap F_i^2 \cap ... \cap F_i^{r-1} \cap F_i^r$: 

\begin{equation}
P(F_i \vert \bigcap_{j=0}^{i-1} F_j) = P(F_i^1 \cap F_i^2 \cap ... \cap F_i^r \vert \bigcap_{j=0}^{i-1} F_j)
\end{equation}

The conditional independence in assumption 3 allows to separate the joint probability into product of individual probabilities: 

\begin{equation}
\begin{split}
P(F_i^1 \cap F_i^2 \cap ... \cap F_i^r \vert \bigcap_{j=0}^{i-1} F_j)
& = P(F_i^1 \vert \bigcap_{j=0}^{i-1} F_j) \cdot P(F_i^2 \vert \bigcap_{j=0}^{i-1} F_j) \cdot ... \cdot P(F_i^r \vert \bigcap_{j=0}^{i-1} F_j) \\
& = \prod_{k=1}^{r} P(F_i^k\vert \bigcap_{j=0}^{i-1} F_j)
\end{split}
\end{equation}

Putting everything together will yield: 

\begin{equation}
\begin{split}
P(F) 
&= P(F_n\cap F_{n-1} \cap ... \cap F_0) \\
&= P(F_n \vert F_{n-1} \cap ... \cap F_0) \cdot ... \cdot P(F_1 \vert F_0) \cdot P(F_0)\\
&= \prod_{i=0}^{n} P(F_i \vert \bigcap_{j=0}^{i-1} F_j) \\
& = \prod_{i=0}^{n} P(F_i^1 \cap F_i^2 \cap ... \cap F_i^r \vert \bigcap_{j=0}^{i-1} F_j) \\
& = \prod_{i=0}^{n} P(F_i^1 \vert \bigcap_{j=0}^{i-1} F_j) \cdot P(F_i^2 \vert \bigcap_{j=0}^{i-1} F_j) \cdot ... \cdot P(F_i^r \vert \bigcap_{j=0}^{i-1} F_j) \\
& = \prod_{i=0}^{n} \prod_{k=1}^{r} P(F_i^k\vert \bigcap_{j=0}^{i-1} F_j)
\end{split}
\end{equation}

Therefore, the formal risk definition, the probability of \emph{not} being able to finish the path, is the probabilistic complement: 

\begin{equation}
\begin{split}
P(\bar{F}) 
&= 1 - P(F) \\
&= 1 - \prod_{i=0}^{n} \prod_{k=1}^{r} P(F_i^k \vert \bigcap_{j=0}^{i-1} F_j) \\
&= 1 - \prod_{i=0}^{n} \prod_{k=1}^{r} (1-P(\bar{F_i^k} \vert \bigcap_{j=0}^{i-1} F_j))
\end{split}
\label{eqn::pfbar}
\end{equation}

In terms of risk representation, the risk of path $P$ is denoted as $risk(P)$ and is equal to $P(\bar{F})$. $P(\bar{F_i^k} \vert \bigcap\limits_{j=0}^{i-1} F_j)$ means the probability of risk $k$ causes a failure at state $i$, given the history of finishing $s_0$ to $s_{i-1}$. It is therefore denoted as the $k$th risk robot faces at state $i$ given that $s_0$ to $s_{i-1}$ were finished: $r_k(\{s_0, s_1, ..., s_i\})$. 

Writing in risk representation form will yield: 

\begin{equation}
risk(P) = 1 - \prod_{i=0}^{n} \prod_{k=1}^{r} (1-r_k(\{s_0, s_1, ..., s_i\}))
\label{eqn::risk_representation}
\end{equation}

This is the proposed probabilistic motion risk indexing to quantify the risk of executing the path. In contrast to the traditional additive state-dependent risk representation, the proposed approach gives a probability value in $[0, 1]$ instead of $[0, \infty]$. It does not require the ill-supported additivity assumption for risk. More importantly, the conditional probability in Eqn. \ref{eqn::pfbar} clearly shows the dependency of risk at certain state on the history, not only the state itself. So the risk the robot is facing at a certain point is not only state-dependent, but also depends on the history leading to this state. For example, if the robot takes a very muddy path to come to a muddy state, the probability of getting stuck at this state is high, due to the mud built up on the wheels or tracks in the history. However, if a clean path was taken, risk at this very same state may be minimum, since clean wheels or tracks can easily maintain sufficient traction. 

\begin{figure}[]
\centering
	\includegraphics[width = 0.9 \columnwidth]{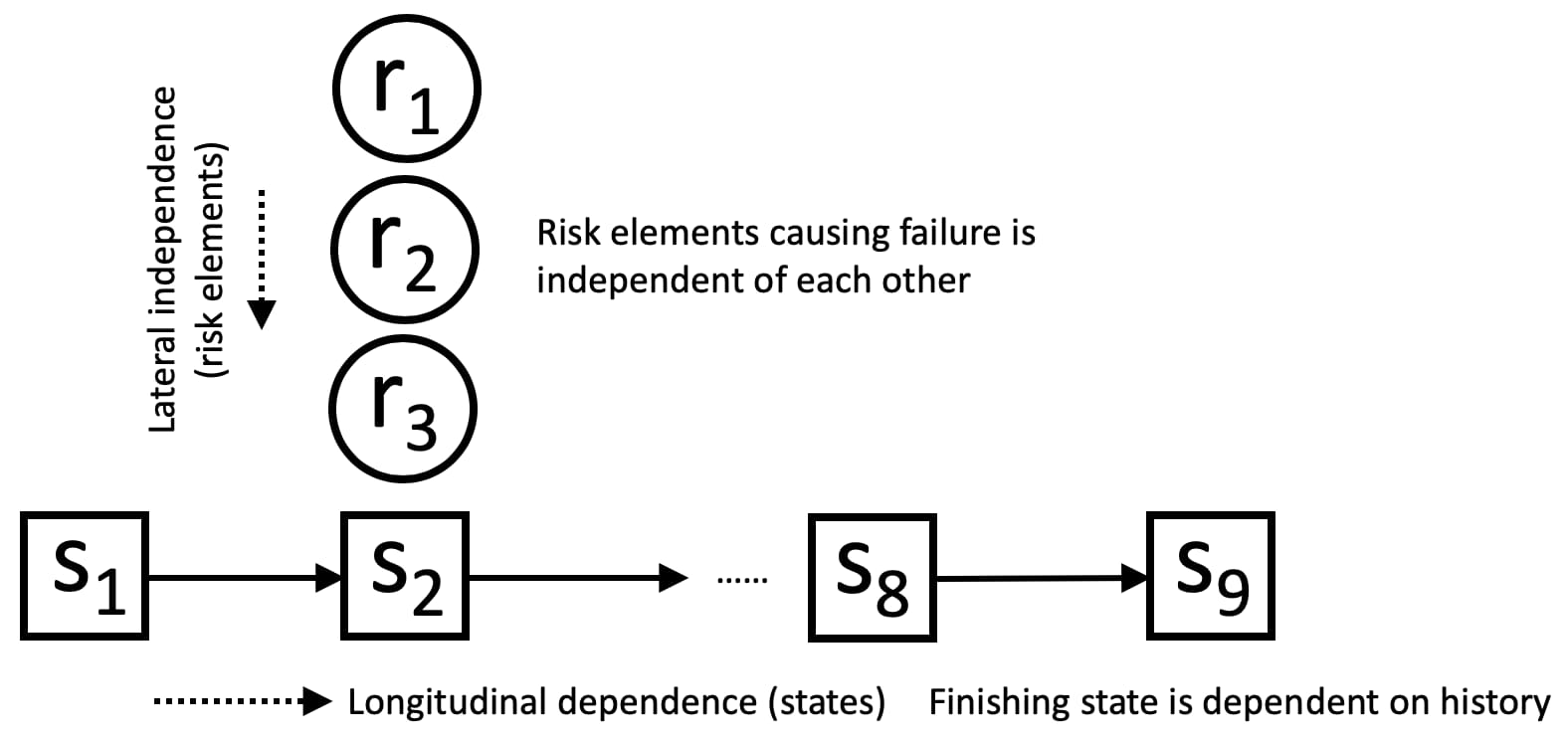}
	\caption{Longitudinal Dependence on History States and Lateral Independence among Risk Elements}
	\label{fig::dependencies}
\end{figure}

Despite the dependencies in the temporal domain, conditional independence among different risk elements at a certain state given the history is still assumed. For instance, if the robot will crash to the closest obstacle is independent of if the robot will tip over due to a sharp turn. This independence assumption matches with the intuition when multiple unrelated risk sources are affecting the robot at the same time. As shown in Fig. \ref{fig::dependencies}, along the direction of the path, risk the robot faces at each individual state is dependent on history (longitudinal dependence), while at each state, the risks caused by different risk elements are independent (lateral independence). 

\section{Risk Elements}
\begin{figure}[]
\centering
	\includegraphics[width = 1 \columnwidth]{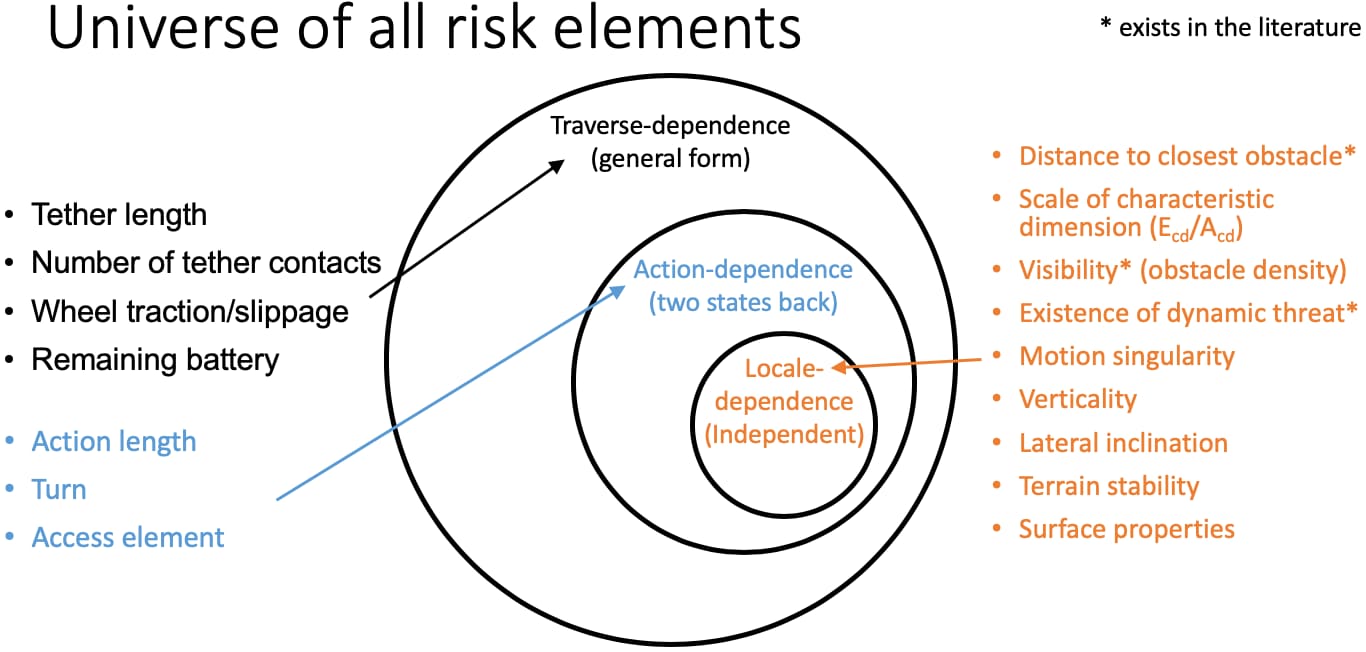}
	\caption{Universe of All Risk Elements}
	\label{fig::universe}
\end{figure}

The formal definition and explicit representation reveal the longitudinal dependence of risk at a certain state on the history. Mathematically speaking, the dependency is on the entire history in general. However, in practice, the dependency of different risk elements may have different depth into the history, e.g. crash to a very close obstacle is only dependent on the closeness of this state to obstacle or crash due to an aggressive turn is only dependent on two states back in the history. In this work, risk elements are divided into three categories: \emph{locale}-dependent, \emph{action}-dependent, and \emph{traverse}-dependent risk elements. Fig. \ref{fig::universe} shows the universe of all risk elements considered in this dissertation, and the categories they belong to. More importantly, the subset/superset relationship between the three categories are displayed: locale-dependence $\subset$ action-dependence $\subset$ traverse-dependence. This section will explain each categories and their own risk elements. 

\subsection{Locale-dependent Risk Elements}

Locale-dependent risk is the most special case in history dependence, since its dependency on history could be entirely relaxed. That is: 

\begin{equation}
P(\bar{F_i^k} \vert \bigcap_{j=0}^{i-1} F_j) = P(\bar{F_i^k})
\end{equation}

The word \emph{locale} connotes the meaning of ``location'', ``position'', or where the robot is currently at. It has similar connotation as the concept of ``state'' in (Cartesian) configuration space, but also emphasizes the relationship with the current proximity of the environment. It is worth to note that the conventional additive risk representation is still referred to as ``state-dependent'', while the proposed formal representation has one of its risk categories named as ``locale-dependent''. 

This category of risk elements has been covered in existing literature under the name of ``location'' or ``state'' and was assumed to be the only type of risk elements. This type of traditional risk elements could be evaluated on the state alone, not depending on history. In the scope of this dissertation, locale-dependent risk elements include distance to closest obstacle, scale of characteristic dimension ($E_{cd}/A_{cd}$) \cite{murphy2014disaster}, visibility (obstacle density), existence of dynamic threat, motion singularity, verticality, lateral inclination, terrain stability, and surface properties.

\subsubsection{Distance to Closest Obstacle}
Distance to closest obstacle is the most straightforward risk element when locomoting in unstructured or confined environments. It has been used extensively in the literature. The closer the robot is to obstacles, the more probability it will crash to the close obstacle, due to the uncertainties in the vehicle itself and environment disturbances. 

However, absolute distance to obstacle alone is not sufficient to determine the risk level of a certain state. In terms of the tethered UAV in this work, flight tolerance of the particular UAV should also be considered. For example, 0.3m from closet obstacle may not be risky for a UAV with 0.1m tolerance, but very risky for one with 0.4m tolerance. 

\subsubsection{Scale of Characteristic Dimension ($E_{cd}/A_{cd}$)}
\begin{figure}[]
\centering
	\includegraphics[width = 0.8 \columnwidth]{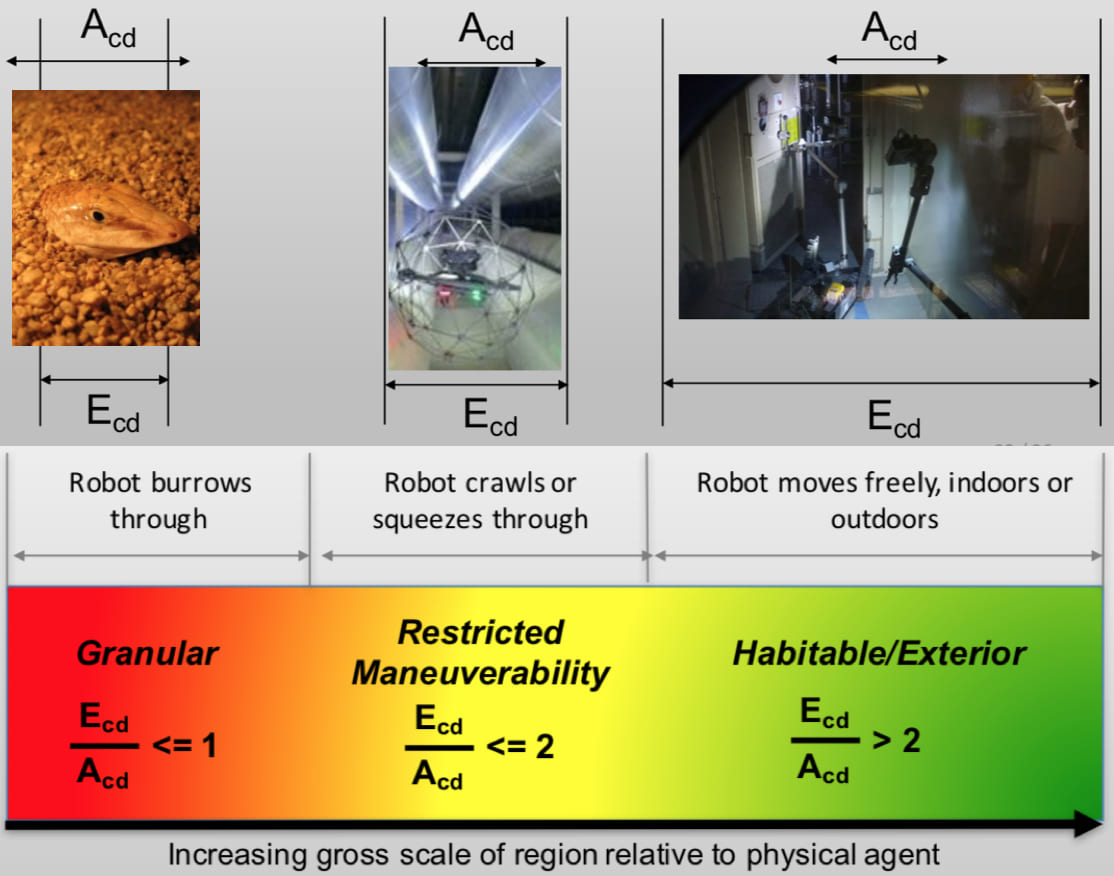}
	\caption{Three Different Scales of Characteristic Dimension}
	\label{fig::scale}
\end{figure}

The scale of characteristic dimension is a definition of small, confined environments used in Disaster Robotics \cite{murphy2014disaster}. It describes general ground robot work envelopes in rubble as a dimensionless number that normalizes the characteristic dimension in entering a void, usually the cross sectional diameter, of the robot agent $A_{cd}$ to the cross section, or characteristic dimension, of work envelope $E_{cd}$. A granular space (Fig. \ref{fig::scale} left) is defined as $E_{cd}/A_{cd} \leq 1$, where the robot must burrow into the work envelope. A restricted maneuverability space (Fig. \ref{fig::scale} middle) is defined as $E_{cd}/A_{cd} \leq 2$, in effect the narrowest cross section of the work envelope is less than twice the cross section diameter of the robot. A tracked ground vehicle would be unlikely to easily turn around in such a relatively narrow space. A habitable/exterior is the least risky, with $E_{cd}/A_{cd} > 2$ (Fig. \ref{fig::scale} right). This characterization of robot's work envelope is independent of the morphology or gait of a robot and therefore comparison of difficulties of different environments to different robots becomes possible \cite{xiao2018review}. Based on the scale of characteristic dimension of a certain state, the probability of the robot getting stuck in this state could be computed. 

\subsubsection{Visibility}
\begin{figure}[]
\centering
	\includegraphics[scale=0.4]{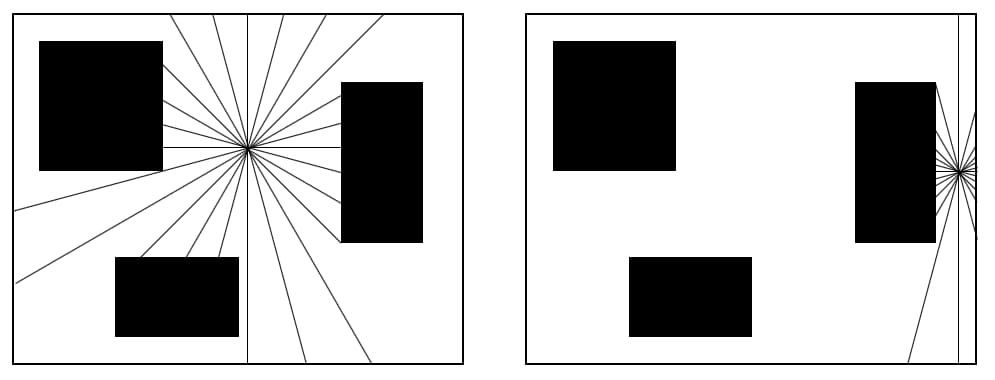}
	\caption{Visibility based on Isovists Lines: left figure represents a state with good visibility while right one with worse. The confinement due to high obstacle density for the state on the right causes more risk. }
	\label{fig::visibility}
\end{figure}

Visibility represents the confinement of a certain state relative to nearby obstacles around it, and is also known as obstacle density. The visibility model casts from each state a set of line of sight, called isovists lines \cite{benedikt1979take}. The isovists lines are defined as the geometry obtained by casting light rays in all directions from a state in the state space. The isovists lines are obtained when the rays are intersected with obstacles or state space boundaries (Fig. \ref{fig::visibility}). The length of each isovists line is summed up and divided by the number of rays to derive the visibility value of a particular state:

\begin{equation}
\label{eqn::visibility}
V = \frac{\Sigma_{i=1}^{n}L_i}{n}
\end{equation}

where $n$ is the number of rays and $L_i$ the length of each isovists line. The greater the visibility value, the less risk the robot is facing at this state. 

\begin{figure}[]
\centering
\subfloat[Risk due to Distance]{\includegraphics[width=0.5\columnwidth]{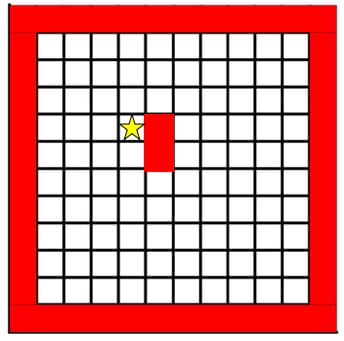}%
\label{fig::distance_compare}}
\hfil
\subfloat[Risk due to Scale]{\includegraphics[width=0.5\columnwidth]{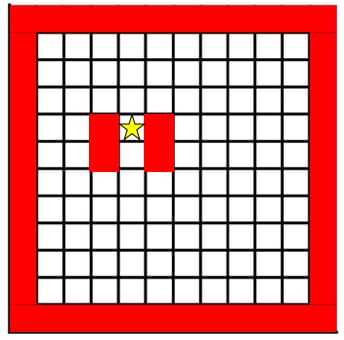}%
\label{fig::scale_compare}}
\subfloat[Risk due to Visibility]{\includegraphics[width=0.5\columnwidth]{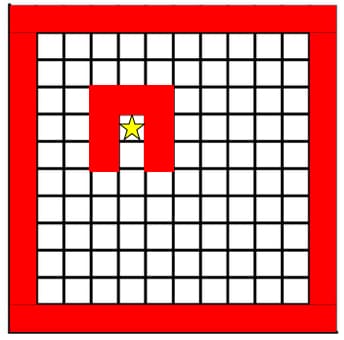}%
\label{fig::visibility_compare}}
\caption{Distance to closest obstacle, scale of characteristic dimension, and visibility are all necessary to capture the risk caused by surrounding obstacles.}
\label{fig::dis_scale_vis}
\end{figure}

The three aforementioned locale-dependent risk elements are all related with obstacles around the robot during locomotion. However, each of them is responsible for different aspects of risk, i.e. crash to obstacle in this case, and is therefore not redundant for a complete risk representation. The yellow star in Fig. \ref{fig::distance_compare} is risky due to the close distance to obstacle. However, the distance to closest obstacle remains the same in Fig. \ref{fig::scale_compare}. The extra risk is caused by the small scale of characteristic dimension. Distance and scale have the same effect in Fig. \ref{fig::visibility_compare}, so visibility is needed to capture the extra risk. This intuitive example shows that visibility is indispensable to comprehensively capture the risk caused by surrounding obstacles. 

\begin{figure}[]
\centering
\subfloat[Same Scale and Visibility, Different Distacne]{\includegraphics[width=0.5\columnwidth]{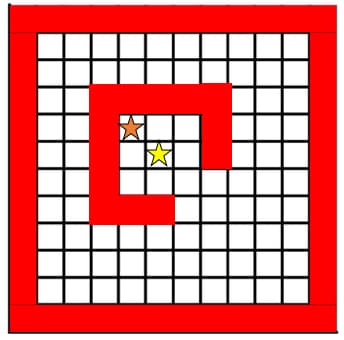}%
\label{fig::distance_compare2}}
\hfil
\subfloat[Same Distance and Visbility, Different Scale]{\includegraphics[width=0.5\columnwidth]{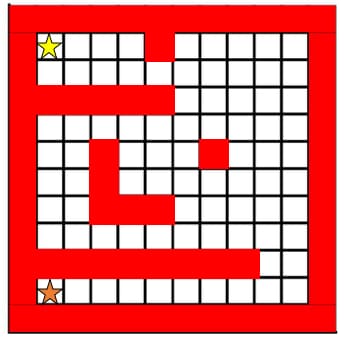}%
\label{fig::scale_compare2}}
\caption{Orange star has higher risk in both scenarios: (a) needs distance to closest obstacle to distinguish same scale of characteristic dimension and visibility, and (b) needs scale of characteristic dimension to distinguish same distance to closest obstacle and visibiity.}
\label{fig::dis_scale_vis2}
\end{figure}

Another two examples are shown in Fig. \ref{fig::dis_scale_vis2}: distance to closest obstacles is needed to render orange star in Fig. \ref{fig::distance_compare2} more risk, since both orange and yellow stars have same scale of characteristic dimension and visibility. In Fig. \ref{fig::scale_compare2}, same distance to closest obstacle and visibility require the inclusion of scale of characteristic dimension in the risk representation to reflect the fact that orange star is riskier than yellow star. 

Fig. \ref{fig::dis_scale_vis} and Fig. \ref{fig::dis_scale_vis2} both prove the necessity to include distance to closest obstacle, scale of characteristic dimension, and visibility at the same time to capture all the aspects of risk caused by surrounding obstacles. 

\subsubsection{Existence of Dynamic Threat}
Existence of dynamic threat captures the potential or frequency a dynamic threat may exist at a certain state. A priori knowledge of the workspace and data-driven approaches can provide necessary information regarding the risk associated with a particular state. For example, a state on the hallway may have more frequent human presence and is therefore riskier than a confined corner. \cite{pereira2011toward, pereira2013risk, krumm2017risk} are examples of using data-driven approaches to capture existence of dynamic threat, using history AIS and traffic data. The higher potential or frequency a dynamic threat may exist in a state, the higher probability the robot may not be able to finish this state, and thus a higher risk value. 

\subsubsection{Motion Singularity}
Motion singularity exists for some robots and for them it presents itself in different forms. Motion singularity exists for articulated manipulator arms moving through tight spaces, when multiple links are aligned. For the tethered visual assistant UAV addressed in this dissertation, motion singularity exists right above the tether reel/UAV ground station. Therefore motion singularity is a locale-dependent risk. Due to the unique tether angle based sensing and controls, Tether azimuth angle is ambiguous and will become unstable when elevation angle is close to 90\degree~(above tether reel). The sensor inaccuracies and flight instability will cause azimuth value to change abruptly between -180\degree~and 180\degree. This problem will become more serious when velocity control is used, which is based on derivative and Jacobian. Therefore, states close to 90\degree~elevation have a higher probability of getting stuck and thus should be deemed risky for tethered UAV. 

\subsubsection{Verticality}
Verticality of a state is risky for those robots, for whom changing elevation is not trivial. While verticality may not be a risk for rotorcraft due to its Vertical Take-Off and Landing (VTOL) capability, it is risky for fixed-wing aircraft, especially in tight spaces. Apparently verticality plays a more vital role for ground robots, who require extra effort to overcome gravity and ascend in its workspace. Therefore if a state has high verticality, relevant robots may have higher probability of being unable to ascend and getting stuck. 

\subsubsection{Lateral Inclination}
Lateral inclination is a property of a state, which can directly cause a ground robot to tip over. It is different than verticality because verticality concerns more about longitudinal movement, such as to steadily ascend on the terrain, while lateral inclination focuses on lateral stability. Under the impact of lateral inclination, the robot should still be able to keep upright and maintain its contact with the terrain along with its traction and mobility. A higher inclination is associated with a higher probability of tipping over, and is therefore riskier. 

\subsubsection{Terrain Stability} 
Terrain stability affects ground robots interacting with the terrain. It is a property of a state on a path on the ground. Unstable terrain may cause the robot get stuck and lead to immobility. The robot then is not able to finish the path. The more stable the terrain is, the less risk a ground robot faces. 

\subsubsection{Surface Properties}
Surface properties only impact those robots who make direct contact with environmental surfaces, such as snake robot squeezing through confined pipelines. Although the representation of surface properties and their effect on robot locomotion have not been extensively studied yet, it is obvious that certain negative surface properties may have higher probability of causing higher friction during the interaction and then getting stuck, or deteriorating the sensing capability, such as refraction or absorption of laser beam, etc. Although proper representation is still unclear, surface properties is included here for completeness. 

\subsection{Action-dependent Risk Elements}
Action-dependent risk is a special case of risk's history dependency, between the general traverse-dependence and the most special locale-dependence. The depth of action-dependent risk elements' history dependency is two states back, such that the finishing of the last two states have impact on the risk the robot is facing at the current state: 

\begin{equation}
P(\bar{F_i^k} \vert \bigcap_{j=0}^{i-1} F_j) = P(\bar{F_i^k} \vert F_{i-2} \cap F_{i-1})
\end{equation}

This category of risk elements usually focuses on the transitions between states, including the effort necessary to initiate the transition and the difference between two consecutive transitions. In the universe of all risk elements covered by this dissertation (Fig. \ref{fig::universe}), action-dependent risk elements include action length, turn, and access element. 

\subsubsection{Action Length}
For the action-dependent risk elements, the ``actions'' do not refer to the actual physical actions the robot takes to go from state to state, but an abstraction of the transition between states: 

\begin{equation}
a_i = s_i - s_{i-1}, i = 1, 2, ..., n
\end{equation} 
 
In a 3D Cartesian space, an action is a 3D vector pointing from the last state to the current state, showing the transition needed to initiate the locomotion. Action length is the 2-norm of this vector and denotes the necessary effort to realize the transition: 

\begin{equation}
\Vert a_i \Vert= \Vert s_i - s_{i-1}\Vert, i = 1, 2, ..., n
\end{equation} 

Apparently the greater the effort to transit between states, the higher probability something will go wrong. An example is that a long path should be riskier than a short path, such as statistical models like MTTF (Mean Time To Failure) capture the possible failure of onboard mechanical and electrical components. In a workspace which takes form as a 2D grid, moving to the diagonally neighboring states should be at least $\sqrt{2}$-times riskier than moving to the directly adjacent neighbors. The inclusion of $s_{i-1}$ demonstrates the one-step dependency into the history, which cannot be directly captured by conventional locale-dependent risk elements. 

\subsubsection{Turn}
\begin{figure}[]
\centering
	\includegraphics[width = 1 \columnwidth]{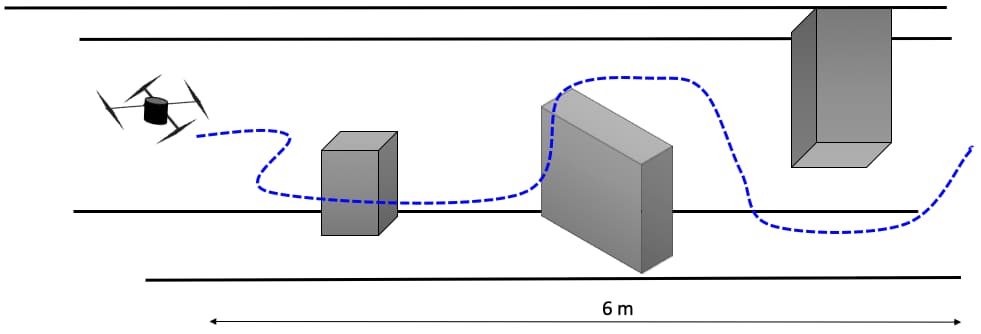}
	\caption{Tortuosity is the number of turns per unit distance, in both horizontal and vertical plane. As shown in the figure, three turns are taken in order to navigate through the 6m course, so the tortuosity value is 3/6=0.5 (adapted from \cite{agarwal2014characteristics}). }
	\label{fig::tortuosity}
\end{figure}

Risk associated with turning comes from the ecological idea of tortuosity (Fig. \ref{fig::tortuosity}): a metric calculated as the number of turns taken by the robot per unit distance \cite{agarwal2014characteristics, murphy2014disaster}. 

This dissertation forgoes the normalization of unit distance and adds quantification on severity of a turn. The risk associated with turning has positive correlation with the severity of a turn: maneuvering aggressively has higher probability of tipping over, motor overheat, loss of localization, etc. Mathematically, the difference between two consecutive actions captures the direction and magnitude of the turn and its 2-norm can give the severity of this turn. Given a feasible path, two steps back into the history is required to compute a turn: 

\begin{equation}
\Vert a_i-a_{i-1} \Vert = \Vert (s_i-s_{i-1})-(s_{i-1}-s_{i-2} \Vert 
\end{equation}

This risk element could not be computed based on a single locale but two states back into the history ($s_i$, $s_{i-1}$, and $s_{i-2}$). The more severe turns the UAV needs to make when navigating along a path, the riskier the motion would be. The risk aware planner should prefer a straight path over a tortuous one in order to minimize the risk caused by turning, both horizontally and vertically. 

\subsubsection{Access Element}
Access element captures the motion between regions within a void that have different sizes, shapes, surface properties, and other environmental conditions in unstructured or confined environments \cite{murphy2014disaster}. Other than the challenges posed by the two different regions themselves, the transition between them usually induces extra risk. For example, moving from a wider pipeline into a narrower one has more risk than the other direction. Entering into a darker area poses risk since proper sensor parameter adjustment needs to be made in a timely and accurate manner during the transition. Transitioning from a straight and open hallway to a stair case at the corner is riskier than the opposite transition. Going from one room to another room through a wider door is less risky than through another narrower door. Proper mathematical quantification to match with the semantic description is necessary, but it is obvious that access element can cause risk during the transition. In general, the risk could be associated with the difference between two states, between which the transition takes place:

\begin{equation}
f(characteristic(s_i) - characteristic(s_{i-1}))
\end{equation}

where $characteristic(\cdot)$ represents the relevant feature of a state regarding the transition, while $f$ maps the difference of the relevant features of two states into some risk level, e.g. adjusting camera ISO properly between regions of different illumination causes risk, and going from less confined to more confined region entails great risk while going the opposite direction does not have risk, etc. By looking at the current locale where $s_i$ is at alone is not enough to capture the risk caused by access element. The inclusion of the history state $s_{i-1}$ shows that access element is action-dependent. 

\subsection{Traverse-dependent Risk Elements}
Traverse-dependent risk is the general form of risk's history dependency, which encompasses both locale-dependent and action-dependent risk elements. The general form has a full depth of history dependency and looks back to the whole traverse from start leading to the current state. Finishing of all the history states has impact on the finishing of the current state: 
 
\begin{equation}
P(\bar{F_i^k} \vert \bigcap_{j=0}^{i-1} F_j) = P(\bar{F_i^k} \vert F_{i-1} \cap F_{i-2} \cap ... \cap F_1 \cap F_0) 
\end{equation}

All risk elements in the universe of risk elements covered by this dissertation (Fig. \ref{fig::universe}) are members of this category. This subsection only focuses on those risk elements, which do not have locale-dependent and action-dependent properties. In the scope of this dissertation, tether length, number of tether contacts, wheel traction/slippage, and remaining battery are discussed. 

\subsubsection{Tether Length}
\begin{figure}[]
\centering
	\includegraphics[width = 0.75 \columnwidth]{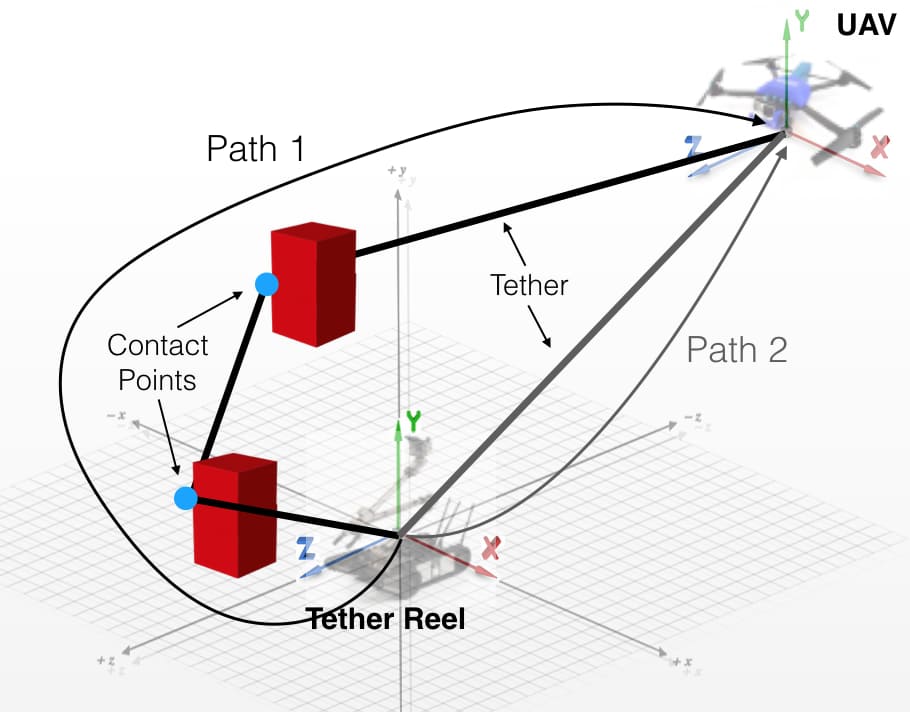}
	\caption{UAV locating at the same state will have different tether length and number of contact points depending on the traverse leading to the state. Blue dots denote tether contact points with red obstacles. Black and grey lines denote two different tether configurations due to different path taken. Path 1 leads to the state with two contact points with the environment and a longer tether, while path 2 does not need any contact point and the tether is much shorter. }
	\label{fig::contact_points}
\end{figure}

While executing tethered motion, connection via a tether between the robot and its base station is required at all time. When moving further away from the base station, the tether prolongs and more of the tether is exposed to the environment instead of being stored and protected in the base station. The tether contact planning and relaxation technique, which will be discussed in detail in Chapter \ref{chapter::low_level}, also shows that the length of tether is not only a function of the robot location alone, but also dependent on the traverse it took to get there, considering the possibility of contact points (Fig. \ref{fig::contact_points}). It also shows that longer tether introduces more uncertainty in localization: the longer the tether is reeled out, the less accuracy the tether-based localization is, due to the increased gravity caused by the tether mass pulling the tether down and away from the straight tether segment assumption. Poor localization are very likely to cause failure to finish path execution. A large portion of the tether being exposed to the unstructured or confined environments also poses more risk through unexpected events. 

Conditioned on the finishing of history states on the entire traverse from start to last state, the length of the tether could be uniquely determined. High probability of not being able to finish the current state is positively correlated with a long tether. Therefore tether length belongs to the general traverse-dependent risk elements. 

\subsubsection{Number of Contact Points}
Locomotion of tethered robot in unstructured or confined environments will inevitably cause tether contact with the environment. When flying a tethered UAV in unstructured or confined environments, allowing tether contact points with the environment will maintain the same reachability space as a tetherless UAV. Chapter \ref{chapter::low_level} will discuss in detail the approach to plan tether contact points. However, tether contact points will also introduce extra risk for multiple reasons. One source of risk is that the material, geometry, or property of the obstacles in the environment may be unclear, so making contact with them could be dangerous. For example, the tether could be cut by sharp edges of the obstacle or some obstacle will leave permanent damage on the tether. Another reason of extra risk is the reduced localization accuracy under the assumption that once formed tether contact points will not move. This assumption makes the localizer and planner practical, but sacrifices the flight precision after the contact point is made. The accuracy will be further deteriorated with increasing number of contact points. Therefore, higher risk will be associated with more contact points. It will be shown in Chapter \ref{chapter::low_level} that the formation of contact points will depend on the traverse, i.e. the same state may include different numbers of contact points based on the traverse leading to the state (Fig. \ref{fig::contact_points}). Therefore, number of contact points falls into the general category of traverse-dependent risk elements. 

\subsubsection{Wheel Traction/Slippage}
\begin{figure}[]
\centering
	\includegraphics[width = 0.75 \columnwidth]{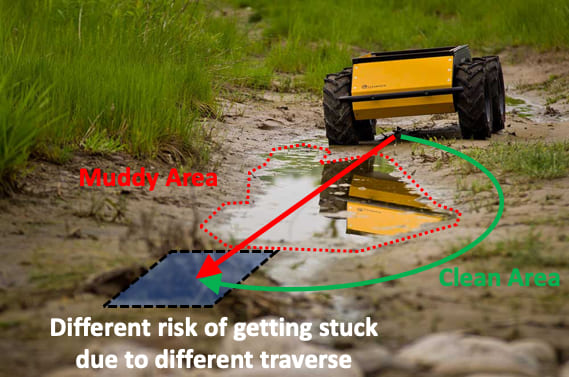}
	\caption{Ground Robot Working in Unstructured Environment: The probability of getting stuck in the blue state is dependent on the traverse the robot took. Taking the red path through muddy area can accumulate mud on the wheels and therefore makes negotiating blue state riskier, while the green path through clean area can maintain enough wheel traction and is therefore safer when coming to blue state. }
	\label{fig::wheel_traction}
\end{figure}

For ground robots locomoting in unstructured or confined environments using either wheels or tracks, maintaining good traction is the key to ensure successful execution of the path. Slippage needs to be by all means avoided, for both not getting stuck and not losing wheel encoder odometry. However, maintaining proper traction and avoiding slippage is not only a matter of the terrain at that state, but also the condition of the wheel: as shown in Fig. \ref{fig::wheel_traction}, in order to finish the tough state occupied with tree branches (blue), if the robot comes through a challenging traverse through a muddy area (red), mud may build up on the wheels and whenever facing challenging terrain again, the probability of slippage is higher. But a detour through the clean area (green) can maintain clear wheels and enough traction to negotiate with and finish the difficult state.

Therefore, wheel traction is a condition of the vehicle dependent on the traverse taken and will affect the probability of finishing or failing the current state. Proper wheel traction/slippage model is necessary to quantify the condition of the wheels, but it is apparent that wheel traction/slippage falls into traverse-dependent risk elements. 

\subsubsection{Remaining Battery}
The last traverse-dependent risk element is remaining battery. As \cite{tiwari2018estimating, tiwari2019unified, tiwari2019orange} showed, battery consumption is crucial in determining if a path could be finished, especially when the path is venturing and exploring unknown environments without the possibility of recharge. The energy model proposed in \cite{tiwari2018estimating, tiwari2019unified, tiwari2019orange} shows that the probability of complete battery depletion depends on various aspects of the traverse the robot took from start. A very intuitive example is that if a longer traverse was taken by the robot to come to a certain state, remaining battery is less and therefore the probability of the robot getting stuck at this state due to battery depletion is higher. Other aspects of the traverse include if the robot ascend or descend during the traverse, the friction coefficient of the terrain on the traverse, etc. This demonstrates that the risk caused by remaining battery is dependent on the entire traverse the robot took. 

\section{Risk Representation}
With the formal definition of risk as the probability of the robot not being able to finish the path, along with three categories of, locale-dependent, action-dependent, and traverse dependent, risk elements, this section explains given a feasible path in an unstructured or confined environment, how risk is represented as a numerical probabilistic value to reason about the risk the robot faces at each individual state and along the entire path. 

Eqn. \ref{eqn::risk_representation} is the basic formulation for risk representation. The risk of executing the entire path $P$ is evaluated based on the contributions each individual risk element (risk element $1$ to $r$) has at each individual state on the path (state $0$ to $n$). Conventional risk representation approaches assumed additivity of risk, i.e. the risk of an entire path is the summation of the risks of individual states. The additivity, however, is not well supported. The result of the conventional risk representation was a risk index in $[0, \infty]$, whose definition and meaning remained unclear. They also only considered locale-dependent risk, ignoring all the dependencies on the finishing of history states, i.e. action-dependent and traverse-dependent risk elements. The proposed approach is grounded on a formal risk definition and uses propositional logic and probability theory to combine the individual effect of risk at state and risk caused by individual risk element into risk of a path. It also considers action-dependent and traverse-dependent risk elements, in addition to locale-dependent risk elements. The output of the proposed risk representation is a risk index exactly as the probability of the robot not being able to finish the path. 

As shown in Eqn. \ref{eqn::risk_representation}, given a state $s_i$, the risk contributed by one risk element $r_k$ is in general dependent on the history states on the traverse $s_0, s_1, ..., s_{i-1}$. The value of this particular risk, as the probability of this risk element $k$ causes failure at this state $i$, could be computed either empirically or theoretically. In the absence of an theoretical approach to compute the probability value, this risk could be calculated based on the extent of the adverse property, e.g. being closer to obstacle, making sharper turn, and having more contact points will have a higher probability of failure at this state. Those probability values could be empirically determined. 

\begin{figure}[]
\centering
	\includegraphics[width = 0.55 \columnwidth]{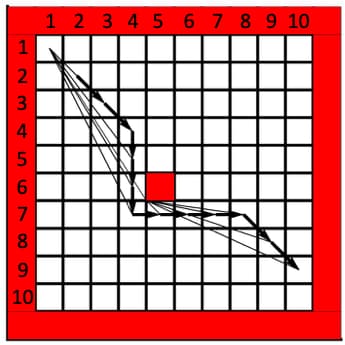}
	\caption{Example Environment and Path for Risk Representation: Red cells indicate obstacles and white cells are free space. The path is composed of an ordered sequence of states, denoted by thick arrows. The thin lines denote the tether configuration at each state, some of which are straight without contact and some have kinks formed as contact points on the obstacle. For convenience, each row and column are numbered. }
	\label{fig::risk_representation_example}
\end{figure}

In order to illustrate risk representation, this section uses the tethered UAV as example, and three representative risk elements are chosen in order to cover all three risk categories and maintain simplicity at the same time. The three example risk elements are distance to closest obstacle as locale-dependent risk element, turn as action-dependent risk element, and number of contact points as traverse-dependent risk element. The workspace is based on the tessellation of 2D Cartesian space, surrounded by obstacles and one extra obstacle in the middle. The workspace and example path to be evaluated is shown in Fig. \ref{fig::risk_representation_example}. For better illustration, other than the index of each state ($0-11$), the subscript also corresponds to the index of rows and columns of the state in the 2D occupancy grid (first and second column in Tab. \ref{tab::risk_representation_table}). 

The path starts from the upper left corner ($s_{22}$) and ends at the lower right corner ($s_{910}$). For each state, all three risk elements are evaluated. Risk caused by distance to closest obstacle is only based on the current locale alone, where the current state locates. The distance value to the closest obstacle is mapped into a risk value empirically, denoting the probability of not being able to finish this state (third column in Tab. \ref{tab::risk_representation_table}). Risk caused by turn is action-dependent, so two states back need to be investigated. Based on the difference of the two consecutive actions, a risk value is empirically assigned (forth column in Tab. \ref{tab::risk_representation_table}). For traverse-dependent number of contact points, by looking back at the entire traverse, the number of contact points could be determined (details of contact planning will be discussed in Chapter \ref{chapter::low_level}). Here we simply assign $0.03$ probability of not finish to states with one contact point and $0$ to those that don't have one (fifth column in Tab. \ref{tab::risk_representation_table}). 

\begin{table}[h!]
\centering
\includegraphics[width = 1 \columnwidth]{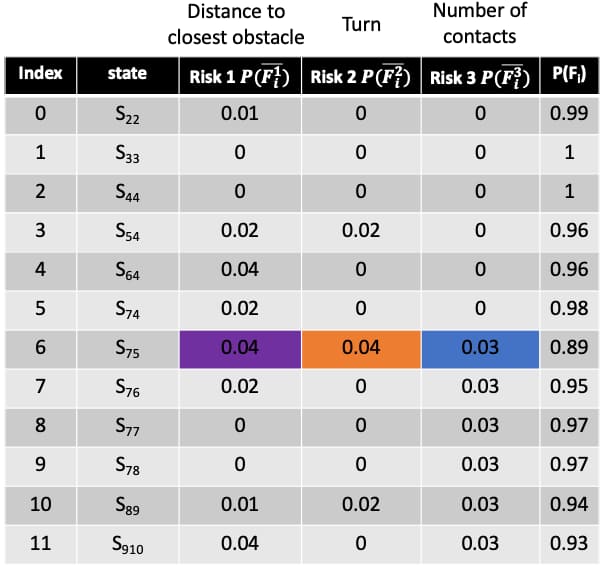}
\caption{Risk Representation for Individual States and Risk Elements}
\label{tab::risk_representation_table}
\end{table}

\begin{figure}[]
\centering
	\includegraphics[width = 0.6 \columnwidth]{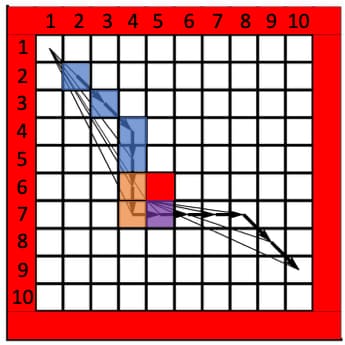}
	\caption{The Sixth State $s_{75}$ as Example to Illustrate Risk Representation: Different color-coded states are used to compute the corresponding color-coded risk values in Tab. \ref{tab::risk_representation_table}. Note that orange blocks and purple block are also blue, and purple block is also orange. }
	\label{fig::risk_representation_example2}
\end{figure}

The sixth state $s_{75}$ on the path is chosen as example to illustrate how the three risk values from three risk elements are computed (color-coded in Fig. \ref{fig::risk_representation_example2} and Tab. \ref{tab::risk_representation_table}). Due to the closeness to the obstacle in the middle, the risk of collision and therefore not being able to finish the state is 0.04. This risk value only needs to be evaluated by the purple block alone, the current state itself. By looking back two states into history ($s_{64}$ and $s_{74}$), the robot moved down first and then makes a sharp 90\degree~turn to move right. Due to the sharpness of the turn, there is 0.04 probability that the robot cannot make the turn and reach $s_{75}$. Note that $s_{75}$ should also be in orange, but due to the overlap with purple the orange is omitted. In terms of contact point, the entire traverse needs to be taken into account (blue blocks), in order to determine how many contact points are formed with this traverse from start. The blue traverse in Fig. \ref{fig::risk_representation_example2} forms one contact point at the lower left corner of the red obstacle in the middle. Therefore the risk due to number of contact points is 0.03 at state $s_{75}$. It is also worth to note that the orange blocks and purple block also have the color blue. If taking another traverse from the right hand side of the red obstacle to come to the same state $s_{75}$, two contact points (upper right and lower right corner of the obstacle) will be formed, instead of one, causing more risk at the same state $s_{75}$. Therefore the entire traverse needs to be considered to determine the risk value associated with number of contact points. 

With all risk values from individual risk elements at individual states computed in Tab. \ref{tab::risk_representation_table}, we can compute the probability of being able to finish each state, shown in the right column ($P(F_i)$). Taking $s_{75}$ as an example again, the probability of finishing $s_{75}$ is the product of the probabilities of all risk elements do not cause failure at this state due to the lateral independence assumption (Eqn. \ref{eqn::risk_representation}): $(1-0.04)\times(1-0.04)\times(1-0.03)=0.89$. In order to finish the path, all the states need to be safely finished. Based on chain rule, the probability of finishing the path is the product of all the entries in the right column: $0.99\times1\times1\times0.96\times0.96\times0.98\times0.89\times95\times0.97\times0.97\times0.94\times0.93=0.62$. Taking the complement will yield the probability of \emph{not} being able to finish the path as $1-0.62=0.38$. This is the risk of the path in Fig. \ref{fig::risk_representation_example}, meaning if the robot executes this path, there is 0.38 probability that the robot is not able to finish the path. 

More examples of the proposed formal risk definition and explicit representation will be shown in Chapter \ref{chapter::high_level}, as results of the high level risk-aware planner. They will also be compared with the conventional additive state-dependent risk representation, along with planning results of conventional risk-aware planners. 

\section{Summary of Risk Reasoning Framework}
Robot motion risk is not well investigated in the literature, especially in terms of the definition, i.e. what is risk, and representation, i.e. how can risk be represented and reasoned about. Without a formal risk definition for general robots in unstructured or confined environments, the ad hoc approaches for risk representation can only address safety and robustness concerns within very specific robots and application scenarios. Based on robot risk in the literature, for most robotic agents locomoting in unstructured or confined environments, risk concerns are in general to safely finish the motion plan, or path, either by remote control or autonomous navigation. To guarantee safety and robustness during locomotion, the possibility of crash or getting stuck needs to be minimized. 

This dissertation proposes a formal definition of robot motion risk as the probability of the robot not being able to finish the path. This formal and general definition is applicable to any robotic agents locomoting in unstructured or confined environments. Therefore risk-aware motion is the motion with maximum probability of being safely finished. An explicit risk representation approach using propositional logic and probability theory is also introduced. The usage of these formal methods reveals that the risk the robot faces at each state along the path is not only a function of that current locale itself, but also dependent on the traverse the robot took from the beginning. By considering not only locale-dependent (distance to closest obstacle, scale of characteristic dimension, visibility, existence of dynamic threat, motion singularity, verticality, lateral inclination, terrain stability, surface properties), but also action-dependent (action length, turn, access element) and traverse-dependent (tether length, number of tether contacts, wheel traction/slippage, remaining battery) risk elements, the proposed risk representation encompasses a comprehensive risk universe in unstructured or confined environments and formally handles their dependencies on the history (longitudinal dependence). The proposed risk universe is the superset to the conventional ad hoc risk functions or treatment of risk as chance constraints within either MDP or RMPC framework. Those approaches only focused on obstacle-related risk sources, such as modeling risk as chance constraints, i.e., the probability of the motion trajectory intersects with geometric obstacles in Cartesian space. Other risk elements outside of the MDP or RMPC's simplified Cartesian state space cannot be properly addressed by those chance constraints, e.g., motor overheat due to aggressive turning, sensor deterioration due to environmental interactions, etc. The inference using probability chain rule in the proposed risk framework also avoids the ill-supported additive assumption of risk along the entire path and other conservative relaxation techniques in the literature. With a simple lateral independence assumption (different risk elements do not affect each others' probability of failure given the history leading to this state) and a formally reasoned longitudinal dependence (finishing states is dependent on history), risk is computed as a single probability value of the robot not being able to finish the path. In comparison, MDP with chance constraints did not encode history states information regarding risk of collision (probability of constraint violation, as intersection between motion trajectory and obstacles) at all, due to its Markovian assumption, while RMPC only had a subset of relevant risk information embedded in Cartesian state space and history implicitly included through system dynamics updates. They either assumed independence on history or relaxed the temporal (between time steps) and spatial (between multiple obstacles) dependencies through conservative approaches such as ellipsoidal relaxation or Boole'e bound. The proposed motion risk framework gives an explicit and intuitive comparison between different motion plans, or paths, for both human and robotic agents. It can be used as a metric to quantify safety for robust robot motion. 
\chapter{APPROACH: HIGH LEVEL RISK-AWARE PATH PLANNER}
\label{chapter::high_level}

Given the formal risk definition and explicit representation approach proposed in Chapter \ref{chapter::risk_representation}, robot motion could be formally and quantitatively reasoned, i.e. given different paths in unstructured or confined environments, their risk could be reasoned, quantified, and compared. This could be used as a metric to quantify safety cost for robust autonomy. To navigate unstructured or confined environments in a risk-aware manner, an autonomous agent is in need of a planning paradigm that can find the least risky path from all its options, equipped with the formal risk definition and explicit representation in Chapter \ref{chapter::risk_representation}.\footnote{The risk-aware planner was preliminarily discussed and published in previous work \cite{xiao2019explicit1}.}

While moving in a risk-aware manner in unstructured or confined environments, the tethered aerial visual assistant also needs to maximize the viewpoint quality along its path so that the primary robot's operator could maintain a good situational awareness in a continuous manner. However, minimizing risk may be at odds with achieving good viewpoint quality rewards. The high level risk-aware planner also needs the capacity to address the tradeoff between risk and reward, i.e. to maximize reward and minimize risk simultaneously. 

This section firstly provides a problem definition and formulates the risk-aware visual assistance problem into a graph-search query. With the problem definition and formulation, finding the most desirable risk-aware visual assistance behavior with optimal reward risk tradeoff becomes searching for an optimal path in terms of utility, the ratio between reward and risk. In order to maintain good viewpoint quality throughout the entire visual assistance process, the reward from good viewpoints are accumulated along the path. While using the risk definition and representation described in Chapter \ref{chapter::risk_representation}, the viewpoint quality reward information is acquired by another separate study, which is not within the scope of this dissertation. 

After that, it is proved that this risk-aware reward-maximizing problem formulated as a graph-search query is well-defined. An exact algorithm is proposed that can guarantee to find optimal solution. But the planner can only work optimally under an unreasonably small graph. In order to find optimal utility path in a graph of practical size, approximate algorithm is necessary. 

This section then presents a two-stage planner to find approximate solution. The upper stage risk-aware planner firstly searches for minimum risk paths to all the states in the workspace, using the risk definition and representation in Chapter \ref{chapter::risk_representation} as the metric to quantify safety cost for each path. The proposed upper stage planner can handle locale-dependent and action-dependent risk elements and plan optimal (least risky) path with existence of these two types of risk elements. The correctness of the algorithm is proved by mathematical induction. For general traverse-dependent risk, however, it cannot guarantee the optimality due to the deep dependency on history up to the start position. Counter example is also shown. The upper stage risk-aware planner is able to plan optimal paths to every other state in the workspace, for up to action-dependent risk elements, and suboptimal paths for traverse-dependent risk elements. The lower stage reward-maximizing planner then takes over, evaluates the reward collected along all risk-aware paths from the upper stage planner, compares the utility as the ratio between reward and risk, and chooses the maximum utility path as the final result.

\section{Problem Definition}
The unstructured or confined workspace of the visual assistant is mapped, tessellated, and assigned relevant risk information for each tessellation. Depending on the dimensionality of interest, the workspace is converted into a 2D or 3D occupancy grid and each tessellation is simply a grid or voxel cell. Risk information regarding all risk elements in Fig. \ref{fig::universe} could be collected, evaluated, and stored in the relevant tessellation(s) in the workspace. For simplicity and practicality, this dissertation focuses on obstacles as the source of risk, since existing mapping technologies and occupancy grid representation are suitable to represent obstacle-occupied vs. free spaces, and a large set of risk elements could be evaluated based on the existence of obstacles. Other risk elements, such as terrain stability, access element, wheel traction/slippage, are not considered in the scope of this dissertation, due to the lack of relevant risk information or irrelevance with the tethered aerial visual assistance. However, when necessary information is available and relevant to the robot of interest, they could be easily added into the a priori map. The author would like to point out that although the risk used in this section adopts the definition and representation in Chapter \ref{chapter::risk_representation} and therefore should take value in $[0, 1]$, just for easier illustration, sometimes risk takes integer values in this section. All the properties of the risk conform with those in Chapter \ref{chapter::risk_representation}, but risk in this section may take a different scale just for simplicity. 

Reward information is assigned to each tessellation as well: motivated by the visual assistance problem, visiting each tessellation of the workspace can provide the primary robot's operator with the viewpoint from that tessellation. Going through a series of good viewpoints should have higher reward than going through a series of bad viewpoints and then a good one. Therefore rewards are ``collected'' along the entire path. Based on the separate study, a viewpoint quality map is generated, with viewpoint quality value assigned to each individual free cell in the map. This is the reward map for the planner. The agent collects reward by reaching the state, and all rewards are summed up from every visited state. The reward could take any arbitrary scale, but the value needs to be consistent within the entire space. 

The agent's sensor and actuation model is assumed to be deterministic, considering all the uncertainties or chance constraints are taken care of by means of the explicit risk representation. In the occupancy grid map, the agent is fully aware of which cell it is currently located and which cells it took to come to the current one. The transition between cells is deterministic, for example, ``up'', ``down'', ``left'', and ``right'' for a 4-connectivity 2D occupancy grid. 

With the environment (map, risk, reward) and agent model defined, The agent plans its actions to navigate through the 3D occupancy grid state space. The agent starts at a given start location, without a predefined goal location. The planner plans the goal location and the path to the goal simultaneously. The planner needs to maximize overall collected reward and minimize encountered risk along the entire planned path. The balance of the tradeoff between reward and risk is reflected as a ratio between the two (reward/risk), named utility. The physical representation of the utility value is the reward collected by taking one unit risk. In other words, the planner finds the best goal-path pair in the entire free space to achieve optimal utility value. 

\section{Approach}
To address the defined problem, the 3D occupancy grid state space is converted to a bi-directional graph to represent all free spaces in the unstructured or confined environment. The conversion could be deterministic or use randomness-based algorithms such as Probabilistic Road Map (PRM) \cite{kavraki1994probabilistic}. Each vertex of the graph corresponds to a viable state in the state space. The reward value of this vertex is the viewpoint quality of the corresponding state. All the edges between neighboring states are bi-directional. The weight of each edge will be assigned as the reward value of reaching the end vertex of that edge. All those weights of the edges will be added up as the collective reward. On the other hand, risk will be evaluated based on the path connecting from the start location to the current vertex, using the risk representation described in Chapter \ref{chapter::risk_representation}. The history dependency and non-additivity of motion risk require the risk to be evaluated during each individual planning step, back tracing from the current vertex to the start in order to determine the risk associated with executing this particular path. The goal of the planner is to find a simple path leading to a goal, whose utility ratio of all collected reward vs. encountered risk on the path from start to goal is maximized. 

\subsection{Exact Algorithm}
First of all, it is proved that this risk-aware reward-maximizing problem is well-defined: assuming a path cannot visit one vertex twice (a simple path), there is a finite number of simple paths given a finite bi-directional graph. Each of those paths is associated with a collective reward value summed up from all the edge weights on the path. Using the risk representation in Chapter \ref{chapter::risk_representation}, a risk value could be evaluated for each path. A utility ratio between the total reward and risk exists for every path. Therefore, among all the finite number of simple paths, there exists one (if not more) path with maximum utility value. This is the optimal solution to the problem. Apparently, brutal force algorithm that enumerates over all possible paths in the graph can find the optimal solution in finite time: firstly, convert state space to $\mathcal{G} = (\mathcal{V}, \mathcal{E})$ with $\mathcal{V} = \{v_1, v_2, ..., v_n\}$ to be the vertex set, and $\mathcal{E} = \{e_1, e_2, ..., e_m\}$ to be all the edges connecting the vertices. $v_{start}$ represents start location. Reward map from the separate study is matched with $\mathcal{V}$ so that a reward value could be computed from a look-up table $rewards$ of any vertex $v_i$. Alg. \ref{alg::exact_algorithm} shows the recursive Depth-First-Search (DFS) based algorithm to recursively find all simple paths: the main function calls Alg. \ref{alg::exact_algorithm} and passes in $\mathcal{G}$, $v_{start}$, $path$ as one single vertex $v_{start}$, the $rewards$ look-up table, and $current\_reward$ as $0$. A discount factor $\gamma$ between $[0, 1]$ is used to determine how much current reward is favored over history rewards. When expanding from vertex $u$ to $v$, it recursively calls itself on vertex $v$ with updated information. 


\begin{algorithm}[!t]
 \caption{Evaluate\_All\_Simple\_Paths}
 \begin{algorithmic}[1]
 \renewcommand{\algorithmicrequire}{\textbf{Input:}}
 \renewcommand{\algorithmicensure}{\textbf{Global Variable:}}
 \REQUIRE $\mathcal{G}$, $u$, $path$, $rewards$, $current\_reward$, $\gamma$
 \ENSURE  $all\_simple\_paths$, $path\_utilities$
 \FOR {each edge $(u, v)\in\mathcal{G}$}
 	\IF {$v \notin path$}
 		\STATE $path\leftarrow path \cup v$
		\STATE $path\_risk\leftarrow evaluate(path)$
		\STATE $current\_reward \leftarrow \gamma*current\_reward+rewards(v)$
		\STATE $utility \leftarrow current\_reward/path\_risk$
		\STATE $path\_utilities \leftarrow path\_utilities \cup utility$
		\STATE $all\_simple\_paths \leftarrow all\_simple\_paths \cup path$
		\STATE Evaluate\_All\_Simple\_Paths ($\mathcal{G}$, $v$, $path$, $rewards$, $current\_reward$, $\gamma$)
		\STATE $path \leftarrow = path \setminus v$
		\STATE $path\_risk\leftarrow evaluate(path)$
		\STATE $current\_reward \leftarrow \frac{(current\_reward-rewards(v))}{\gamma}$
	\ENDIF
 \ENDFOR
 \end{algorithmic}
\label{alg::exact_algorithm}
 \end{algorithm}

Although the exact algorithm is guaranteed to eventually find optimal path from $v_{start}$ in $G$, it enumerates over all existing simple paths in the graph and will become computationally intractable when the scale of the graph increases. So it is only practical on state space of very small scale. 

\subsection{Approximate Algorithm}
Due to the intractable computational complexity the exact algorithm has to address graphs of normal size, an approximate algorithm is proposed to solve the problem of scale in a reasonable amount of time. The approximate algorithm is divided into two stages. The upper stage plans minimum-risk path from start location to every other state using a search algorithm similar to Dijkstra's approach. For a graph of $V$ vertices, the upper stage planner computes $V-1$ minimum risk paths to other vertices. The lower stage planner then computes the overall collected rewards on those $V-1$ paths, compares with the utility of staying at the start location, and finally picks the one with maximum utility value. 

\subsubsection{Upper Stage Risk-aware Planner}
The upper stage risk-aware planner searches for minimum risk paths to all other vertices in the graph. It is worth to note that the upper-stage planner is also a stand-alone risk-aware planner. Without the existence or consideration of rewards in the planning problem, this risk-aware planner could be used by itself to find least-risky path, i.e. the path with minimum probability of not being finished. Therefore this risk-aware planner is a stand-alone tool useful for finding safe path between two points. 

Traditional risk-aware planner adopted search-based methods and treated risk as additive cost. Algorithms such as Dijkstra's or A* have been extensively used to find minimum-risk (minimum-cost) path between point A and point B. For them, risk of the entire path $P$ is only the summation of the risk of each individual state $s_i$: 

\begin{equation}
risk(P) = \sum \limits_{i=0}^{i=1}r(s_i)
\end{equation}

Formulating risk as static cost dependent only on state alone can maintain desired properties such as additivity and substructure optimality. However, the explicit representation in Chapter \ref{chapter::risk_representation} gives: 

\begin{equation}
risk(P) = 1 - \prod_{i=0}^{n} \prod_{k=1}^{r} (1-r_k(\{s_0, s_1, ..., s_i\}))
\label{eqn::risk_representation_chap4}
\end{equation}

This risk representation has neither additivity nor locale-dependency, and therefore does not have substructure optimality. The risk robot faces at state $i$ is not well-defined on $s_i$, but can take different values depending on the traverse taken $\{s_0, s_1, ..., s_{i-1}, s_i \ ]$. 

\begin{figure}[]
\centering
	\includegraphics[width = 0.6 \columnwidth]{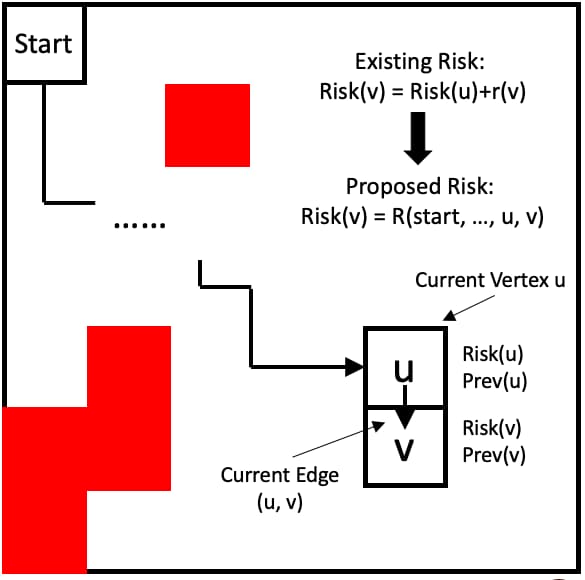}
	\caption{Risk at a certain state is dynamically changing depending on the traverse the robot took to come to the state. So risk at a certain state is not well defined on the locale where this state locates alone, but the entire traverse. Therefore the risk of the entire path is not simply the risk of the subpath plus the risk of this state evaluated by the state alone. The whole path needs to be evaluated (reprinted from \cite{xiao2019explicit1}).}
	\label{fig::dynamical}
\end{figure}

In terms of the impact of those differences on the planner, an intuitive visual example is shown in Fig. \ref{fig::dynamical}: when traditional approaches expand from vertex $u$ to vertex $v$, the risk of the path from start to $v$ is simply the sum of the risk of the subpath from start to $u$ and the risk at $v$. The risk at $v$ is simply well defined on vertex $v$. However, using the proposed risk representation (Eqn. \ref{eqn::risk_representation_chap4}), the risk of vertex (state) $v$ is not well defined by only looking at $v$ alone. The dependency on the history requires the risk at $v$ to be evaluated based on the entire traverse (start, ..., $u$, $v$). 

\begin{figure}[]
\centering
	\includegraphics[width = 0.6 \columnwidth]{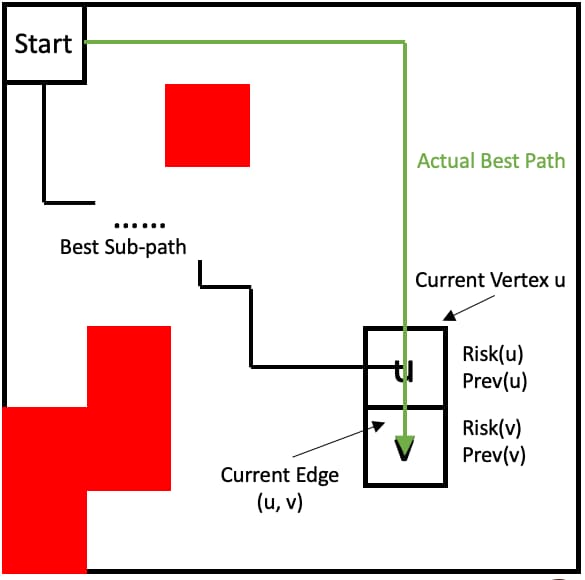}
	\caption{Due to the dependency on history and non-additivity of the proposed risk representation, optimal substructure does not hold. Therefore a subpath of an optimal path may not be optimal (reprinted from \cite{xiao2019explicit1}).}
	\label{fig::directional}
\end{figure}

Due to the dynamically changing state risk dependent on history, the problems loses optimal substructure, therefore scenarios such as the one shown in Fig. \ref{fig::directional} may occur: traditional approaches based on substructure optimality, so if the optimal path to $v$ passes through $u$, the subpath to $u$ is also guaranteed to be optimal to $u$. However, due to the risk representation in Eqn. \ref{eqn::risk_representation_chap4}, this may not be necessarily true. Given the green path is the best path to $v$, the subpath of the green path to $u$ is not the best path to $u$. The black path actually is. The optimal path to $v$ is from a different direction to $u$ as the optimal path to $u$. 

\begin{figure}[]
\centering
	\includegraphics[width = 1 \columnwidth]{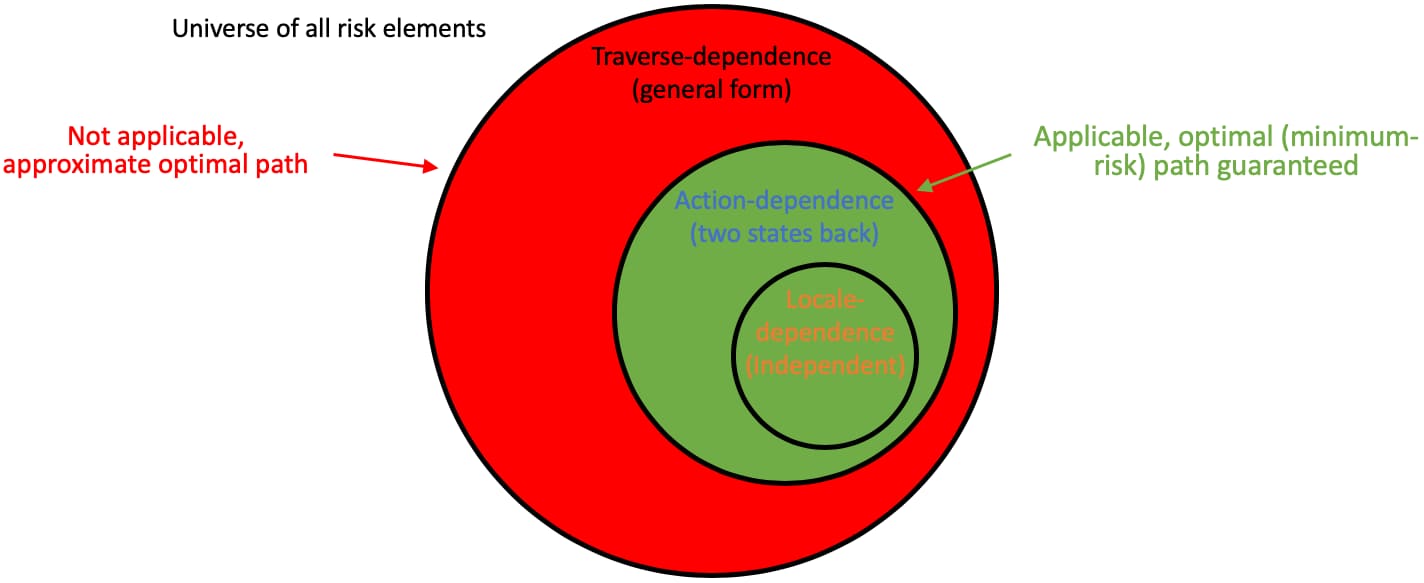}
	\caption{Applicability of the Proposed Risk-aware Planner: Locale-dependent and action-dependent risk elements, not traverse-dependent risk elements}
	\label{fig::plan_applicability}
\end{figure}

Therefore due to the risk definition and representation presented in Chapter \ref{chapter::risk_representation} the planning problem at hand has at least two issues, preventing from the usage of traditional search algorithms: the risk being dynamical (dependent on history) and directional (lost optimal substructure). Considering the three categories of risk elements, locale-dependency, action-dependency, and traverse-dependency, a new algorithm is designed to optimally address the first two categories of risk elements, by looking back two states into history during planning, at the cost of more computation. Traverse-dependent risk elements, however, cannot be guaranteed to be optimally addressed, since it is the most general form of risk and the look-back has to be into the entire history to guarantee optimality. Only those risks caused by two steps back into the traverse could be handled. Fig. \ref{fig::plan_applicability} shows the applicability of the proposed risk-aware planner. 

Similar to Dijkstra's algorithm, the risk-aware planner deploys a search-based method that computes minimum risk paths to every other vertices in the graph, but conforms the newly proposed risk definition and representation. Two major modifications from the original Dijkstra's are designed to accommodate the special properties of the proposed risk framework: since our risk representation does not have substructure optimality, we add directional components to each vertex and minimum risk path reaching each vertex from each directional component is computed. This is the directional part and provides the opportunity of two-step look-back to the planner. The second difference is dynamical risk evaluation due to the non-additivity and history dependency of the risk elements in the explicit risk representation (Chapter \ref{chapter::risk_representation}). The risk from \textit{start} via \textit{u} to \textit{v} is not simply risk(\textit{u}) + r(\textit{v}). Here, the risk of getting \textit{v} needs to be re-evaluated based on the path from \textit{start} via \textit{u} to \textit{v} dynamically using the explicit risk representation proposed in Chapter \ref{chapter::risk_representation}. 

Let $\mathcal{V} = \{v_1, v_2, ..., v_n\}$ to be the vertex set, and $\mathcal{E} = \{e_1, e_2, ..., e_m\}$ to be all the edges connecting the vertices. To accommodate the history dependency of action-dependent risk elements, each vertex is further represented by $v_i = (D_i^{(1)}, D_i^{(2)}, ..., D_i^{(c)})$, where $D_i^{(j)}$ represents the direction from which $v_i$ is reached. They memorize the two-step history information to be used when being expanded in the future. The total number $c$ is the connectivity of $v_i$, as the number of incoming edges reaching $v_i$. For each direction reaching $v_i$, $D_i^{(j)}$ is defined as $D_i^{(j)} = (r_i^{(j)}, PD_i^{(j)})$, where $r_i^{(j)}$ is the risk of reaching $v_i$ from direction $D_i^{(j)}$ starting from start vertex $v_{start}$, and $PD_i^{(j)}$ is the previous direction of reaching the previous vertex, in other words, previous direction of two steps back. All the directions of all vertices $D_i^{(j)}$ compose the superset of all directions $\mathcal{D} = \{D_i^{(j)}|i=1, 2, ..., n\}$ and $j$ is a variable for different vertices depending on how many directions (edges) are leading to the vertex. The graph is defined as $\mathcal{G} = (\mathcal{V}, \mathcal{E})$. The algorithm is shown in Alg. \ref{alg::risk-aware}. It finds the minimum-risk directional component in the graph (line 5) and expands the vertex which this directional component belongs to (line 6 - line 15). After expanding all the neighbors, this directional component is marked visited (line 16). When all directional components are visited, the final minimum-risk path to each vertex is selected from its minimum-risk directional components (line 18 - 22).

 
 \begin{algorithm}[!t]
 \caption{Risk-aware Path Planner}
 \begin{algorithmic}[1]
 \renewcommand{\algorithmicrequire}{\textbf{Input:}}
 \renewcommand{\algorithmicensure}{\textbf{Output:}}
 \REQUIRE $\mathcal{G}$, $v_{start}$
 \ENSURE  Risk-aware paths to all vertices other than $v_{start}$
  \STATE $\forall D_i^{(j)} \in \mathcal{D}$ set $r_i^{(j)} \leftarrow \infty$ and $PD_i^{(j)} \leftarrow NULL$
  \STATE For $v_{start}$, set $r_{start}^{(j)}\leftarrow 0$ in all $D_{start}^{(j)}$
  \STATE Initialize visited set to $\mathcal{R}\leftarrow\{\}$
  \WHILE {$\mathcal{R}\neq\mathcal{D}$}
  	\STATE pick vertex $v_u$ with smallest $r_u^{(i)}$ where $D_u^{(i)}\notin\mathcal{R}$
	\FOR {each edge $(v_u, v_v)\in\mathcal{E}$}
		\STATE $path_u^{(i)}\leftarrow backtrack(D_u^{(i)}$)
		\STATE $path_v(i) \leftarrow path_u^{(i)} \cup \{v_v\}$
		\STATE $path\_risk_v(i)\leftarrow evaluate(path_v(i))$
		\STATE $current\_min\_risk\leftarrow v_v.D_v^{(j)}.r_v^{(j)}$, where $D_v^{(j)}$ corresponds to reaching $v_v$ from $v_u$
		\IF {$path\_risk_v(i)<current\_min\_risk$}
			\STATE $v_v.D_v^{(j)}.r_v^{(j)}\leftarrow path\_risk_v(i)$
			\STATE $v_v.D_v^{(j)}.PD_v^{(j)} \leftarrow D_u^{(i)}$
		\ENDIF
	\ENDFOR
	\STATE $\mathcal{R}\leftarrow \mathcal{R} \cup \{D_u^{(i)}\}$
  \ENDWHILE 
  \FOR {each $v_i \in \mathcal{V}$}
  	\STATE pick $D_i^{(j)}$ with the smallest $r_i^{(j)}$
	\STATE $risk_i \leftarrow r_i^{(j)}$
	\STATE $path_i \leftarrow backtrack(D_i^{(j)})$
  \ENDFOR
 \RETURN all $path_i$ with $risk_i$
 \end{algorithmic}
 \label{alg::risk-aware}
 \end{algorithm}
 
\begin{figure}[]
\centering
	\includegraphics[width = 0.6 \columnwidth]{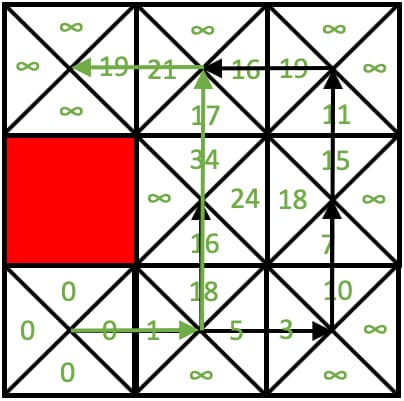}
	\caption{Example of the Risk-aware Planner: Each grid represents a tessellation (vertex) in the workspace with red indicating obstacle. Start location is at the lower left corner. The partition of each tessellation into four directional components is due to the four-connectivity assumption. The numbers indicate the minimum risk the robot faces to reach the vertex through the directional component from start location. Optimal path to each tessellation is shown in black arrow, except the one to the upper left tessellation in green to highlight the efficacy of the proposed algorithm. }
	\label{fig::toy_example}
\end{figure}

One example is displayed in Fig. \ref{fig::toy_example}. The 2D Cartesian workspace is tessellated into a three-by-three 2D occupancy grid, with one red obstacle on the left. The robot starts at the lower left corner. The map is converted to a graph with vertices and edges. The map is assumed to have four-connectivity, i.e. the robot can move up, down, left, or right. Therefore each vertex is augmented into four different directional components (with edge and corner cases treated similaly for simplicity). Applying the proposed risk-aware algorithm can find minimum risk paths to all directional components, from which minimum risk path to each vertex (tessellation) is selected. The optimal path to each vertex is shown in black arrows, except the optimal path to the upper left tessellation shown in green. While the optimal path to the upper middle tessellation is the black one going through the entire right hand side of the map, the optimal path to the upper left tessellation is the green path, passing through the upper middle tessellation, where substructure optimality does not hold: the green path is an optimal path which would not be possible to find using traditional planner. 
 
\begin{figure}[]
\centering
\subfloat[Path 1 (Result of Conventional Planner)]{\includegraphics[width=0.5\columnwidth]{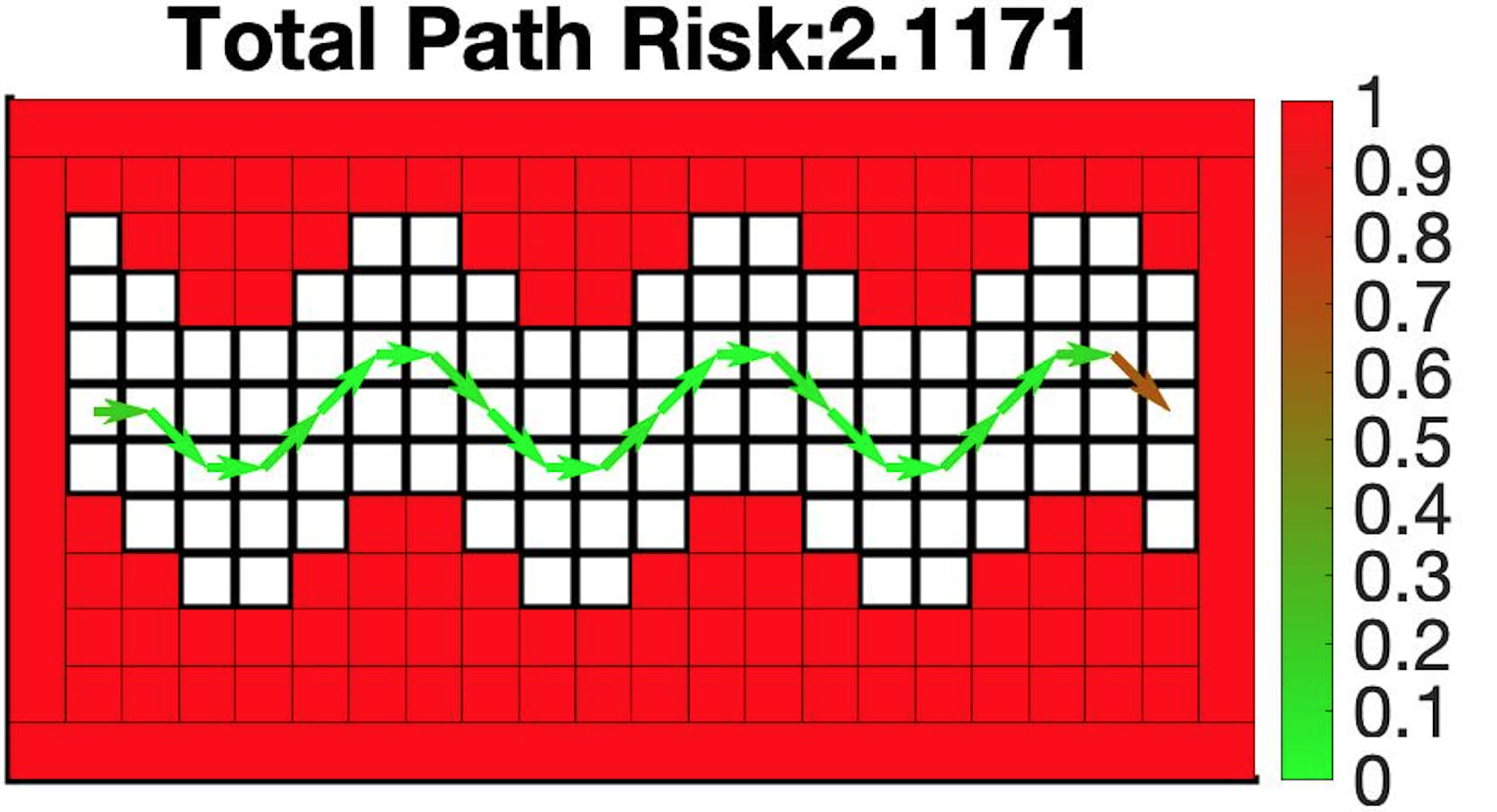}%
\label{fig::path11}}
\hfill
\subfloat[Path 2 (for Comparison)]{\includegraphics[width=0.5\columnwidth]{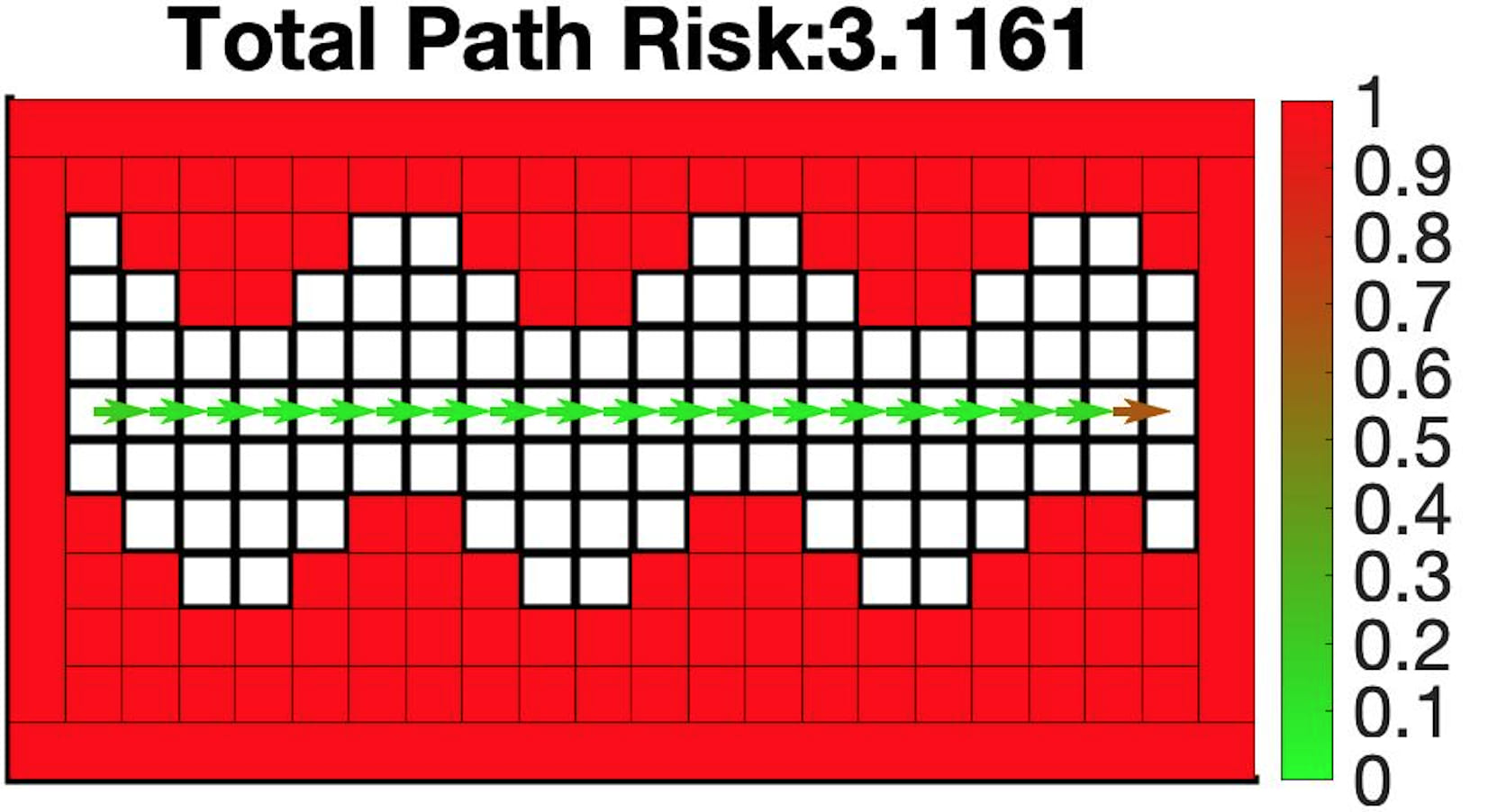}%
\label{fig::path12}}
\caption{Conventional Planner with Additive State-dependent Risk}
\label{fig::conventional}
\end{figure}

\begin{figure}[]
\centering
\subfloat[Path 1 (for Comparison)]{\includegraphics[width=0.5\columnwidth]{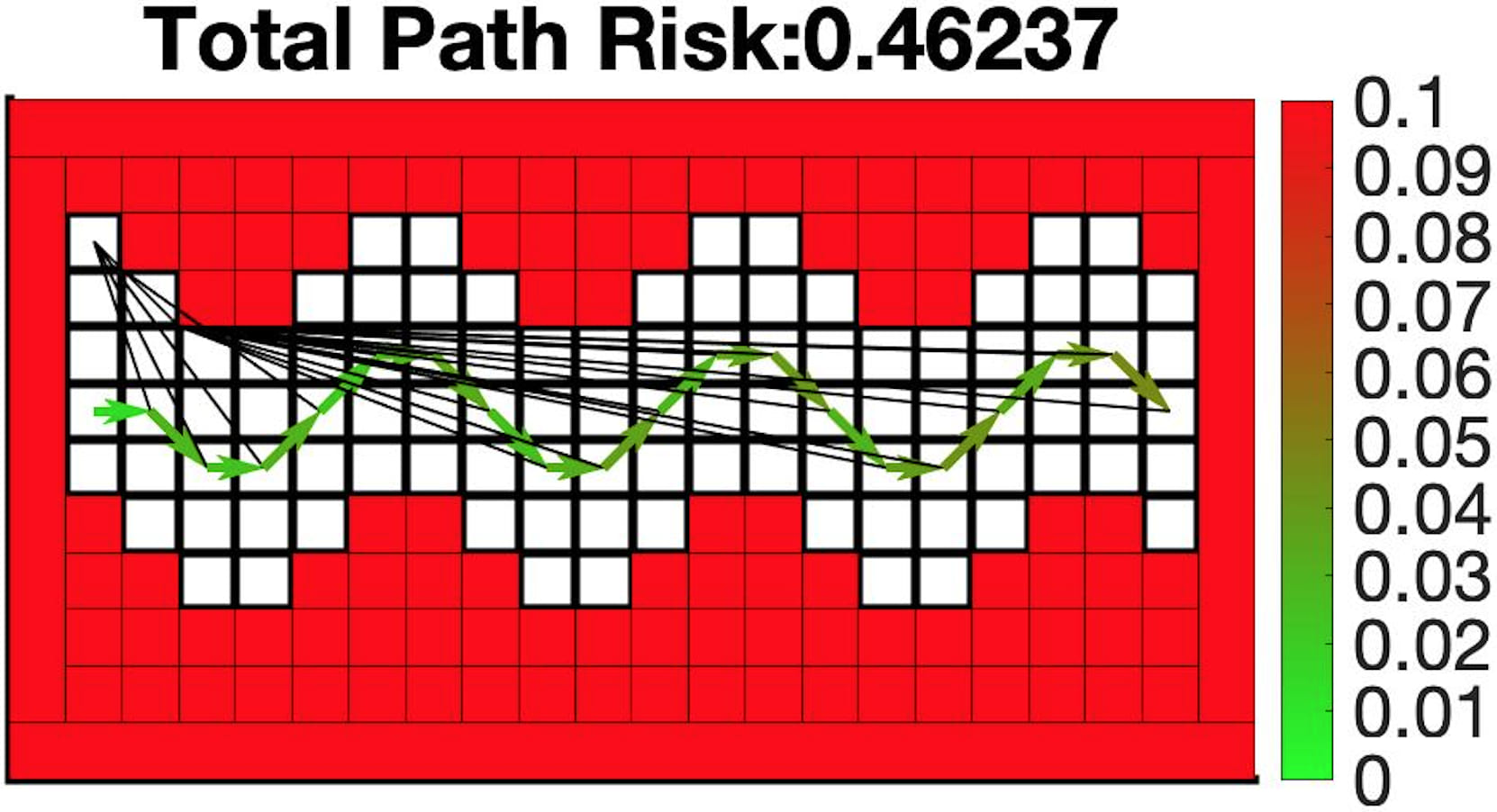}%
\label{fig::path21}}
\hfill
\subfloat[Path 2 (Result of Proposed Planner)]{\includegraphics[width=0.5\columnwidth]{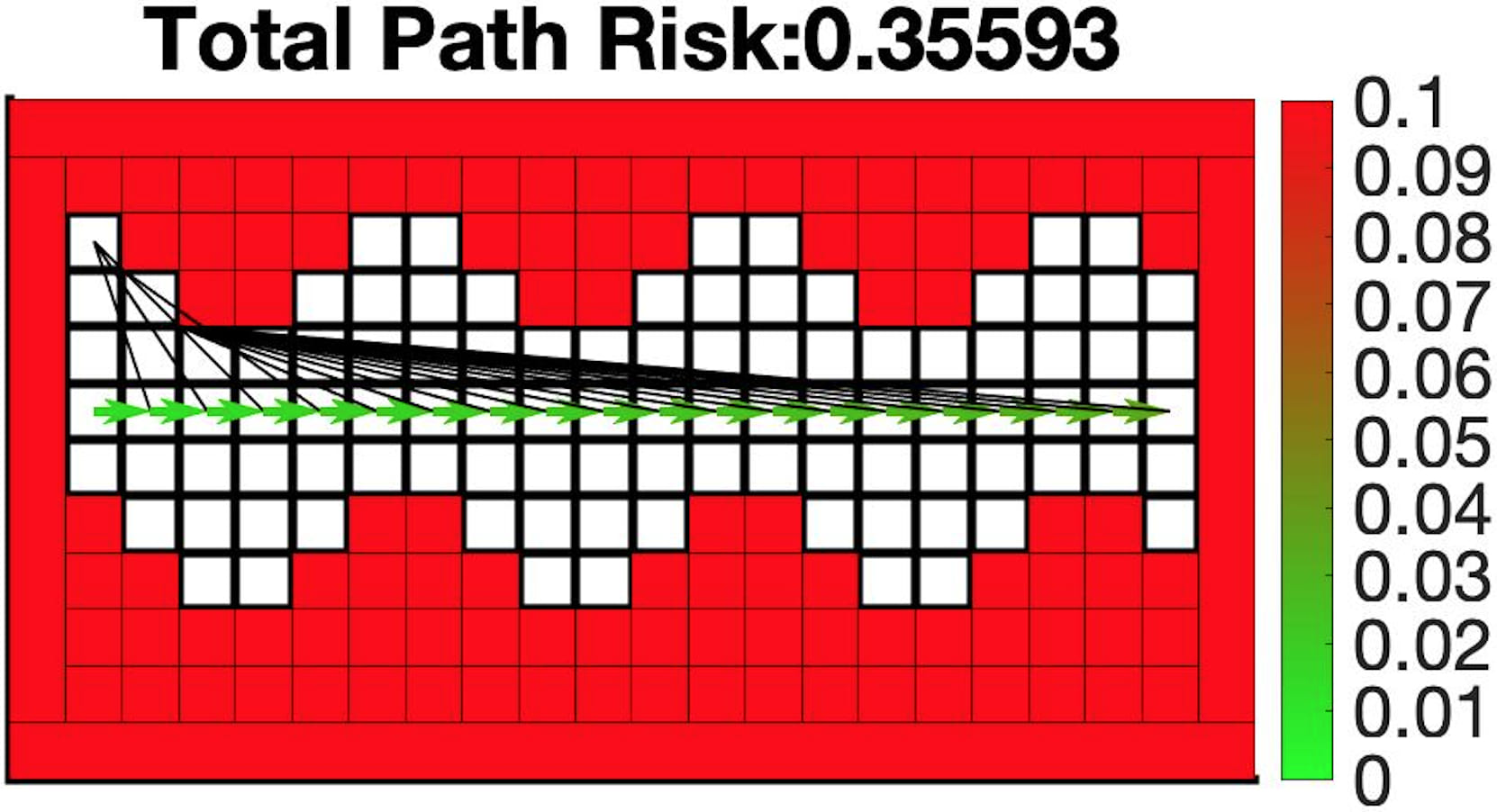}%
\label{fig::path22}}
\caption{Proposed Planner with Probabilistic History-dependent Risk}
\label{fig::proposed}
\end{figure}

Examples using the proposed risk-aware planner are shown in Fig. \ref{fig::proposed}. As comparison, results of conventional risk-aware planner based on additive state-dependent risk are presented in Fig. \ref{fig::conventional}. For easy illustration, the workspace is assumed to be a 2D occupancy grid, with each grid either be free (white) or occupied (red). The color of the arrows indicates the risk the robot faces at each state and the color map is displayed on the right. The robot starts from the left of the map and the goal is going to the right. Six risk elements are chosen as examples from the three risk categories: distance to closest obstacle and visibility from locale-dependent risk elements, action length and turn from action-dependent risk elements, and tether length and number of tether contacts from traverse-dependent risk elements. 

For distance to closest obstacle and visibility, fuzzy logic similar to the approach used in \cite{soltani2004fuzzy} is used: a fuzzy membership function first computes a membership value, which is then proportionally converted to a probability value. The rationale behind this is the closer a state is to obstacle or the lower visibility the state has, the higher probability that the robot is not able to finish this state. The proportional assumption of the probability to the fuzzy membership value is simple and straightforward, but more sophisticated probabilistic model could be used to capture more complex risk relationship. The one important property is locale-dependency. 

For action length, the risk (probability) value is proportional to the norm of the difference between the last and current state. Turning is the difference between two actions, which is further the difference between two states. So second to last, last and current state are taken into account. The risk (probability) value is proportional to the norm of the difference between two actions. Again, we assume an easy linear relationship between risk and action length or turning magnitude. 

Tether length and number of contacts are specific risk elements for tethered vehicles, for example, our tethered aerial visual assistant. Both of them are traverse-dependent and the methods to compute them are discussed in detail in the low level motion suite in the following chapter (Chapter \ref{chapter::low_level}). The tether planning techniques can output the length and number of contacts. We simply assume the probability of failure (risk) is proportional to the length and number. 

Fig. \ref{fig::path11} shows the result of conventional risk-aware planner using additive state-dependent risk. Due to the assumption of state-dependency, action length, turn, and tether length, number of contacts cannot be properly addressed by the planner. The only possible risk elements are distance to closest obstacle and visibility, which are evaluated based on state alone. Their risk values at each state are combined using normalization and weighted sum (identical weights for both risk elements in the examples) and summed up along the entire path. Using this approach, the planner will find the path shown in Fig. \ref{fig::path11}, since this is the minimum risk path according to the conventional additive state-dependent risk representation and could be found by traditional search-based algorithms, such as Dijkstra's or A*. The path shown in Fig. \ref{fig::path12}, however, will be neglected, since it is supposed to have a higher risk according to the additive state-dependent risk representation. 

Fig. \ref{fig::path22} shows the result of the proposed risk-aware planner using probabilistic history-dependent risk. All six risk elements from all three risk categories could be properly addressed by the proposed planner, with the optimality of locale-dependent and action-dependent risk elements guaranteed but traverse-dependent risk elements not guaranteed. The risk at each state is now formulated as the probability of robot not being able to finish the state, displayed in color. The probability of not being able to finish the path, as risk index of the path, is computed using propositional logic and probability theory presented in Chapter \ref{chapter::risk_representation}. The two-step look-back in the proposed risk-aware planner makes sure that history dependencies of risk up to actions could be addressed optimally. The traverse-dependent risk elements, however, are only suboptimal, or in other words, optimal up to two states in the history of the traverse, not the entire history to the start. As shown by Fig. \ref{fig::proposed}, the risk aware-planner is willing to sacrifice distance to closest obstacle and visibility (locale-dependent risk elements) for shorter action length, less aggressive turn (action-dependent risk elements), and shorter tether length, few tether contacts (traverse-dependent risk elements).

\begin{figure}[]
\centering
\subfloat[Robot squeezes through to minimize action-dependent and traverse-dependent risk elements]{\includegraphics[width=0.5\columnwidth]{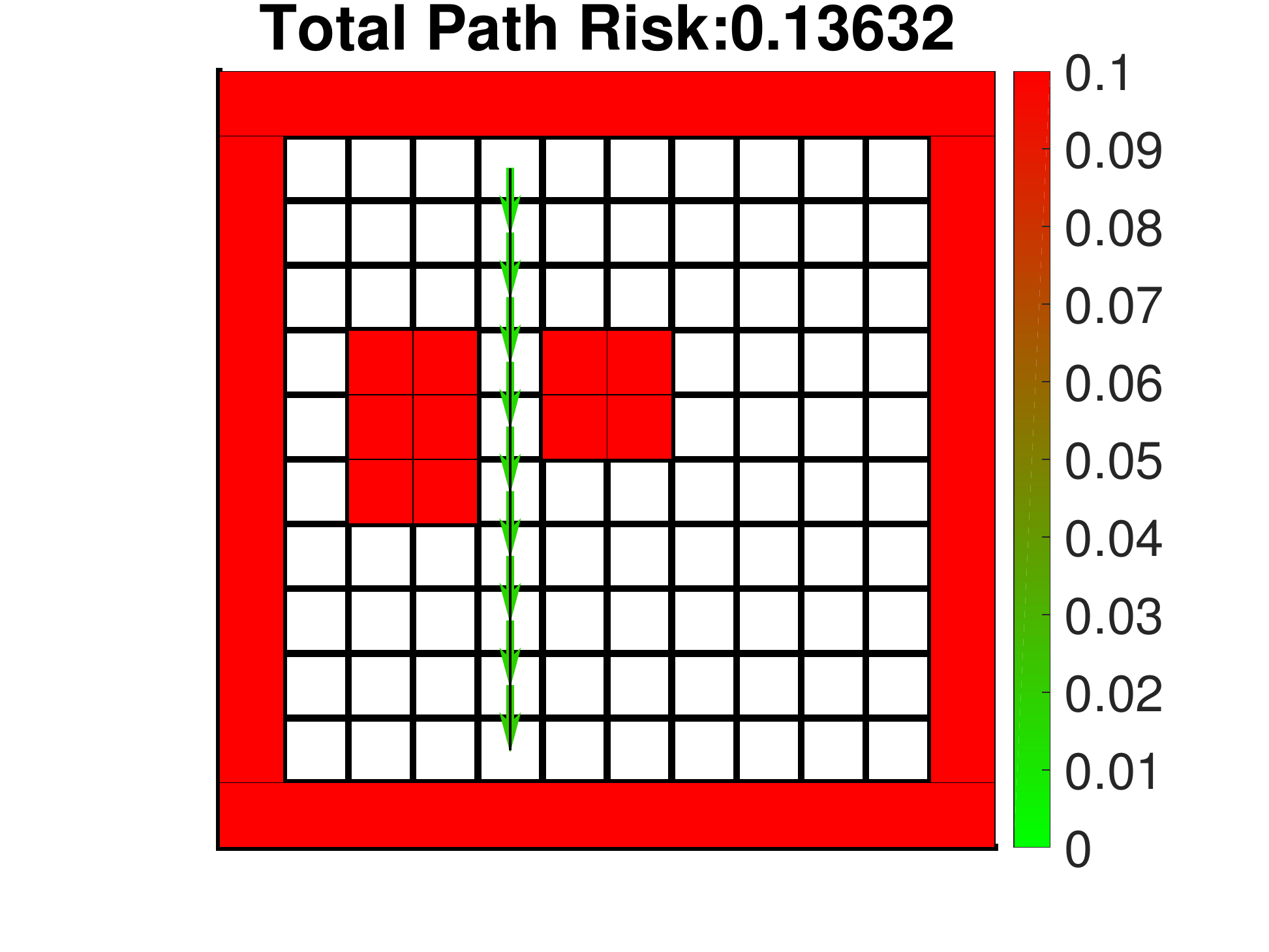}%
\label{fig::in_between}}
\hfill
\subfloat[Robot makes detour to optimize for locale-dependent risk elements only]{\includegraphics[width=0.5\columnwidth]{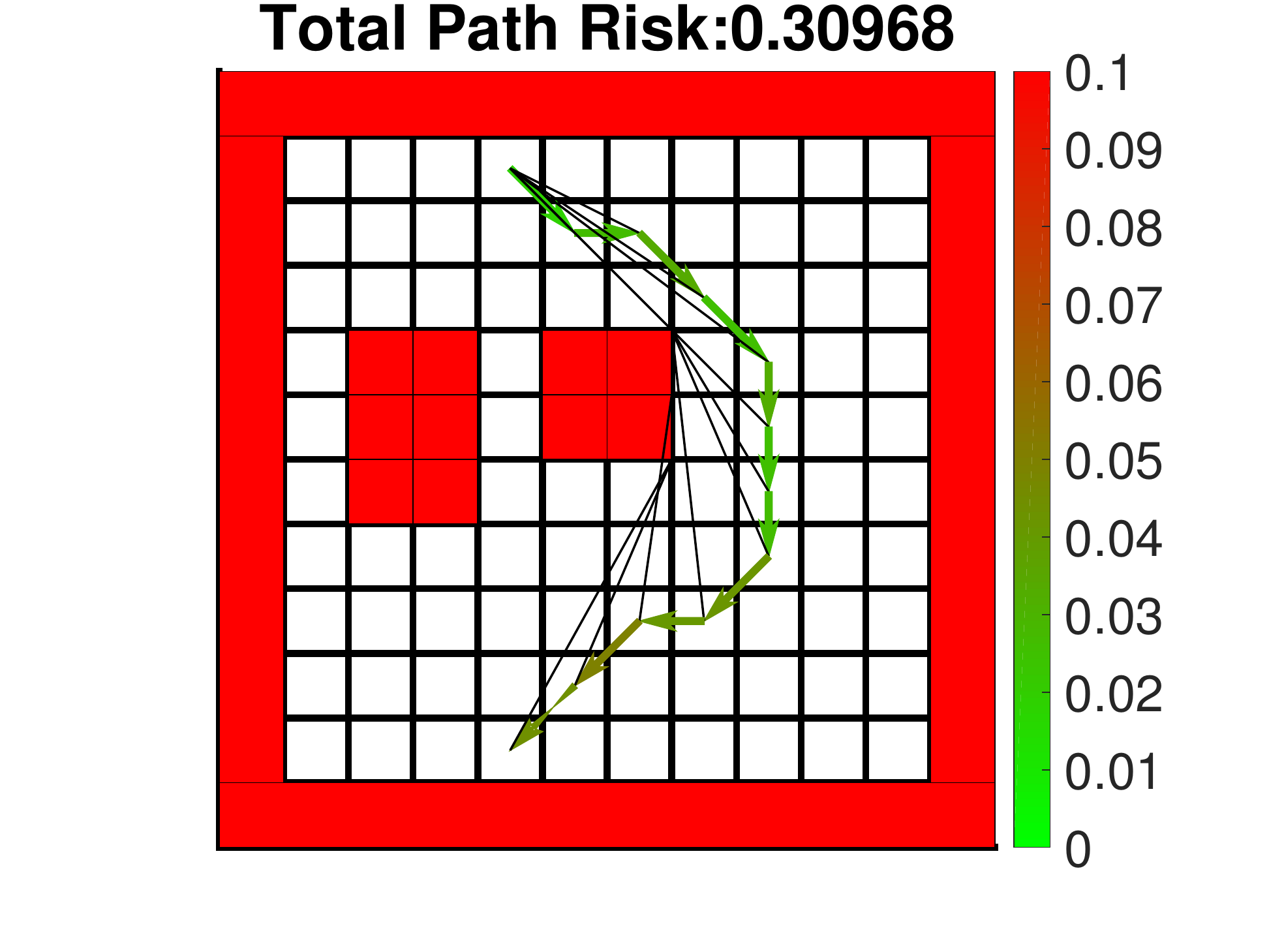}%
\label{fig::around}}
\caption{Path Risk Associated with Paths Found by Proposed and Conventional Planners}
\label{fig::squeeze_around}
\end{figure}

Another set of examples are shown in Fig. \ref{fig::squeeze_around}. The proposed planner finds path shown in Fig. \ref{fig::in_between}, where squeezing through the narrow passage between obstacles compromises distance to closest obstacle and visibility (locale-dependent risk elements), but optimizes action length, turn, tether length, and number of tether contacts. The optimal path found by the conventional planner (Fig. \ref{fig::around}) by only considering state-dependent risk elements, however, has a higher probabilistic risk index. Using the formal risk definition and explicit representation, path on the right has 0.31 probability of not finishing, while path on the left only has 0.14. Therefore the path found by the conventional planner is riskier than the one found by the proposed risk-aware planner. 

\begin{figure}[]
\centering
	\includegraphics[width = 0.9 \columnwidth]{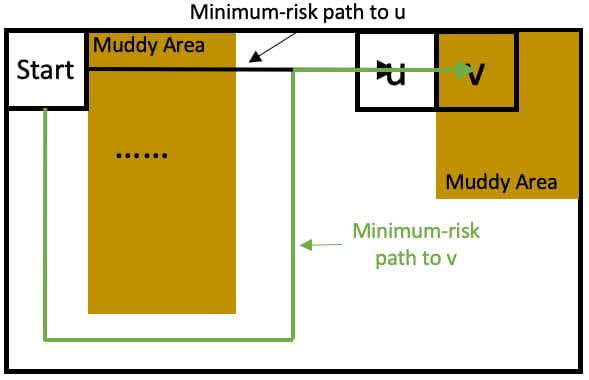}
	\caption{Example of the Proposed Planner Fails to Find Minimum Risk Path due to Traverse-dependency}
	\label{fig::not_traverse}
\end{figure}

One example of why the proposed risk-aware planner cannot optimally address traverse-dependent risk elements is shown in Fig. \ref{fig::not_traverse}. Take wheel traction/slippage as an example of traverse-dependent risk element and assume the robot has two muddy areas to negotiate with in the workspace: the minimum risk path to $u$ could be the black path shown in Fig. \ref{fig::not_traverse}, since $u$ is in a clean area and the mud built up on the robot wheels would not cause significant risk at $u$. However, if the robot keeps venturing into $v$, which is another muddy area, the mud built up on the wheels from the first muddy area may cause major risk and the robot has very high probability of getting stuck at $v$. The green path becomes less risky, since the risk associated with the extra length and turns are justified by keeping clean wheels and reliable traction. However, the green path can never been found by the proposed risk-aware planner, since two-step look-back (from $v$ looking back to $u$ and the state left to $u$) cannot cover sufficient depth into history to find the green path. Therefore, for traverse-dependent risk element, only risk caused by the last two steps could be properly addressed, in the similar way as how action-dependent risk element is addressed. 

It is easy to show that the algorithm is optimal for up to action-dependent risk elements. Like the inductive proof in Dijkstra's algorithm where the minimum cost to each vertex is found whenever this vertex is closed, our risk-aware planner guarantees that each directional component has minimum risk when being closed: because action-dependent risk can only cause different results within two steps in the history, the substructure optimality actually exists before two vertices back. All the candidate minimum-risk paths have optimal substructure before two steps in the history, so all of them are properly addressed when evaluating the directional component. Using directional components actually embeds memory of one step back and one step in the future, so action-dependent risk value could be uniquely determined when expanding on a directional component. Therefore the optimality with respect to locale-dependent and action dependent-risk elements is guaranteed. 

let $r(D)$ be the minimum risk from direction $D$ to the vertex it belongs to, found by the algorithm, and let $\delta(D)$ be the actual minimum risk from $v_{start}$ to $D$. We want to show that $r(D) = \delta(D)$ for every directional component $D$ at the end of the algorithm. We use induction to prove the correctness. Note that the multiplicativity in the risk representation in Eqn. \ref{eqn::risk_representation_chap4} could be transformed to additivity by taking the logarithm form. But the risk at a certain state is still dependent on history. In particular, for up to action-dependent risk elements, two previous states back are necessary to evaluate the risk at the current state. 

\textbf{Proof by Induction:} 

\emph{Based case $|R| = 1$:} Since $R$ only grows in size, the only time $|R| = 1$ is when $R = \{v_{start}\}$ and $r(v_{start}) = 0 = \delta(v_{start})$, which is correct. 

\emph{Inductive step:} Assume that for each $x \in R, r(x) = \delta(x)$. This is our inductive hypothesis. It needs to be shown that if a directional component $u$ is added to $R$ so $R' = R \cup \{u\}$, we have for each $x \in R', r(x) = \delta(x)$. Due to the inductive hypothesis, we only need to prove $r(u) = \delta(u)$.

Using directional components instead of vertex itself can uniquely determine the risk caused by action-dependent risk: when expanding on a directional component, one step back into the history is embedded in the direction itself, while the next step is the vertex which the current edge connects to. So for the next state, history from two states back is memorized and therefore the additional additive term in each step in the logarithm form of Eqn. \ref{eqn::risk_representation_chap4} will be only $r_k({s_{i-2}, s_{i-1}, s_i})$. Note taking logarithm can change multiplicativity to additivity. In this sense, the whole risk-aware planning problem can be converted to Dijkstra's search based on directional components. 

Suppose for a contradiction that the minimum risk path from $v_{start}$-to-$u$ is $Q$ and has length

\begin{equation}
l(Q)<r(u)
\end{equation}

$Q$ starts in $R$ and at some point leaves $R$ to get to $u$ which is not in $R$. Let $xy$ be the first edge along $Q$ that leaves $R$. Let $Q_x$ be the $v_{start}$-to-$x$ subpath of $Q$. It is important to note that $x$ is not a vertex, but a directional component, just like $u$ and $y$. This fact guarantees that which state was before $x$ was implicitly memorized by $x$ as a directional component. That state, $x$, and $y$ compose the three states relevant to action-dependency, $s_{i-2}, s_{i-1}, s_i$. Therefore the transformation from multiplicativity to additivity through logarithm of Eqn. \ref{eqn::risk_representation_chap4} can give: 

\begin{equation}
l(Q_x) + l(xy) \leq l(Q)
\end{equation}

Note $l(xy)$ is well defined based on action-dependency. $l(Q_x)$ is the minimum risk to $x$ based on substructure optimality before two states back in the history. 

Since $r(x)$ is the risk of the minimum risk $v_{start}$-to-$x$ path by the inductive hypothesis, $r(x) \leq l(Q_x)$, we have 

\begin{equation}
r(x) + l(x, y) \leq l(Q)
\end{equation}

Since $y$ is the directional component adjacent to $x$, $r(y)$ must have been updated by the algorithm when expanding $x$, so 

\begin{equation}
r(y) \leq r(x) + l(xy)
\end{equation}

Finally, since $u$ is currently expanded by the algorithm (before $y$), $u$ must have smaller risk than $y$:

\begin{equation}
r(u) \leq r(y)
\end{equation}

Combining all these inequalities gives us: 

\begin{equation}
r(x)+l(xy) \leq l(Q) < r(u) \leq r(y) \leq r(x) + l(xy)
\end{equation}

which means

\begin{equation}
r(x) < r(x) 
\end{equation}

This is the contradiction. Therefore, no less risk path $Q$ exists and so $r(u)=\delta(u)$.  \hfill $\blacksquare$

Therefore we prove that the risk-aware planner is correct with respect to action-dependent risk elements. The transformation from multiplicativity to additivity using logarithm and the two-state history embedded by directional components to ensure risk's well-definedness are necessary for the proof.

\begin{figure}[]
\centering
	\includegraphics[width = 1 \columnwidth]{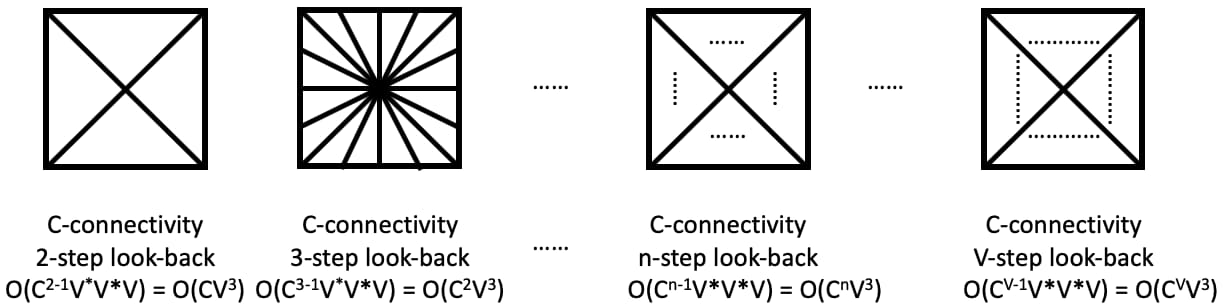}
	\caption{Potential Extension of the Proposed Risk-aware Planner: Trading more computation for deeper history dependency (Graphical illustrations assume 4-connectivity as example)}
	\label{fig::dependency_computation}
\end{figure}

\begin{figure}[]
\centering
	\includegraphics[width = 1 \columnwidth]{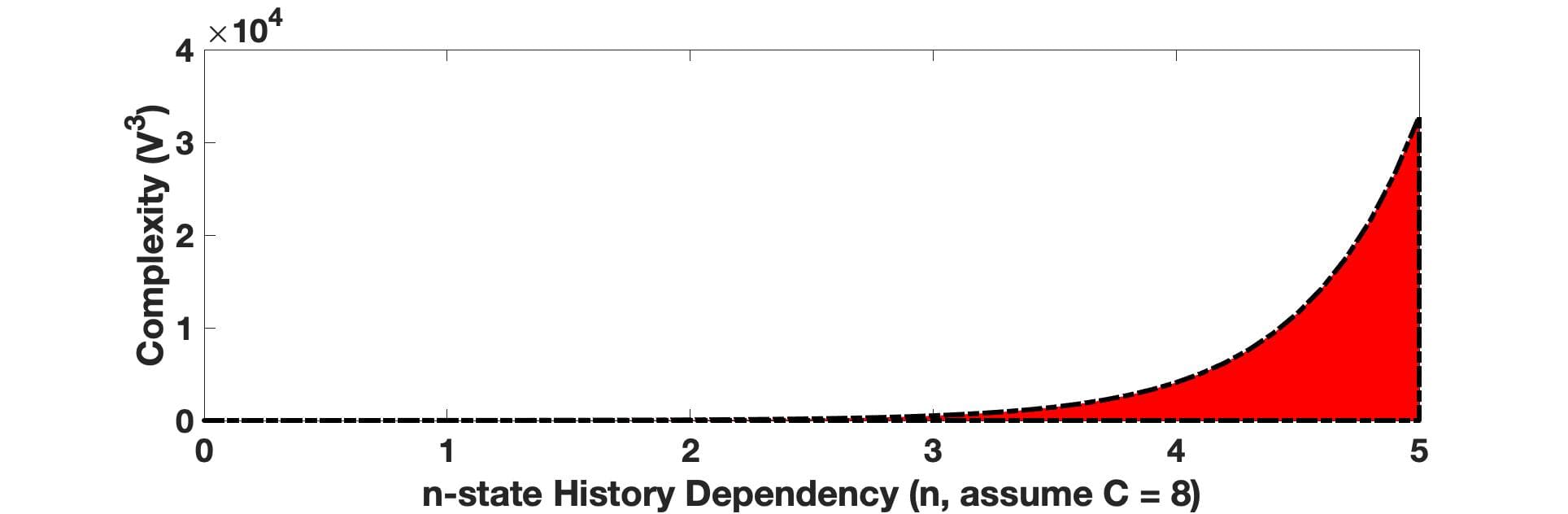}
	\caption{Complexity as Function of History Dependency Depth (Graph generated assuming 8-connectivity as example)}
	\label{fig::history_complexity}
\end{figure}

Although it is shown above that the proposed risk-aware planner is not optimal with respect to general traverse-dependent risk elements, tt is possible that look-back into more steps in the history can direct the planner closer to the true optimal path, but at the cost of computation. Fig. \ref{fig::dependency_computation} shows the potential extension of the proposed risk-aware planner in order to be able to address more depth in history dependency. The proposed risk-aware planner looks two-step back into the history and therefore augments every original vertex into $C$ directional components (as the four partitions in the left state of Fig. \ref{fig::dependency_computation}, assuming $C=4$). If three-step look-back is necessary, the original vertex could be augmented into $C^2$ directional components (as the sixteen partitions in the second to left state of Fig. \ref{fig::dependency_computation}, assuming $C=4$). By the same token, an arbitrary number $n$-step look-back requires $C^{n-1}$ directional components. The deepest possible history dependency is $V$ steps, as the longest simple paths have $V$ vertices (here, $V$ is trivially equivalent to $V-1$) and the complexity would be $\mathcal{O}(C^VV^3)$. The complexities of the proposed risk-aware path planner (2-step look-back), potential 3-step look-back planner, general risk-aware planner ($n$-step look-back), and omnipotent risk-aware planner with full history dependency are shown in Fig. \ref{fig::dependency_computation}. The omnipotent risk-aware planner with full history dependency is supposed to guarantee optimality with even traverse-dependent risk elements, but the computation is intractable (Fig. \ref{fig::history_complexity}). 

Fig. \ref{fig::history_complexity} shows the exponential increase in computation with increasing depth into history dependency. The connectivity is assumed to be 8 as an example to generate the graph. 

\subsubsection{Lower Stage Reward-maximizing Planner}
Given an ensemble of minimum-risk paths computed by the upper stage risk-aware planner, the lower stage planner maximizes reward based on the path utility. The upper stage planner provides minimum risk paths to all vertices in the graph. The lower stage planner then picks the most rewarding path among them by maximizing the utility value defined as the ratio between total collected reward and encountered risk on the path (Alg. \ref{alg::lower_stage_planner}).
 
\begin{algorithm}[]
 \caption{Lower Stage Maximum Reward Planner}
 \begin{algorithmic}[1]
 \renewcommand{\algorithmicrequire}{\textbf{Input:}}
 \renewcommand{\algorithmicensure}{\textbf{Output:}}
 \REQUIRE \textit{ensemble of minimum risk paths}, \textit{reward map}
 \ENSURE  sub-optimal utility path from \textit{start} 
  \FOR {each \textit{path} from \textit{start} to \textit{v} in \textit{ensemble}}
	\STATE Compute overall collected reward
	\STATE utility[\textit{path}] $\leftarrow$ reward[\textit{path}]/risk[\textit{path}]
  \ENDFOR
  \STATE Compute utility of staying at \textit{start} as a unit \textit{path}
   \RETURN \textit{path} with maximum utility value
 \end{algorithmic}
 \label{alg::lower_stage_planner}
 \end{algorithm}
 
 Alg. \ref{alg::lower_stage_planner} iterates over a subset of paths of Alg. \ref{alg::exact_algorithm}. Instead of iterating over all possible simple paths in the graph from start location,  Alg. \ref{alg::lower_stage_planner} only looks at minimum risk path to each vertex. This is more computationally efficient but may sacrifice optimality. Given a minimum risk path \textit{p} from \textit{start} to \textit{v} which encounters risk value of 10 and collects 20 reward, the utility value of this path is 2. However, there may exist a slightly riskier path \textit{p'} from \textit{start} to \textit{v}, which has risk value 11 and therefore is neglected by the upper level planner. This path, however, may have a reward value of 25, making the utility to be 25/11 = 2.27 (>2). Apparently \textit{p} is suboptimal considering the existence of \textit{p'}. 
 
\begin{figure}[]
\centering
\subfloat[Exact Solution]{\includegraphics[width=0.5\columnwidth]{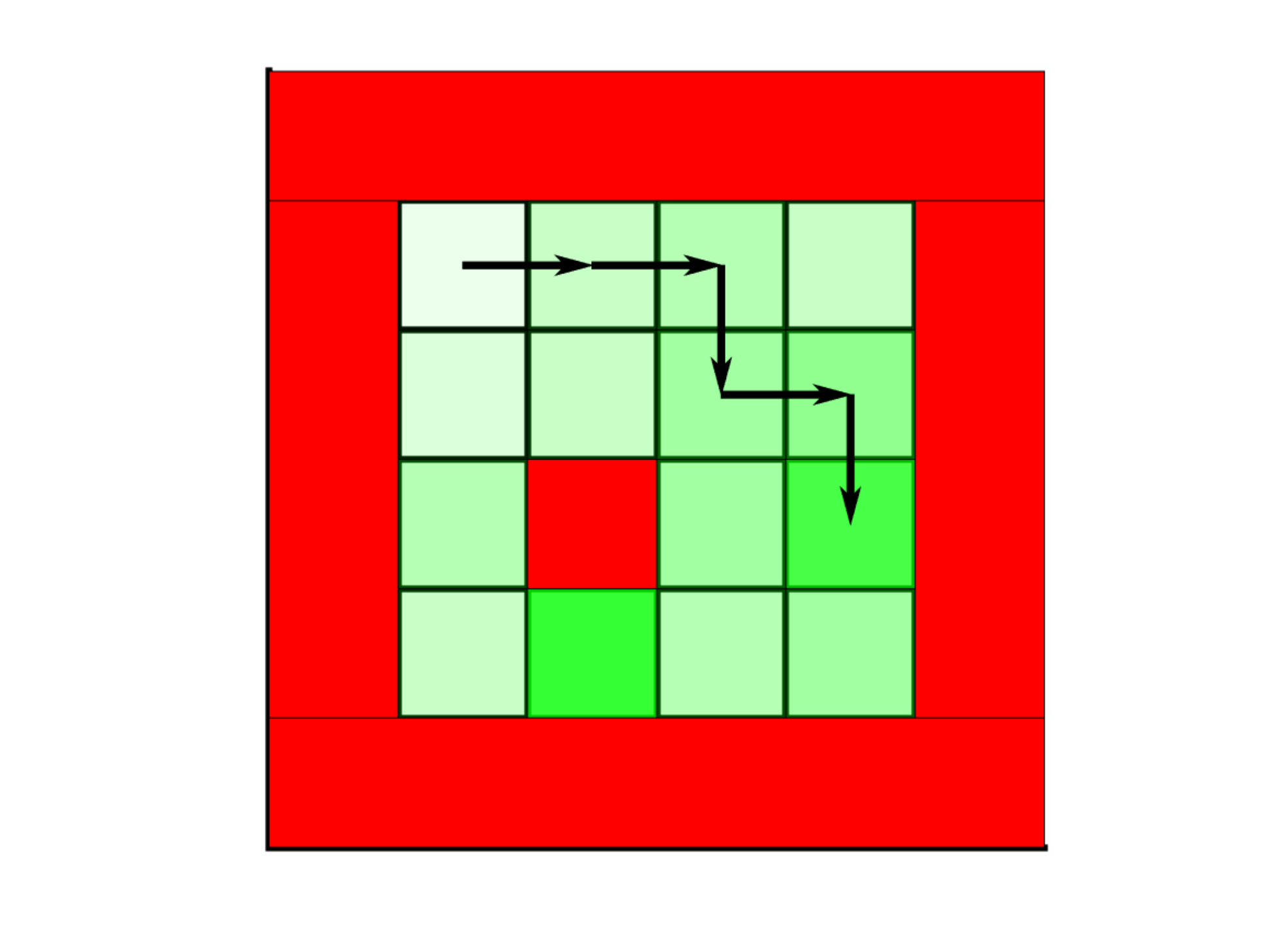}%
\label{fig::exact_4x4}}
\hfill
\subfloat[Approximate Solution]{\includegraphics[width=0.5\columnwidth]{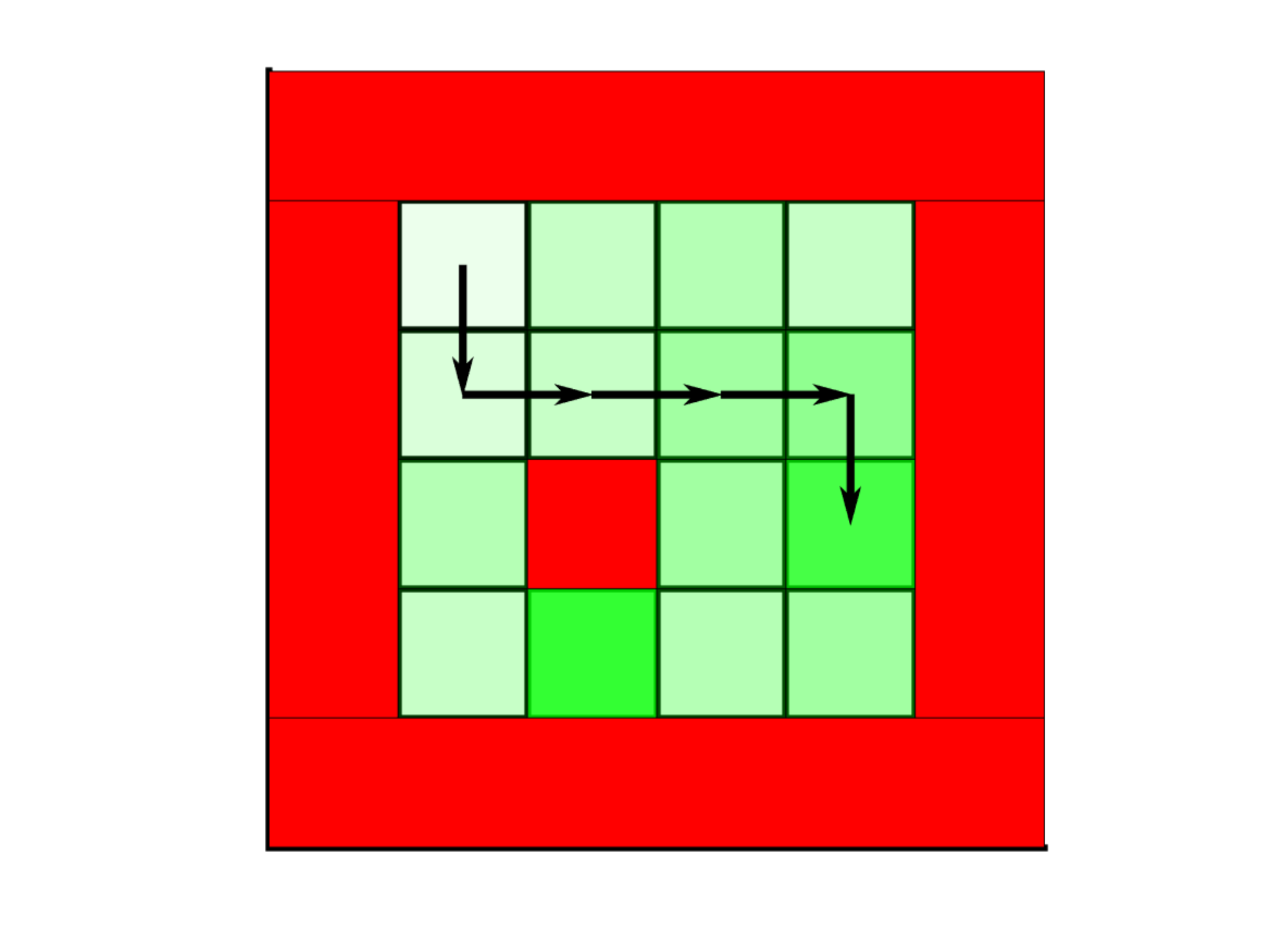}%
\label{fig::approximate_4x4}}
\caption{Comparison between Exact and Approximate Solutions (Reprinted from \cite{xiao2019explicit1})}
\label{fig::exact_approximate}
\end{figure}

Fig. \ref{fig::exact_approximate} shows the comparison between exact and approximate solutions when difference exists. Red cells indicate obstacles and greenness of the free cells represents the reward values. The darker the greenness is, the higher the reward value. The robot starts at the upper left corner and wants to find an optimal path and goal location in terms of utility, the ratio between reward and risk. As shown in Fig. \ref{fig::approximate_4x4}, the final suboptimal path is found as the minimum-risk path to the same goal location as found by the exact algorithm, but the optimal path (Fig. \ref{fig::exact_4x4}) is neglected by the approximate algorithm. Fig. \ref{fig::exact_approximate} indicates that it is actually worth the extra risk caused by the extra turn to collect more rewards along the path. But the approximate algorithm only looks at the ensemble of minimum-risk paths. 

\begin{figure}[]
\centering
\includegraphics[width=1\columnwidth]{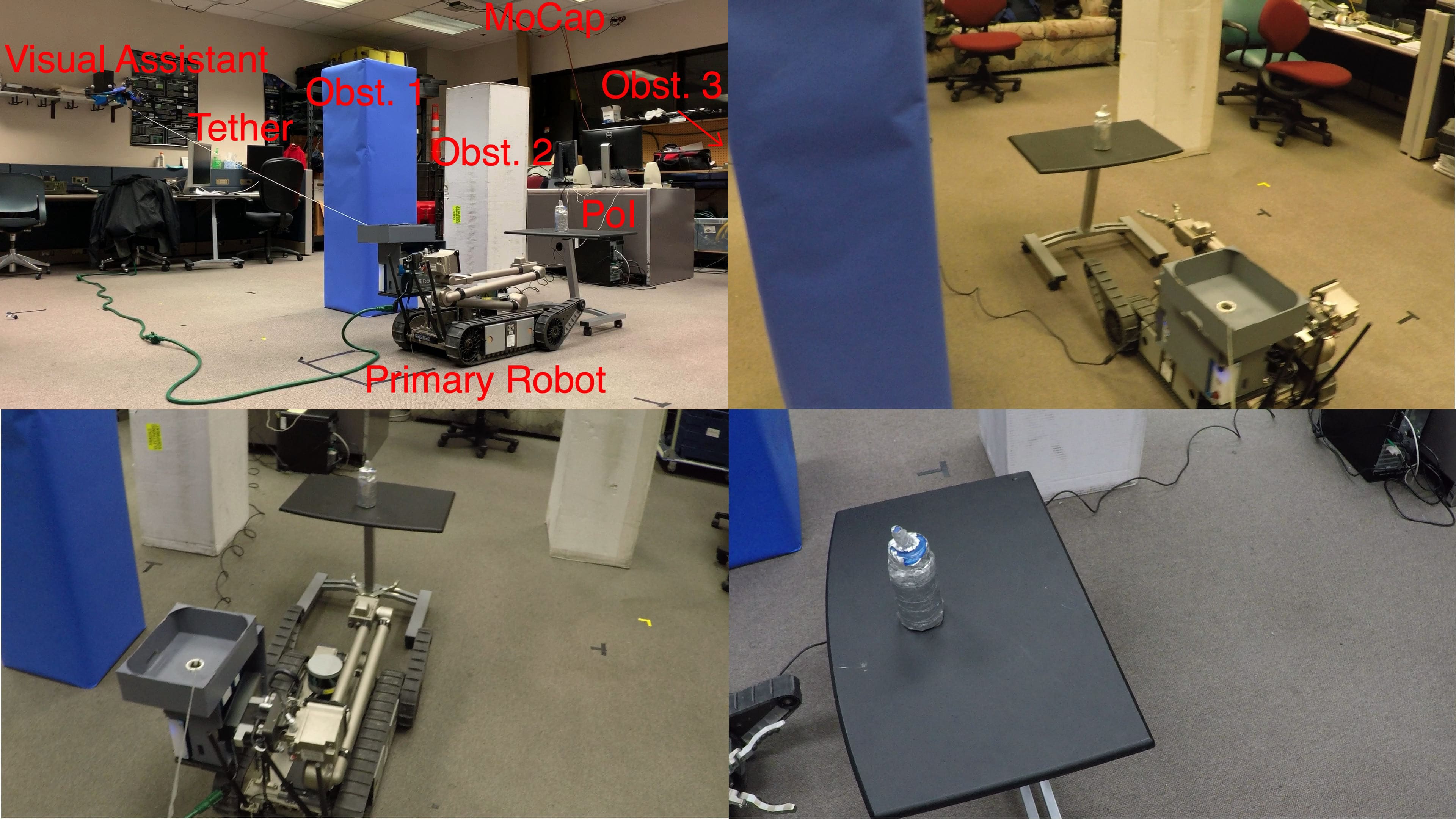}
\caption{Upper Left: Third person (external) view of the physical environment. Upper Right, Lower Left and Lower Right: Accumulated rewards in terms of visual assistance video feed (snapshots) along the entire risk-aware maximum-utility path in chronological order (reprinted from \cite{xiao2019explicit1}).}
\label{fig::all_snaps}
\end{figure}

The upper left figure in Fig. \ref{fig::all_snaps} shows a real world example with three obstacles in the workspace. For simplicity the workspace is limited into 2D at the UAV's nominal flight altitude (1.5m). The objective of the tethered aerial visual assistance is to provide good viewpoints toward a Point of Interest (PoI) in the middle, the sensor the primary robot is supposed to retrieve. 

\begin{figure}[]
\centering
\includegraphics[width=0.6\columnwidth]{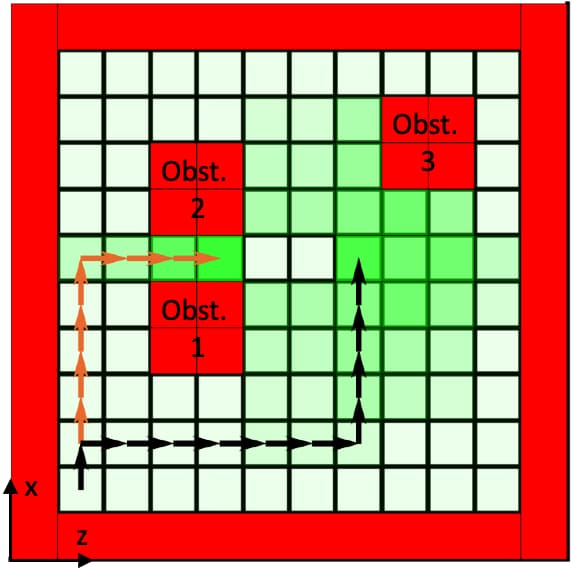}
\caption{Approximate Algorithm Solution in Real World: To observe the action taking place in the middle two white cells, the planner finds a path which minimizes all categories of risk elements and collects good rewards along the entire path. The orange path only aims at the best rewarding state but faces large risk (reprinted from \cite{xiao2019explicit1}).}
\label{fig::appriximate_example}
\end{figure}

The solution found by the approximate algorithm is shown in Fig. \ref{fig::appriximate_example}. 
The suboptimal path found by the approximate algorithm minimizes all three categories of risk elements by going through wide open spaces between obstacles and making as few turns and tether contacts as possible. The reward collected along the entire path is maximized simultaneously. The best viewpoints, shown in green between the two obstacles, do not worth to go to due to the risk of going through tight spaces. Although this example is shown in 2-D, this algorithm works in any dimensions with any vertex connectivity. 

\begin{figure}[]
\centering
\includegraphics[width=1\columnwidth]{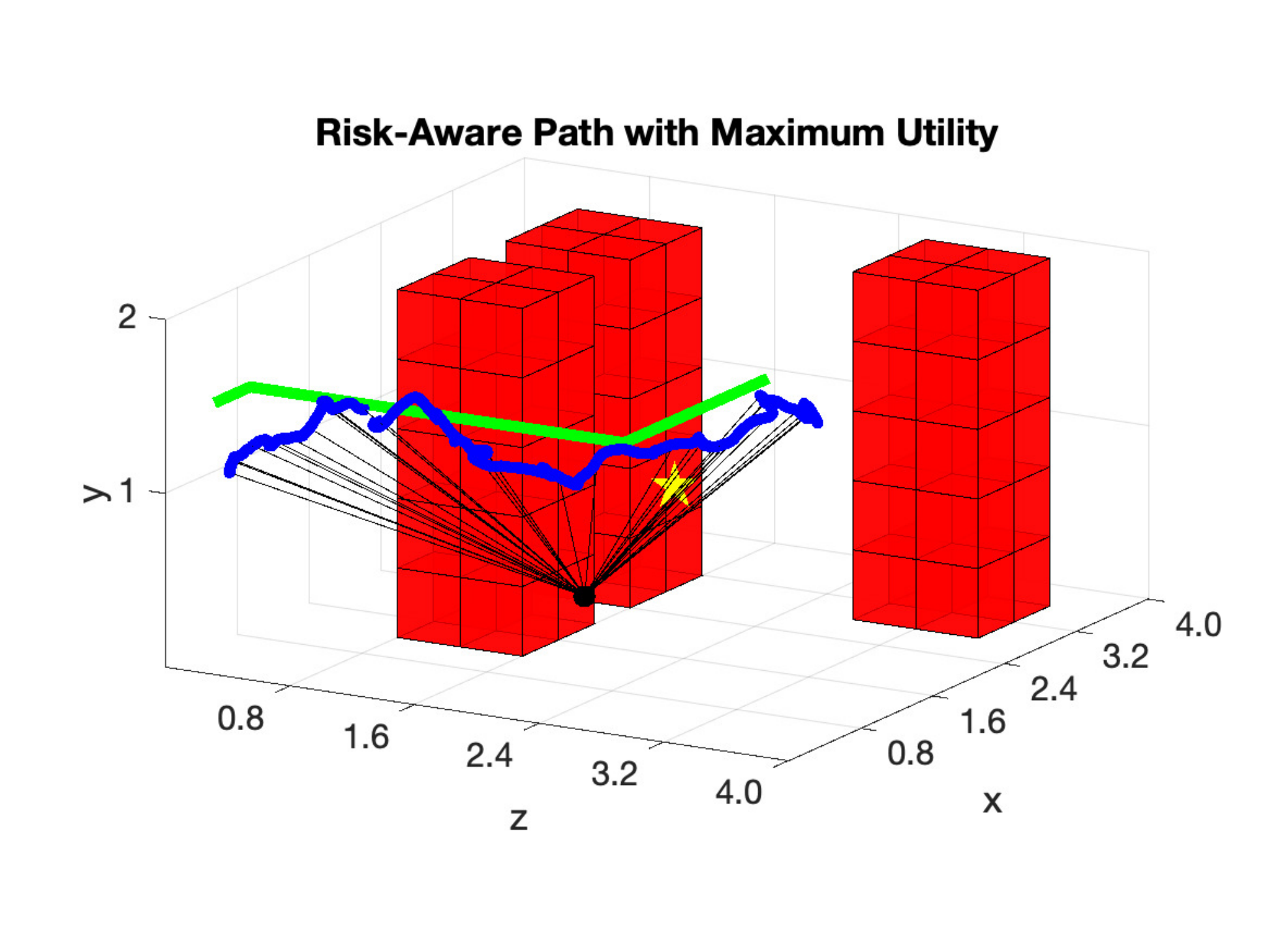}
\caption{Risk-aware Path with Maximum Utility Value Executed on a Physical Tethered UAV: Red voxels represent obstacles. Yellow star is the visual assistance PoI. The planned path is shown in green while the physically executed path in blue (reprinted from \cite{xiao2019explicit1}).}
\label{fig::experiment_approximate}
\end{figure}

The black path shown in Fig. \ref{fig::appriximate_example} is implemented on the tethered aerial visual assistant, Fotokite Pro, using the low level motion suite to be described in Chapter \ref{chapter::low_level}. The physical demonstration aims at showing the proposed risk-aware reward-maximizing planner being used on real robot, with its actually encountered motion risk in physical environments (Fig. \ref{fig::experiment_approximate}) and real-world collected reward in terms of visual assistance quality (Fig. \ref{fig::all_snaps}). 

The physical demonstration is conducted in a motion capture studio to ground-truth the visual assistant's actual motion. The studio is equipped with 6 OptiTrack Flex 13 cameras running at 120Hz. The 1280$\times$1024 high resolution cameras with a 56\degree~Field of View provide less than 0.5mm positional error and cover the whole 4$\times$4$\times$2m space. Although the original planner is shown in 2-D for easy illustration, the physical demonstration is conducted in 3-D space, with the same 3 obstacles distributed in the map. The mission for the teleoperated ground robot is to pick up a sensor in front with visual assistance from the tethered UAV. The visual Point of Interest (PoI) is therefore defined as the sensor, shown as the yellow star in Fig. \ref{fig::experiment_approximate}.

As shown in Fig. \ref{fig::appriximate_example}, the most rewarding state (best viewpoint) is to the left of the PoI (from ground robot's point of view). A traditional planner would plan a path leading to the optimal viewpoint (shown in orange). However, considering the fact that the best viewpoint locates between two obstacles and the path leading to it goes through the narrow passage between obstacles and map boundary (also treated as obstacle) and contains tether contact, it does not worth to take the risk. Our risk-aware reward-maximizing planner, on the other hand, could balance the trade-off between reward and risk. The approximate algorithm compares the utility value of the minimum-risk path leading to the optimal viewpoint with other candidate paths, and chooses the one with optimal utility among all minimum-risk paths. The planned path (green) and actual path (blue) in Fig. \ref{fig::experiment_approximate} maintain a maximum distance to closest obstacle and also a good visibility value and therefore a low locale-dependent risk along the way, while making only two turns with zero tether contact to minimize action-dependent and traverse-dependent risk. 

\section{Summary of High Level Risk-aware Path Planner}
Based on the proposed formal risk definition and explicit representation in Chapter \ref{chapter::risk_representation}, this chapter addresses the problem of risk-aware planning, i.e. how to plan minimum risk path given the proposed probabilistic risk framework. It also looks at how to maintain good mission reward along the path when reward and risk may be at odds. 

Motivated by the visual assistance problem, this chapter firstly defines and formulates it into a well-defined graph search problem. Using the risk framework in chapter \ref{chapter::risk_representation} and viewpoint reward from a separate study, the tradeoff is defined as the ratio between accumulated reward from all states  and encountered risk along the path. It is proved that the risk-aware reward-maximizing problem converted to a graph-search query is well-defined and an exact algorithm proposed. But only solutions for small-scale maps are practical in terms of computation time. This motivates the approximate algorithm, which is divided into upper stage risk-aware planner and lower stage reward maximizer. 

As one of the main contributions of this dissertation, the risk-aware planner finds minimum risk path based on the newly proposed risk definition and representation, as the probability of the robot not being able to finish the path. Adoption of this new risk definition and representation poses issues to the planner due to the problem's non-additivity and history dependency. These directly cause lost substructure optimality and conventional risk-aware planners cannot address risk without this kind of properties. The proposed risk-aware planner augments the existing vertex (state) into multiple directional components depending on the connectivity of the vertex and evaluates the path risk coming from those directions. In addition to the directional part, the planner also evaluates the risk dynamically based on the entire traverse from start, since the proposed risk is no longer additive and state-dependent. The two-step look-back of the proposed risk-aware planner allows it to plan optimally for locale-dependent and action-dependent risk elements. The correctness of the algorithm is proved by mathematical induction. However, traverse-dependent risk elements have full depth dependency on the history up to the start. Therefore their optimality cannot be guaranteed. If necessary, potential approaches of deeper history look-up is briefly discussed, but at the cost of exponentially increasing computation. In addition to the differences due to non-additivity and history dependency, the proposed risk-aware planner also looks for minimum risk path in an absolute sense, instead of a feasible path within a probability bound, such as a threshold of constraint violation chance. For example, Chance constraints based methods within either MDP or RMPC framework only found path with a failure probability (chance of constraint violation) less or equal to a certain manually defined threshold, which is difficult to determine in practice. When locomoting in unstructured or confined spaces, the most important objective is to safely finish the path, therefore a paradigm to find the absolutely minimum-risk path is of interest to robot path planning. The proposed risk-aware planner partially solves this problem by providing optimal path up to action-dependent risk elements, and points the directions to address deeper history dependency. 

The lower stage reward maximizing planner works on the ensemble of minimum risk paths provided by the upper stage risk-aware planner. It computes the utility value for each path as the ratio between collected reward and encountered risk. The suboptimal solution is chosen as the optimal utility path among all minimum risk paths. It prevents the planner from blindly going to the best rewarding state, but also taking other slightly less rewarding states into account. 

Until now, a high-level risk-aware reward-maximizing path is planned, using either the exact algorithm by brutal force to guarantee optimality or the approximate algorithm to be computationally tractable and scalable at the cost of sub-optimality. This high level path will be implemented on a tethered UAV in unstructured or confined environments, as discussed in the following chapter. 
\chapter{APPROACH: LOW LEVEL TETHERED MOTION SUITE}
\label{chapter::low_level}
Based on the formal risk definition and explicit representation, the high level risk-aware path planner could plan the path for the visual assistant so that motion risk is minimized, while considering the viewpoint quality reward. The risk-aware path takes form of an ordered sequence of 3D waypoints. In order to realize this risk-aware path on a tethered visual assistant UAV, this section introduces a suite of low level motion sensor, controllers, planners, executor, and servomechanism, which take advantage of the tether while minimizing the negative effect brought by the tether to UAV motion in unstructured or confined environments. The low level motion suite is composed of (1) a tether-based UAV localizer, which works in indoor-GPS denied environments and has negligible computational overhead, (2) two sets of different tether-based motion primitives to enable free flight of the tethered UAV in Cartesian space, (3) two different tether planning techniques that mitigate the disadvantage of tether in unstructured or confined environments, along with a motion executor that can handle tethered motion with contact points with the environments, and (4) a reactive visual servoing approach which can maintain a constant 6-DoF configuration of the visual assistant to a pre-defined target. 

\section{Tether-based Localization}
\begin{figure}[]
\centering
	\includegraphics[scale=0.3]{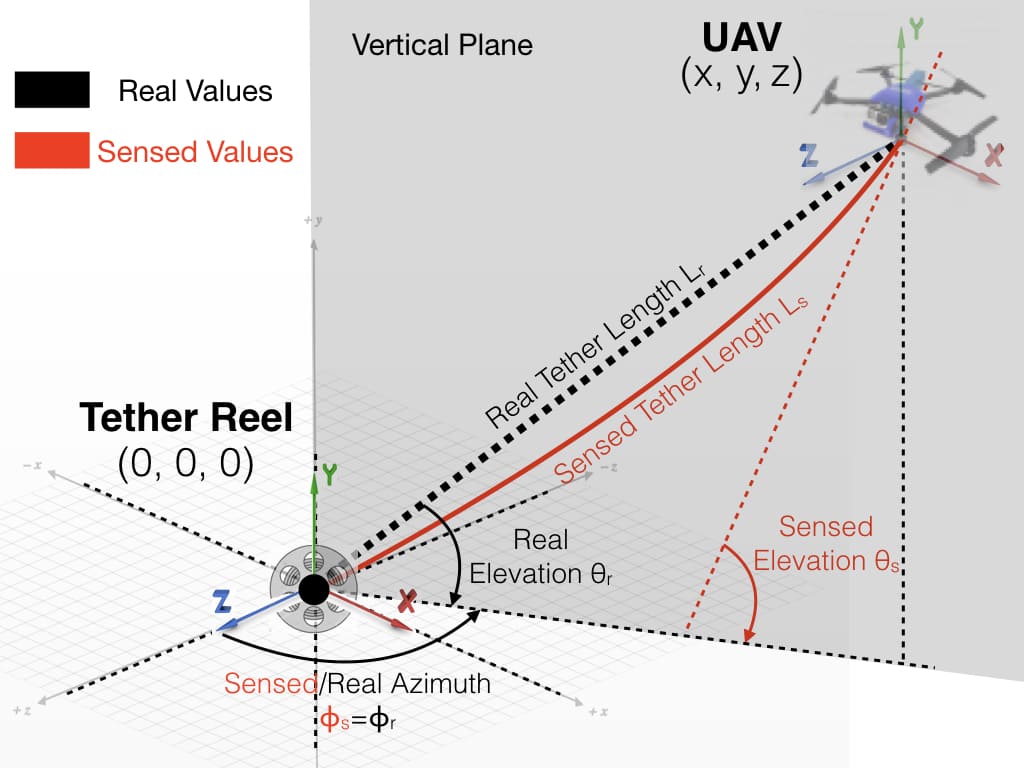}
	\caption{Tethered UAV is localized using tether-based sensory feedback including tether length, azimuth, and elevation angle. The preliminary localizer assumes sensed tether length, azimuth, and elevation angles are in fact the real values shown in black. In practice, this invalid assumption, shown in red, especially with longer tether, is compensated with a reasonable offset by a mechanics model (reprinted from \cite{xiao2018indoor}).}
	\label{fig::real_and_sensed}
\end{figure}

The proposed UGV/UAV team is expected to work in indoor GPS-denied unstructured or confined environments. The UAV is also assumed to be light weight and can not carry heavy computation payload for vision-based localization, such as visual odometry or SLAM. In this research, a new sensor modality for indoor localization of a tethered UAV is proposed, which utilizes tether-based feedback to avoid using GPS and vision-based sensing. The input for the localizer is tether length from tether reel encoder, tether azimuth (horizontal angle of tether) and elevation (vertical angle of tether) from a piezoelectric deformation tether angle sensor (Fig. \ref{fig::real_and_sensed}). This section will start with a preliminary localizer based on polar-to-Cartesian coordinate transformation and point out the problems. A mechanics model is then introduced to solve the problem and achieve more accurate localization.\footnote{This approach was discussed and published in previous work \cite{xiao2018indoor}.}

\subsection{Preliminary Localizer}
The preliminary localizer assumes the tether is always taut and straight, so the tether length, azimuth, and elevation values can take the real values shown in black in Fig. \ref{fig::real_and_sensed}. So the localization of the tethered UAV is just a transformation from polar coordinates to Cartesian space: 

\begin{equation}
\label{eqn::polar2euclidean}
\left\{\begin{matrix}
x =& L_r cos \theta_r sin \phi_r \\ 
y =& L_r sin \theta_r \\ 
z =& L_r cos \theta_r cos \phi_r
\end{matrix}\right.
\end{equation}

In Eqn. \ref{eqn::polar2euclidean}, $L_r$ is the imaginary real tether length, $\theta_r$ is the elevation angle, and $\phi_r$ is the azimuth angle. The preliminary localizer assumes that the tether is always taut and therefore straight, so the actual sensed values ($L_s$, $\theta_s$, $\phi_s$) from the sensors are in fact $L_r$, $\theta_r$, and $\phi_r$. Those sensed values are directly used in Eqn. \ref{eqn::polar2euclidean}.

However, this straight tether assumption does not hold all times, especially when the tether is long and forms an arc instead of a straight line due to increased gravitational force (shown in red in Fig. \ref{fig::real_and_sensed}). So the localization accuracy is deteriorated due to invalid straight tether assumption and the deterioration increases with increasing tether length. 

\subsection{Tether Deformation Model}
In order to solve the problem in the preliminary localizer, a tether deformation model is developed in order to compensate the error between the real and sensed values (Fig. \ref{fig::real_and_sensed}). We start with a free body diagram analysis, which calculates the force from tether pulling the UAV down (tether tension) using UAV configuration and sensory feedback. Based on that we further introduce our tether deformation model. The model takes calculated tether tension and sensed elevation angle as input, and outputs the real elevation angle and tether length. 

\subsubsection{Free Body Diagram}
\begin{figure}[]
\centering
\includegraphics[scale=0.3]{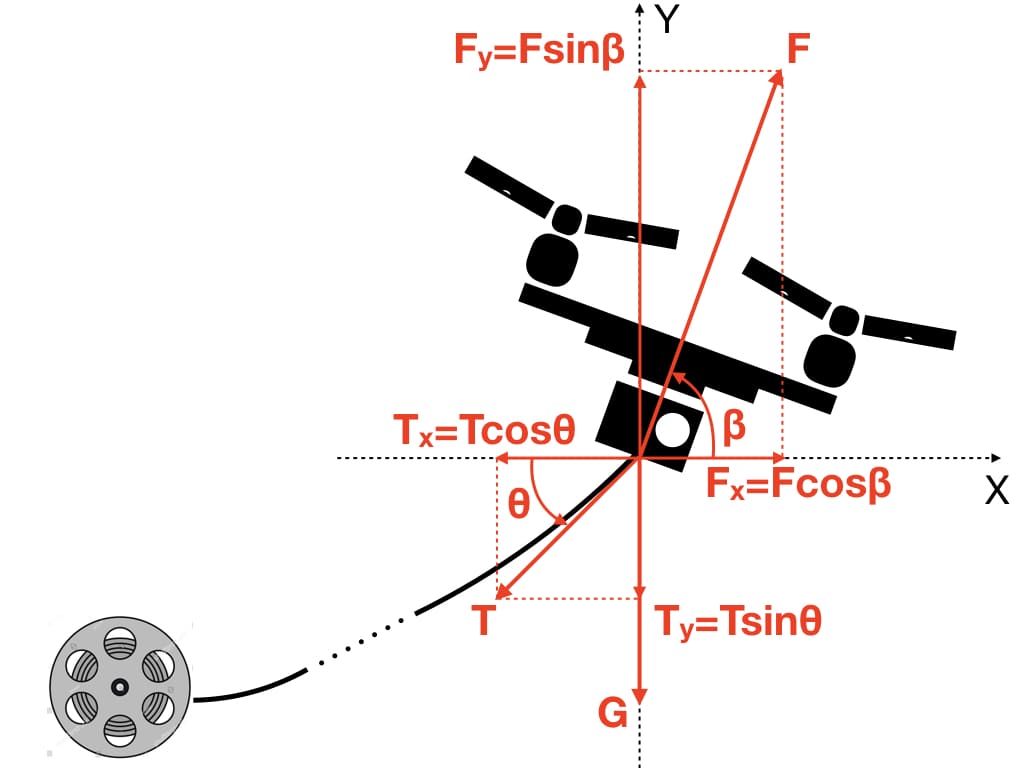}
\caption{Free Body Diagram of the Tethered UAV (Reprinted from \cite{xiao2018indoor})}
\label{fig::free_body_diagram}
\end{figure}

A UAV flying with a taut tether needs to hover with an angle with respect to the horizontal plane since the propellors need to provide a horizontal force to balance the horizontal component of tether tension acting on the UAV. As shown in Fig. \ref{fig::free_body_diagram}, we denote the force created by the propellers $F$, tether tension $T$, and the gravity of the UAV $G$. The angles of $F$ and $T$ with respect to the horizontal plane is denoted as $\beta$ and $\theta$, respectively. $\theta$ is simply the sensed elevation angle described above. 

To balance both $x$ and $y$ directions, we have
\begin{equation}
\label{eqn::free_body}
\left\{\begin{matrix}
Fcos\beta=Tcos\theta \\
Fsin\beta = Tsin\theta+G
\end{matrix}\right.
\end{equation}

$\beta$ is available from UAV onboard IMU and $\theta$ from tether angle sensor. By solving the equations, we can calculate $F$ and $T$:
\begin{equation}
\label{eqn::F_and_T}
\left\{\begin{matrix}
F=\frac{G}{sin\beta-tan\theta cos\beta}\\
T = \frac{Gcos\beta}{sin\beta cos\theta-tan\theta cos\theta cos\beta}
\end{matrix}\right.
\end{equation}

\subsubsection{Mechanics Model}
In order to describe the shape of the tether, catenary curve is considered. The catenary curve was first introduced by Leibniz, Huygens and Johann Bernoulli in 1691 and it has been widely used in predicting geometric response of hanging ropes, chains and cables under the force of gravity. Assuming the gravity is uniform, and the two free ends of the tether are hanged on the same altitude. Based on symmetry, only half of the catenary is needed for analysis. 

\begin{figure}[]
\centering
\includegraphics[scale=0.4]{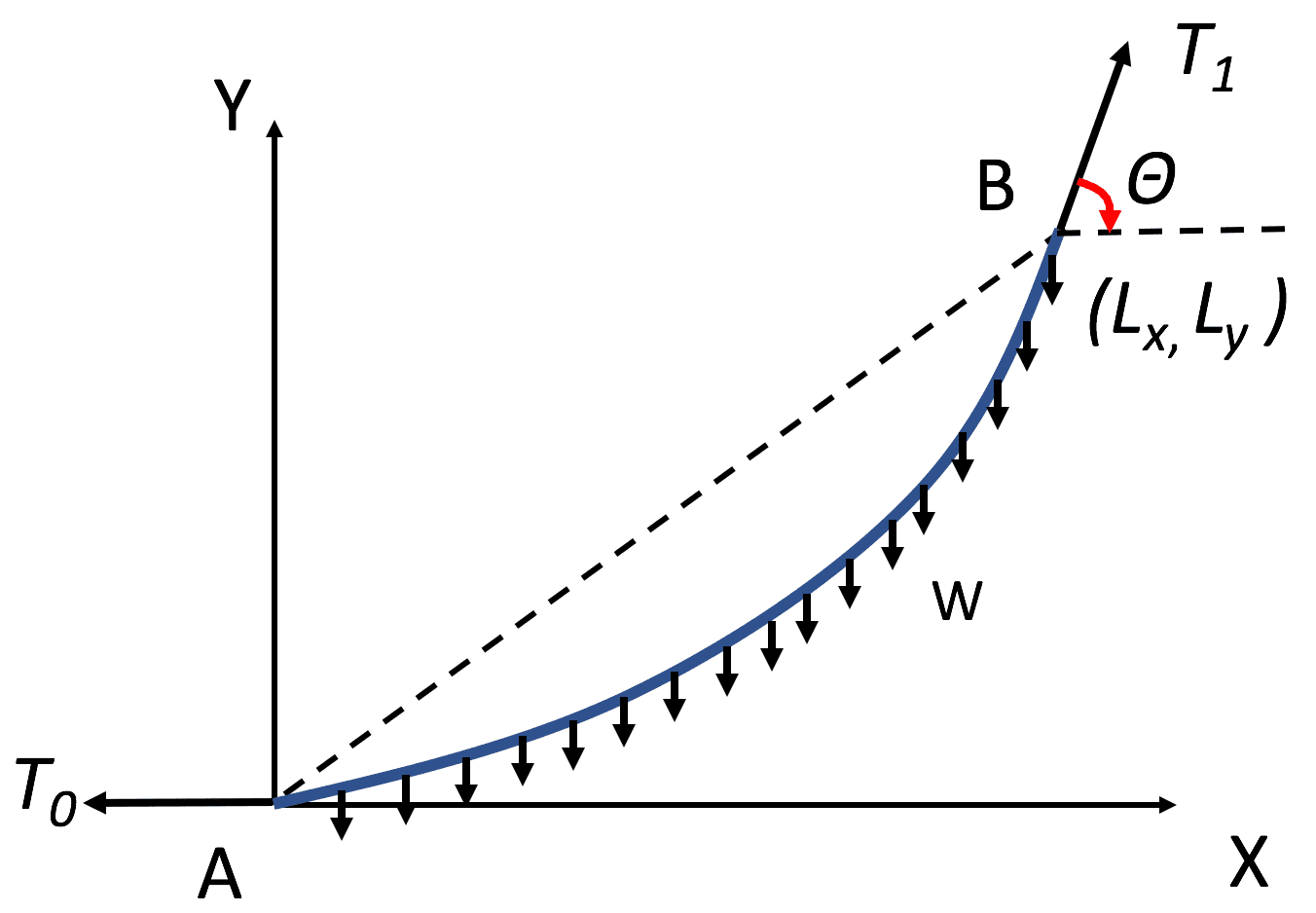}
\caption{Free Body Diagram of UAV Tether (Reprinted from \cite{xiao2018indoor})}
\label{fig::catenary}
\end{figure}

Fig. \ref{fig::catenary} shows the free body diagram of the tether. Point B is the UAV end where the tether is suspended, while point A is the axisymmetric end. The tether is subject to gravity $W$, which is assumed to be uniformly distributed as shown. The tension acting on the tether is noted as $T_0$ and $T_1$, where $T_0$ is at end A and $T_1$ at end B. $T_1$ is the reaction force of $T$ acting on the UAV, as mentioned in the last free body diagram, so they have same magnitude but opposite directions. By applying equilibrium equations to the tether, we can get:

\begin{equation}
\label{eqn::catenary_eqn_1}
\left\{\begin{matrix}
\Sigma F_x = 0 \quad \Rightarrow \quad T_1 cos\theta - T_0 = 0\\
\Sigma F_y = 0 \quad \Rightarrow \quad T_1 sin\theta - W = 0
\end{matrix}\right.
\end{equation}

where $\theta$ is the same departure angle of tension $T_1$ with respect to the horizontal axis, which we can measure through tether angle sensor. By rewriting Eqn. \ref{eqn::catenary_eqn_1} we have: 

\begin{equation}
\label{eqn::catenary_eqn_2}
\left\{\begin{matrix}
T_1 cos\theta &=&T_0\\
T_1 sin\theta &=&\rho Lg
\end{matrix}\right.
\end{equation}

where $\rho$ is the linear density of the tether, $g$ is the gravitational acceleration and $L$ is the total tether length. 

\begin{figure}[]
\centering
\includegraphics[scale=0.4]{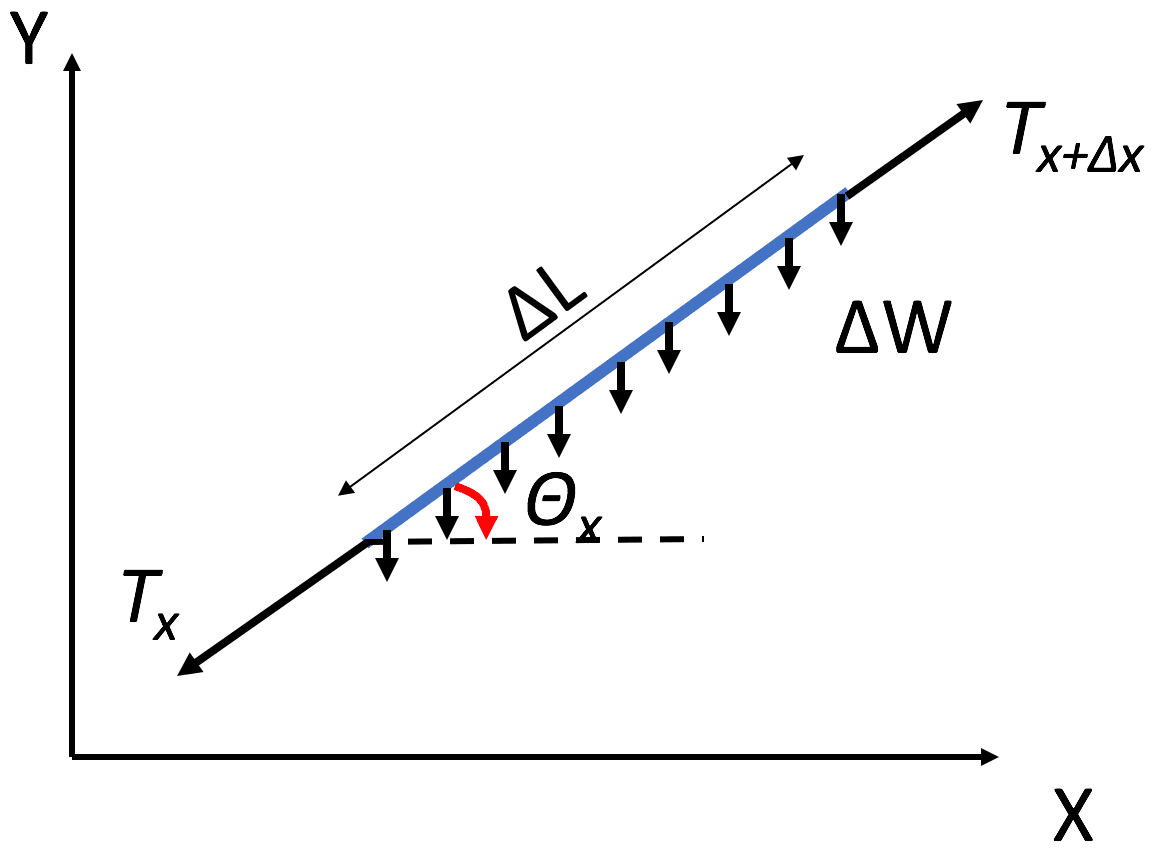}
\caption{Free Body Diagram of Tether Segment (Reprinted from \cite{xiao2018indoor})}
\label{fig::catenary_segment}
\end{figure}

A closer look into the free body diagram of one small piece of tether segment is shown in Fig. \ref{fig::catenary_segment}. The equilibrium equations are: 

\begin{equation}
\label{eqn::catenary_eqn_3}
\left\{\begin{matrix}
\Sigma F_x = 0  \Rightarrow&  T_{x+\Delta x} cos\theta_{x+\Delta x} - T_xcos\theta_x &=& 0\\
\Sigma F_y = 0  \Rightarrow&  T_{x+\Delta x} sin\theta_{x+\Delta x} - T_xsin\theta_x-\Delta W &=& 0
\end{matrix}\right.
\end{equation}

where $T_x$ and $T_{x+\Delta x}$ are the tension and $\theta_x$ and $\theta_{x+\Delta x}$ the angle with respect to the horizontal axis at both ends of the segment. For the second equation in Eqn. \ref{eqn::catenary_eqn_3}, move $\Delta W$ to the right hand side and divide both sides by $\Delta x$, the left hand side is the definition of derivation of $T_xsin\theta_x$: 

\begin{equation}
\label{eqn::catenary_eqn_4}
\frac{d}{dx}(T_xsin\theta_x) = \frac{d}{dx}W_x
\end{equation}

$T_x$ can be expressed as a function of $T_0$ and $\theta_x$:

\begin{equation}
\label{eqn::catenary_eqn_5}
T_x = \frac{T_0}{cos\theta_x}
\end{equation}

The geometry of the tether segment gives us:

\begin{equation}
\label{eqn::catenary_eqn_6}
\left\{\begin{matrix}
dL=\sqrt{d_x^2+d_y^2}=\sqrt{1+(\frac{dy}{dx})^2}dx\\
tan\theta_x= \frac{dy}{dx}
\end{matrix}\right.
\end{equation}

Substituting $T_x$ in Eqn. \ref{eqn::catenary_eqn_4} with Eqn. \ref{eqn::catenary_eqn_5}, we get: 

\begin{equation}
\label{eqn::catenary_eqn_7}
\frac{d}{dx}(T_0tan\theta_x) = \frac{d}{dx}W_x = \rho gdL
\end{equation}

And by substituting Eqn. \ref{eqn::catenary_eqn_6} into Eqn. \ref{eqn::catenary_eqn_7}, we have:

\begin{equation}
\label{eqn::catenary_eqn_8}
\frac{d}{dx}(\frac{dy}{dx}) = \frac{\rho g}{T_0}\sqrt{1+(\frac{dy}{dx})^2}dx
\end{equation}

The solution to Eqn. \ref{eqn::catenary_eqn_8} is the catenary curve and can be expressed as: 

\begin{equation}
\label{eqn::catenary_eqn_9}
y = a cosh\frac{x}{a}
\end{equation}

where $a=\frac{T_0}{\rho g}$ is a coefficient that depends on tension $T_0$, tether linear density $\rho$, and gravitational acceleration $g$. In order to get the coordinate of the UAV end, we take the derivative of Eqn. \ref{eqn::catenary_eqn_9}:

\begin{equation}
\label{eqn::catenary_eqn_10}
\frac{dy}{dx}=sinh\frac{x}{a}
\end{equation}

By comparing Eqn. \ref{eqn::catenary_eqn_10} and Eqn. \ref{eqn::catenary_eqn_6}, we get:

\begin{equation}
\label{eqn::catenary_eqn_11}
tan\theta_x=sinh\frac{x}{a}
\end{equation}

Eqn. \ref{eqn::catenary_eqn_11} is a general form and specifically at the UAV end B, $\theta_x$ is the departure angle $\theta$, and $x$ coordinate is equal to $L_x$:

\begin{equation}
\label{eqn::catenary_eqn_12}
tan\theta=sinh\frac{L_x}{a}
\end{equation}

Based on Eqn. \ref{eqn::catenary_eqn_12} and catenary curve, the $x$ and $y$ coordinates of the UAV end B takes the form:

\begin{equation}
\label{eqn::catenary_eqn_13}
\left\{\begin{matrix}
L_x =& aln(tan\theta+\sqrt{tan^2\theta+1})\\
L_y =& a cosh\frac{L_x}{a}-acosh0
\end{matrix}\right.
\end{equation}

The real elevation angle $\theta_r$ and tether length $L_r$ would be corrected as:

\begin{equation}
\label{eqn::catenary_eqn_13}
\left\{\begin{matrix}
\theta_r =& atan(\frac{L_y}{L_x})\\
L_r =& \sqrt{L_x^2+L_y^2}
\end{matrix}\right.
\end{equation}

while real azimuth angle $\phi_r$ is still equal to sensed value $\phi_s$. Using arc length equation, we could compute the actual sensed length of the curved tether $L_s$: 

\begin{equation}
\label{eqn::catenary_eqn_14}
\begin{split}
L_s =& \int_0^{L_x}\sqrt{1+f'(x)^2}dx\\
   =& \int_0^{L_x}\sqrt{1+sinh^2\frac{x}{a}}dx\\
   =& asinh\frac{L_x}{a}
\end{split}
\end{equation}

\subsection{Implementation}
The mechanics model described above is implemented on the tethered aerial visual assistant, Fotokite Pro, with its provided Software Development Kit (SDK).  Details about the implementation of the proposed localization method are discussed. 

\subsubsection{Sensory Input}
\textbf{Tether Length $L_s$}
Using the encoder reading of the tether reel from the SDK, we are able to calculate the relative tether length with respect to its initial state. If we initialize tether length to be zero, the absolute tether length equals to the relative value. 

\textbf{Tether Angles $\theta_s$ and $\phi_s$}
The SDK also provides two tether angle measurements, azimuth and elevation, measured directly from the attachment point of the tether to the vehicle. Elevation angle is with respect to gravity and azimuth to initialization. We convert the elevation angle based on our definition with respect to horizontal plane. 

\textbf{Vehicular Lean Angle $\beta$}
The UAV configuration is given in the form of a quaternion $\bf{z} = a + b \bf{i} + c \bf{j} + d \bf{k}$. The normal vector pointing up $\bf{n_v} = [0, 1, 0]^T$ expressed in the vehicle frame could be transformed to the global frame by multiplying the corresponding rotation matrix $R$ of the quaternion: 

\begin{equation}
\label{quaternion}
n_g = [x_g, y_g, z_g]^T = R*n_v
\end{equation}

So the lean angle is 

\begin{equation}
\label{lean}
\beta = arcsin(\frac{y_g}{\sqrt{x_g^2+y_g^2+z_g^2}})
\end{equation}

\subsubsection{Length and Angle Correction}
We take the sensed elevation angle $\theta_s$ and computed lean angle $\beta$ to compute tension $T=T_0$ using Eqn. \ref{eqn::F_and_T}. Here the UAV weighs 6N. Then we feed them into the model described above. Tether linear density $\rho$ is measured to be 0.0061kg/m. Eqn. \ref{eqn::catenary_eqn_13} gives the final corrected elevation angle $\theta_r$. The control for tether length need to be based on the arc length $S$ (Eqn. \ref{eqn::catenary_eqn_14}). Azimuth angle remains the same.

\subsubsection{Navigation}
In order to navigate the UAV to target point $(x, y, z)$, we compute the desired elevation $\theta_d$ and azimuth $\phi_d$ values by: 

\begin{equation}
\label{eqn::euclidean2polar}
\left\{\begin{matrix}
\theta_d = &arcsin(\frac{y}{\sqrt{x^2+y^2+z^2}})\\
\phi_d = &atan2(\frac{x}{z})
\end{matrix}\right.
\end{equation}

The desired tether arc length $L_d$ is given by Eqn. \ref{eqn::catenary_eqn_14}. By comparing the desired values with corrected current sensory input, three individual positional PID controllers are implemented to drive $L_r$, $\theta_r$, and $\phi_r$ to their desired values, which will be discussed in detail in next section and compared with velocity control. 

\subsection{Summary of Tether-based Localization}
This section presents a novel indoor localization scheme for UAVs operating with a quasi-straight tether. This localizer is based on polar-to-Cartesian coordinates conversion and uses tether sensory information including tether length, elevation and azimuth angles. More importantly, a mechanics model is built to quantify the inevitable tether deformation when the tether is long and pulled down by gravity and therefore forming an arc instead of an ideal straight line. This model is expected to be capable of correcting the measured elevation angle and tether length and thus improve localization accuracy. The localization accuracy will no longer depend on tether length. The experiments to validate this hypothesis will be presented in Chapter \ref{chapter::experiments}.  

\section{Tether-based Motion Primitives}
With an improved indoor localizer with very little computation necessary, this section proposes two different tether-based UAV motion primitives, reactive feed-back based position control and model-predictive feedforward velocity control. Both controllers translate the motion commands in Cartesian space ($x$, $y$, $z$) into tether-based coordinates (tether length, azimuth, elevation).\footnote{This approach was discussed and published in previous work \cite{xiao2019benchmarking}.}

The path plan is given by any type of high-level path planner, such as the risk-aware planner discussed in Chapter \ref{chapter::high_level}, in the form of an ordered sequence of 3D waypoints. The execution of the path is to navigate the tethered UAV along this waypoint sequence in order. \cite{lupashin2013stabilization} presented the controller for elevation and azimuth angles, while tether length is regulated by the tether reel motor. This provides us with our controller input for both motion primitives: change rate of tether length $\dot{L}$, elevation $\dot{\theta}$, and azimuth $\dot{\phi}$. The tethered UAV is by default stabilized around a new equilibrium with regard to a taut tether. The feedback from the tether includes the length computed by tether reel motor encoder, tether elevation and azimuth angle perceived by the piezoelectric deformation sensor mounted on the connecting point between tether and UAV. Using the localizer presented above, sensed tether-based feedback is corrected and therefore the real values are treated equally as sensed values. This is our sensory feedback of the system: $L_{s}$, $\theta_{s}$, and $\phi_{s}$. 

\subsection{Position Control}
In the position control, we want to utilize the control over change rate of tether length ($\dot{L}$), elevation angle ($\dot{\theta}$), and azimuth angle ($\dot{\phi}$) to realize UAV airframe translational motion in terms of position in 3D Cartesian space using the onboard position feedback. 

The transformation from the tether-based coordinates to Cartesian coordinates is expressed in Eqn. \ref{eqn::polar2euclidean}. The inverse mapping could be easily derived and gives us the desired tether variables: 

\begin{equation}
\label{eqn::eucledean2polar}
\left\{\begin{matrix}
L_d =& \sqrt{x^2+y^2+z^2}\\ 
\theta_d =& arcsin\frac{y}{\sqrt{x^2+y^2+z^2}}\\ 
\phi_d = &atan2(\frac{x}{z})
\end{matrix}\right.
\end{equation}

Given a 3D waypoint on a pre-defined path, Eqn. \ref{eqn::eucledean2polar} maps the Cartesian $x$, $y$, and $z$ values into tethered-based $L$, $\theta$, and $\phi$ values. For position control, three independent PD controllers use the three tether input to drive the tether into the desired configuration. Let $e_L$, $e_\theta$, and $e_\phi$ to be the error between desired and sensed value of the three tether variables:

\begin{equation}
\label{eqn::pos_errors}
\vv{\bm{e}}_{\bm{(L, \theta, \phi)}} = \begin{bmatrix} e_L, e_\theta, e_\phi\end{bmatrix}^T= \begin{bmatrix} L_d, \theta_d, \phi_d\end{bmatrix}^T - \begin{bmatrix} L_s, \theta_s, \phi_s\end{bmatrix}^T
\end{equation}

Our control variable $\vv{\bm{u}} = \begin{bmatrix}\dot{L}, \dot{\theta}, \dot{\phi}\end{bmatrix}^T$ are computed by 

\begin{equation}
\label{eqn::pos_controls}
\vv{\bm{u}}
=\vv{\bm{K}}_{\bm{P}}
\vv{\bm{e}}_{\bm{(L, \theta, \phi)}}
+
\vv{\bm{K}}_{\bm{D}}
\dot{\vv{\bm{e}}}_{\bm{(L, \theta, \phi)}}
\end{equation}

where $\vv{\bm{K}}_{\bm{P}}$ and $\vv{\bm{K}}_{\bm{D}}$ are the corresponding proportional and derivative gains: 

\begin{equation}
\label{eqn::kp}
\vv{\bm{K}}_{\bm{P(D)}}=
\begin{bmatrix}
K_{P(D)L}\quad K_{P(D)\theta} \quad K_{P(D)\phi} 
\end{bmatrix}
\end{equation}

Applying $\vv{\bm{u}}$ based on error feedback $\vv{\bm{e}}_{\bm{(L, \theta, \phi)}}$, the system is driven to the desired values. When an acceptance radius is reached around a certain waypoint, the position controller moves on to the next waypoint until the whole sequences is finished. 

\subsection{Velocity Control}
Given the fact that the three PD controllers work independently, it is expected that the position control will achieve unpredictable motion between waypoints. With this in mind, velocity control is proposed to achieve smoother and straighter motion. Based on Eqn. \ref{eqn::polar2euclidean}, the Jacobian matrix of the system could be derived: 

\begin{equation}
\label{eqn::jacobian}
\vv{\bm{\dot{x}}}
=\bm{J} 
\vv{\bm{u}}
\end{equation}

where $\vv{\bm{\dot{x}}}=\begin{bmatrix} \frac{dx}{dt}, \frac{dy}{dt}, \frac{dz}{dt} \end{bmatrix}^T$, $\vv{\bm{u}} = \begin{bmatrix}\dot{L}, \dot{\theta}, \dot{\phi}\end{bmatrix}^T$, and 

\begin{equation}
\label{eqn::j}
\bm{J} = 
\begin{pmatrix}
cos\theta sin\phi & -Lsin\theta sin\phi & Lcos\theta cos\phi\\ 
sin\theta & Lcos\theta & 0\\ 
cos\theta cos\phi & -Lsin\theta cos\phi & -Lcos\theta sin\phi
\end{pmatrix}
\end{equation}

The velocity vector $\vv{\bm{\dot{x}}}$ could be computed by a vector pointing from the current sensed position $\begin{bmatrix} x_s, y_s, z_y \end{bmatrix}^T$ (Eqn. \ref{eqn::polar2euclidean}) to the desired waypoint $\begin{bmatrix} x_d, y_d, z_d \end{bmatrix}^T$: 

\begin{equation}
\label{eqn::norm}
\vv{\bm{\dot{x}}} = 
\alpha \frac{\begin{bmatrix} x_d, y_d, z_d \end{bmatrix}^T - \begin{bmatrix} x_s, y_s, z_y \end{bmatrix}^T}{ \lVert \begin{bmatrix} x_d, y_d, z_d \end{bmatrix}^T - \begin{bmatrix} x_s, y_s, z_y \end{bmatrix}^T \rVert}
\end{equation}

where $\alpha$ is a scalar constant defining the length of the vector, or the absolute speed value of the UAV. So the input $\vv{\bm{u}}$ could be computed by: 

\begin{equation}
\label{eqn::jacobian2}
\vv{\bm{u}}
=\bm{J}^{-1} \vv{\bm{\dot{x}}}
\end{equation}

Velocity control aims at the current desired waypoint from the current position at every single time step, and the three control variables in $\vv{\bm{u}}$ are coupled to assure smooth and straight motion. However, when $\theta = 90^\circ$, the Jacobian loses rank and singularity occurs. In fact, even for manual control, the tethered UAV can hardly fly right across the top of the tether reel ($\theta = 90^\circ$). Therefore, $\theta = 90^\circ$ should be avoided when using velocity control.

\subsection{Summary of Tether-based Motion Primitives}
This section presents two tether-based motion primitives to enable autonomous tethered UAV motion given pre-computed path plans. The two motion primitives are either based on three independent PID controllers or the system's inverse Jacobian matrix to compute control commands in the form of change rate of tether length, elevation, and azimuth angles. Both motion primitives are expected to be able to translate 3D motion in Cartesian space into tether-based motion commands, and therefore realize free-flight in 3D Cartesian space on a tethered UAV. The experiments to validate this hypothesis will be presented in Chapter \ref{chapter::experiments}.

\section{Tether Planning and Motion Execution}
This section proposes two different methods to handle the existence of tether during flight in unstructured or confined environments. Tether planning is necessary since unlike free-flying UAVs, in tethered flight obstacles may come in the way of not only the UAV itself, but also its tether. The first method maintains a straight tether all the time and does not allow tether contact with the environment. The second one plans tether contact point(s) and relaxes unnecessary one(s) depending on planned flight trajectory. It also presents the motion executor to execute tethered flight with (or without) the existence of contact points on physical tethered UAV.\footnote{These two tether-handling approaches were discussed and published in previous work \cite{xiao2018motion}.}

\subsection{Reachable Space Reduction via Ray Casting}
In contrast to conventional UAVs, tethered UAVs have to maintain connection with its ground station via a tether. If the tether needs to remain taut and straight, no contacts are allowed with the environment. This constraint reduces the reachable space. Obstacles cannot locate between the UAV and its tether reel, since otherwise the tether in between would touch the obstacles. Based on this idea, this planner uses a ray casting approach from tether reel to obstacles in order to identify spaces in the configuration space, which are feasible for the UAV alone, but not with a tether. For voxels on the ray, those between reel center and obstacles are still open, while those beyond obstacles are blocked. 

\begin{figure}[]
\centering
\includegraphics[width=0.9\columnwidth]{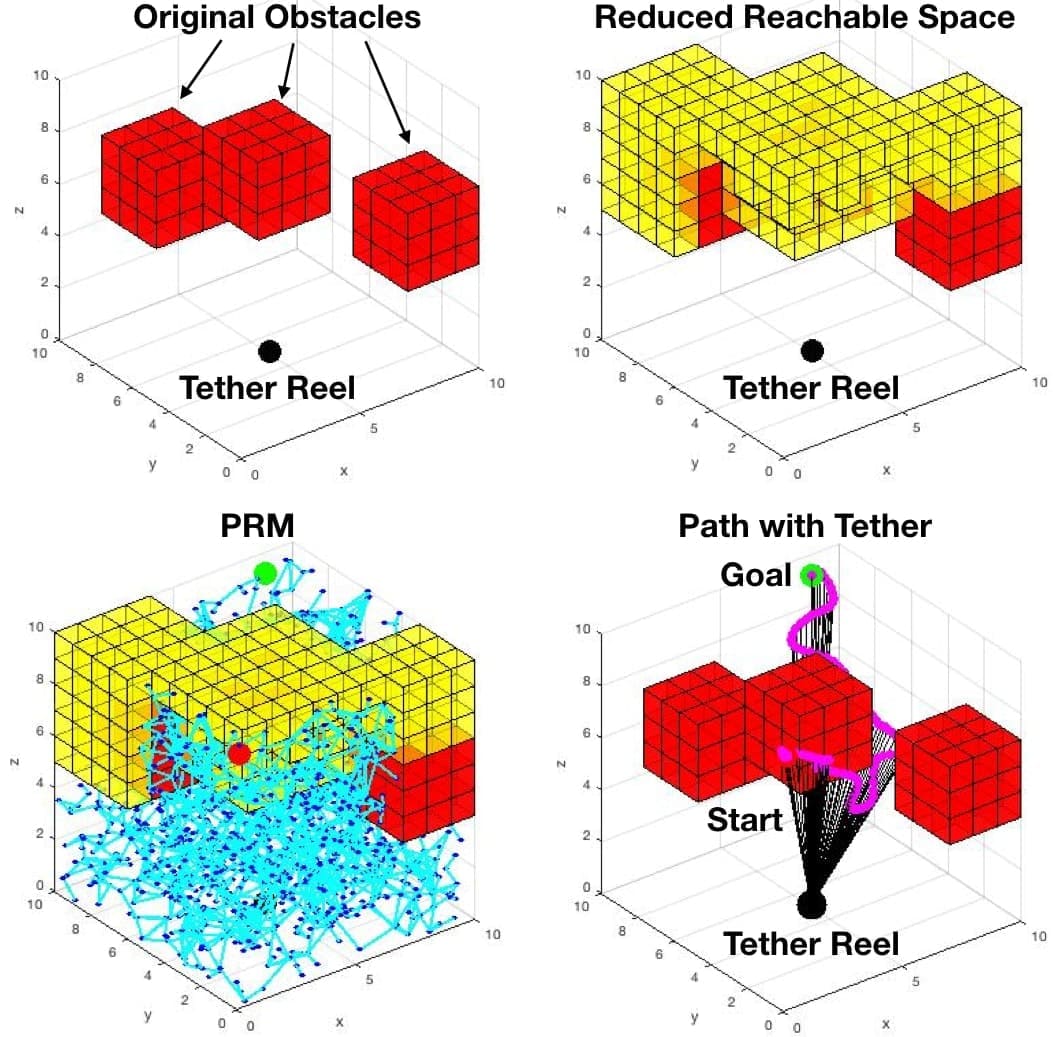}
\caption{Reachable space is reduced by ray casting from tether reel to original obstacles (in red). Yellow voxels are non-reachable space due to tether. UAV path is planned using PRM (cyan) in the reduced reachable space. The UAV has to go beneath the obstacles since a direct straight path above all obstacles is not allowed by the existence of tether (reprinted from \cite{xiao2018motion}).}
\label{fig::ray_casting}
\end{figure}

After reachable space reduction, remaining space is completely free even with respect to the tether. In this research, we could use this reduced map as input to our high level path planner (Chapter \ref{chapter::high_level}), which then plans an executable path in the reduced reachable space. An illustration of reachable space reduction from original map is shown in Fig. \ref{fig::ray_casting}. In fact, any path planning algorithm which works in 3D could be applied between any two points in the reduced space. Here, Probabilistic Road Map (PRM) \cite{kavraki1996probabilistic} is used to plan an executable path in the reduced space. 

Ray casting has a complexity proportional to the number of obstacles $\mathcal{O}(o)$, which is the only extra complexity introduced in addition to the high-level planner. It prepares the map in advance for the high-level planner. The complete algorithm pipeline is illustrated in Fig. \ref{fig::ray_casting}. 

\subsection{Contact(s) Planning and Relaxation}
As we can see in Fig. \ref{fig::ray_casting}, the free space is largely reduced by ray casting, leaving only a subset of the original free space reachable with a tether. A different algorithm with tether contact point(s) planning is presented here. It automatically plans the tether contact point(s) when the robot locates in the originally free but actually occupied spaces after reduction (yellow voxels in Fig. \ref{fig::ray_casting}). It also has the capability to relax the contact point(s) when the UAV returns to the post-reduction free space. It assumes that once a contact point is formed, it doesn't move unless being relaxed. The algorithm is outlined in Algorithm \ref{alg::contact_planning}.

\begin{algorithm}[]
 \caption{Contact Point(s) Planning and Relaxation}
 \begin{algorithmic}[1]
 \renewcommand{\algorithmicrequire}{\textbf{Input:}}
 \renewcommand{\algorithmicensure}{\textbf{Output:}}
 \REQUIRE \textit{map}, \textit{path}, \textit{tether\_origin}
 \ENSURE  executable motion plan: waypoints with contact points
  \STATE Initialize \textit{CP\_stack} with \textit{tether\_origin}
  \STATE Attach \textit{tether\_origin} to all waypoints \textit{WP}s on \textit{path}
  \STATE \textit{relax\_flag} = 0 
  \FOR {every \textit{WP} on \textit{path}}
  \STATE \textit{curent\_contact} = \textit{CP\_stack} top \textit{CP}
  \IF {(\textit{CP\_stack} has more than one CPs)}
  \STATE \textit{last\_contact} = second \textit{CP} from \textit{CP\_stack} top
  \STATE \textit{collision\_flag} = CheckCollision (\textit{last\_contact}, \textit{WP}, \textit{map})
  \IF {\textit{collision\_flag} == 0}
  \IF {ObstacleConfined (\textit{curent\_contact}, \textit{last\_contact}, \textit{WP}, \textit{map})}
  \STATE \textit{relax\_flag} = 0
  \ELSE
  \STATE \textit{relax\_flag} = 1 // contact relaxation
  \ENDIF
  \ELSIF {\textit{collision\_flag} == 1} 
  \STATE \textit{relax\_flag} = 0
  \ENDIF
  \ENDIF
  \IF {\textit{relax\_flag} == 1}
  \STATE pop \textit{CP\_stack}
  \STATE attach new \textit{CP\_stack} top to all following \textit{WP}s
  \STATE  \textit{relax\_flag} = 0
  \ELSIF {\textit{relax\_flag} = 0}
  \IF {CheckCollision (\textit{current\_contact}, \textit{WP}, \textit{map})}
  \STATE push new \textit{CP} to \textit{CP\_stack} // contact planning 
  \STATE attach new \textit{CP\_stack} top to all following \textit{WP}s
  \ENDIF
  \ENDIF
  \ENDFOR
 \RETURN all \textit{WP}s along with their \textit{CP}s
 \end{algorithmic}
 \label{alg::contact_planning}
 \end{algorithm}
 
 \begin{figure}[]
\centering
\subfloat[Original Configuration Space with Current and Last Contact Points]{\includegraphics[width=0.485\columnwidth]{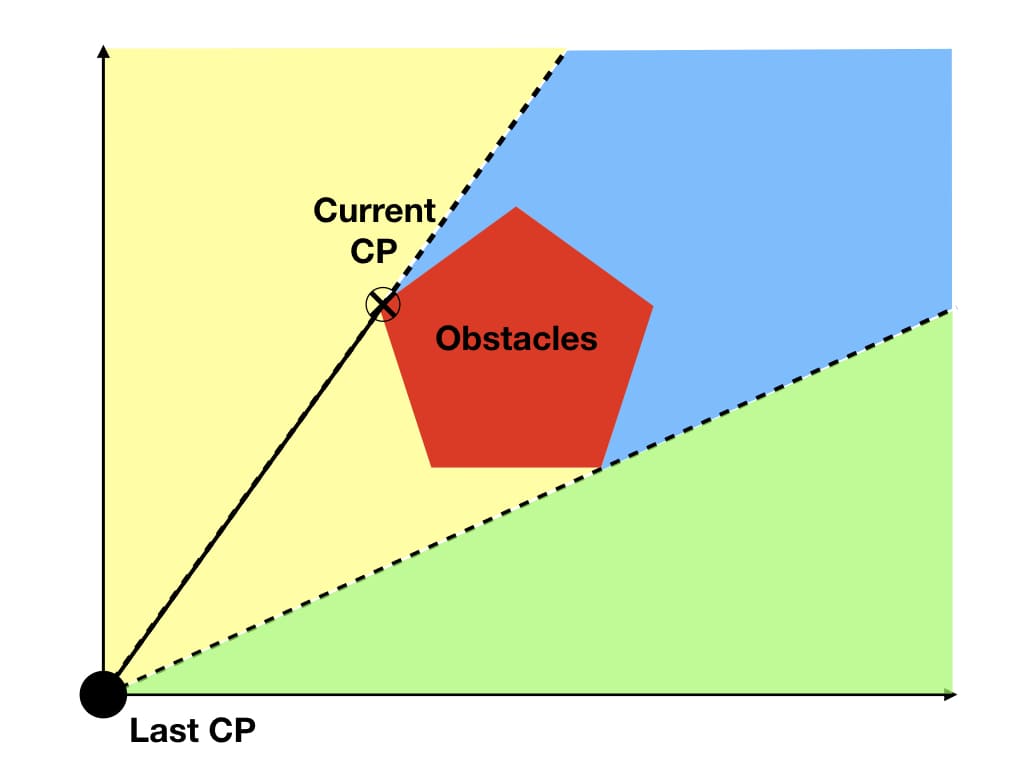}%
\label{fig::original}}
\hspace{0.015\columnwidth}
\subfloat[Current Contact Point Relaxed due to No Collision and Obstacles Not Being Confined]{\includegraphics[width=0.485\columnwidth]{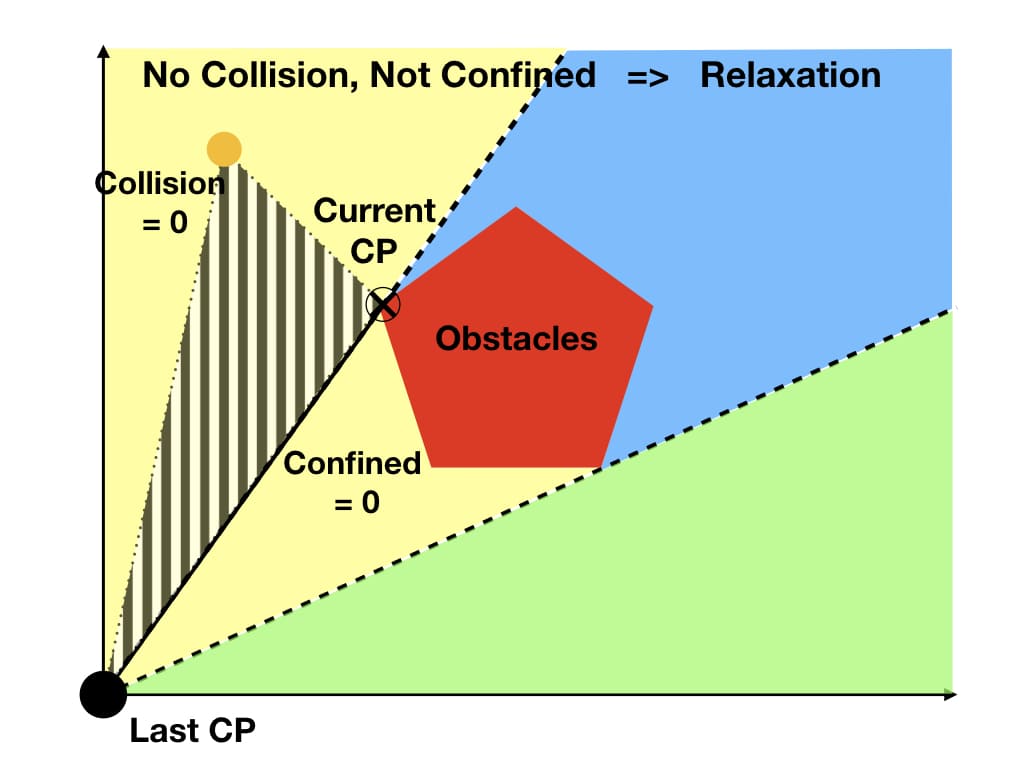}%
\label{fig::relaxation}}
\hfil
\subfloat[Current Contact Point Not Relaxed due to No Collision and Obstacles Being Confined]{\includegraphics[width=0.485\columnwidth]{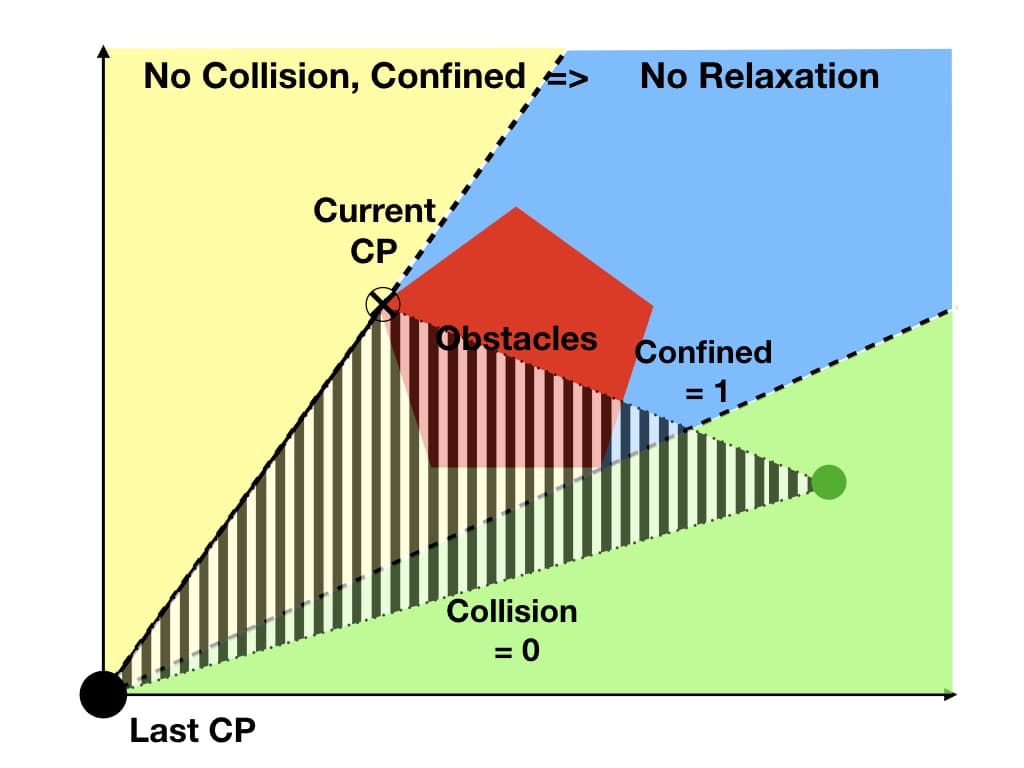}%
\label{fig::no_relaxation1}}
\hspace{0.015\columnwidth}
\subfloat[Current Contact Point Not Relaxed due to Collision]{\includegraphics[width=0.485\columnwidth]{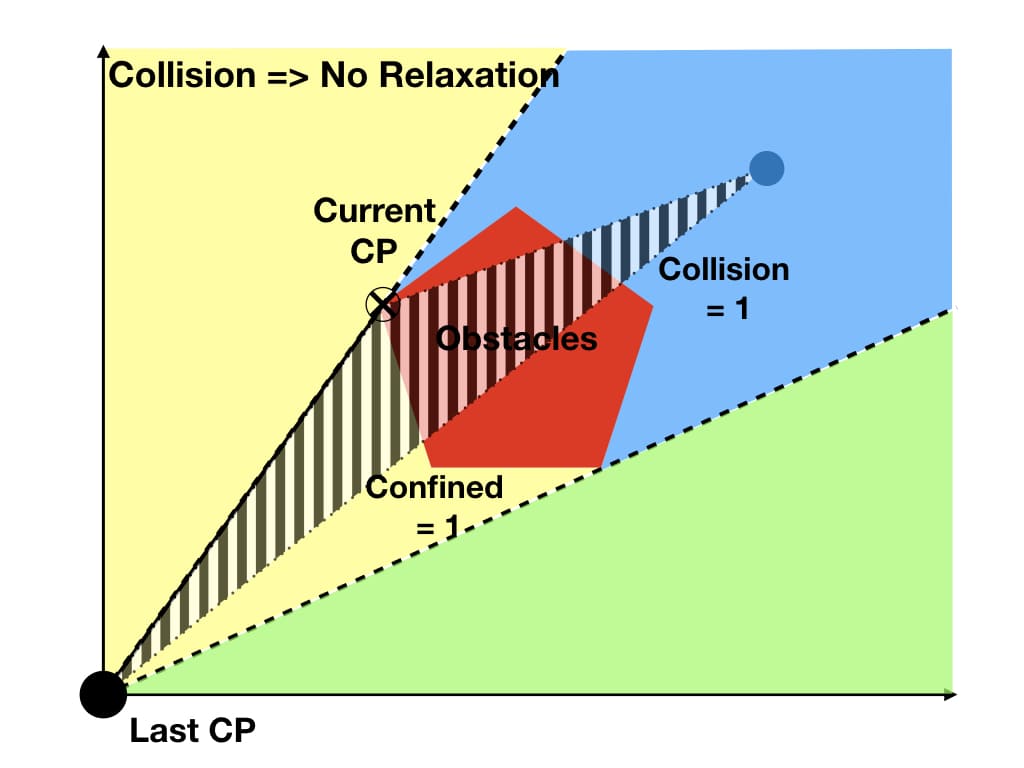}%
\label{fig::no_relaxation2}}
\caption{2-D Representation of the Tether Relaxation Scheme: Based on \textit{CheckCollision} between last contact point and current waypoint with the map, \textit{ObstacleConfined} checks if any obstacles are confined within the triangle formed by waypoint, last and current contact points (reprinted from \cite{xiao2018motion}).}
\label{fig:relaxation}
\end{figure}
 
The algorithm takes in as input the original map, a path planned by the high level planner, and the coordinates of tether origin (reel). We use a stack (\textit{CP\_stack}) to keep track of all current active tether contact points with the environment. Line 2 initializes contact points of all waypoints along the path to be the original tether reel center (\textit{tether\_origin}). \textit{relax\_flag} indicates if contact point relaxation is necessary. The rest of the algorithm plans the contact point for each individual waypoint. Line 6 to 18 determines if it is necessary to relax the current contact point. It first checks whether the robot is located at a waypoint directly reachable from the last contact point (line 8). If true (\textit{collision\_flag} == 0), it is possibly necessary to relax the current contact point (\textit{CP}), depending on if obstacle is confined within the triangle formed by the waypoint, last and current contact points. If false (\textit{collision\_flag} == 1), relaxation is not necessary. The actual relaxation is implemented in line 19 to 22. The current \textit{CP} is popped from \textit{CP\_stack}, and the last contact point is assigned to all subsequent waypoints. If relaxation is not necessary, line 24 checks if it's necessary to form a new contact point. If yes, the new \textit{CP} is pushed into \textit{CP\_stack} and all following waypoints are assigned the new contact point. 

\textit{CheckCollision (point A, point B, map)} draws a line between \textit{point A} and \textit{Point B} and see if any points on the line intersect with any obstacles in \textit{map}. If there is no collision, \textit{ObstacleConfined (point A, point B, point C, map)} further checks if any obstacle in \textit{map} is confined in the triangle formed by \textit{point A}, \textit{point B}, and \textit{point C}. A 2-D illustration of the tether relaxation pipeline is shown in Fig. \ref{fig:relaxation}. The 3D version works on the projection onto x-y, y-z, and x-z planes. To be 3-dimensionally confined, obstacle needs to be 2-dimensionally confined in all three projection planes. 

Both \textit{CheckCollision} and \textit{ObstacleConfined} have a complexity proportional to the number of obstacles $\mathcal{O}(o)$. Assuming the high-level planner path consists of $p$ waypoints, the whole algorithm's complexity is $\mathcal{O}(po)$. 

The result of the algorithm is an executable 6-D motion plan composed of 3D waypoints (from high-level path planner) along with corresponding 3D contact points. If the tether is not touching the environment, contact point is treated as the tether reel center. 

\subsection{Motion Executor}
\begin{figure}[]
\centering
\includegraphics[width=0.9\columnwidth]{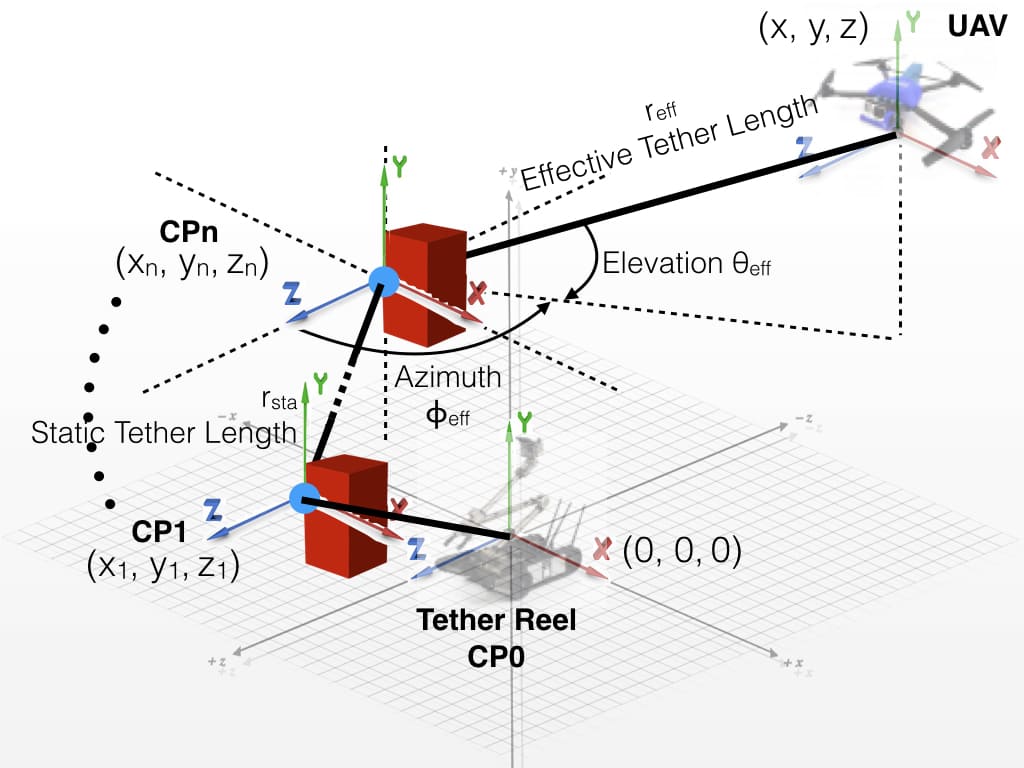}
\caption{Motion Executor Interpretation: Tether contact points are saved in a stack, where the latest contact point locates at the top. Tether is divided into several straight line segments, whose lengths are saved and associated with each contact point. UAV airframe's translational motion control is based on the planned relative coordinates of the UAV with respect to the last contact point (reprinted from \cite{xiao2018motion}). }
\label{fig::executor}
\end{figure}

How to execute the planned motion on the tethered UAV is discussed, with a focus on handling not only waypoints but also contact point(s). The 6-dimensional motion plan is parsed by the online motion executor. The UAV is commanded to reach every single waypoint along the path. UAV is treated here as a mass point and thus only positional movement is considered. The vehicle position control still uses tether length $L$, elevation $\theta$, and azimuth $\phi$. The position of the vehicle could be represented in polar coordinate system (Fig. \ref{fig::executor}). Given a certain $x$, $y$, and $z$, $L$, $\theta$, and $\phi$ could still be derived from Eqn. \ref{eqn::eucledean2polar}. 

However, since in this work multiple contact points are allowed,  all three control parameters, $L$, $\theta$, and $\phi$, are not relative to the tether reel (origin), but to the last contact point $CP_n$. Here, we use a stack to store all contact points. Whenever the motion executor reaches a new contact point, it pushes it into the stack. It also saves the current static tether length ($r_{sta}$) from the reel to this contact point. It is termed as static since this portion of the tether remains static based on our assumption that formed contact points don't move unless being relaxed. Whenever a contact point is relaxed, the motion executor pops it from the stack and reduces the static tether length by the corresponding segment length. So we have:

\begin{equation}
\label{eqn::static_tether_length}
L_{sta} =  \sum_{0}^{n-1}\sqrt{(x_{i+1}-x_i)^2+(y_{i+1}-y_i)^2+(z_{i+1}-z_i)^2}
\end{equation}

Since we have all our controls with respect to the last formed contact point (top of stack), we have effective values relative to this point: 

\begin{equation}
\label{eqn::effective_controls}
\left\{\begin{matrix}
L_{eff} = &\sqrt{(x-x_n)^2+(y-y_n)^2+(z-z_n)^2}\\
\theta_{eff} = &arcsin(\frac{y-y_n}{\sqrt{(x-x_n)^2+(y-y_n)^2+(z-z_n)^2}})\\
\phi_{eff} = &atan2(\frac{x-x_n}{z-z_n})
\end{matrix}\right.
\end{equation}

So the desired controls are: 
\begin{equation}
\label{eqn::desired_controls}
\left\{\begin{matrix}
L = &L_{eff} + L_{sta}\\
\theta = &\theta_{eff}\\
\phi = &\phi_{eff}
\end{matrix}\right.
\end{equation}

The desired values of $L$, $\theta$, and $\phi$ are regulated by the positional PID controller based on the sensory feedback from the UAV (reel encoder and tether angle sensors). An acceptance radius $R_{acc}$ is defined so that whenever the UAV reaches a ball with radius $R_{acc}$ around the desired waypoint, this waypoint is treated as reached and the executor moves on to the next waypoint. Velocity controller in Eqn. \ref{eqn::jacobian} and \ref{eqn::jacobian2} could also be applied, but all Cartesian coordinates and tether commands must be computed in the frame defined by the last contact point as well.

The motion executor doesn't need to discriminate between two different motion planners. The waypoint file from the ray casting approach could also be 6-dimensional, with all contact points to be the tether reel, namely the origin of the global coordinate system. This also applies to the non-contact path segment(s) from the contact planning approach.

\subsection{Summary of Tether Planning and Motion Execution}
This section presents two motion planning methods and a motion executor to navigate a tethered UAV in confined spaces with obstacles. Both motion planning methods are expected to allow tethered UAV to negotiate with obstacle-occupied spaces, even with the existence of the tether. The reachable space reduction approach forbids the UAV from entering spaces which is inaccessible to a straight tether and does not allow any interaction between the tether and environment. The contact(s) planning and relaxation approach maintains the same reachability space by allowing contact point(s) between the tether and environment. Along with the proposed motion executor, tethered flight could be implemented using the shifting-origin strategy with or without the existence of contact point(s). Quantitative experimental results and the advantages and disadvantages of both approaches will be presented through physical experiments in Chapter \ref{chapter::experiments}.

\section{Visual Servoing}
Based on the above-mentioned low level motion suite components, this section describes the proposed reactive visual assisting behavior: 6-DoF visual servoing of a pre-defined Point of Interest (PoI). This is a stand-alone visual assistance approach and is complementary to the deliberate high level risk-aware path planner. It takes in as input the live video stream of the visual assistant's camera, and issues motion commands to servo the PoI in the camera's 6-DoF configuration space.\footnote{This approach was discussed and published in previous work \cite{xiao2017visual}.}

\subsection{Servoing Approach}
\begin{figure}[]
\centering
\includegraphics[width=0.9\columnwidth]{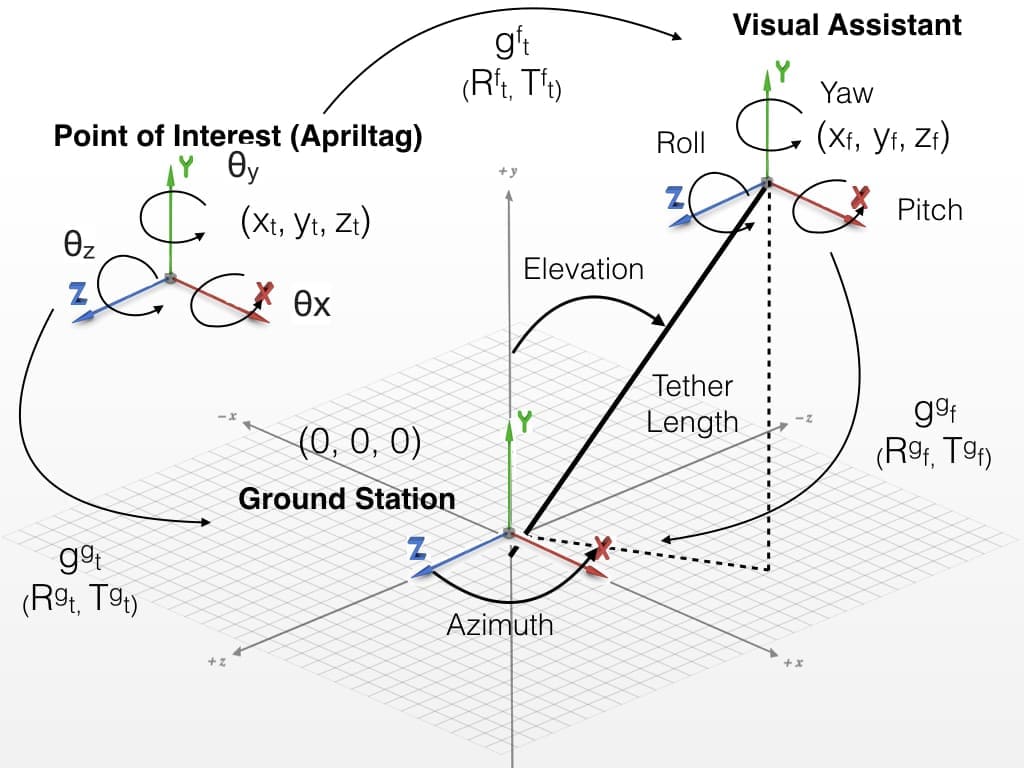}
\caption{Three Coordinate Systems (Reprinted from \cite{xiao2017visual})}
\label{fig::coordinates}
\end{figure}

The three coordinates systems used in the visual servoing are illustrated in Fig. \ref{fig::coordinates}. The ground station is fixed to the primary robot and is defined as the inertial frame. For convenience, horizontal plane is defined as $zx$ plane, while $y$ axis is pointing up vertically. This easily aligns with the camera frame. Tether length, elevation, and azimuth are defined with respect to the axes in the inertial frame. The position of the visual assistant $\begin{bmatrix} x_f,&y_f,&z_f, \end{bmatrix}^T$ is determined by Eqn. \ref{eqn::polar2euclidean}. Since the translation from the vehicle Center of Mass (CoM) to the camera origin is negligible, the vehicle frame and camera frame are treated equivalently. The rotational components  $\begin{bmatrix} yaw,&pitch,&roll \end{bmatrix}^T$ is the rotation angle with respect to the $y$, $x$, and $z$ axis, respectively. By the same token, the PoI frame, represented by an AprilTag \cite{olson2011apriltag}, is defined with respect to the camera frame. The homogeneous transformations from the visual assistant to ground station, and from AprilTag to UAV, are respectively defined as:
\begin{equation}
\label{eqn::ggf}
\bf{g^g_f} = 
\begin{bmatrix}
\bf{R^g_f} & \bf{T^g_f}\\ 
 0& 1 
\end{bmatrix}, \quad
\bf{g^f_t} = 
\begin{bmatrix}
\bf{R^f_t} & \bf{T^f_t}\\ 
 0& 1 
\end{bmatrix}
\end{equation}

where $\bf{R}$ and $\bf{T}$ denote the rotation matrix and translation vector. 

Assuming a point in the AprilTag frame to be $\bf{\bar{q}_t} = \begin{bmatrix} x_t,&y_t,&z_t \end{bmatrix}^T$,  we could apply the following coordinate system transformation to express it in the ground station frame, where $\bf{g^g_f}$ could be derived by the current flight status and $\bf{g^f_t}$ is given by the AprilTag tracking system:

\begin{equation}
\label{eqn::transfer}
\bf{\bar{q}_g} = \bf{g^g_t}\boldsymbol{\cdot} \bf{\bar{q}_t}=\bf{g^g_f}\boldsymbol{\cdot}\bf{g^f_t}\boldsymbol{\cdot} \bf{\bar{q}_t}
\end{equation}

In order to observe the PoI from a fixed pose, independent of how the PoI moves in the free space, our visual servoing controller should maintain a constant $\bf{g^f_t}$, the homogeneous transformation from the PoI to UAV frame. We denote this desired transformation as $\bf{g^f_t*}$. The controllable states in the system are $\begin{bmatrix} x,&y,&z,&yaw,&pitch,&roll \end{bmatrix}^T$, which determine $\bf{g^g_f*}$, our desired vehicle configuration. So we have an alternative way to express  $\bf{\bar{q}_g}$: 

\begin{equation}
\label{eqn::qg*}
\bf{\bar{q}_g} = \bf{g^g_t}\boldsymbol{\cdot} \bf{\bar{q}_t}=\bf{g^g_f*}\boldsymbol{\cdot} \bf{g^f_t*} \boldsymbol{\cdot} \bf{\bar{q}_t}
\end{equation}

Combining Eqn. \ref{eqn::transfer} and Eqn. \ref{eqn::qg*} the desired vehicle configuration could be calculated: 

\begin{equation}
\label{controls}
\bf{g^g_f*} =  \bf{g^g_f}\boldsymbol{\cdot}\bf{g^f_t}\boldsymbol{\cdot}  \bf{g^f_t*}^{-1}
\end{equation}

While $\bf{g^g_f}$ could be computed by UAV onboard telemetry and $\bf{g^f_t*}$ is pre-defined desired point of view, $\bf{g^f_t}$ is given by the AprilTag tracking system \cite{olson2011apriltag}. 6-DoF configuration of the tag could give the homogeneous transformation. Based on different teleoperation tasks, different desired observing positions and orientations $\bf{g^f_t*}$ could be easily defined. For example, the desired observing pose could be chosen as zero rotation and 10 unit distance shifted away from PoI plane. As a result, the AprilTag will always locate in the middle of the image frame, facing straight toward the camera with a proper frontal size. 

$\bf{g^g_f*} $ could be further decomposed to $\bf{R^g_f*}$ and $\bf{T^g_f*}$, from which the controls for the UAV could be derived. 

It is worth to note that the tracking of PoI does not necessarily need to be through vision-based methods. Any form of 6-DoF tracking of the PoI can serve as visual servoing input. For example, if the PoI is defined as the primary robot's manipulator gripper, the 6-DoF configuration of the gripper could be easily tracked by the motor encoders on the manipulator arm and segment dimensions. The offset from the manipulator base to the tether reel center needs to be compensated. 

The status updates (tether length $r$, elevation $\theta$, azimuth $\phi$, and quaternion representing the vehicle orientation) of the current sensed vehicular configuration is used to compute $\bf{g^g_f}$. The gimbal pitch and roll, however, need to be estimated using integration since it is not provided by the current version of Fotokite SDK firmware. 

Using $\bf{g^g_f}$, along with $\bf{g^f_t}$ from AprilTag and predefined $\bf{g^f_t*}$, $\bf{g^g_f*}$ is computed from Eqn. \ref{controls}. After computing the desired transformation between the vehicle and ground station, $\bf{g^g_f*}$ is translated into the visual assistant's configuration space: 

\begin{center}
$\begin{bmatrix} x*,&y*,&z*,&yaw*,&pitch*,&roll* \end{bmatrix}^T$
\end{center}

The first four dimensions are controllable by the vehicle, while the gimbal is responsible for the last two. $\begin{bmatrix} x*,&y*,&z* \end{bmatrix}^T$ is controlled by the position controller discussed above. Six PID controllers are used to drive those six independent variables to the desired value. 

Given the fact that camera roll will cause disruptive motion in the video stream, although the roll of the POI is tracked, the actual gimbal roll is not controlled. This assures that the video feed from the visual assistant is always upright, which is desirable for the operator. 

\subsection{Summary of Visual Servoing}
This section provides a stand-alone visual assistance approach, reactive visual servoing, and is complementary to the deliberate high level risk-aware path planner. Using a fiducial marker as the visual servoing Point of Interest, the assistant UAV is able to track the PoI's full state space. Based on a predefined desired viewpoint configuration, the visual servoing algorithm computes the coordinate system transformation and can control the assistant's vehicle and camera pose to maintain a constant 6-DoF relative position and orientation with respect to the PoI. This proposed approach is expected to allow reactive 6-DoF visual servoing of a visual PoI using a tethered UAV, including translational $x$, $y$, and $z$ in 3D Cartesian space, and rotational yaw, pitch, and roll in camera orientation. The experiments of the visual servoing approach will be presented in Chapter \ref{chapter::experiments}.

\section{Summary of Low Level Tethered Motion Suite}
This section presents a complete low level motion suite that can implement the planned risk-aware motion of the visual assistant on a physical tethered UAV. Despite the fact that this motion suite is developed in the context of a risk-aware tethered aerial visual assistant, it is applicable to any tethered UAV flying in indoor GPS-denied, unstructured or confined environments, especially with possible interference between tether and obstacles. This opens up another regime of indoor aerial locomotion: tethered flight. This motion suite aims at resolving the issues brought in by the tether in cluttered spaces and still maintaining the advantages of tether, such as prolonged UAV power durations and satisfying tethered safety requirements for mission-critical operations. 

The motion suite starts with a novel but simple localization scheme based on a mechanics model of the tether and tether sensory feedback. It does not require much computational overhead and does not rely on GPS or other exteroception. Tether has been effectively utilized with the proposed localizer, in addition to power considerations and safety concerns. Actual localization accuracy improvement will be demonstrated via physical experiments in a MoCap studio in Chapter \ref{chapter::experiments}. Two different tether-based motion primitives are presented to enable free flight of the tethered UAV in 3D Cartesian space. Motion commands in Cartesian space are translated using PID positional control or Jacobian-based velocity control into tether-based commands. The navigation accuracy will be benchmarked for both motion primitives: position control requires high waypoint density to guarantee precise motion while velocity control works with sparse waypoints but is sensitive to singularity (shown in Chapter \ref{chapter::experiments}). Despite the low smoothness of position control on dense waypoints, it is still recommended for field use due to its superior robustness. Two different tether planning approaches are also introduced, one without and one with the possibility of tether contact point(s) with the environment. Following Chapter \ref{chapter::experiments} will show that the first reachable space reduction via ray casting method trades reachability for accuracy, while the second contact planning and relaxation aims at maintaining the same reachability as a tetherless UAV. The strong invariant tether contact position assumption, however, causes increased navigational error with increased number of tether contact points. The motion executor to execute waypoints along with planned contact points is also introduced and will be included in the physical experiments. Lastly, as complement to the entire risk-aware deliberate visual assistance approach, a reactive visual servoing method is also presented, which has the potential to allow the visual assistant to react to the movement of a trackable visual stimuli and maintain a constant 6-DoF configuration to that PoI. Physical experiments of the visual servoing approach will be presented in Chapter \ref{chapter::experiments}.

\chapter{EXPERIMENTS}
\label{chapter::experiments}
This chapter presents experiments for all the individual components in the low level motion suite described in Chapter  \ref{chapter::low_level}. The individual experimental results will be presented and discussed. The aim of the experiments in this chapter is to validate the theoretical approaches proposed in Chapter \ref{chapter::low_level} on a physical tethered UAV. It needs to be pointed out that the risk reasoning framework in Chapter \ref{chapter::risk_representation} is deduced through formal methods such as propositional logic and probability theory. For the risk-aware planner in Chapter \ref{chapter::high_level}, its suboptimality with respect to traverse-dependent risk element is shown by example and its optimality up to action-dependent risk element is proved by inductive reasoning. Therefore no experiments are necessary to validate the risk reasoning framework (Chapter \ref{chapter::risk_representation}) and risk-aware planner (Chapter \ref{chapter::high_level}). 

All the experiments conducted in this chapter use a tethered UAV platform, Fotokite Pro, from Perspective Robotics AG \cite{fotokitewebsite}. The UAV is a quad-rotorcraft and is equipped with an onboard camera with a 2-DoF gimbal (pitch and roll). The camera's yaw is controlled dependently by the vehicular yaw. The experiments use the SDK provided by Fotokite Pro, including the sensory feedback, tether length, azimuth and elevation angles, and change rate control over these three control parameters. 

\section{Tether-based Localization Experiments}
Experiments for the tether-based localization approach described in Chapter \ref{chapter::low_level} are presented.\footnote{Detailed experimental results were presented and published in previous work \cite{xiao2018indoor}.}

\subsection{Hypothesis and Metrics}
The hypothesis for the experiments of the tether-based localization is \emph{the proposed approach can improve localization accuracy compared to the preliminary localizer based on straight tether assumption and the new accuracy is no longer dependent on tether length}. The metric used is the \emph{average localization error}, as the distance between the localization result and its ground truth location. It is also compared with tether length. 

\subsection{Experiments}
\begin{figure}[]
\centering
\includegraphics[width=0.8\columnwidth]{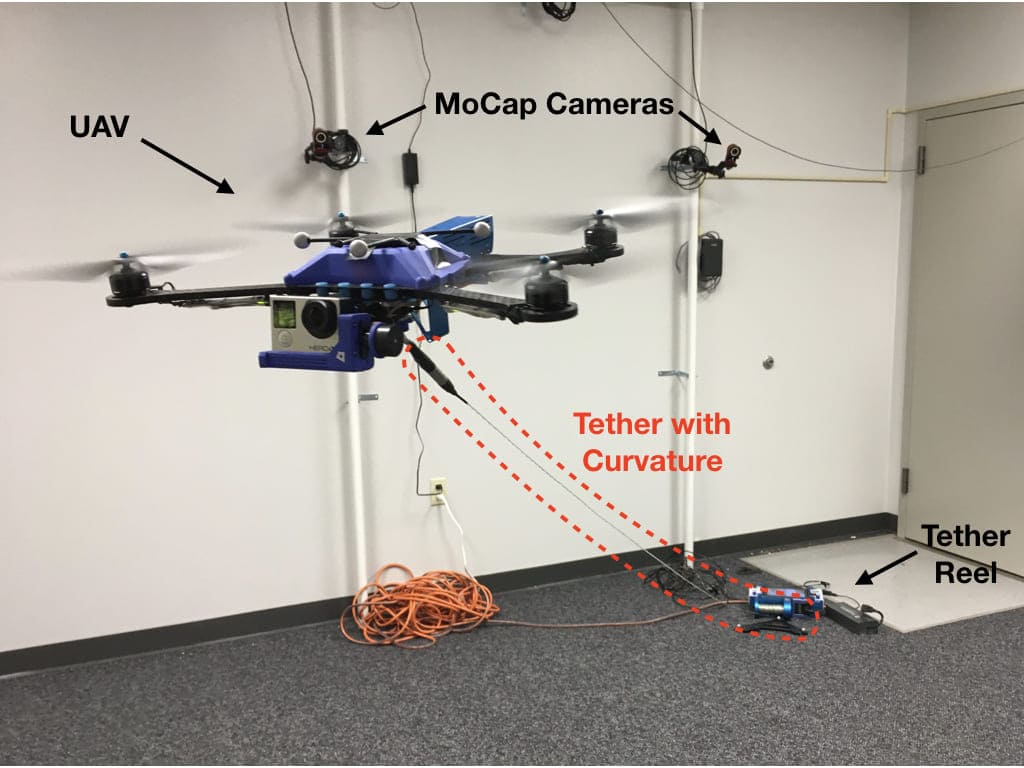}
\caption{Experimental Setup: UAV flying in a motion capture studio with a tether pulled down by gravity (reprinted from \cite{xiao2018indoor})}
\label{fig::fotokite_studio}
\end{figure}

The experiments are conducted in a motion capture studio to capture motion ground truth (Fig. \ref{fig::fotokite_studio}). The studio is equipped with 12 OptiTrack Flex 13 cameras running at 120 Hz. The 1280$\times$1024 high resolution cameras with a 56\degree~Field of View provide less than 0.3mm positional error and cover the whole 2.5$\times$2.5$\times$2.5m space. 

\begin{figure}[]
\centering
\includegraphics[width=1\columnwidth]{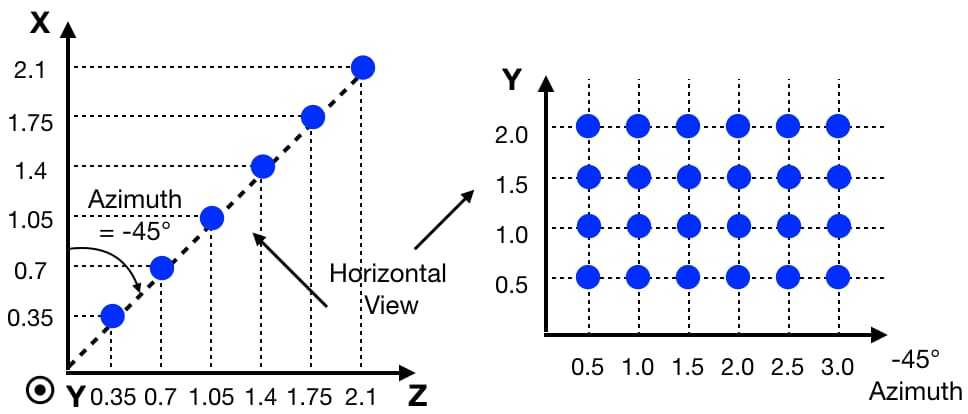}
\caption{Experimental points are chosen within a horizontal plane with -45\degree azimuth angle. Within this plane, points are distributed over a grid with 0.5m interval (reprinted from \cite{xiao2018indoor}). }
\label{fig::experiments}
\end{figure}

Since our approach does not affect azimuth angle, we fix the experiments at a constant azimuth. All experiment points are chosen on a horizontal plane with  -45\degree azimuth angle (Fig. \ref{fig::experiments} left). Within this plane, points are located on a grid pattern with an interval of 0.5m (Fig. \ref{fig::experiments} right). We fly the UAV, Fotokite Pro, to and hover at each individual experiment points, using the preliminary and our proposed localizer. Ground truth positional data is recorded by the motion capture system. 

\begin{figure}[]
\centering
\subfloat[3D Illustration of Experimental Results]{\includegraphics[width=0.5\columnwidth]{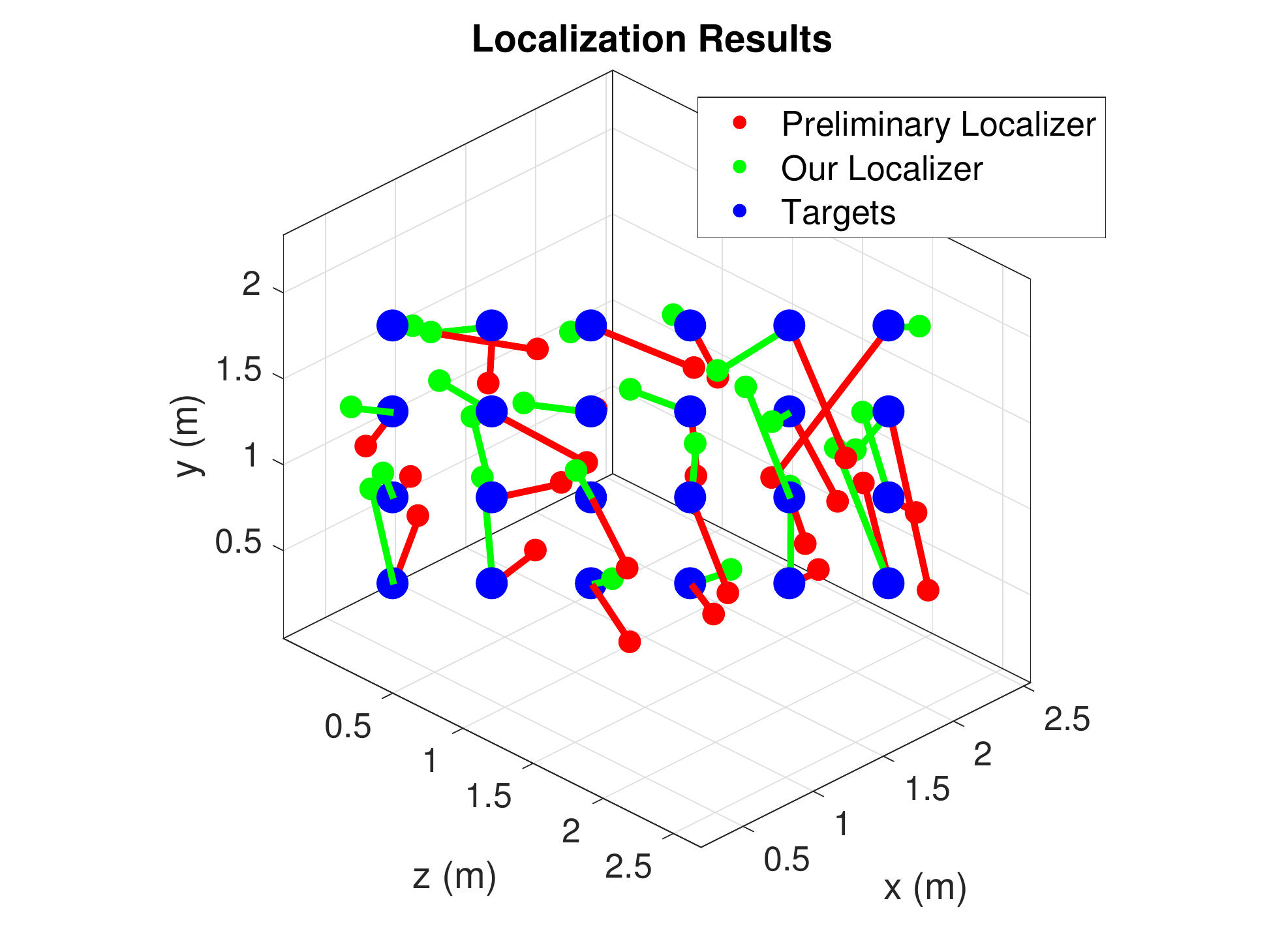}%
\label{fig::all_results1}}
\hfil
\subfloat[-45\degree Azimuth Plane Horizontal View]{\includegraphics[width=0.5\columnwidth]{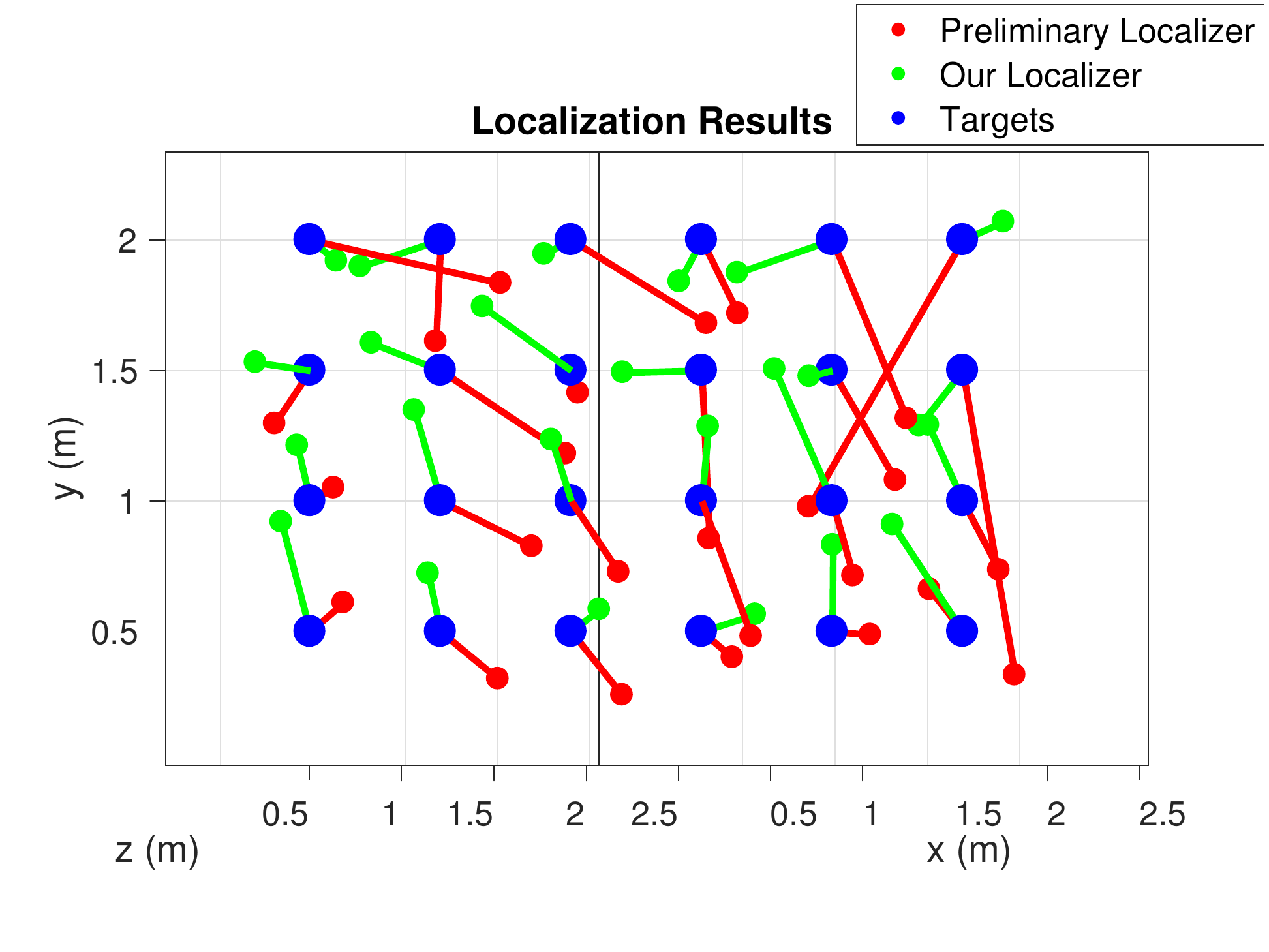}%
\label{fig::all_results2}}
\subfloat[Top Down View]{\includegraphics[width=0.5\columnwidth]{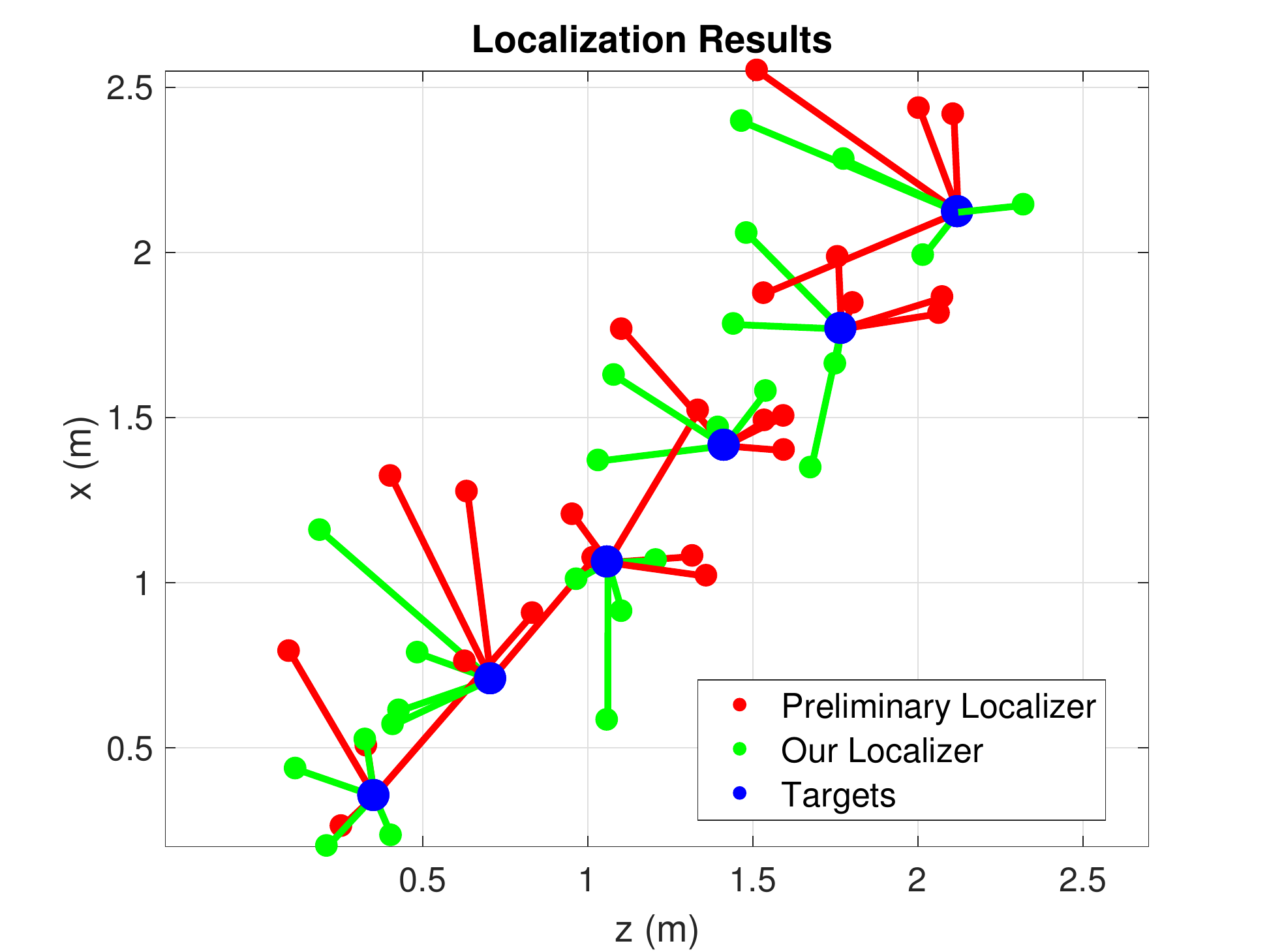}%
\label{fig::all_results3}}
\caption{Localization Results from 48 Experimental Trials: Blue points designate the target points the UAV should localize and hover at. Red points are the localization results using the preliminary localizer. Green points are resulted by our proposed mechanics-based approach. Small straight line segments connect localization results with their corresponding targets (adapted from \cite{xiao2018indoor}).}
\label{fig::results}
\end{figure}

48 localization trials are performed with 48 ground truth positions collected, 24 using preliminary localizer and the other 24 our new approach. Due to the turbulence created by the propellors in a confined indoor studio space, the UAV wobbles at the target location. So for each data point, we record the motion of Fotokite as a rigid body using the motion capture system for 5 seconds after it stabilizes at the target location. Average value is taken over the 600 tracked points (5 seconds at 120 Hz). All 48 localized points along with the 24 target points from our physical experiments are displayed in Fig. \ref{fig::results}. While blue points designate the ideal target points where the UAV should localize and hover at, red and green points denote localization results from the preliminary and our proposed localizer, respectively. Red and green straight line segments illustrate the correspondence between localization results and target point. 

\subsection{Discussions}
As Fig. \ref{fig::all_results1} shows, green straight line segments are usually shorter than the red ones connected with the same blue points. This means improved localization accuracy using our proposed localizer. A closer look into the experimental results are shown in Fig. \ref{fig::all_results2} and \ref{fig::all_results3}. Fig. \ref{fig::all_results2} is the perpendicular view toward -45\degree azimuth plane. It could be observed that red points are always lower than blue points. This is the reason caused by the invalid straight tether assumption (Fig. \ref{fig::real_and_sensed}). The real elevation angle is always smaller than the sensed value, so given a certain tether length the preliminary localizer thinks the UAV were at a higher position, but in fact it's lower. Green points achieved by our new model are distributed around blue points, with a smaller distance. This shows that our proposed localizer overcomes the problem caused by invalid straight tether assumption and reduces the localization error in the vertical direction. Fig. \ref{fig::all_results3} shows the top town view of the 48 trials and directly illustrate the localization accuracy in the horizontal plane. There is not much difference to be observed between red and green points with respect to the blue ones since our localizer doesn't deal with azimuth angle correction. The slightly denser distribution within 0 and -45\degree is due to tether azimuth and vehicular yaw angle initialization error. Overall speaking, the improvement of localization accuracy is summarized in Tab. \ref{tab::average_tracking_error}.

\begin{table}[]
\centering
\caption{Average Localization Error (Reprinted from \cite{xiao2018indoor})}
\label{tab::average_tracking_error}
\begin{tabular}{|c|c|c|c|}
\hline
                                                                           & \textbf{\begin{tabular}[c]{@{}c@{}}Preliminary \\ Localizer\end{tabular}} & \textbf{\begin{tabular}[c]{@{}c@{}}Proposed\\ Method\end{tabular}} & \textbf{Improvement} \\ \hline
\textbf{\begin{tabular}[c]{@{}c@{}}Average \\Localization \\ Error (m)\end{tabular}} & 0.5335                                                                    & 0.3675                                                             & 31.12\%              \\ \hline
\end{tabular}
\end{table}

\begin{figure}[]
\centering
\includegraphics[width=0.8\columnwidth]{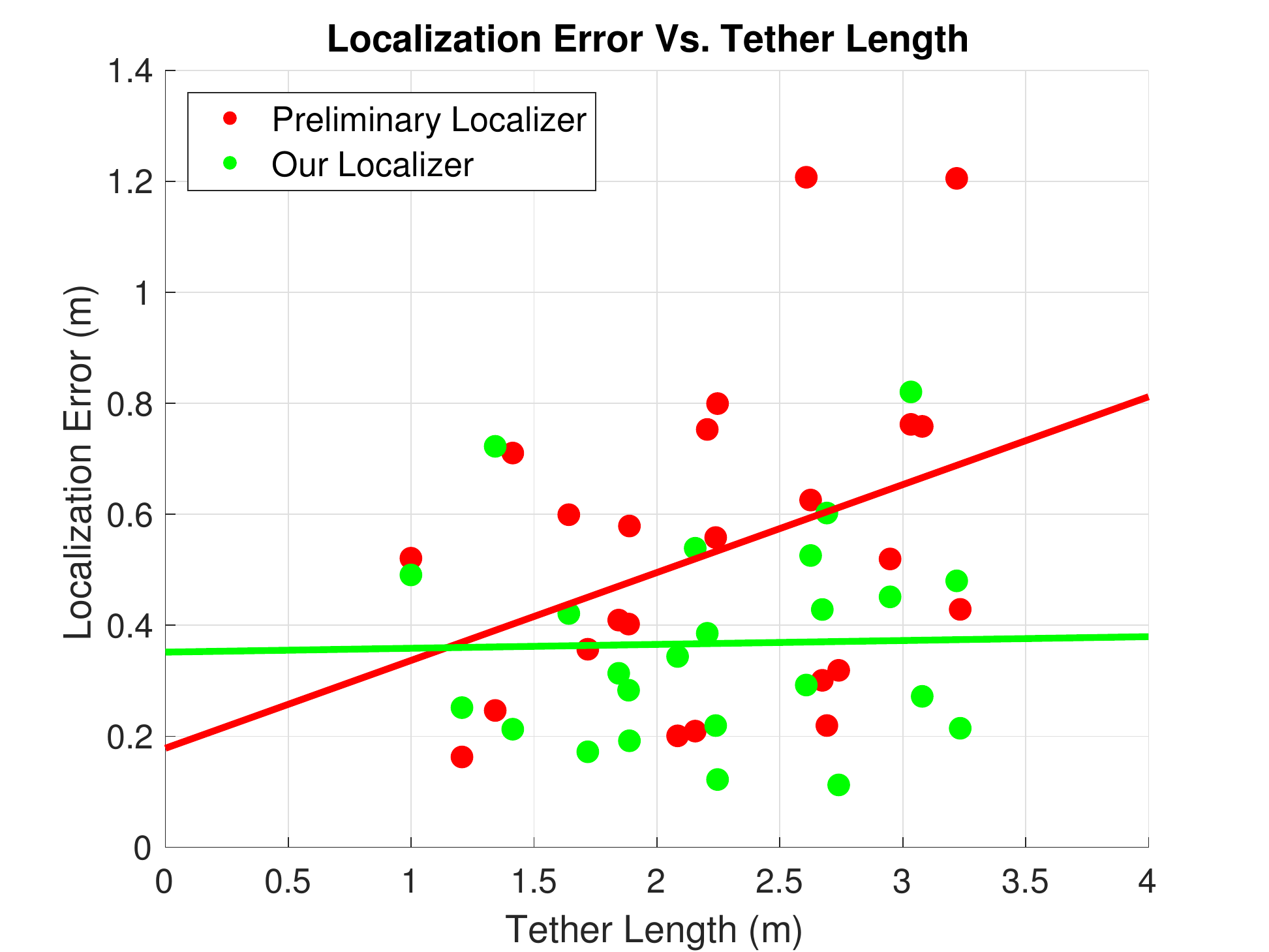}
\caption{Localization Error in Terms of Tether Length: Preliminary localizer's error (red) increases with longer tether, while tether length does not have a significant effect on our proposed method (green) (reprinted from \cite{xiao2018indoor}).}
\label{fig::errors}
\end{figure}

Due to the invalid straight tether assumption, accuracy of the preliminary localizer is deteriorated with increasing tether length. Fig. \ref{fig::errors} looks into this effect in detail. In general, localization error from the preliminary localizer is worse (red points) than our proposed method (green points). A line is fitted to the results of each localizer using linear regression. As we can see, the red line indicates that longer tether length has a significant negative effect on localization accuracy, while our proposed method is not sensitive to increasing tether length. Our proposed localizer has a limited average localization error within 0.4m. This is the best hovering stability achievable by Fotokite's built-in controller measured by experiments \cite{xiao2018motion}. 

\subsection{Summary of Tether-based Localization Experiments}
The improved localization accuracy is demonstrated by experiments on the physical tethered UAV, Fotokite Pro. The results indicate that our model is able to ameliorate localization accuracy by 31.12\% and effectively eliminate the negative effect of increased tether length on localization result. The average localization error achieved by our proposed method is limited within the hovering stability tolerance of our particular UAV platform. 

\section{Tether-based Motion Primitives Experiments}
Experiments for the tether-based motion primitives approach described in Chapter \ref{chapter::low_level} are presented.\footnote{Detailed experimental results were presented and published in previous work \cite{xiao2019benchmarking}.}

\subsection{Hypothesis and Metrics}
The hypothesis for the experiments of the tether-based motion primitives is \emph{the proposed approach can realize free-flight in 3D Cartesian space on a tethered UAV using both tether-based motion primitives}. The metric used is \emph{success/fail} of path execution and \emph{flight accuracy/navigation error}, as the distance between the actual UAV location and its planned ideal location. The \emph{path smoothness}, measured as average turning angle between UAV locations, is also analyzed and discussed.

\subsection{Experiments}
The two proposed motion primitives are tested in a MoCap studio to quantify their flight performance using two sets of experiments. 

The experiments are conducted in a motion capture studio to capture motion ground truth. In the studio, 6 OptiTrack Flex 13 cameras run at 120 Hz. The 1280$\times$1024 high resolution cameras with a 56\degree~Field of View provide less than 0.5mm positional error and cover a whole 4$\times$4$\times$2.5m space. Eight infrared reflective markers are attached and evenly distributed on all sides of the UAV so that the UAV could be captured even if some of the markers are blocked by the aerial frame itself. 

\begin{figure}[]
\centering
\includegraphics[width=0.8\columnwidth]{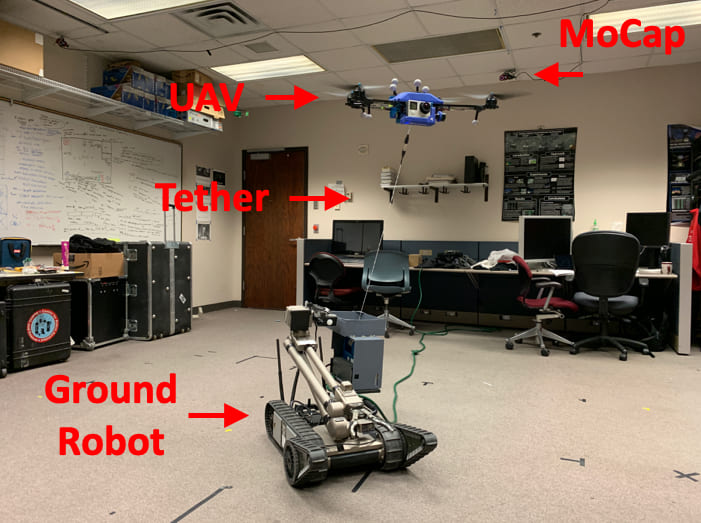}
\caption{Tethered UAV Flying in MoCap Studio (Reprinted from \cite{xiao2019benchmarking})}
\label{fig::mocap}
\end{figure}

During the physical tests, the acceptance radius for each waypoint is set to 0.4m. That is, when the UAV is within 0.4m from the current waypoint (localized by onboard sensing only), it is considered that the UAV reaches that particular waypoint and it moves on to the next one. This is the best localization accuracy achievable by the UAV's onboard sensory feedback measured by experiments. Fig. \ref{fig::mocap} shows the tethered UAV flying in the MoCap studio. 

\begin{figure}
\centering
\subfloat[Auxiliary View]{\includegraphics[width=0.5\columnwidth]{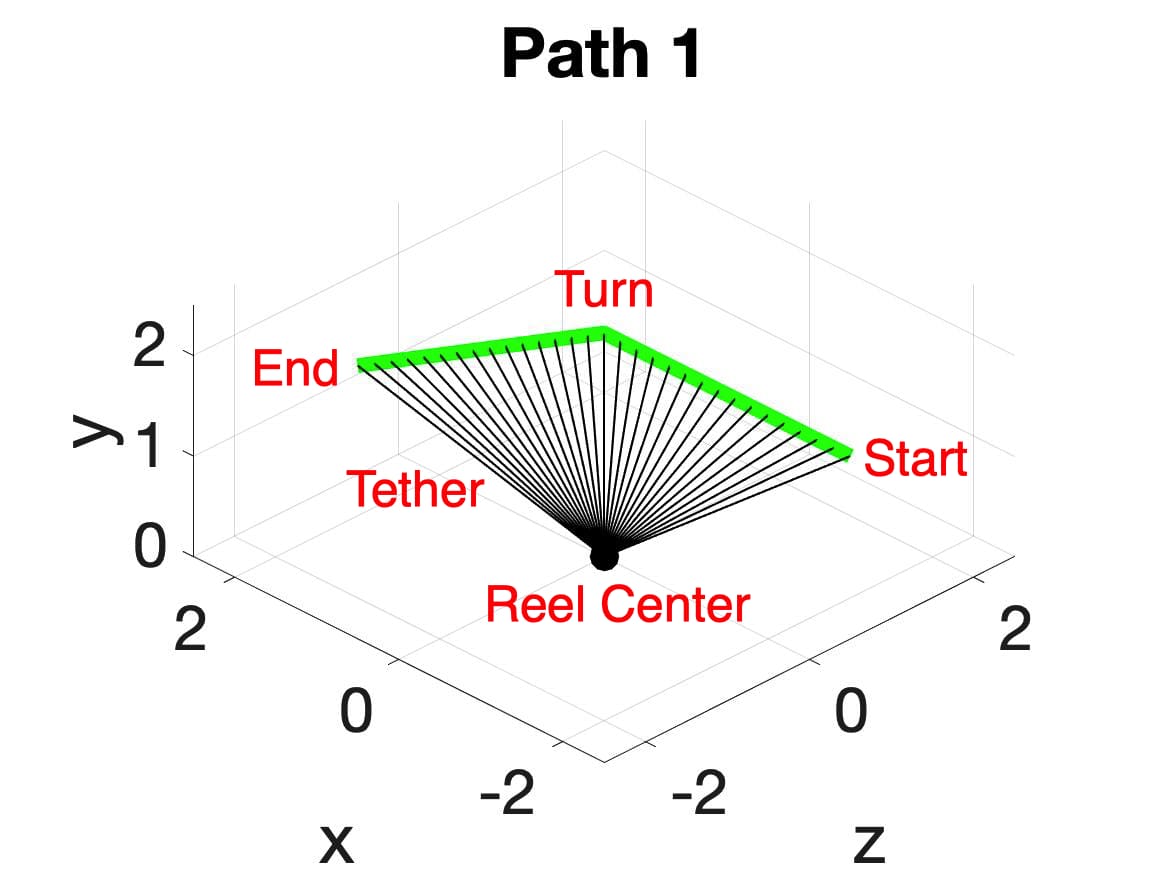}%
\label{fig::auxiliary}}
\subfloat[Front View]{\includegraphics[width=0.5\columnwidth]{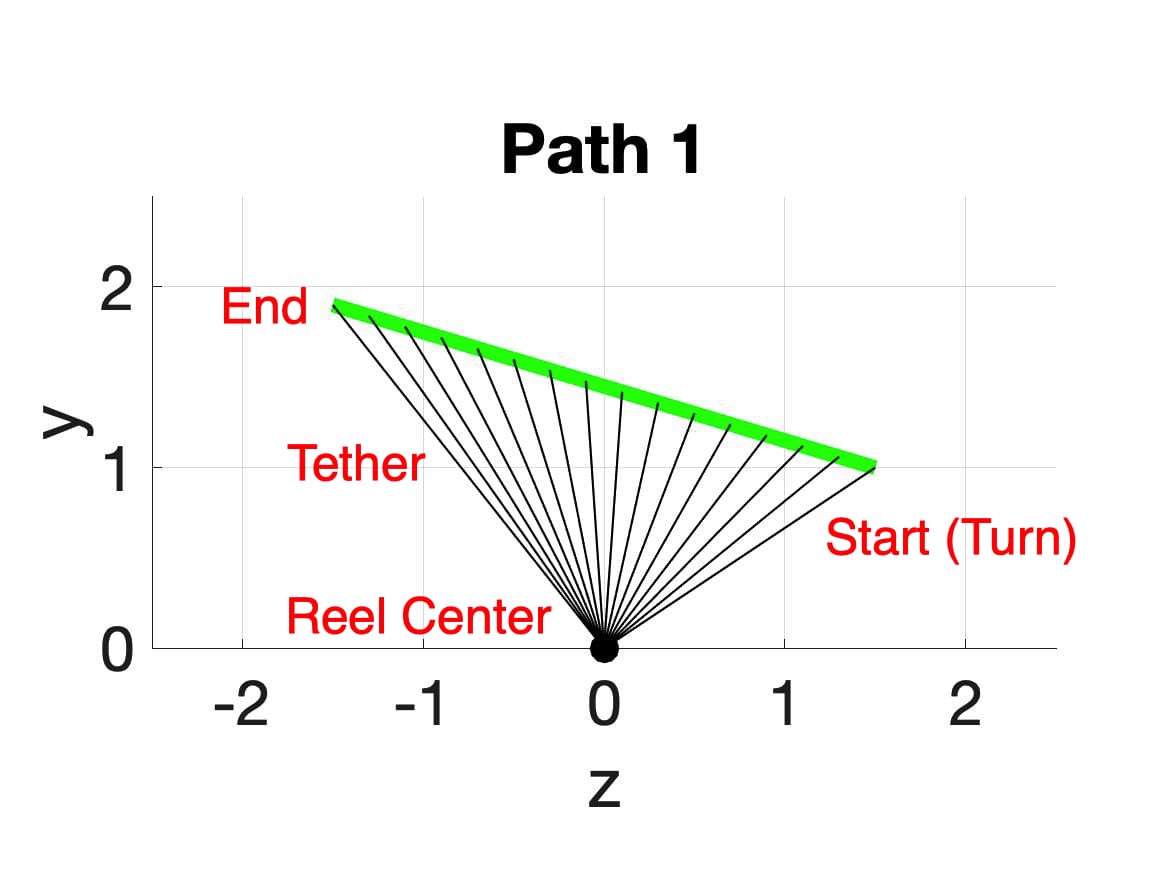}%
\label{fig::front}}\\
\subfloat[Top View]{\includegraphics[width=0.5\columnwidth]{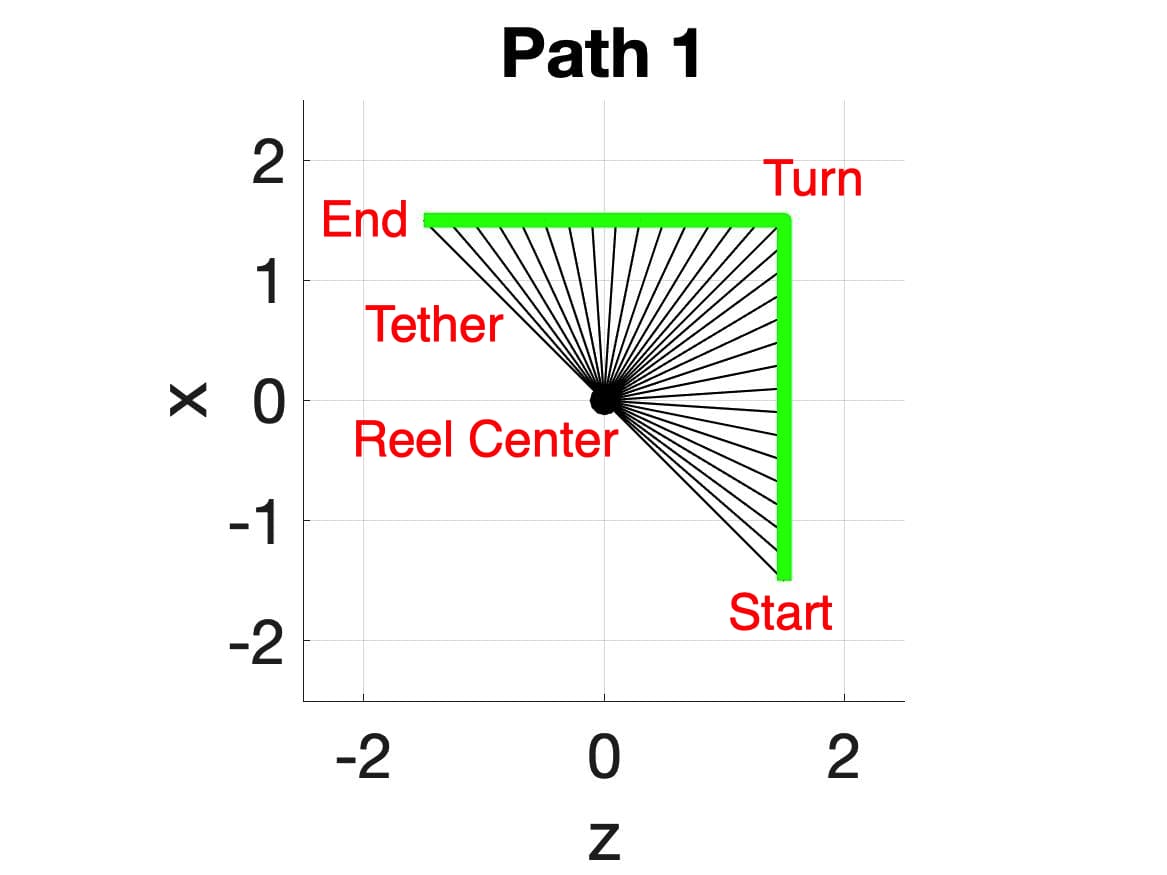}%
\label{fig::top}}
\subfloat[Side View]{\includegraphics[width=0.5\columnwidth]{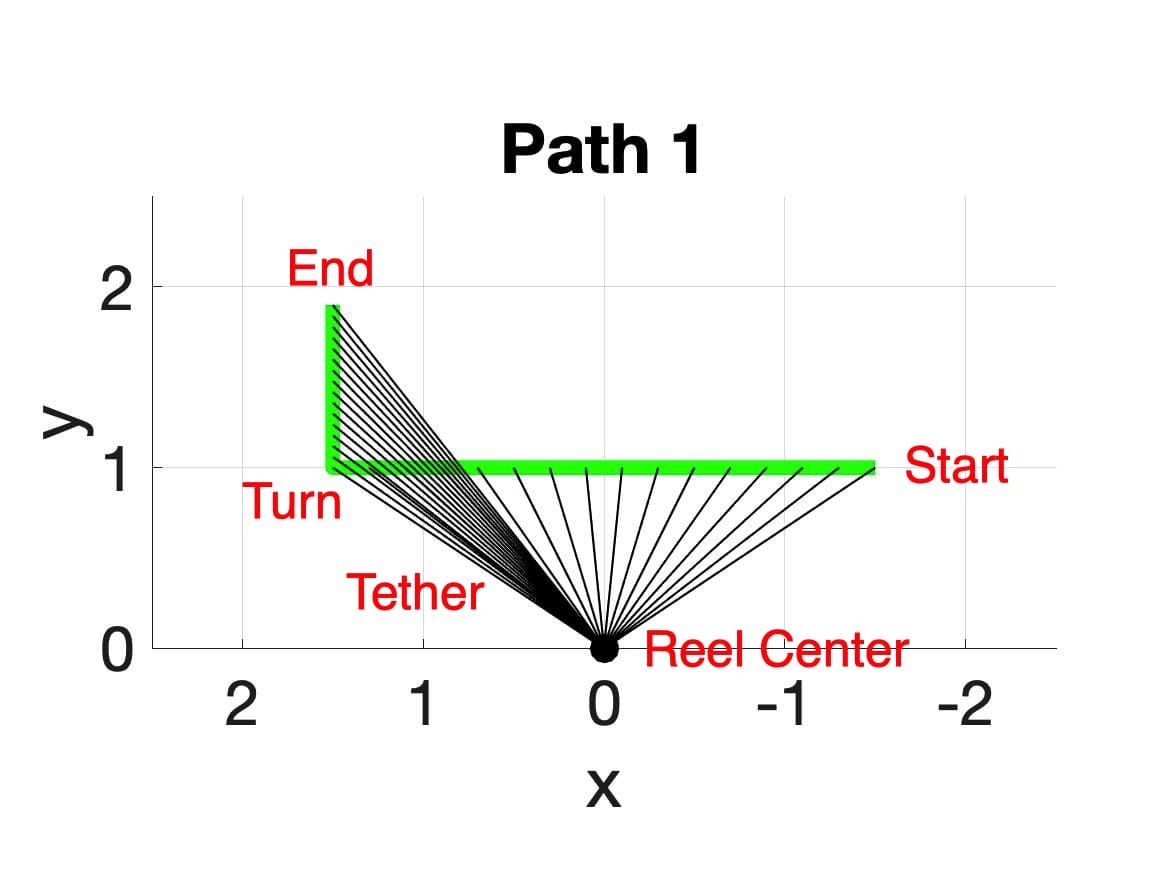}%
\label{fig::side}}
\caption{Different Views for Experiment 1 (Reprinted from \cite{xiao2019benchmarking})}
\label{fig::path1_views}
\end{figure}

Executing a straight line path may be trivial for free flying UAVs, but the straightness and accuracy of the path execution is of importance to tethered UAVs. In the first set of experiments, we first test a flight path consisting of a 3m horizontal and an ascending straight path (3m projection length on horizontal plane) connected by a 90\degree~turn (Fig. \ref{fig::path1_views}). We test both motion primitives on path plans with five different waypoint densities. That is, from dense to sparse, the intervals between two consecutive waypoints projected in the horizontal plane are 0.2m, 0.5m, 1m, 1.5m, and 3m. Therefore the numbers of waypoints for each path plan are 31, 13, 7, 5, and 3, respectively, denoting the same path. For each waypoint density, six repetitive trials are executed, three of which using position control and other three using velocity control. A second set of experiments is conducted on a straight line path passing above the tether reel center from the first to third quadrant in the x-z plane. This set of experiments shows the inability of velocity control near singularity and the improvement of flight accuracy with denser waypoints using position control. Since the orientation control of the UAV is not the focus of this research, the yaw is not explicitly controlled during path execution. 

\subsection{Discussions}
\begin{figure}
\centering
\subfloat[Position 0.2m]{\includegraphics[width=0.33\columnwidth]{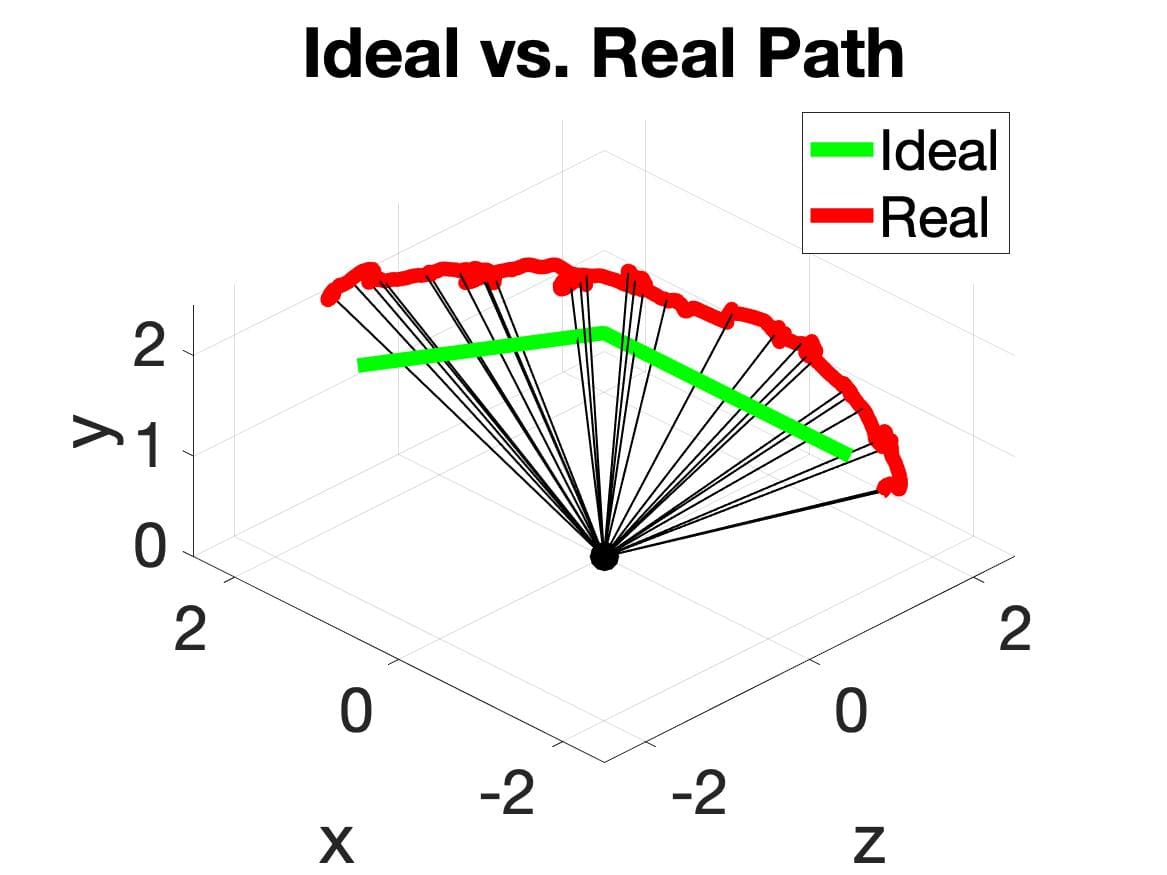}%
\label{fig::pos0.2}}
\subfloat[Position 0.5m]{\includegraphics[width=0.33\columnwidth]{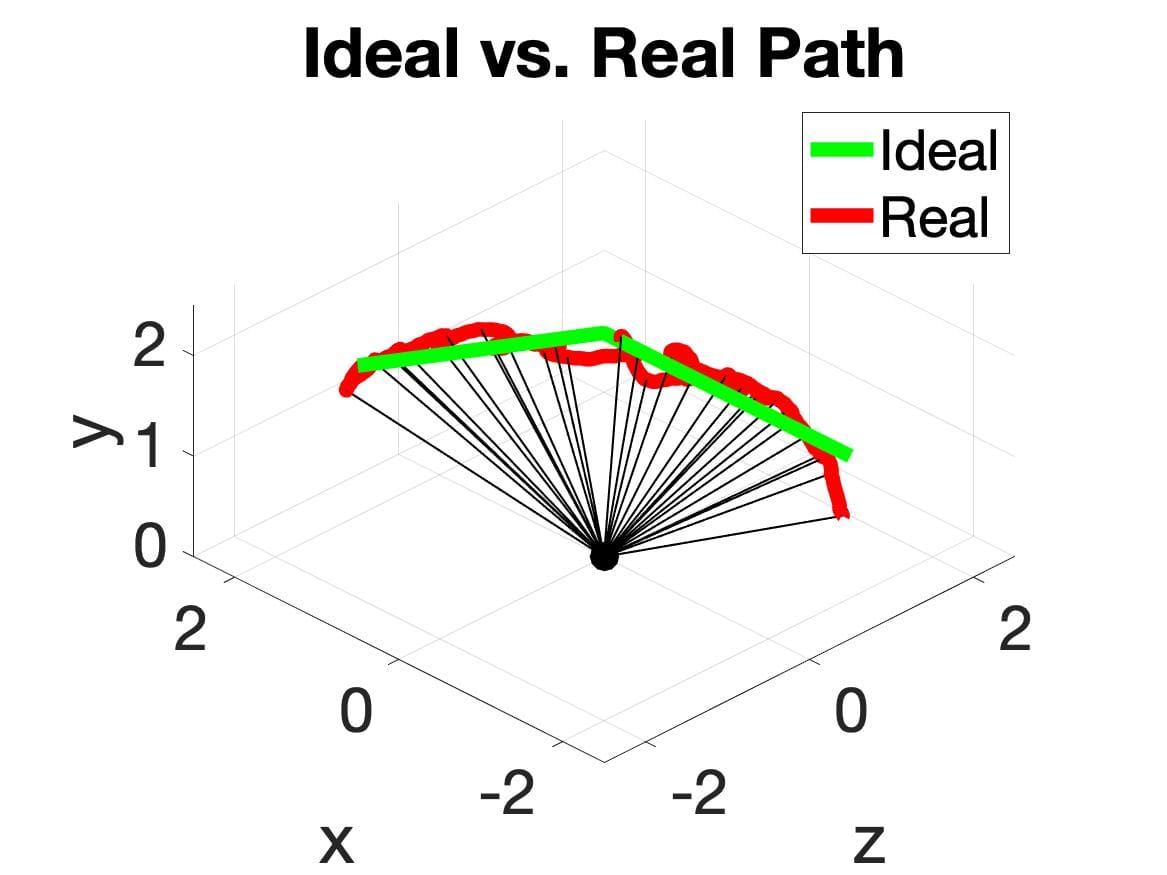}%
\label{fig::pos0.5}}
\subfloat[Position 1m]{\includegraphics[width=0.33\columnwidth]{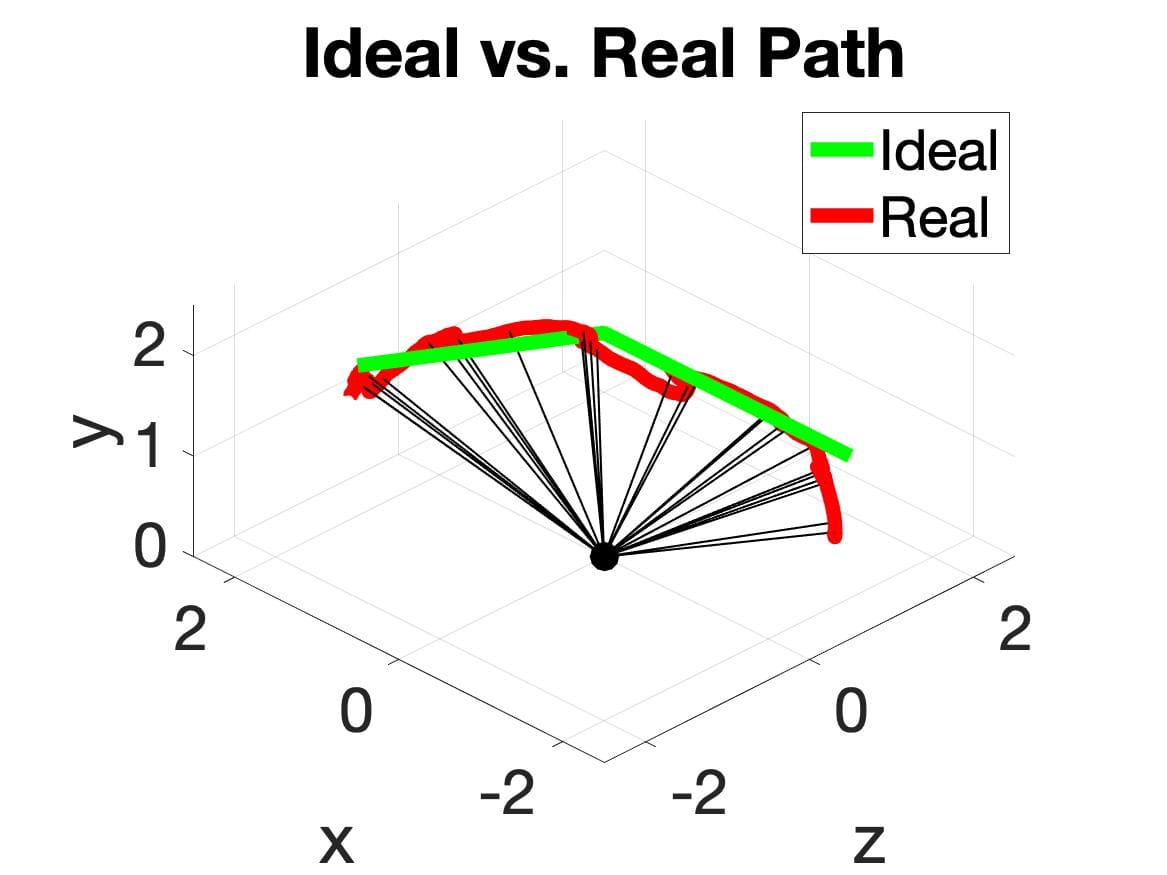}%
\label{fig::pos1}}\\
\subfloat[Position 1.5m]{\includegraphics[width=0.33\columnwidth]{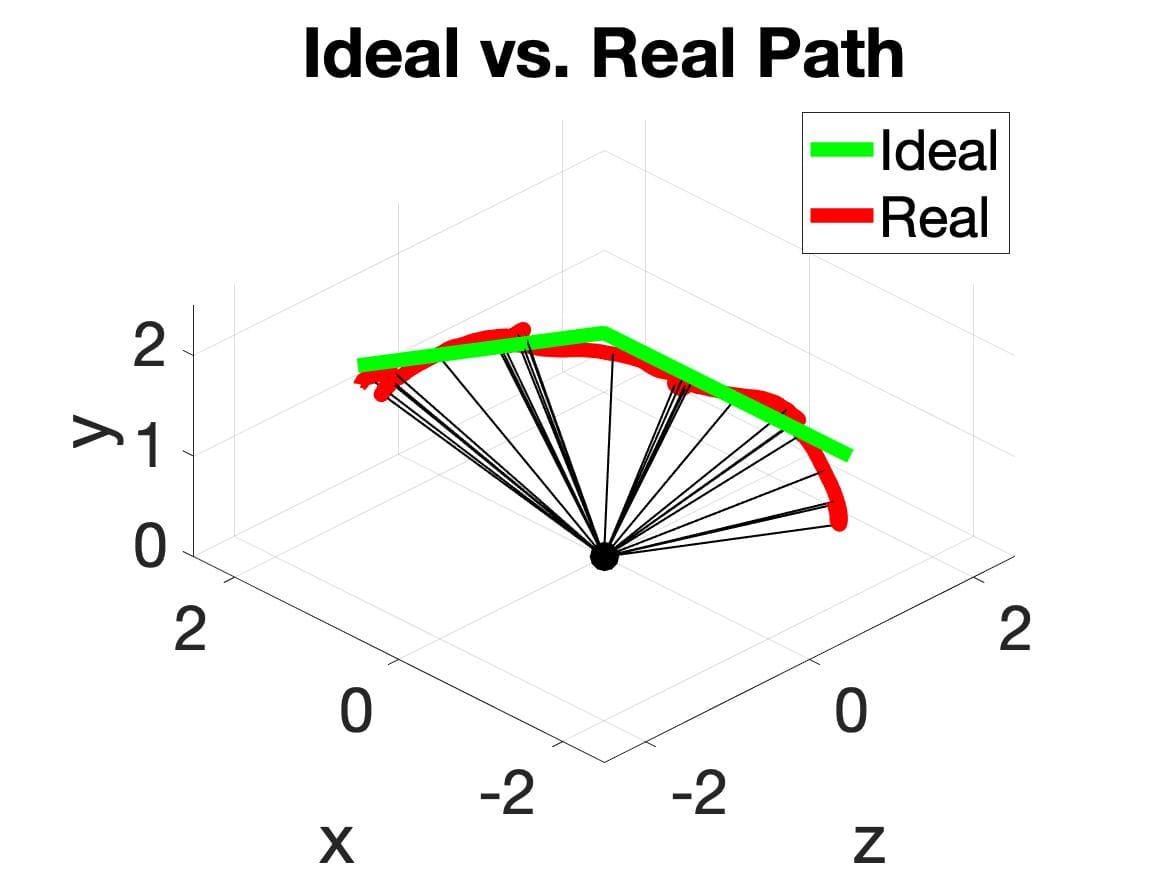}%
\label{fig::pos1.5}}
\subfloat[Position 3m]{\includegraphics[width=0.33\columnwidth]{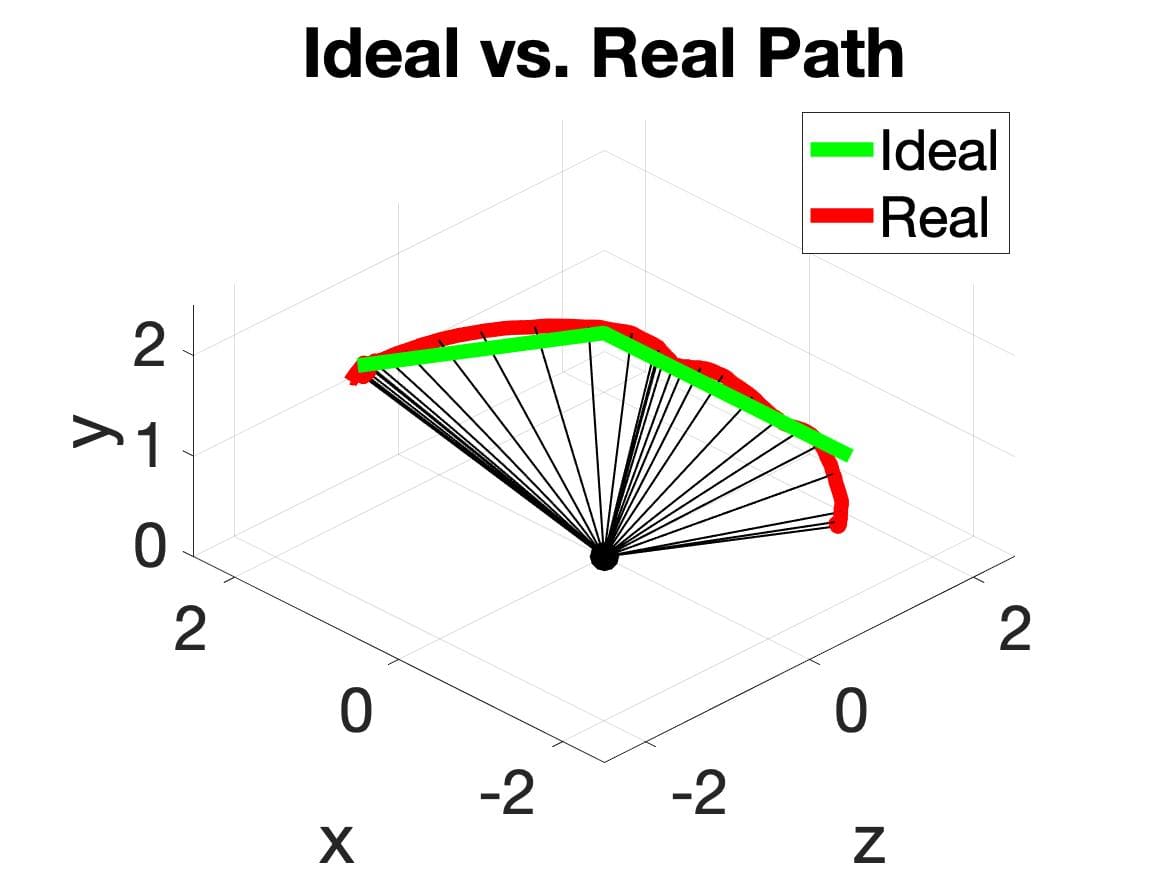}%
\label{fig::pos3}}\\
\subfloat[Velocity 0.2m]{\includegraphics[width=0.33\columnwidth]{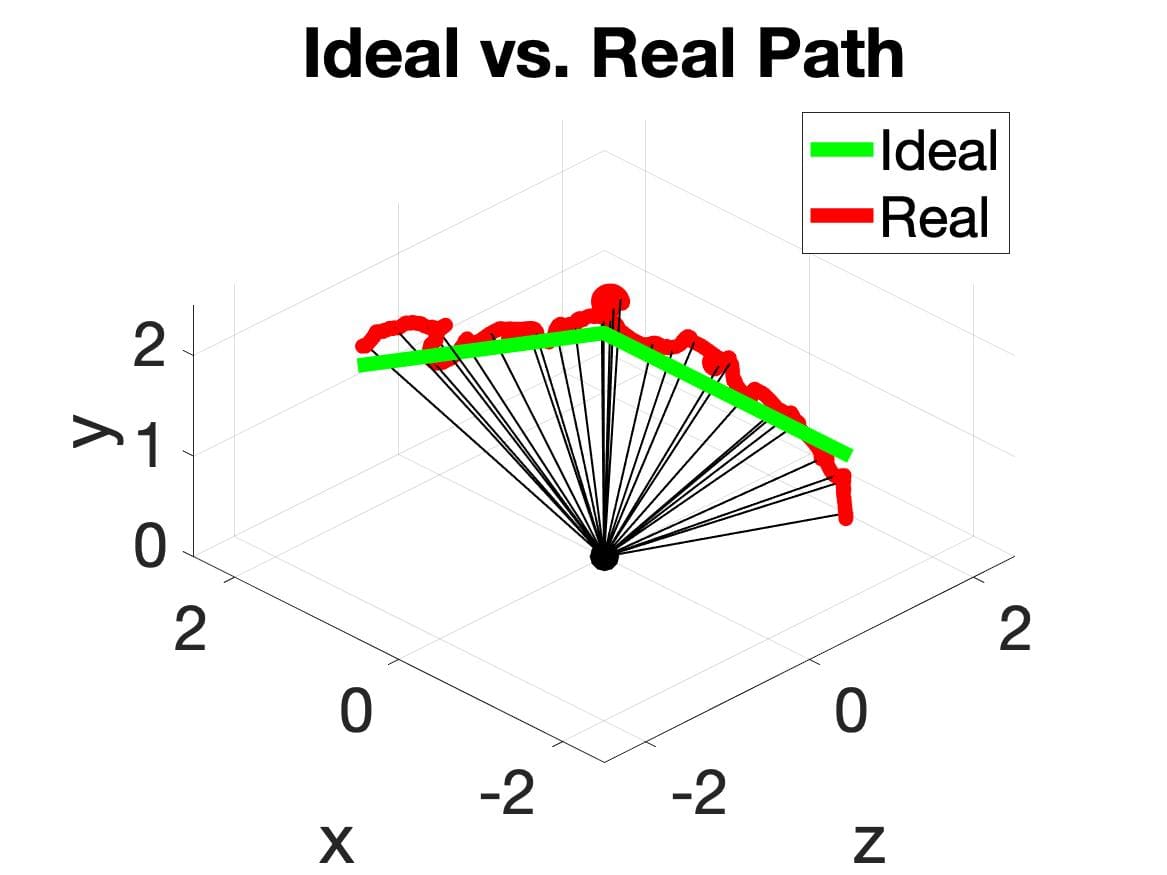}%
\label{fig::vel0.2}}
\subfloat[Velocity 0.5m]{\includegraphics[width=0.33\columnwidth]{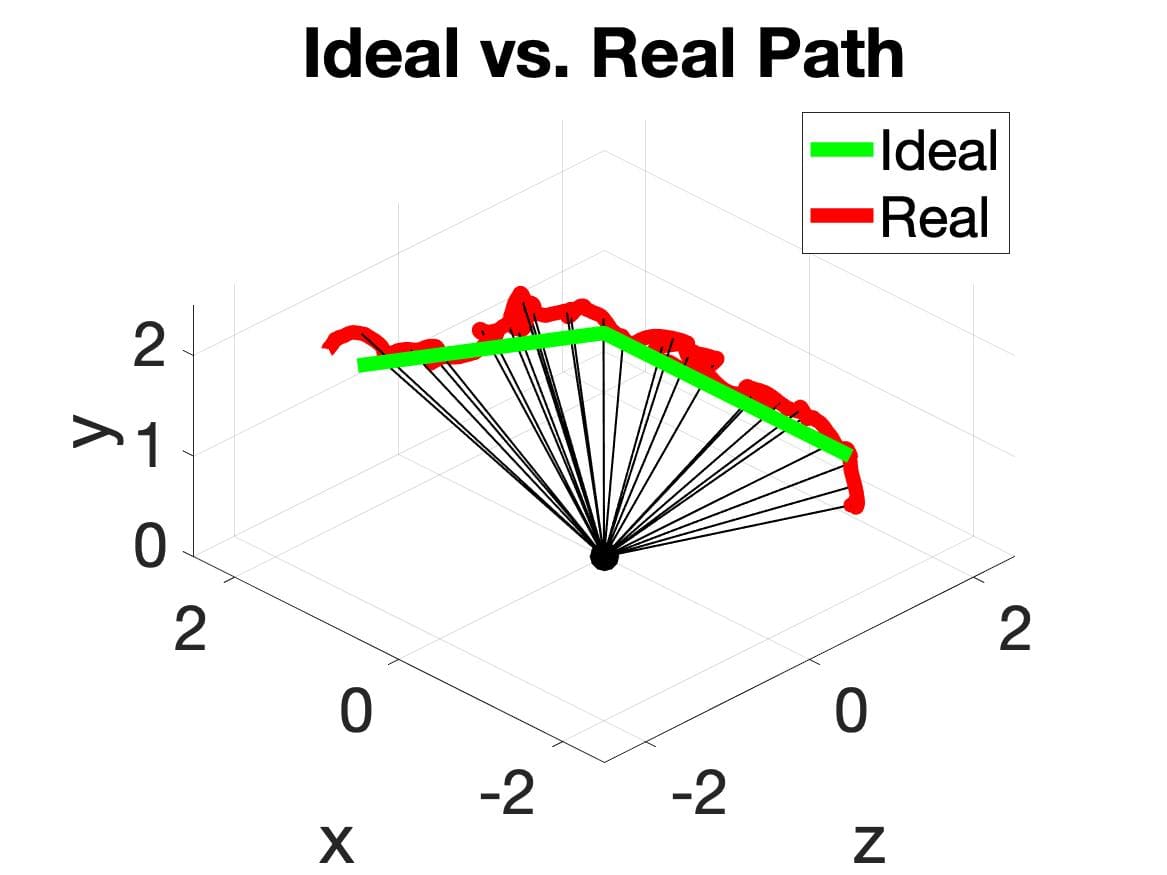}%
\label{fig::vel0.5}}
\subfloat[Velocity 1m]{\includegraphics[width=0.33\columnwidth]{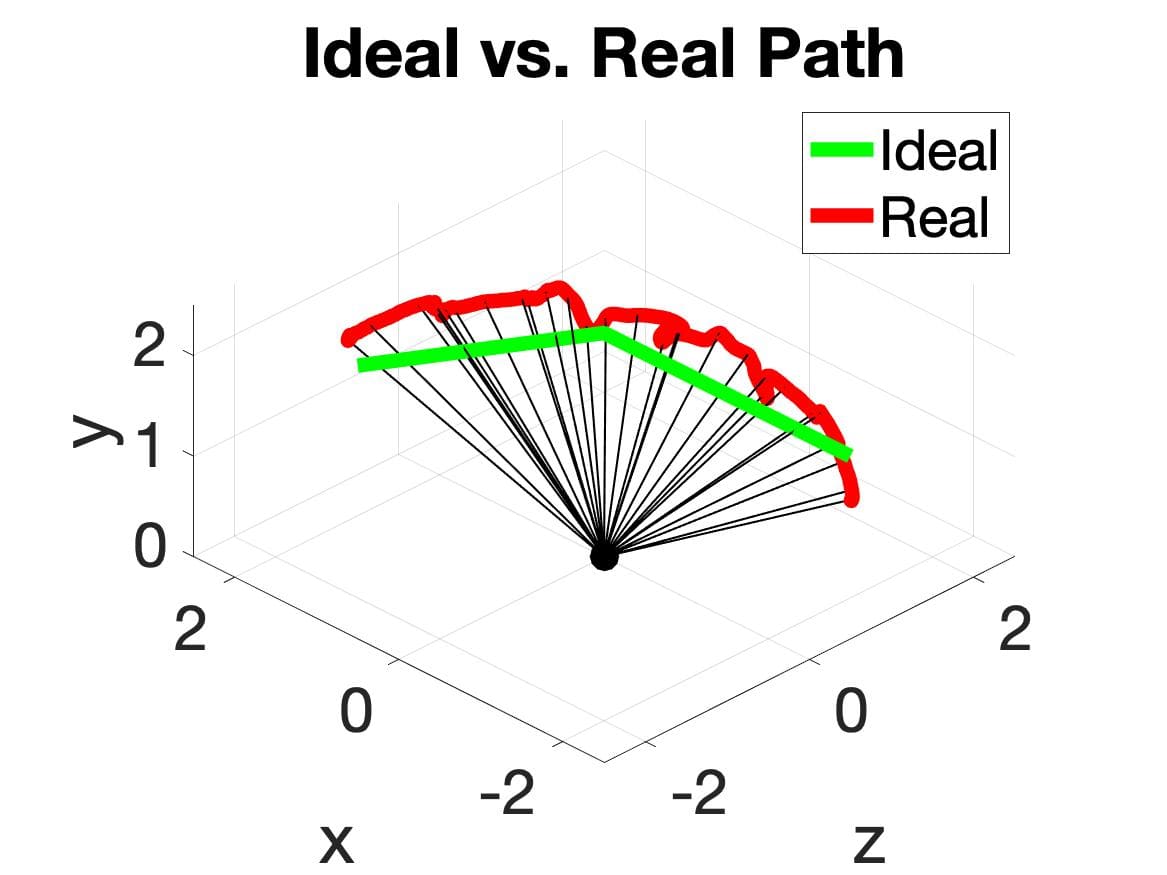}%
\label{fig::vel1}}\\
\subfloat[Velocity 1.5m]{\includegraphics[width=0.33\columnwidth]{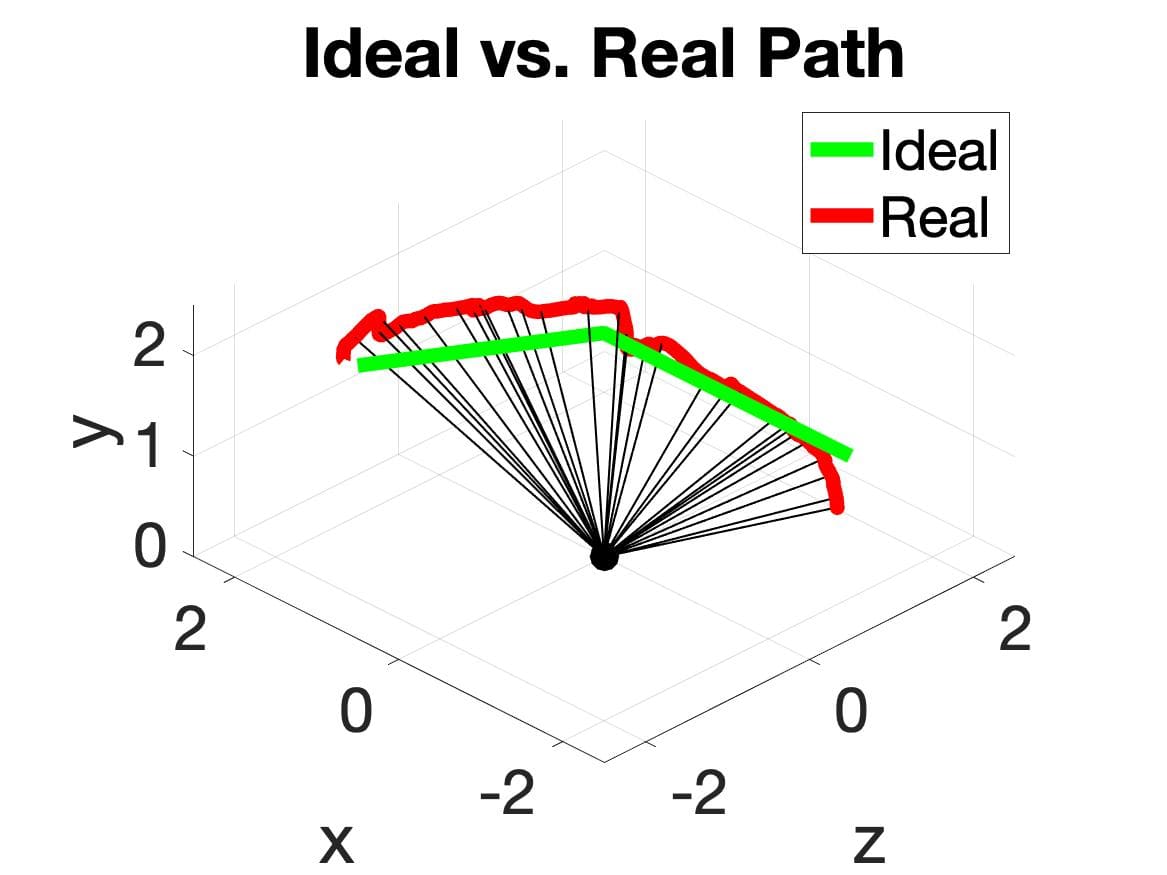}%
\label{fig::vel1.5}}
\subfloat[Velocity 3m]{\includegraphics[width=0.33\columnwidth]{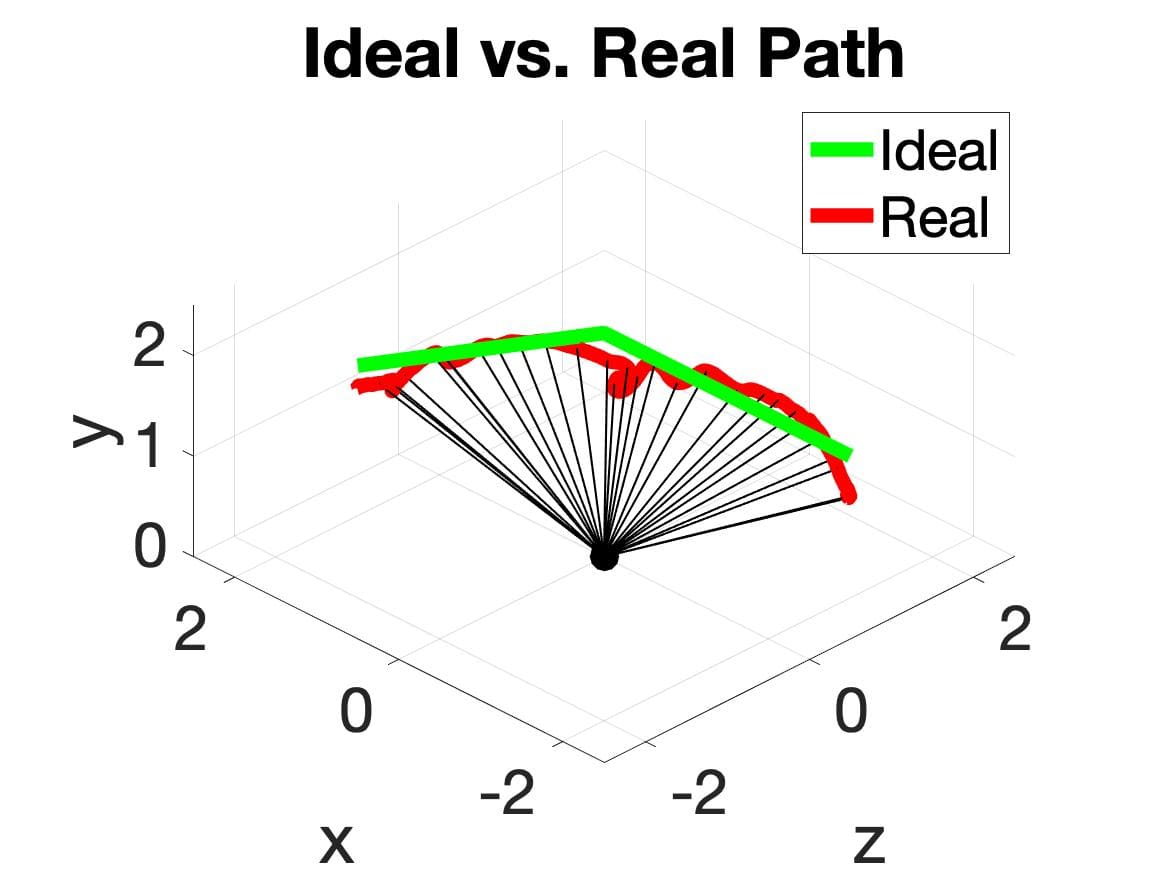}%
\label{fig::vel3}}
\caption{Experimental Results (Adapted from \cite{xiao2019benchmarking})}
\label{fig::prim_experiments}
\end{figure}

Among the total 30 trials, one example trial is randomly selected for each density and each motion primitive and is shown in Fig. \ref{fig::prim_experiments}. 

\begin{figure}[]
\centering
\includegraphics[width=0.5\columnwidth]{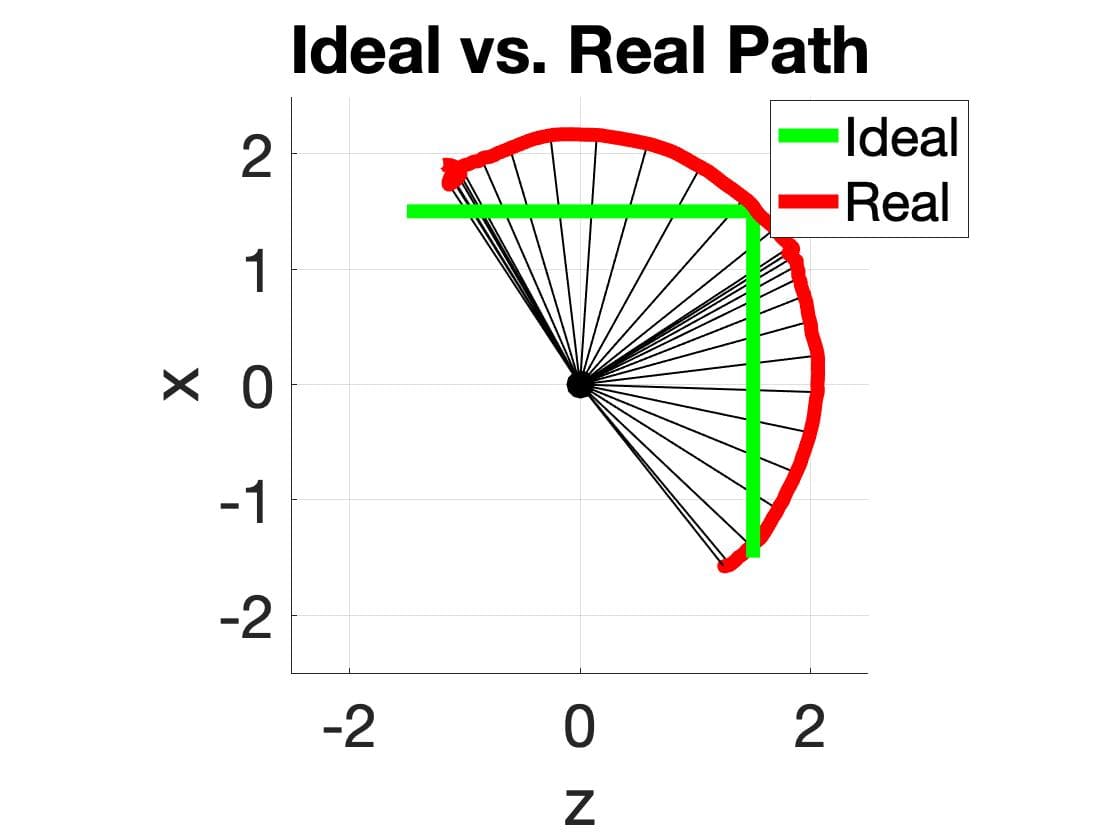}
\caption{Top View of Position Control on 3m Interval (Reprinted from \cite{xiao2019benchmarking})}
\label{fig::pos_3_1_up}
\end{figure}

\begin{figure}[]
\centering
\includegraphics[width=0.5\columnwidth]{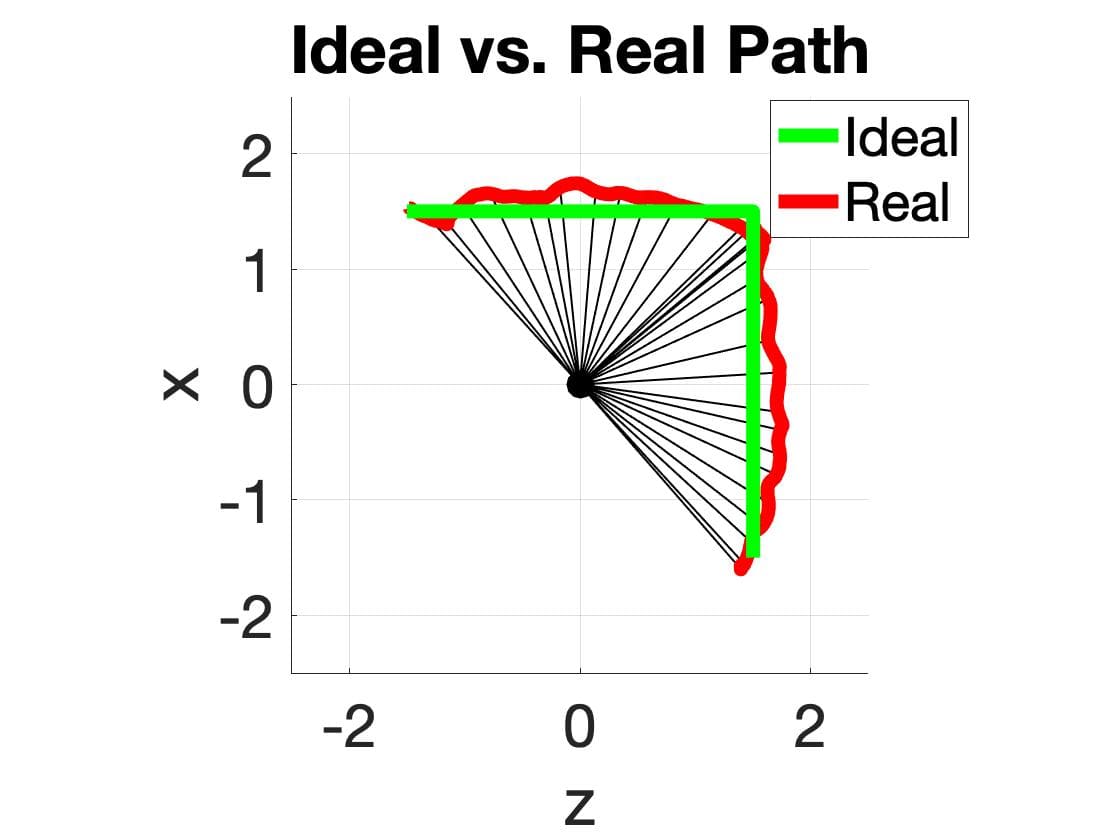}
\caption{Top View of Velocity Control on 3m Interval (Reprinted from \cite{xiao2019benchmarking})}
\label{fig::vel_3_3_up}
\end{figure}

One closer look into the two motion primitives on 3m interval is shown in Fig. \ref{fig::pos_3_1_up} and Fig. \ref{fig::vel_3_3_up}. These clearly demonstrate the problem with position control on path plans with sparse waypoints: since the three tether variables are controlled independently, the trajectory between two consecutive waypoints are non-deterministic. In Fig. \ref{fig::pos_3_1_up} and Fig. \ref{fig::vel_3_3_up}, only three waypoints are used to defined the start (lower right), turn (upper right), and end (upper left) point of the path. Apparently the first two points have the same tether length, therefore the position controller does not change the tether length at all and makes an arc-like trajectory instead of a straight line (Fig. \ref{fig::pos_3_1_up}). The end point has slightly longer tether length due to the increase in elevation, and the UAV executes a similar path. This does not happen in velocity control (Fig. \ref{fig::vel_3_3_up}) due to the coordination through system Jacobian. It is expected that position control may perform better with dense waypoints. 

\begin{figure}[]
\centering
\includegraphics[width=0.6\columnwidth]{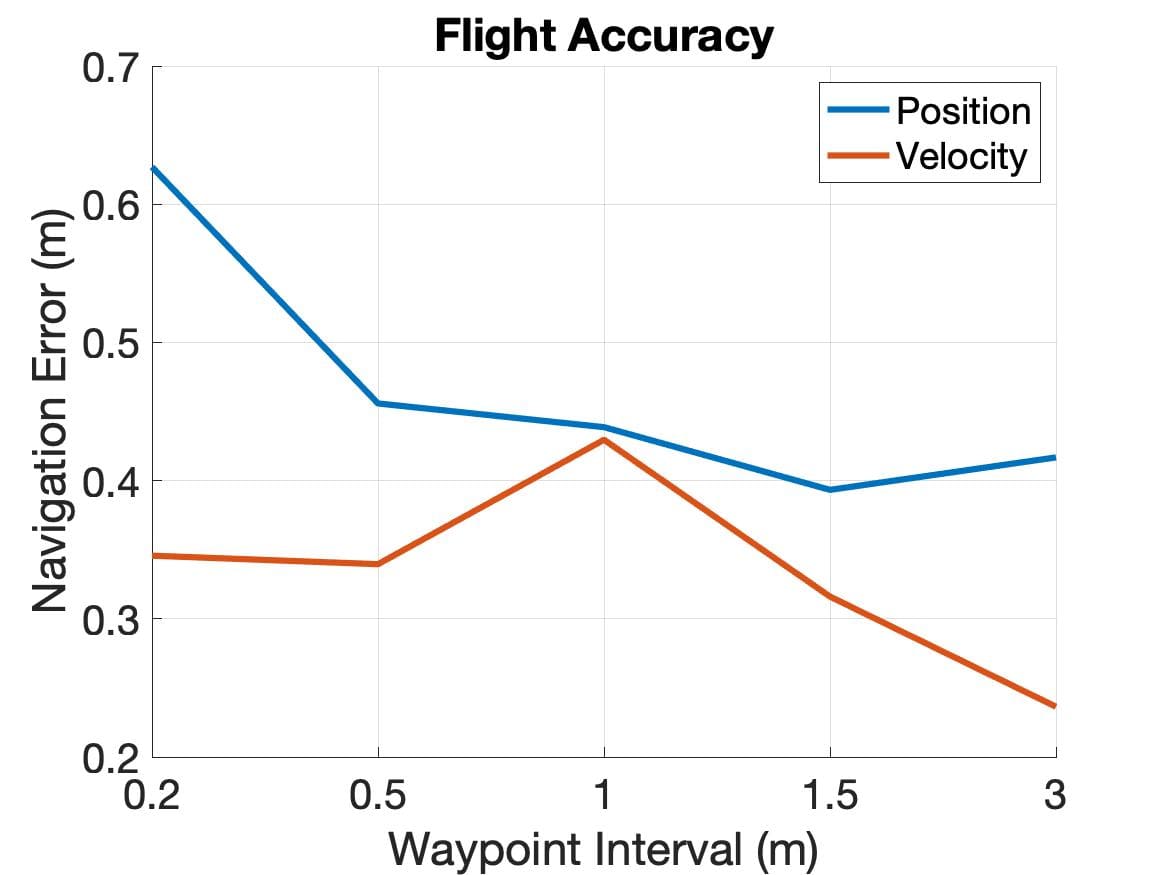}
\caption{Flight Accuracy in Terms of Cross Track Error (Reprinted from \cite{xiao2019benchmarking})}
\label{fig::flight_accuracy}
\end{figure}

\begin{figure}[]
\centering
\includegraphics[width=0.6\columnwidth]{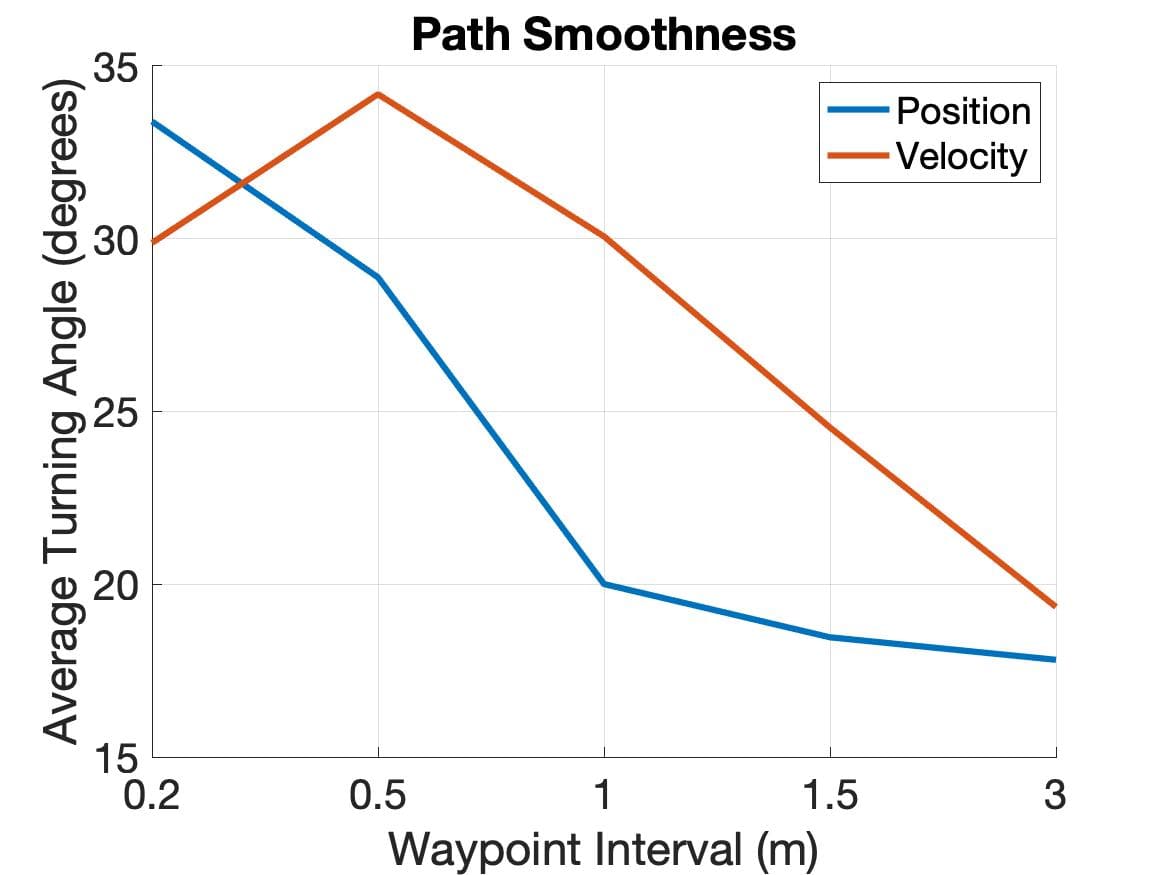}
\caption{Path Smoothness in Terms of Angular Difference (Reprinted from \cite{xiao2019benchmarking})}
\label{fig::path_smoothness}
\end{figure}

The average flight accuracy (error) and path smoothness are analyzed in Fig. \ref{fig::flight_accuracy} and Fig. \ref{fig::path_smoothness} . Flight accuracy is defined as the cross track error between the real and ideal trajectory. Path smoothness is the average angular difference between two vectors connecting two pairs of consecutive waypoints. Path is smoother with sparser waypoints for both motion primitives. This is because when executing sparse waypoints, both motion primitives are aiming at a farther waypoint, instead of focusing on some waypoint in the vicinity. The ``short-sightedness'' caused by dense waypoints will introduce instability to the controller, such as overshoot by trying too hard to converge to the ideal path. With sparse waypoints, on the other hand, both controllers act using ``line-of-sight'', aiming at the path ahead of the UAV and avoiding over-compensation. Flight accuracy for velocity control in Fig. \ref{fig::flight_accuracy} is not very sensitive to waypoint density. One surprising result of flight accuracy is for position control: instead of increasing error with sparser waypoints, error actually decreases. Upon examination of the captured trajectories, it is found out that the expected error caused by the independent control of the three tether variables (Fig. \ref{fig::pos_3_1_up}) is in the same range as the UAV flight tolerance 0.4m. Therefore, even with a path plan with dense waypoints, the expected better accuracy is actually canceled by the large tolerance value around all waypoints. 

\begin{figure}
\centering
\subfloat[Position 0.2m]{\includegraphics[width=0.33\columnwidth]{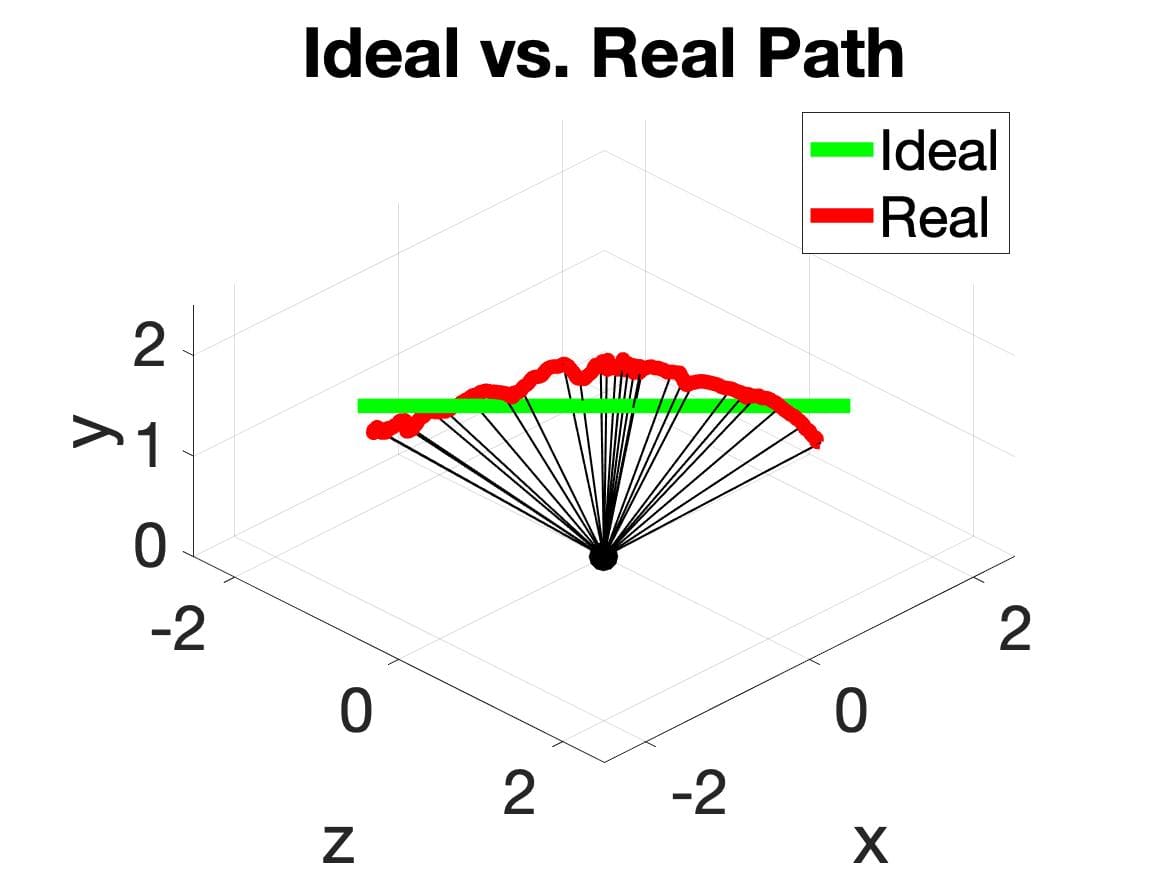}%
\label{fig::pos_02_1_2}}
\subfloat[Position 0.5m]{\includegraphics[width=0.33\columnwidth]{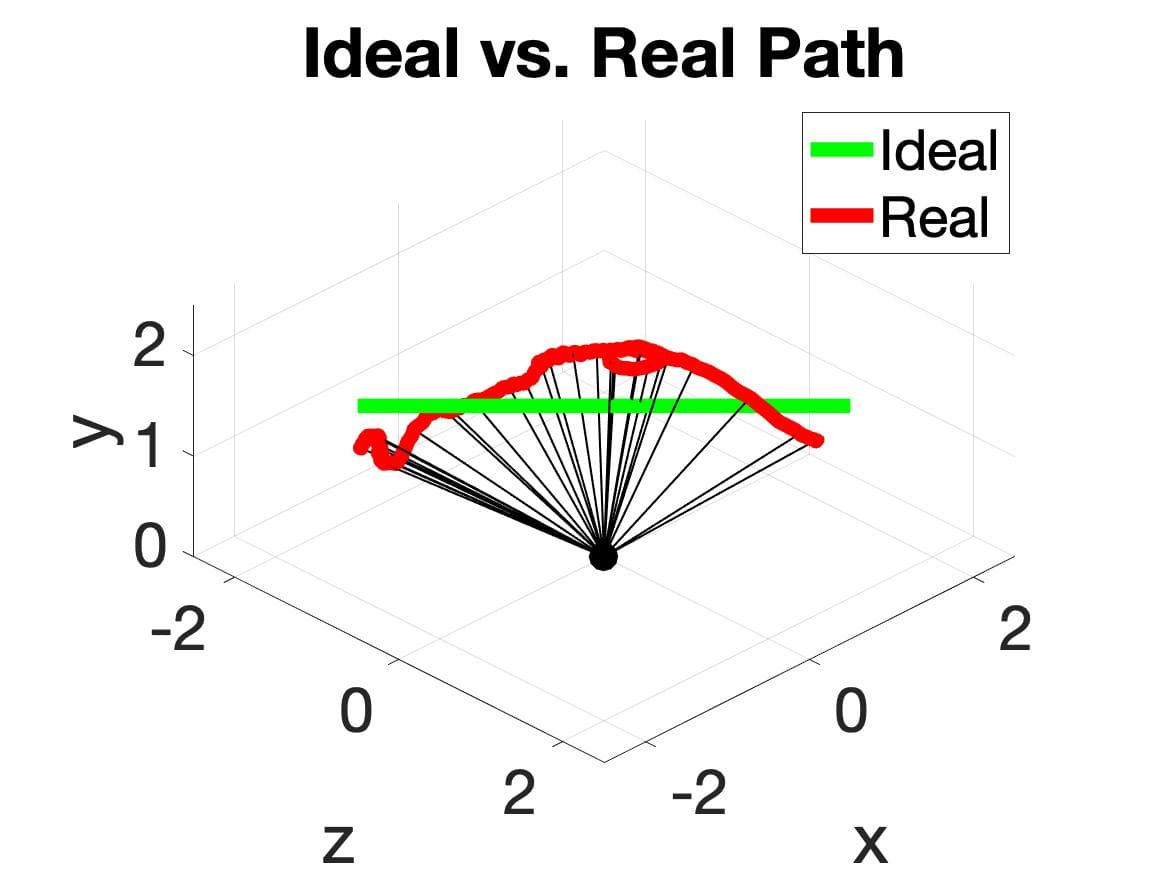}%
\label{fig::pos_05_2_2}}
\subfloat[Position 1m]{\includegraphics[width=0.33\columnwidth]{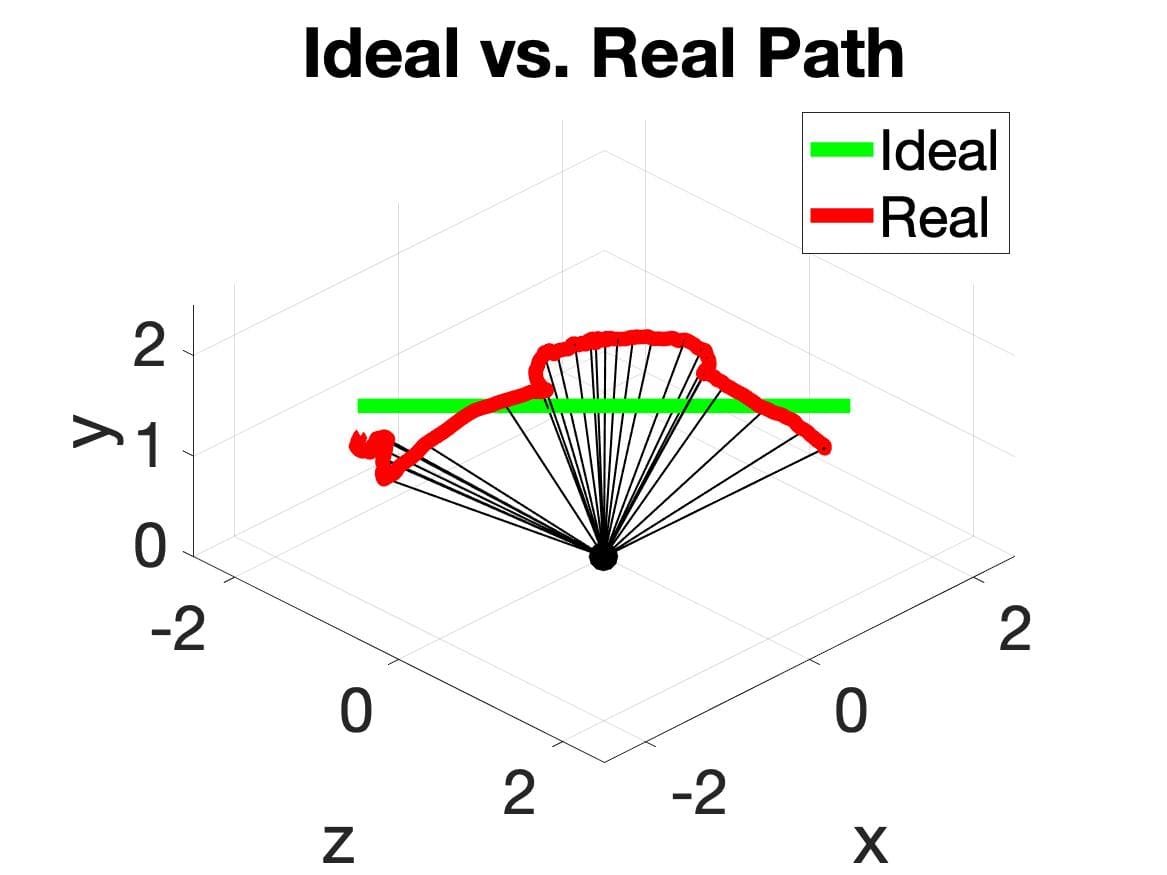}%
\label{fig::pos_1_1_2}}\\
\subfloat[Position 1.5m]{\includegraphics[width=0.33\columnwidth]{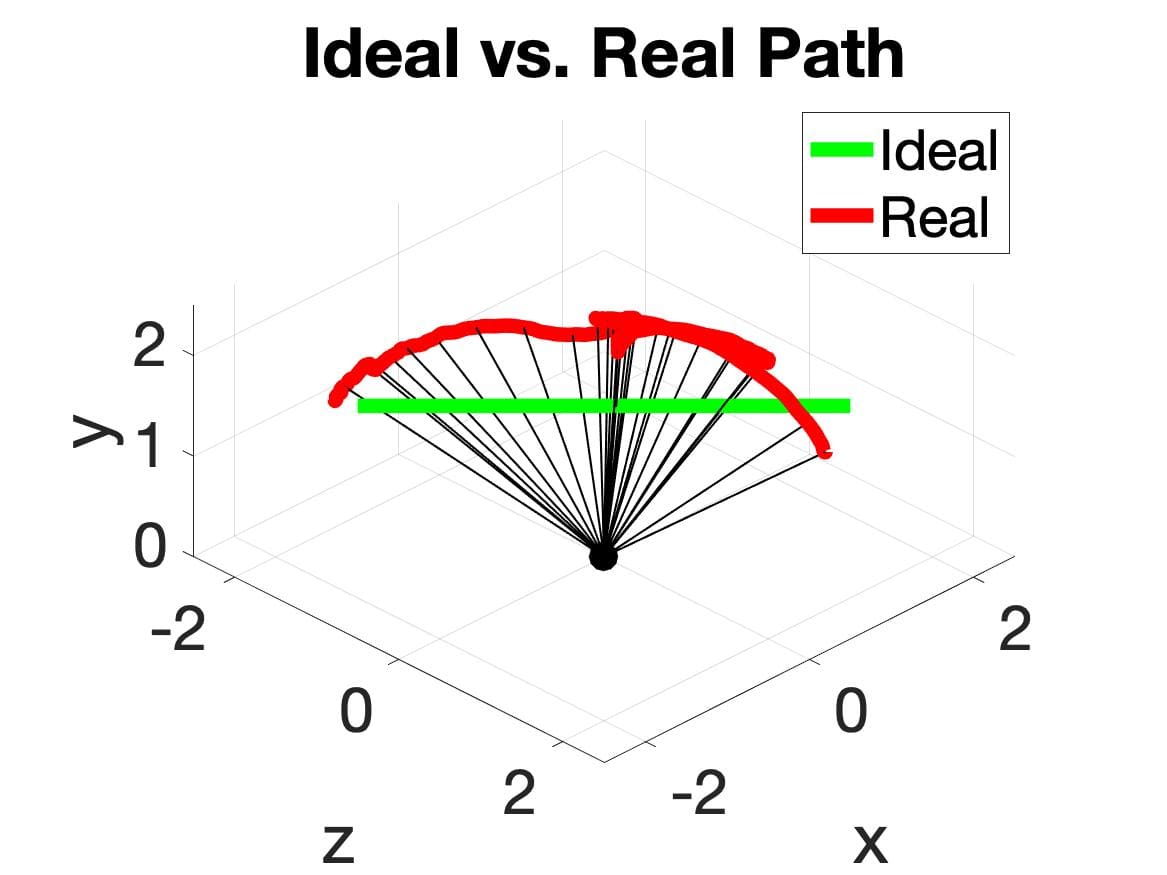}%
\label{fig::pos_15_2_2}}
\subfloat[Position 3m]{\includegraphics[width=0.33\columnwidth]{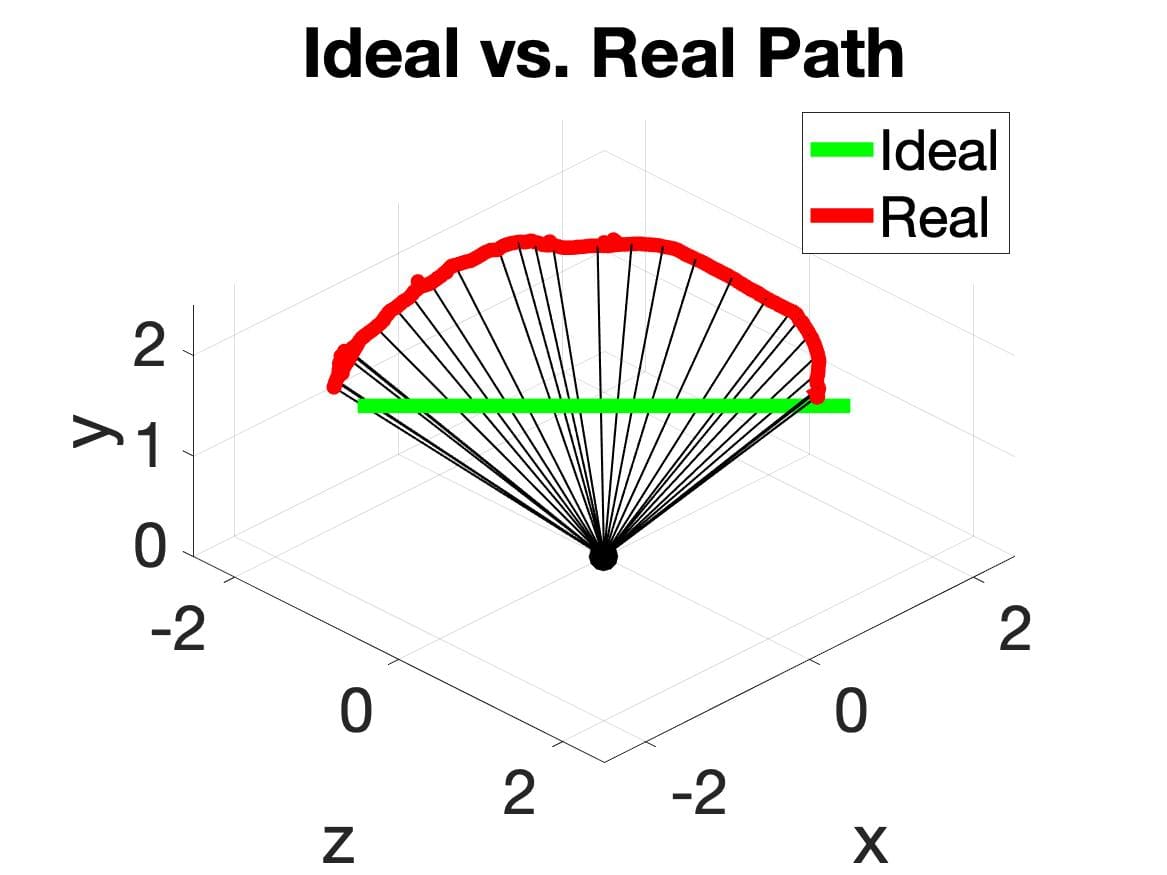}%
\label{fig::pos_3_1_2}}
\caption{Position Control Results (Adapted from \cite{xiao2019benchmarking})}
\label{fig::experiments2}
\end{figure}

Therefore, another set of experiment is conducted, with the focus on benchmarking the effect of waypoint density on position control accuracy. The path is designed to be horizontal and pass diagonally above the tether reel to amplify the effect of incoordination between control variables. Due to the singularity above the tether reel of the Jacobian matrix of velocity control, the UAV is inevitably trapped at the singularity when coming close to it. For velocity control, regions above the tether reel with 90\degree~elevation and indeterministic azimuth need to be avoided. Three position control trials are executed for each of the five waypoint densities, with one trial shown in Fig. \ref{fig::experiments2}.

\begin{figure}[]
\centering
\includegraphics[width=0.6\columnwidth]{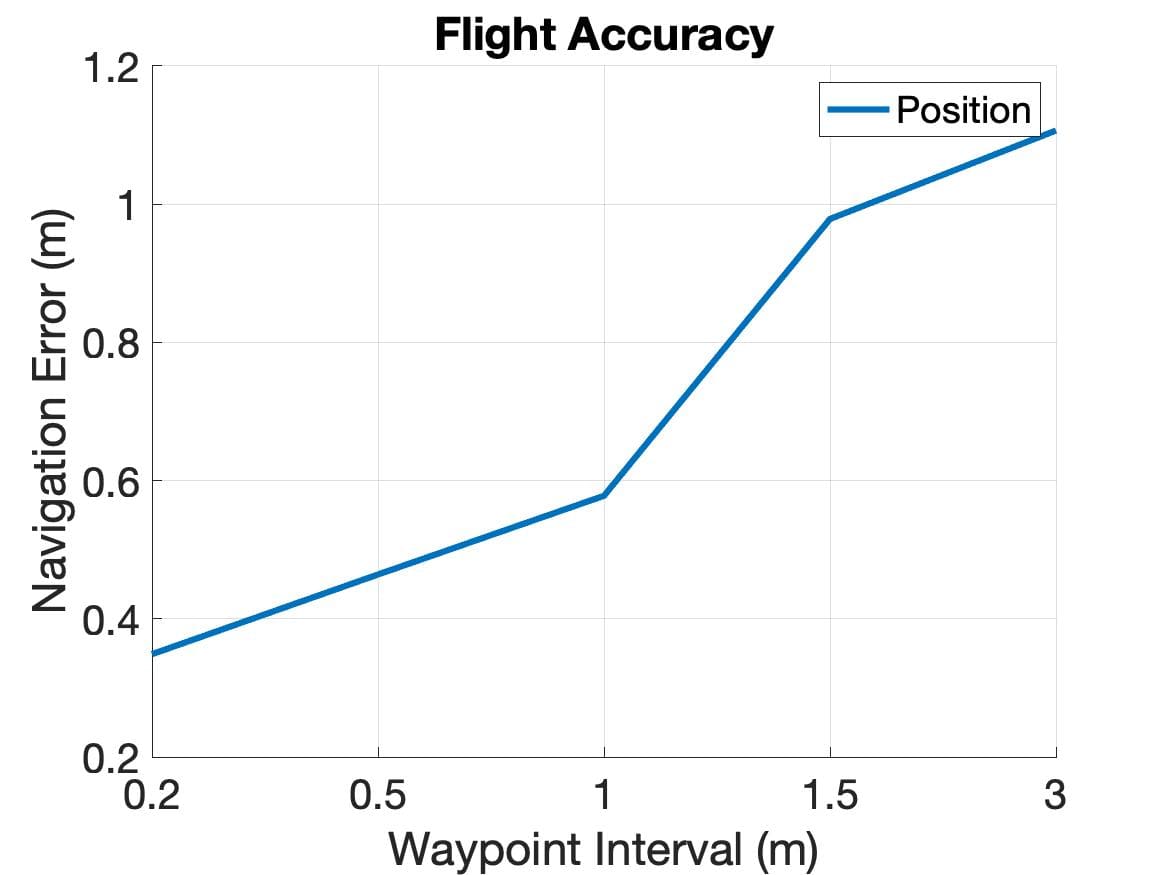}
\caption{Flight Accuracy for Path 2 (Reprinted from \cite{xiao2019benchmarking})}
\label{fig::flight_accuracy2}
\end{figure}

In this set of experiments, it clearly shows that position control accuracy decreases with sparser waypoints (Fig. \ref{fig::flight_accuracy2}). From left to right in Fig. \ref{fig::experiments2}, the UAV deviates more and more from the ideal path, due to the lack of guidance between two consecutive waypoints. In the extreme case on the right hand side where only two waypoints denote the start and end position of the path, the UAV forms a semicircle-shaped trajectory instead of the intended straight line path. 

From the results of both experimental sets, position control works better with dense waypoints in terms of flight accuracy. This is because of the independent control over the three tether variables between waypoints. And denser waypoints provide extra guidance in between. However, denser waypoints also introduce jittery motion of the UAV since the shortsightedness causes overshoot so the path smoothness is no longer guaranteed. Proper waypoint density should be sufficiently dense to constrain the nondeterministic motion between waypoints while sparse enough to generate a smooth path. On the other hand, velocity control's accuracy is not very sensitive to the waypoint density, thanks to the coordination among the three tether variables using system's inverse Jacobian matrix. Similar to position control, smoothness of the path will be deteriorated by increasing waypoint density. Therefore, when using velocity control, sparse path plan is desirable as long as the critical points on the path is uniquely described by a minimum amount of waypoints. However, velocity controller could be trapped by singularity above the tether reel center, causing certain path to be not executable. Those areas need to be avoided when using velocity control only. An alternative approach is to use a composite controller which mostly uses velocity control but switches to position control when the UAV comes close to singularity. 

\subsection{Summary of Tether-based Motion Primitives Experiments}
Both motion primitives, position control and velocity control, are implemented on a tethered UAV in a MoCap studio, in order to validate the hypothesis that tethered UAV can achieve free-flight in 3D Cartesian space and benchmark their control performance with respect to different path plans. Path smoothness prefers paths with sparse waypoints for both motion primitives. However, position control's flight accuracy depends on proper waypoint density, which can provide extra guidance to minimize motion error between waypoints due to the independency of the three sub-controllers. The sparsity of waypoints is not an issue for the velocity control, thanks to the controller coordination enabled by the Jacobian matrix. But singularity exists for the velocity controller, where elevation angle is 90\degree~and azimuth is impossible to determine. Areas close to the top of the tether reel should be avoided using velocity control. 

\section{Tether Planning and Motion Execution Experiments}
Experiments for the tether planning and motion execution approach described in Chapter \ref{chapter::low_level} are presented.\footnote{Detailed experimental results were presented and published in previous work \cite{xiao2018motion}.}

\subsection{Hypothesis and Metrics}
The hypothesis for the experiments of the tether planning and motion execution is \emph{the proposed approach can enable tethered flight in obstacle-occupied environments}. The metric used is \emph{success/fail} of path execution with tether and \emph{flight accuracy/navigation error}, in term of cross track error between the actual and planned UAV trajectory. The reduction (percentage) in reachable space and flight accuracy with respect to the number of contact point(s) are also analyzed and discussed. 

\subsection{Experiments}
The purpose of the experiments is proof of concept of our two tether-handling motion planning algorithms, reachable space reduction by ray casting and contact point(s) planning and relaxation, and the usage of the motion executor. By running experiments on physical robots, we wanted to show that our motion planners can navigate the UAV between two points in the corresponding free space of each planner. The trial completion was determined based on the UAV's onboard localization. By running our two motion planning algorithms, we also wanted to demonstrate different reachability sets achievable by the two planners. It was computed as a percentage of reachable spaces in the whole map by offline computation based on the obstacles in the map. Finally, navigation accuracy in terms of cross track error was presented by comparison between planned paths and executed paths. The latter was captured by a ground truth motion capture (MoCap) system. 

\begin{figure}[]
\centering
\includegraphics[width=1\columnwidth]{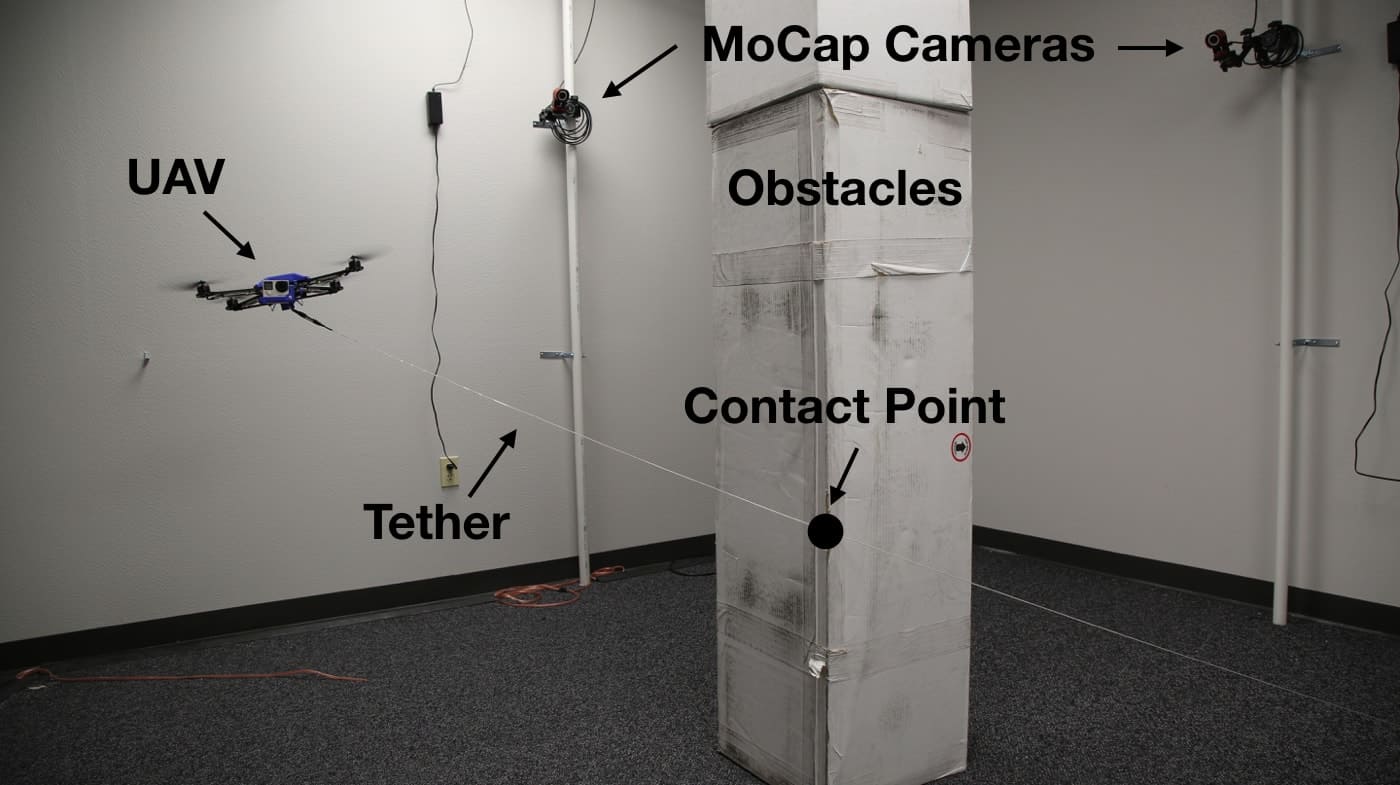}
\caption{UAV flying with one tether contact point in the MoCap studio (reprinted from \cite{xiao2018motion})}
\label{fig::mocap_mp}
\end{figure}

Our experiments were conducted in a motion capture studio to capture motion ground truth. The studio is equipped with 12 OptiTrack Flex 13 cameras running at 120 Hz. The 1280$\times$1024 high resolution cameras with a 56\degree~Field of View provide less than 0.3mm positional error and cover the whole 3.3$\times$3.3$\times$2.97m space. The high number of cameras guarantee that the UAV could be captured even if the markers were blocked by the obstacles from some cameras. We used obstacles made of cardboard, which formed a 0.33$\times$0.33$\times$0.297m vertical shaft and located in the middle of the experimental environment. The choice of cardboard was to guarantee safe tether contact. This configuration of obstacles blocked most direct passages between different regions in the map, and was particularly difficult for a tethered UAV to navigate through. Fotokite Pro was used as our tethered UAV. The online motion executor executed the offline motion plan from the two algorithms. During the physical tests, the acceptance radius $R_{acc}$ was set to 0.4m. This is the best localization accuracy achievable by Fotokite's sensory feedback measured by experiments. Fig. \ref{fig::mocap_mp} shows the tethered UAV flying in the MoCap studio. 

\begin{figure}[]
\centering
\subfloat[Reduced Reachable Space by Ray casting]{\includegraphics[width=0.5\columnwidth]{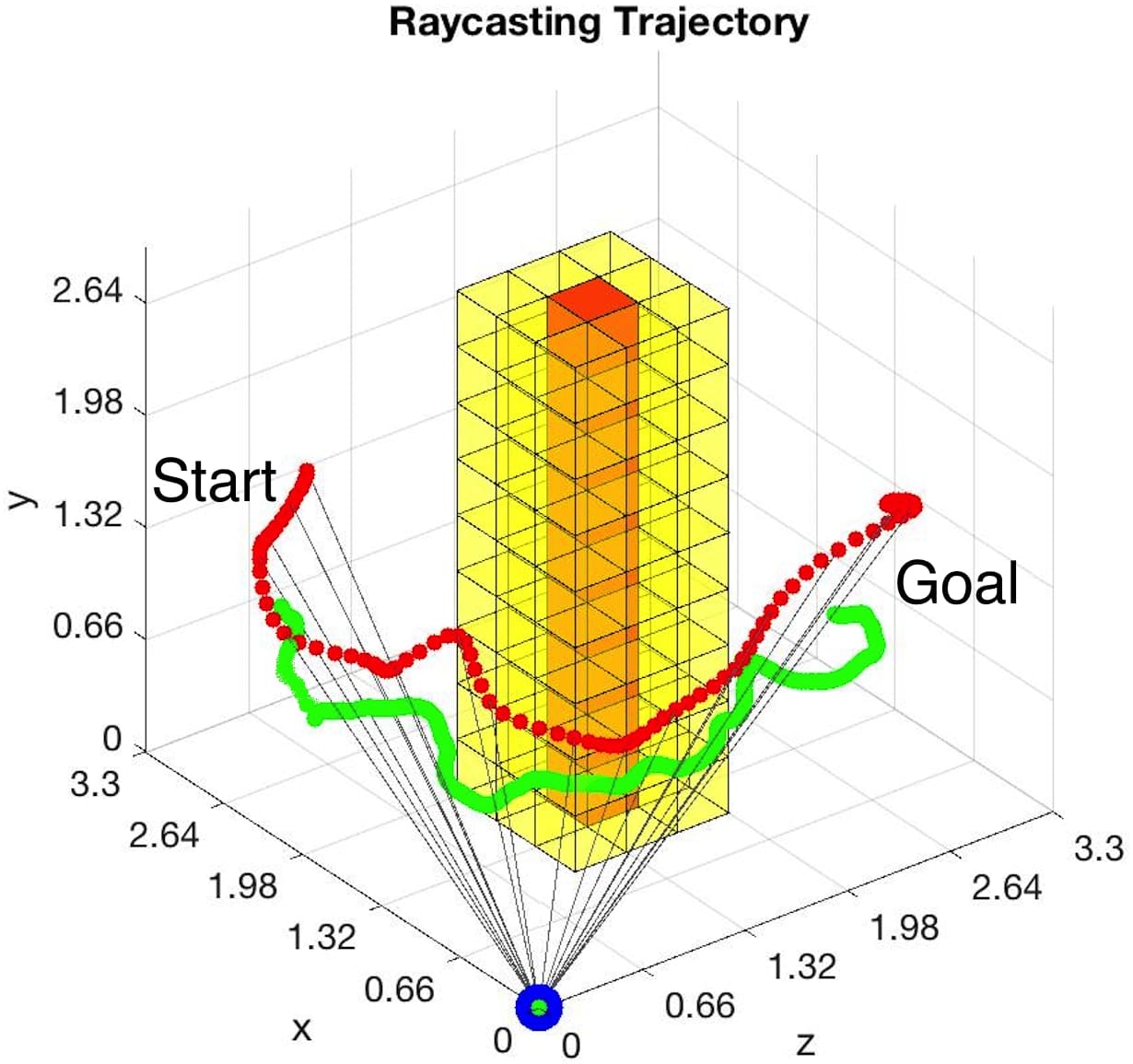}%
\label{fig::raycasting_traj}}
\subfloat[Contact Point with Relaxation]{\includegraphics[width=0.5\columnwidth]{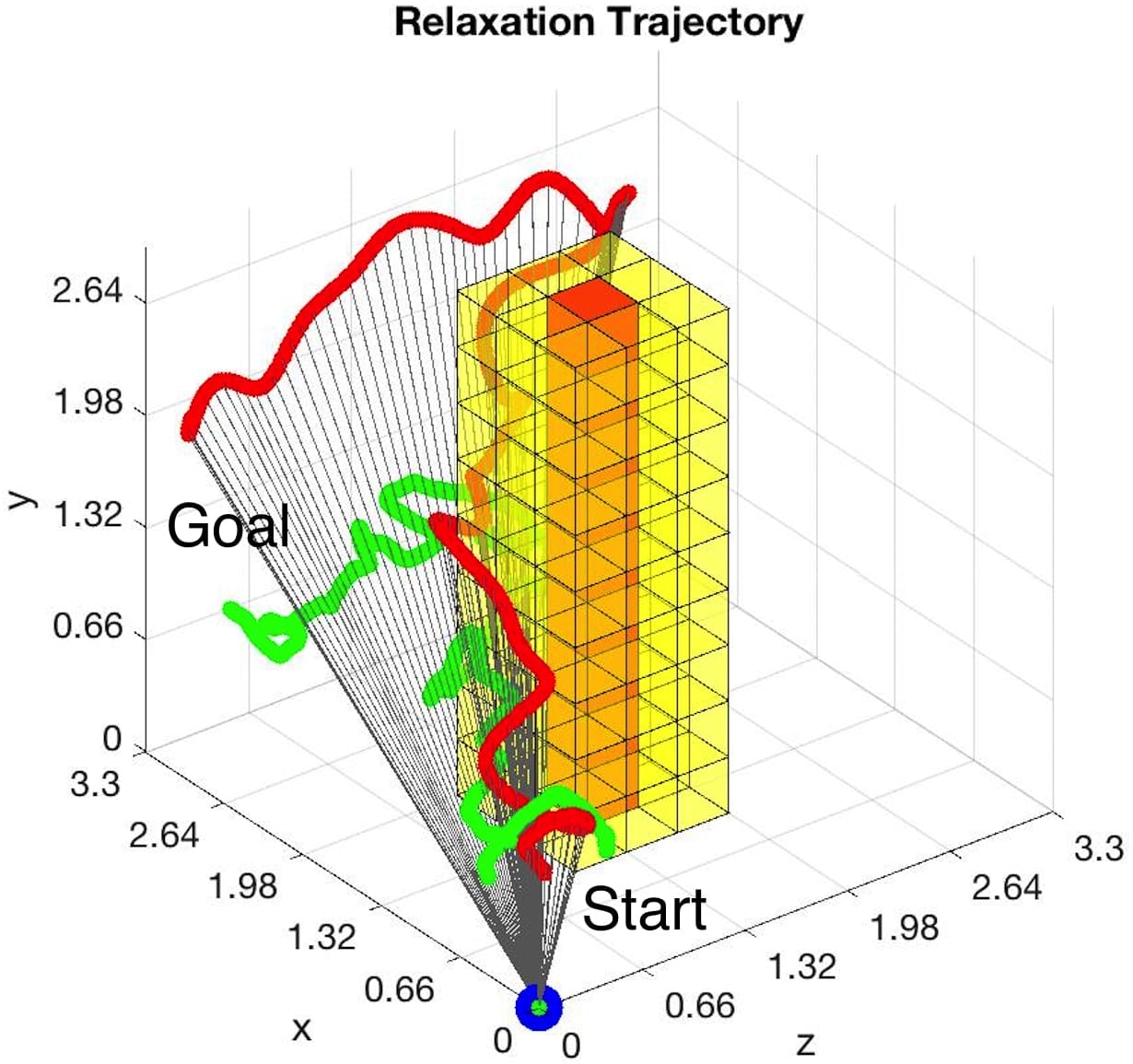}%
\label{fig::relaxation_traj}}\\
\subfloat[Two Contact Points]{\includegraphics[width=0.5\columnwidth]{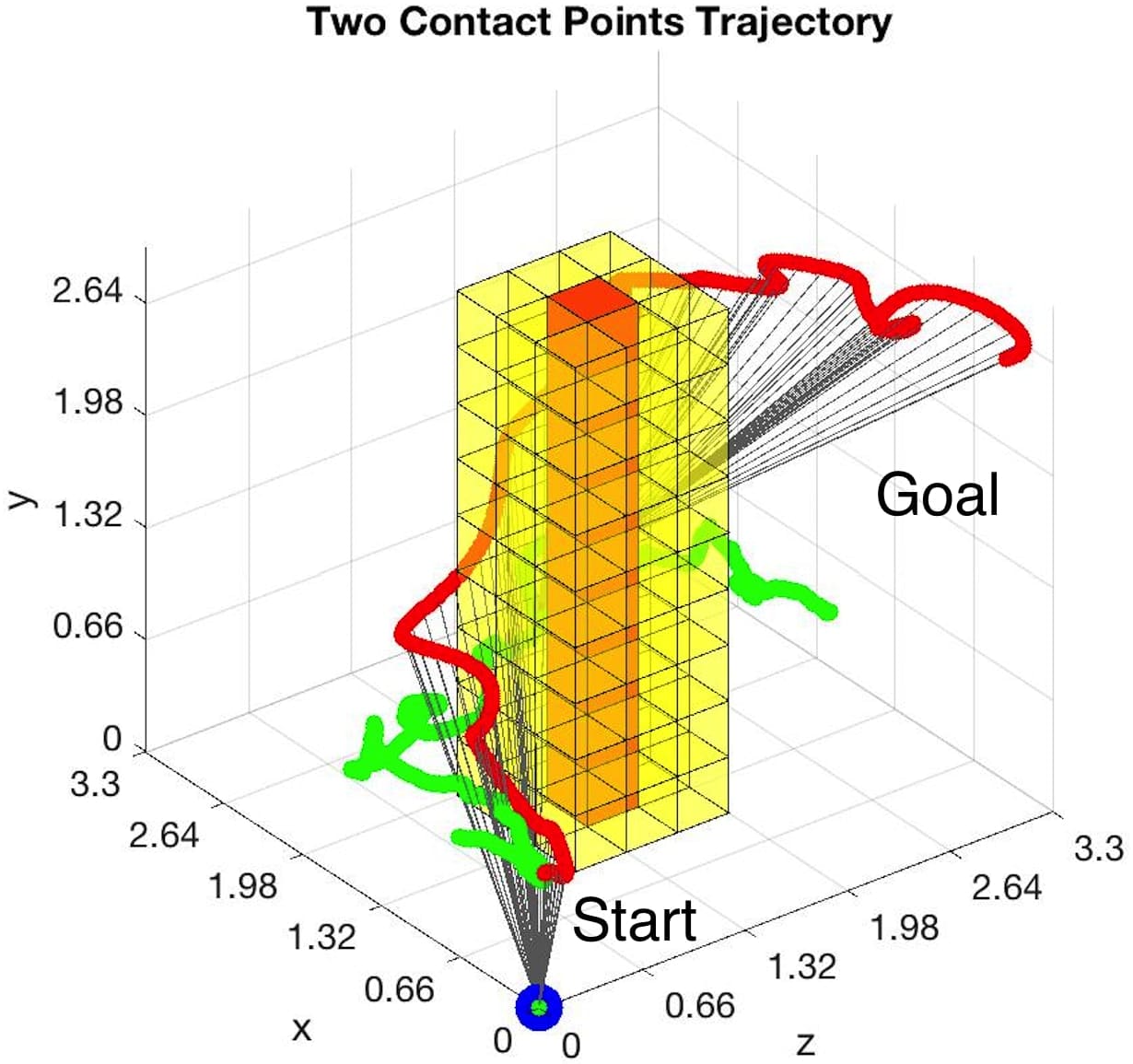}%
\label{fig::two_contact_traj}}
\caption{Three Different Paths Planned (Red) and Executed (Green): Red voxels represent the obstacles and yellow voxels are the occupied spaces due to map inflation. Red path is the off-line computed motion plan and green one is the actual path captured by OptiTrack motion capture system (adapted from \cite{xiao2018motion}).}
\label{fig::results}
\end{figure}

In order to validate our motion planning algorithms, we conducted three sets of experiments on the tethered UAV: 

\begin{itemize}
\item Moving in free space after reachable space reduction using ray casting (Fig. \ref{fig::raycasting_traj})
\item Returning to free space by relaxing previously formed contact point (Fig. \ref{fig::relaxation_traj})
\item Entering non-reachable space with a straight tether by planning two contact points (Fig. \ref{fig::two_contact_traj})
\end{itemize}

Since the two different motion planners are dealing with different configuration spaces, i.e. reduced and original reachable spaces, we cannot replicate the same navigation task (same \textit{start} and \textit{goal}) for both of them. For the first set of experiments, we manually chose pairs of \textit{start} and \textit{goal} in the reduced reachable space. The obvious direct paths between the pairs were not executable due to the tether. The ray casting motion planner needed to come up with an alternative path to circumvent the obstacles to remain a straight tether. For the second set of experiments, we manually chose a tuple of \textit{(start, middle point, goal)} in the original free space. The \textit{middle point} located at a position where one contact point was necessary to reach. The \textit{Goal} located at a position where no contact was necessary. So the robot had to form and then relax the contact point to reach the final target during the flight. For the third set of experiments, we manually chose a tuple of \textit{(start, middle point, goal)}. The \textit{middle point} located at a position where one contact point was necessary to reach. The \textit{Goal} located at a position where two contacts were necessary. So the robot had to form two contact points in a row to reach the final target during the flight. 

Based on the given map, we obtained two different reachable spaces from the two motion planners. The ray casting method reduced navigable space from the original free space while contact(s) point planning and relaxation kept the whole free space intact. 

We totally performed 21 trials. Two trials were discarded due to the UAV platform hardware failure and one was discarded due to the UAV flying out of the range of the MoCap system. We obtained 18 planned paths with way points and contact points (CPs for ray casting were simply tether reel) and corresponding 18 executed paths captured at 120Hz, six trials for each set. 

\subsection{Discussions}
The results of the experiments using the two tether planning algorithms along with the motion executor are discussed here. 

\subsubsection{Comparison of the Two Algorithms:}

Ray casting works in post-reduction free spaces and the UAV cannot reach spaces blocked by ray casting. Contact point(s) planning can navigate to spaces which are not reachable with a straight tether. Tether can be properly relaxed when UAV returns to original free spaces. Multiple contact points could be formed and handled. For this particular set up, ray casting can reach 60\% of the whole free space, and contact point(s) planning can reach 100\%. A 40\% reduction of reachable space was observed for ray casting to maintain a straight tether. Contact point planning has greater reachability since the UAV is de facto tetherless, but tether contact may not be acceptable in all domains. There is an open issue as to whether the tether would break or would damage the environment. Ray casting has an extra complexity of $\mathcal{O}(o)$. Contact point(s) planning has $\mathcal{O}(po)$ due to the extra work load to plan and relax contact points. 

\subsubsection{Insights on Implementation:}

All 18 trials were completed based on the UAV sensor feedback. However, the onboard localization error accumulates during flight, so position estimation is not precise. Fig. \ref{fig::results} shows three example trials. Fig. \ref{fig::raycasting_traj} shows the execution of the path generated by ray casting method. Fig. \ref{fig::relaxation_traj} and Fig. \ref{fig::two_contact_traj} use contact point(s) planning and relaxation. To be noticed is that the tether can pass through yellow voxels (inflation) and contact points can only be formed on the surfaces/edges of the red voxels (obstacles). In Fig. \ref{fig::raycasting_traj}, originally free spaces behind the obstacles are blocked by ray casting, so the robot has to forgo the short path behind the obstacles and circumvent from the front in order to maintain a straight tether through the whole flight. Fig. \ref{fig::relaxation_traj} shows the robot firstly navigates to the far end of the map, where the tether has to touch the obstacles. One contact point is planned, which is thereafter relaxed, since the robot flies back to the non-contact space and reaches the final destination with a straight tether. As we can see, the navigation accuracy decreases significantly after making the contact. In Fig \ref{fig::two_contact_traj}, two contact points are planned along the way. Although the last portion of the path is reachable directly from the tether reel, it still keeps the two contact points since obstacles are confined within the triangle formed by the waypoint, current and last contact points (Fig. \ref{fig::no_relaxation1}). 

\begin{figure}[]
\centering
\includegraphics[width=0.745\columnwidth]{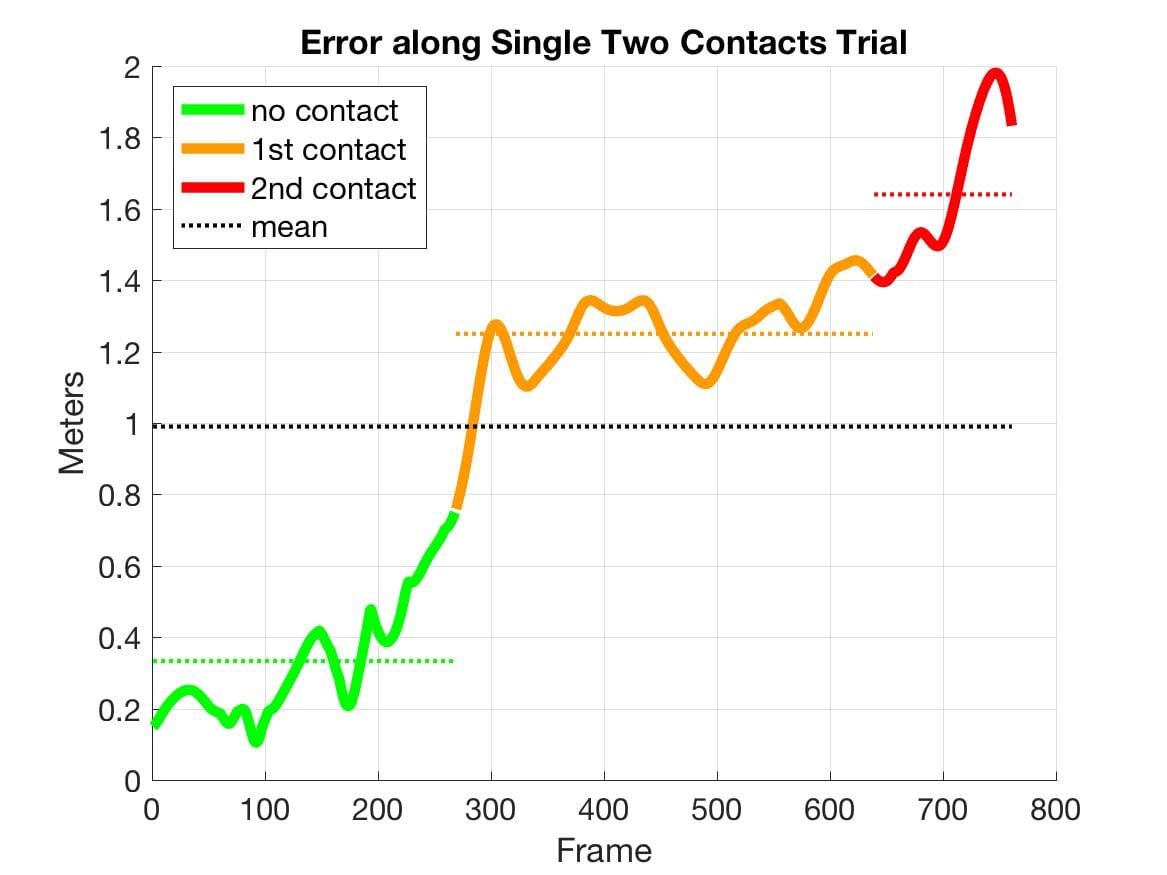}
\caption{Navigational Error along an Example Trial with Two Contact Points (Reprinted from \cite{xiao2018motion})}
\label{fig::contact_errors}
\end{figure}

An examination on all 18 trials (Tab. \ref{tab::errors}) of the accuracy in terms of cross track error shows that contact point(s) planning has a larger error, which is even more significant with two contact points. We presume that this is due to the error introduced by contact point position, which will accumulate with increased number of contacts made. We further investigate this presumption by looking into the segmented accuracy for different contact points (Fig. \ref{fig::contact_errors}). When no contact is made, the accuracy (0.4198m) is comparable to the ray casting result in Tab. \ref{tab::errors}. The average error increases to 1.3602m at one contact and 1.7634 at two. The increased positional error is because of two reasons: (1) Due to the lack of contact point positional feedback, the actual contact point may differ from the original motion plan at initial touch. This will shift the navigation space in the next region. (2) The assumption of fixed contact point may not hold all the time, so the contact point will move slightly during flight. This process adds random noise into the system. These two sources of error will accumulate and further deteriorate the navigation accuracy. The three stages of error profile in Fig. \ref{fig::contact_errors} clearly indicate the impact of increased number of contact points: the navigational precision is less satisfactory when more contact points are formed. 

\begin{table}[]
\centering
\caption{Mean Cross Track Errors (meter) (Reprinted from \cite{xiao2018motion})}
\label{tab::errors}
\begin{tabular}{|c|c|c|c|}
\hline
              & \textbf{Raycasting} & \textbf{\begin{tabular}[c]{@{}c@{}}1 contact \\ w/ relaxation\end{tabular}} & \textbf{\begin{tabular}[c]{@{}c@{}}2 contacts \\ w/o relaxation\end{tabular}} \\ \hline
\textbf{1}    & 0.6963              & 1.0005                                                                      & 0.9900                                                                       \\ \hline
\textbf{2}    & 0.5644              & 0.9587                                                                      & 0.9933                                                                       \\ \hline
\textbf{3}    & 0.5355              & 0.8407                                                                      & 1.1895                                                                       \\ \hline
\textbf{4}    & 0.4105              & 0.9940                                                                      & 1.1173                                                                       \\ \hline
\textbf{5}    & 0.6026              & 1.0146                                                                      & 1.1539                                                                       \\ \hline
\textbf{6}    & 0.5298              & 0.9653                                                                      & 1.1212                                                                       \\ \hline
\textbf{Mean} & 0.5565              & 0.9623                                                                      & 1.0942                                                                       \\ \hline
\end{tabular}
\end{table}

\subsection{Summary of Tether Planning and Motion Execution Experiments}
The reachable space reduction approach by ray casting provides the best navigational accuracy, but with the price of a smaller reachable space. Contact point planning allows the robot to navigate in all original free spaces as if it were tetherless. It also enables contact relaxation when necessary. However, this approach compromises motion accuracy with increased number of contact points. Two reasons were presented and errors were analyzed. The results indicate that the motion planners and executor provide an alternative way of UAV localization and navigation in indoor cluttered environments using a taut tether. They also alleviate the challenges caused by managing a tether in obstacle-occupied spaces. However, a trade-off between reachable volume and navigational accuracy exists, so full coverage of the free configuration space and high motion precision cannot be achieved at the same time.

\section{Visual Servoing Experiments}
Experiments for the visual servoing approach described in Chapter \ref{chapter::low_level} are presented.\footnote{Detailed experimental results were presented and published in previous work \cite{xiao2017visual}.}

\subsection{Hypothesis and Metrics}
The hypothesis for the experiments of the visual servoing is \emph{the proposed approach can reactively maintain a constant 6-DoF configuration with respect to a visual 6-DoF trackable stimuli}. The metric used is \emph{success/fail} of path execution with tether and \emph{translational and rotational errors} in all 6 DoFs, in term of $x$, $y$, $z$, yaw, pitch, and roll. Different statistics for the errors are presented and discussed. 

\subsection{Experiments}
\begin{figure}[]
\centering
\includegraphics[width=1\columnwidth]{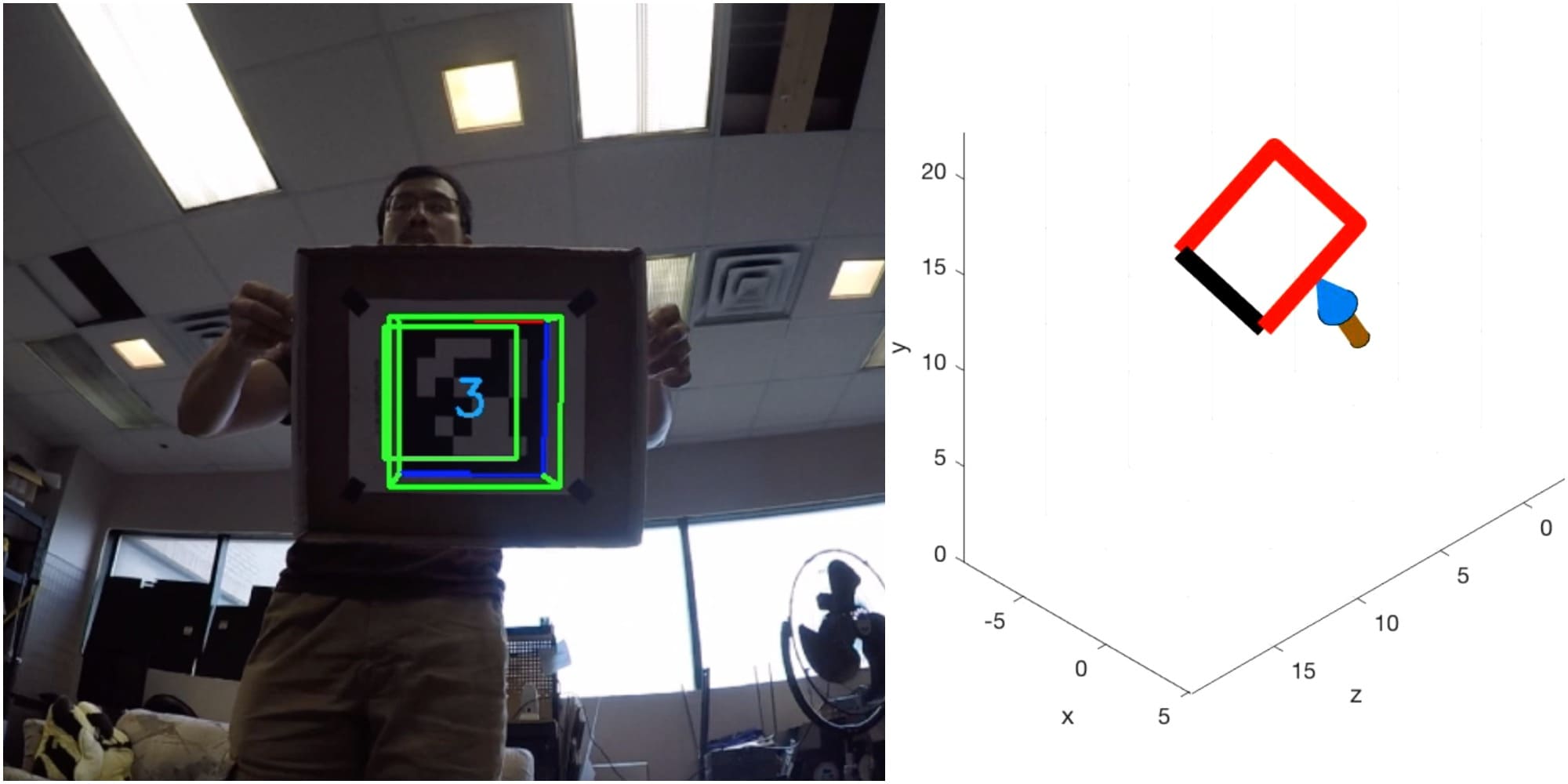}
\caption{One Example Time Step of Visual Servoing (Pitch-up). Arrow represents camera's optical axis, and box is AprilTag (with black side down) (reprinted from \cite{xiao2017visual}).}
\label{pitch_up}
\end{figure}

In order to test the controllability of the visual assisting system in as much as possible of its entire configuration space, the tethered UAV ground station (tether reel) is placed in the middle of the experimental environment. The test is conducted in indoor lab environment without obstacles. AprilTag is moved in a random path but covers all four quadrants of the space. All the length units are in AprilTag unit (1 unit = 8.5 cm) and angular units in radians. Fig. \ref{pitch_up} shows an example time step of visual servoing interface and its actual pose in world frame. In the left hand side, the small green box represents the desired tag pose and the colored box (blue, red and green lines) is the currently detected tag pose (the large green box is only to visualize 6-DOF tag tracking with depth information). The currently detected pose box should converge to the desired pose box with some disturbances caused by vehicle oscillation. It is not enough that the two squares are co-centered, the four lines of the two squares should also overlap with each other, indicating that not only POI position, but also depth and orientation are servoed.  

\subsection{Discussions}
\begin{figure}[]
\centering
\includegraphics[width=1\columnwidth]{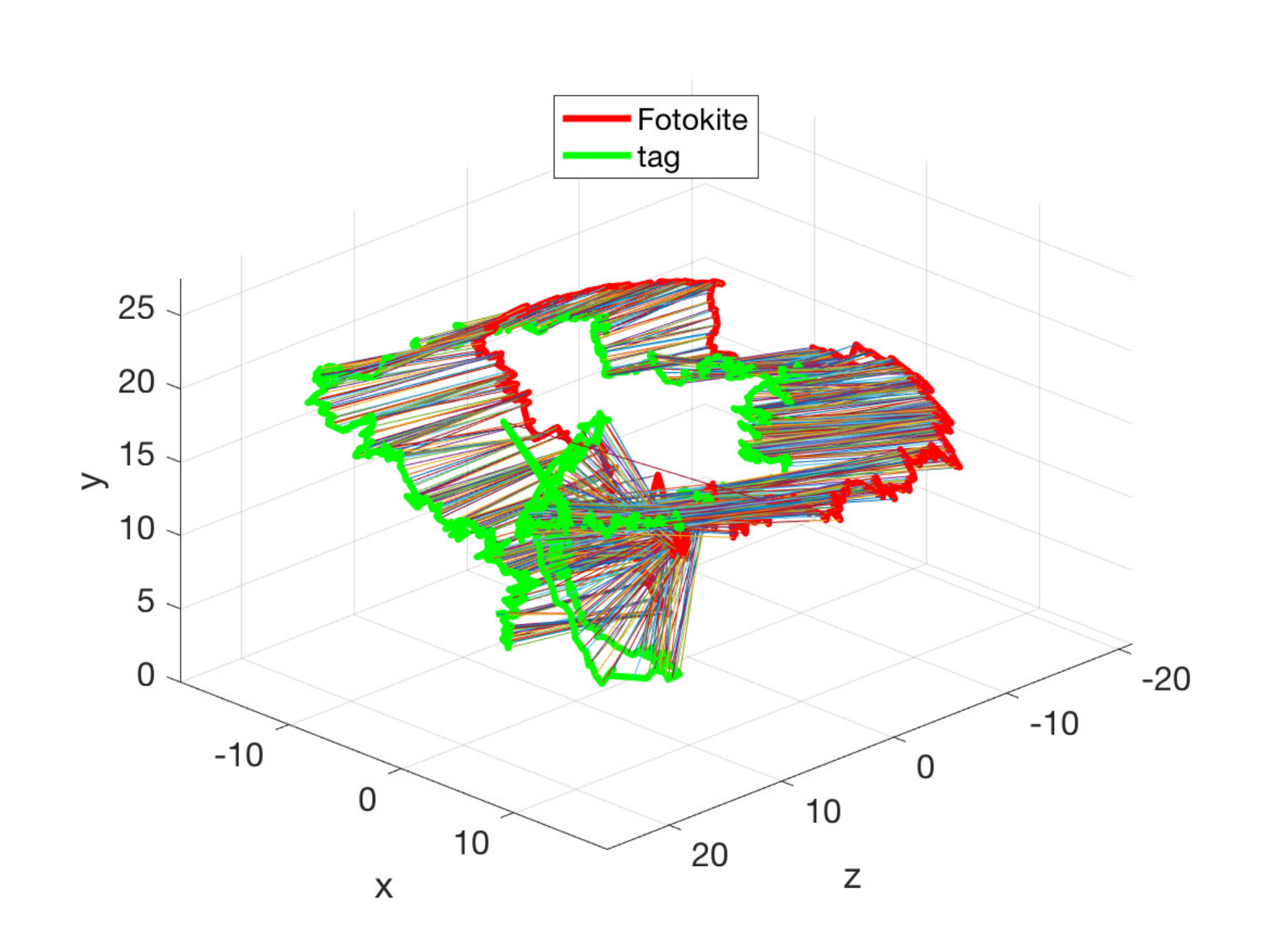}
\caption{Trajectory of the POI (Green) and Visual Assistant (Red): Colorful lines connect the origin of the two frames and indicates the constant relative position and orientation from the visual assistant to the POI (reprinted from \cite{xiao2017visual})}
\label{traj}
\end{figure}

A continuos visual assisting trial is displayed in Fig. \ref{traj}. As we can see, the red trajectory (Fotokite) follows the green trajectory (POI) and Fotokite is maintaining a constant relative position and orientation to the POI. 

\begin{figure}[]
\centering
\includegraphics[width=1\columnwidth]{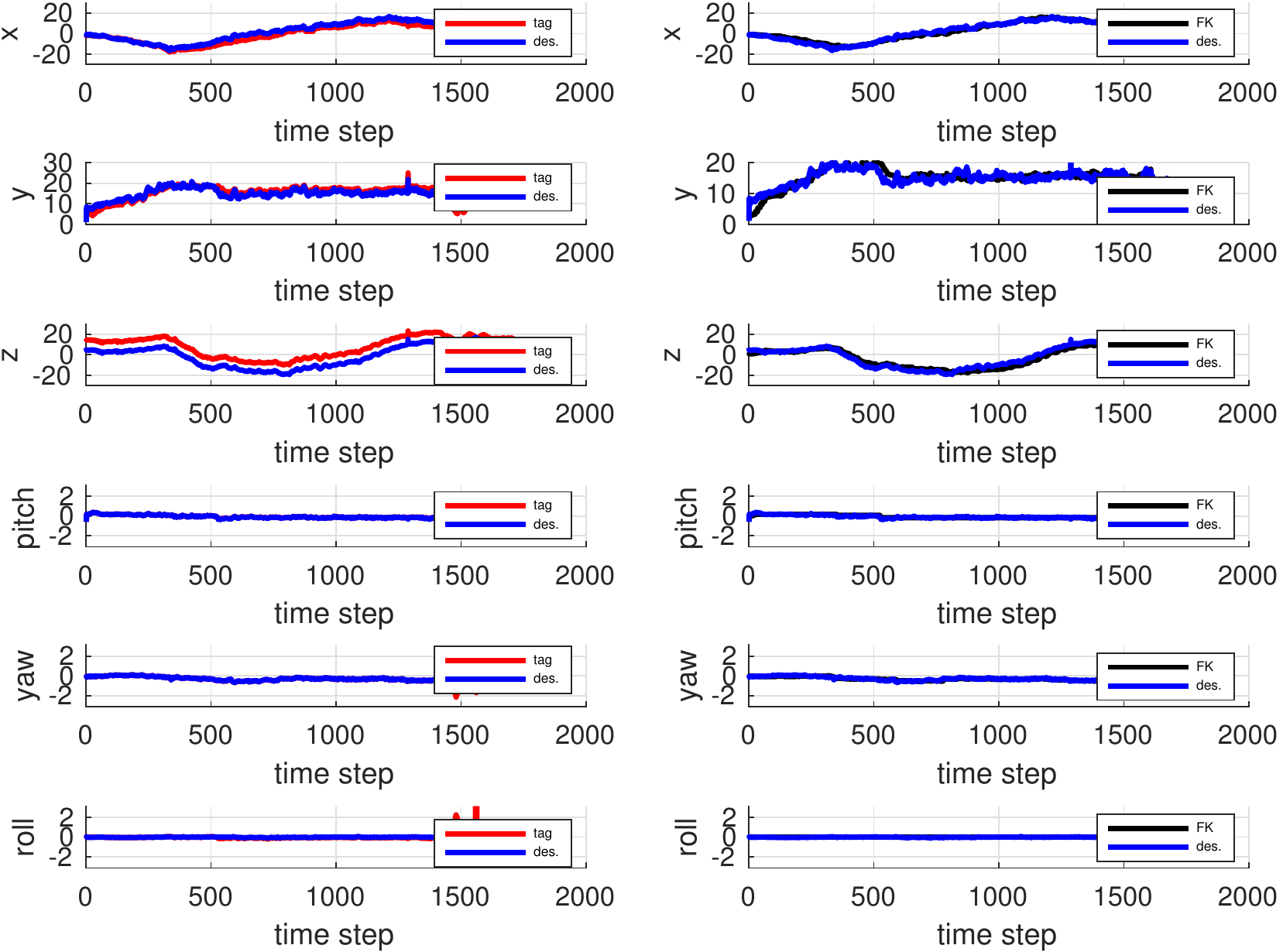}
\caption{$x$,  $y$, $z$, $pitch$, $yaw$, and $roll$ of POI, Desired and Actual Visual Assistant Configuration: $x$,  $y$, and $z$ are in AprilTag units while $pitch$, $yaw$, and $roll$ are in radians (reprinted from \cite{xiao2017visual})}
\label{prof}
\end{figure}

A closer look into the performance of the same trial is demonstrated in Fig \ref{prof}. On the left hand side, the profile $\begin{bmatrix} x,&y,&z,&pitch,&yaw,&roll \end{bmatrix}^T$ of the AprilTag and desired vehicle configuration is compared. The profiles are apart by the desired $\bf{g^f_t*}$. On the right hand side, the desired and actual vehicle configuration is compared. The two profiles for each state space dimension match with each other, indicating that the visual servoing algorithm is directing Fotokite to the desirable configuration to provide visual assistance. 

\begin{figure}[]
\centering
\includegraphics[width=1\columnwidth]{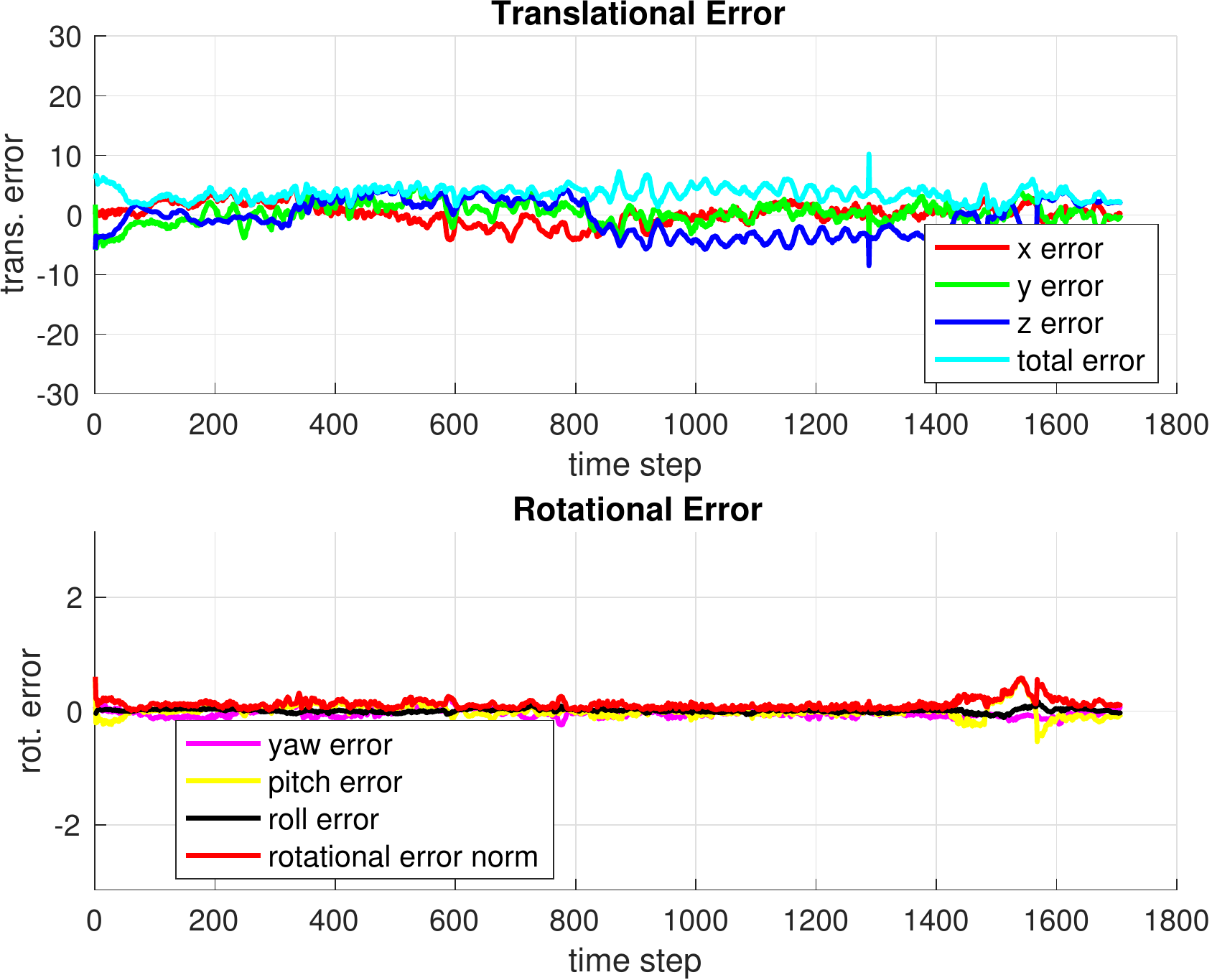}
\caption{Error of Translational and Rotational Motion: Translational error is in AprilTag units while rotational error is in radians (reprinted from \cite{xiao2017visual})}
\label{error}
\end{figure}

The error between the desired and actual pose is further investigated in Fig. \ref{error}. The mean, root mean square, maximum, and standard deviation of the error for the 3 translations, 3 rotations, total euclidean distance, and rotational norm are summarized in Tab. \ref{meanmaxstd}. The results indicate that, despite some disturbances caused by UAV's aerial oscillation, sensing inaccuracies and noises, the overal visual servoing process can maintain a relatively constant 6-DOF pose from the POI to the visual assistant's camera frame. 

\begin{table}[]
\centering
\caption{Mean, Root Mean Square, Maximum, and Standard Deviation of Servoing Error (Reprinted from \cite{xiao2017visual})}
\label{meanmaxstd}
\begin{tabular}{|c|c|c|c|c|c|c|c|c|}
\hline
                                                                      & \textbf{x} & \textbf{y} & \textbf{z} & \textbf{Euclidean} & \textbf{Yaw} & \textbf{Pitch} & \textbf{Roll} & \textbf{Rot. Norm} \\ \hline
\textbf{Mean}                                                         & 0.2216    & 0.2473     & -0.4885    & 3.5961            & -0.0335      & 0.0032        & 0.0080       & 0.1197                  \\ \hline
\textbf{RMS}                                                         & 1.6998    & 1.7042    & 2.8976            & 3.7669     & 0.0812        & 0.1175       & 0.0324   & 0.1464                \\ \hline
\textbf{Max}                                                          & 4.6605     & 5.5776    & 8.4954     & 10.2147             & 0.2905       & 0.5976        & 0.1935        & 0.6035                   \\ \hline
\textbf{\begin{tabular}[c]{@{}c@{}}SD\end{tabular}} & 1.6858     & 1.6866     & 2.8570     & 1.1217             & 0.0739       & 0.1174        & 0.0314        & 0.0844                   \\ \hline
\end{tabular}
\end{table}

\subsection{Summary of Visual Servoing Experiments}
Visual servoing experimental trials have been conducted on physical robot and the performance is quantified and analyzed in terms of control errors in all 6 DoFs. The results indicate that the proposed visual servoing approach is able to successfully drive the visual assistant to a moving PoI, while maintaining the desired 6-DoF pose for the operator's observation. This could be used as a complement to the deliberate risk-aware planning approach.

\section{Summary of Experiments}
This chapter presents all experiments conducted in the scope of this dissertation to validate the relevant proposed approaches. Note that the risk reasoning framework is deducted by formal methods such as propositional logic and probability theory while the risk-aware planner's optimality up to action-dependent risk elements is shown by mathematical induction and the suboptimality for traverse-dependent risk elements shown by example, experiments on physical robot platform is not necessary for them.

Therefore, the physical experiments presented in this chapter focus on each individual components in the low level motion suite from Chapter \ref{chapter::low_level}. Results are demonstrated, analyzed, and discussed. In general, all the components in the low level motion suite are validated using physical experiment results: the tether-based localizer can improve localization accuracy in comparison to the preliminary localizer and reduce the negative effect of tether length on localization accuracy. The two motion primitives can translate motion in 3D Cartesian space into motion commands in tether space so that the tethered UAV could execute any possible paths planned for normal UAVs. The pros and cons of the two motion primitives are discussed. The two motion planning techniques to handle tether in obstacle-occupied spaces are also proved to be working, enabling tethered flight in the vicinity of obstacles. It has also been shown through experiments that maximum reachability and navigation accuracy cannot be achieved simultaneously. Lastly, the complementary reactive visual servoing approach is experimented and the results show that this approach can reactively maintain a constant 6-DoF configuration of the visual assistant to a trackable visual PoI. 

With all the proposed low level components working as expected, the tethered aerial vehicle is teamed up with a teleoperated ground robot, as a marsupial heterogeneous robot team, to assist with better third person viewpoint for the primary robot's teleoperator. An integrated demonstration of the entire robot team will be presented in the next chapter.

\chapter{INTEGRATED DEMONSTRATION}
\label{chapter::integrated_demonstration} 
 
With the theories and implementations in the previous chapters, this chapter presents an integrated demonstration conducted using all the contributions in the scope of this dissertation: the formal risk definition and representation, the risk-aware planner that considers both motion risk and mission reward, and the components in the low level tethered motion suite. It puts together all the proposed approaches in a real-world unstructured or confined environment. The purpose of the integrated demonstration is three-fold: 1. validate the proposed risk definition and representation using real-world physical robot path execution, 2. implement the components in the low level tethered motion suite to enable tethered flight in real unstructured or confined environments, and 3. showcase the better third person viewpoint achieved by the autonomous visual assistant in comparison with that from onboard camera only. The three goals of the experiments will be explained in detail in this chapter. 

The integrated demonstration uses the proposed risk representation and risk-aware planning for two different paths in a real-world unstructured or confined environment and conducts twenty experimental trials using the tethered aerial visual assistant to validate the proposed risk framework. It is hypothesized that \emph{the proposed theoretical risk representation could reflect the results of physical experiments in real-world unstructured or confined environments}. The metric used is the \emph{success/failure rate} of the path execution. Since the same physical tethered UAV platform and the same low level motion suite are used to execute the two paths in the same unstructured or confined environment, factors such as differences in hardware, implementation, and environment are eliminated. The only difference left is the difference in paths and their corresponding risk index values. Therefore the comparison of executions of the two paths is only focused on their different risk index values. The selection of two paths with two different risk values is sufficient to demonstrate if the pass/failure rate of the physical path executions could be reflected by the risk representation. The theoretical and physical results along with the findings from the integrated demonstration are, presented, analyzed and discussed in detail in order to realize the three goals mentioned earlier. The experiments described in this chapter are from a macroscopic view, so it is a proof-of-concept demonstration of the entire working visual assistance system. Due to the lack of data collection apparatus (such as MoCap) in the real-world environment, quantitative analysis of the UAV motion is not the focus of these experiments in this chapter. For those, readers could refer to the experiments presented in Chapter \ref{chapter::experiments}, which are conducted in controlled and engineered lab environment with sophisticated infrastructure for data collection (such as MoCap). 

\section{Implementation: The Co-robots Team}
All the aforementioned and validated approaches are combined and implemented on a co-robots team: a teleoperated ground primary robot, an autonomous tethered aerial visual assistant, and a human operator of the primary robot under the visual assistance of the aerial vehicle.\footnote{The co-robots team was introduced and published in previous work \cite{xiao2019autonomous}.}

\subsection{Teleoperated Ground Robot}
In the co-robots team, the primary robot is a teleoperated Endeavor PackBot 510 (Fig. \ref{fig::team} upper left). PackBot has a chassis with two main differential treads that allow zero radius turn and maximum speed up to 9.3 km/h. Two articulated flippers with treads are used to climb over obstacles or stairs (up to 40\degree). PackBot's three-link manipulator locates on topic of the chassis, with an articulated gripper on the second link and an onboard camera on the third. The manipulator can lift 5kg at full extension and 20kg close-in. Motor encoders on the arm provide precise position of the articulated joints, including the gripper, the default visual assistance point of interest. Four onboard cameras provide first-person-views, but are all limited to the robot body. On the chassis, a Velodyne Puck LiDAR constantly scans the 3-D environments, providing the map for the co-robots team to navigate through. Currently, the autonomous 3D SLAM has not been fully implemented and integrated to the whole system yet. So for this integrated demonstration, a complete map of the workspace is built in advance and given to the team. Four BB-2590 batteries provide up to 8 hrs run time. 

\subsection{Autonomous Visual Assistant}
The same tethered UAV, Fotokite Pro, as used in all the experiments in Chapter \ref{chapter::experiments}, is used as the autonomous aerial visual assistant (Fig. \ref{fig::team} lower left) in the integrated demonstration. Being paired with the teleoperated primary ground robot, it could be deployed from a landing platform mounted on the ground robot's chassis. The onboard camera with the 2-DoF gimbal (pitch and roll) coupled with vehicular yaw is used for visual assistance. Although the tether is supposed to allow the UAV share battery with the ground robot in order to match the run time of both aerial and ground vehicles, the power sharing hardware has not been implemented in this integrated demonstration. The UAV is still using its own battery mounted on its ground station transmitted via the tether. The experiments use the SDK provided by Fotokite Pro, including the sensory feedback, tether length, azimuth and elevation angles, and change rate control over these three control parameters. 

\subsection{Human Operator}
\begin{figure}
\centering
\subfloat[PackBot uPoint Controller Interface]{\includegraphics[width=0.43\columnwidth]{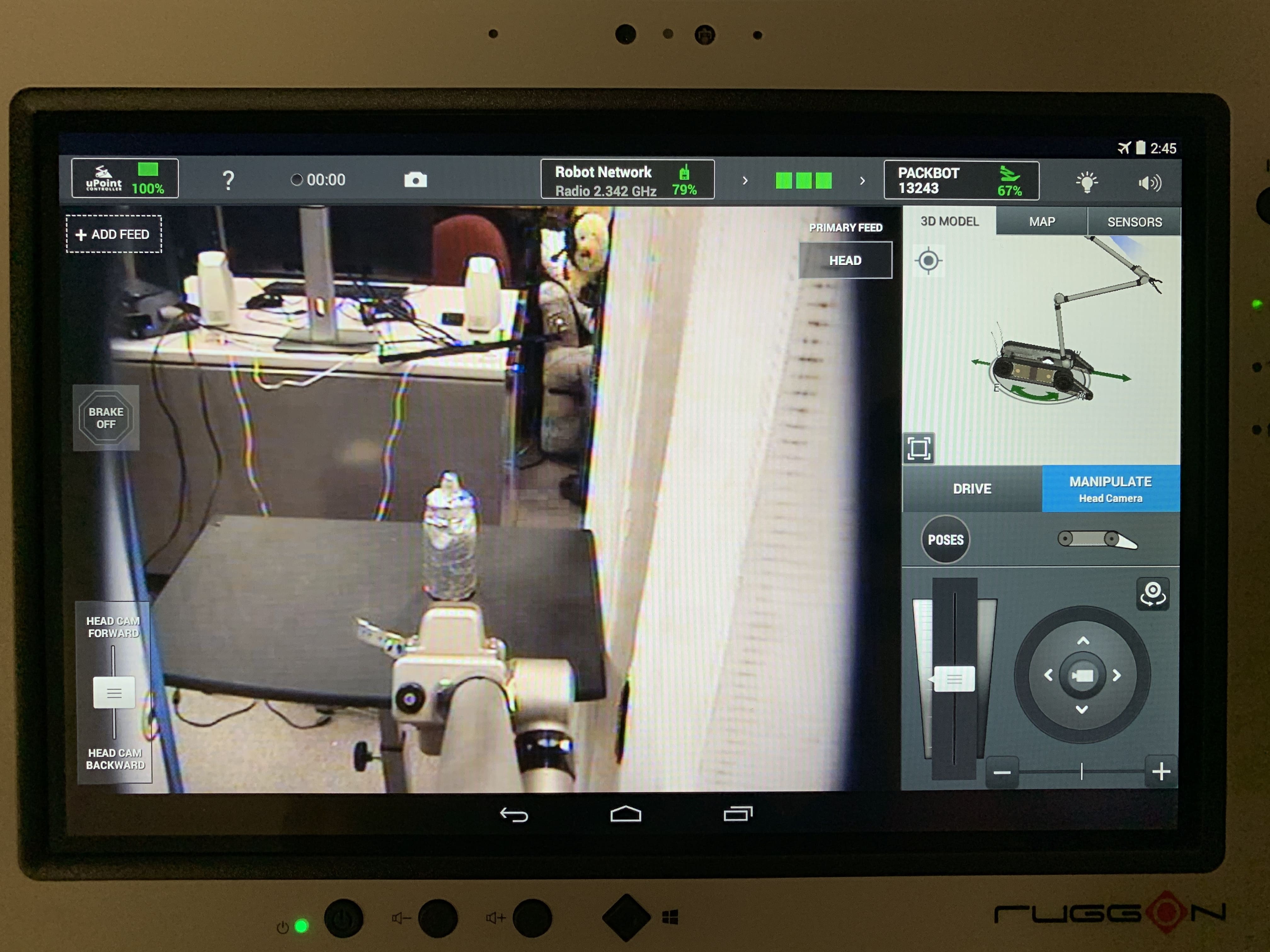}%
\label{fig::upoint}}
\subfloat[Visual Assistant Interface]{\includegraphics[width=0.57\columnwidth]{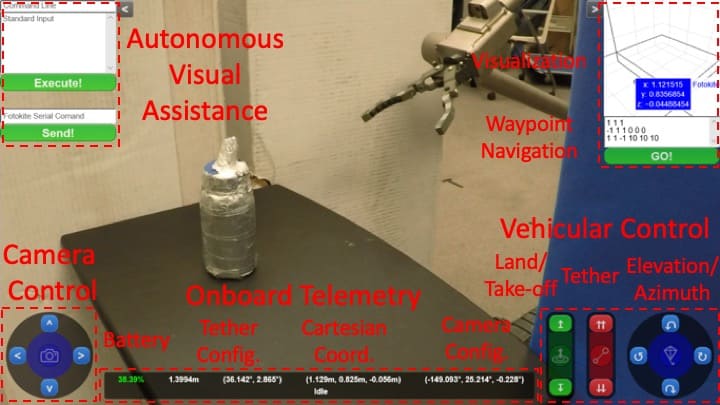}%
\label{fig::mickie}}
\caption{Interfaces with the Human Operator (Reprinted from \cite{xiao2019autonomous})}
\label{fig::interfaces}
\end{figure}

The human operator teleoperates the primary ground robot with the visual assistance of the UAV. In addition to the default PackBot uPoint controller with onboard first-person-view, the visual feedback from the visual assistant's onboard camera is also available to provide improved viewpoints. For example, the visual assistant could move to a location perpendicular to the teleoperation action, providing extra depth perception to the operator. The visual assistant could be either manually controlled or automated. This integrated demonstration focuses on autonomous visual assistance, where autonomous path execution sheds lights on the risk the robot faces. The paths are pre-planned based on a given 3D map and then autonomously executed by the UAV. No attention from the teleoperator is necessary for the control of the aerial visual assistant. The teleoperator only needs to remotely control the primary ground robot to conduct teleoperation mission under the visual assistance of the autonomous tethered UAV. The uPoint teleoperation and visual assistance interfaces are shown in Fig. \ref{fig::interfaces}. 

\subsection{System Architecture}
\begin{landscape}
\begin{figure}[!h]
   \centering
    \includegraphics[width=1\columnwidth]{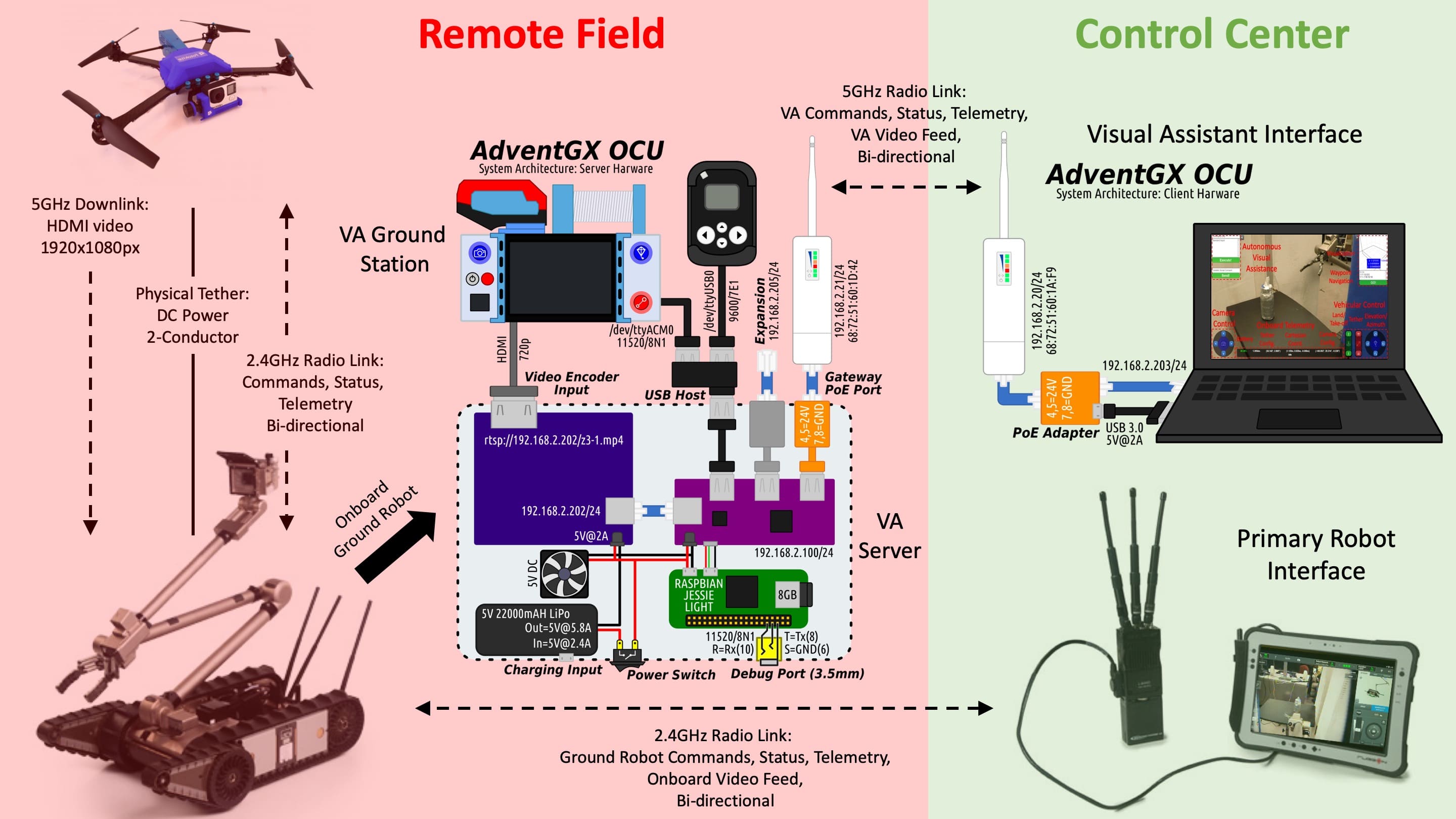}
    \caption{System Architecture}
    \label{fig::implementation_system_architecture}
\end{figure}
\end{landscape}

The architecture diagram of the system deployed in the integrated demonstration is shown in Fig. \ref{fig::implementation_system_architecture}. The entire system locates in two separate locations, the remote field where the teleoperation mission is conducted and the control center where the teleoperator is physically located. The communication between those two locations are through multiple bi-directional radio links. 

\subsubsection{In Remote Field}
The tethered aerial visual assistant's ground station (with tether reel) is mounted on the primary ground robot. The UAV is powered via the physical tether, while onboard commands, status, and telemetry are transmitted by 2.4GHz bi-directional radio link. The HDMI video from the visual assistant's camera is streamed via 5GHz radio downlink to the ground station. The primary robot, Packbot, is teleoperated by 2.4GHz bi-directional radio link for controls and video. 

A server built in collaboration with AdventGX is connected with the UAV's ground station via HDMI and USB cables, for video encoding and UAV control (via Fotokite SDK), respectively.  The black RadEye SPRD spectroscopic radiation detector is not used in the integrated demonstration. The server is composed of a video encoder (shown in purple in Fig. \ref{fig::implementation_system_architecture}) and a Raspberry Pi computer (shown in green in Fig. \ref{fig::implementation_system_architecture}). The video encoder encodes the HDMI video output from the ground station so it could be transmitted wirelessly to the visual assistant's Operator Control Unit (OCU). The Raspberry Pi computer connects to the ground station via USB and uses Fotokite SDK to receive sensor data and send control commands. All video and telemetry data are ported to an antenna and transmitted to the visual assistant's OCU wirelessly via a 5GHz radio link. The link is bi-directional, whose other direction is used for transmitting control commands from the OCU to the server. The server is self-powered with its own battery and cooled by a 5V DC fan. 

All the executables of the UAV controls are compiled from C++ code and then stored on the server's Raspberry Pi computer. Direct tether, vehicle and camera commands (teleoperation) are ported directly from the OCU to the server and then to the Fotokite ground station. Autonomous flight, including landing, taking-off, autonomous waypoint navigation and visual assistance, is triggered by the OCU, and then the corresponding executable on the server is called. 

\subsubsection{In Control Center}
In the control center, the teleoperator uses the uPoint interface to control the primary ground robot via 2.4GHz bi-directional radio link. The current system has a separate OCU for the visual assistant, which is planned to be integrate with the ground robot interface in the future. The visual assistant OCU is a laptop connected with an antenna for 5GHz bi-directional radio link. The interface is implemented on a specific web socket and could be displayed via web browser. As mentioned earlier, the interface allows both teleoperation and autonomous navigation. In this integrated demonstration, two paths are pre-planned based on a given 3D map, saved in the OCU, and uploaded to the server. The autonomous navigation executable on the server called by the OCU command takes the path as input argument. The teleoperator only needs to select the autonomous navigation and all the following process for the visual assistant is carried out autonomously. 

\section{Experimental Environment}
\begin{figure}[]
\centering
	\includegraphics[width =0.7 \columnwidth]{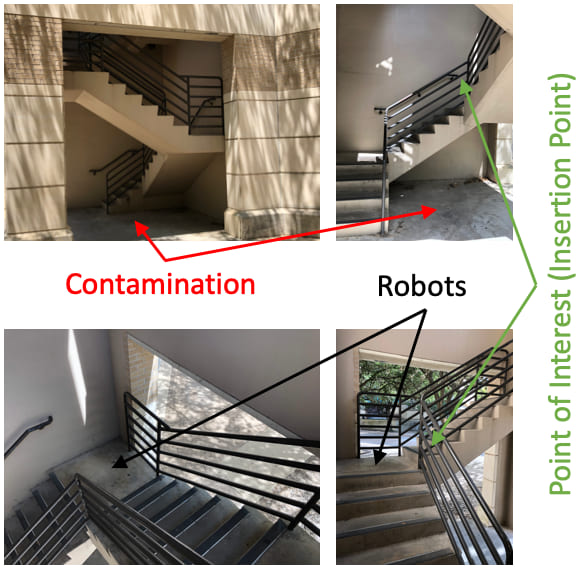}
	\caption{Different Views of the Experimental Environment: Contamination locates at the bottom of the staircase. The robot team can reach the second level. The task is to teleoperate the primary ground robot to insert the manipulator arm between the railings (Point of Interest), release the gripper, and drop the sensor into the pool of contamination. Teleoperation of the dexterous manipulation requires good viewpoint provided by the visual assistant. }
	\label{fig::staircase}
\end{figure}

\begin{figure}[]
\centering
	\includegraphics[width =0.7 \columnwidth]{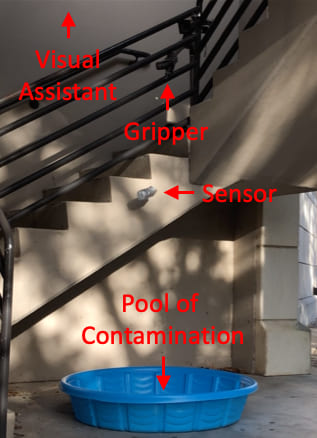}
	\caption{Teleoperated Ground Robot Dropping Sensor Through Railings to the Pool with the Help of the Visual Assistant}
	\label{fig::drop}
\end{figure}

The experiments are conducted in a real-world environment, which resembles the environment encountered by search and rescue robot and personnel in Fukushima Daichi nuclear disaster response. The unstructuredness and confinedness of the experimental environment are similar to those in the actual nuclear power plant. The environment is located in a staircase, at the bottom of which exists a pool of contamination (Fig. \ref{fig::staircase}). The marsupial robot team is able to reach the second level of the staircase. The task is to drop a sensor into the contamination so that radioactivity strength could be measured. The practice at Fukushima was to use Packbot's gripper to hold the sensor, teleoperate the manipulator arm to insert the gripper with sensor between the staircase railings, then release the gripper and drop the sensor into the pool of contamination at the bottom. In Fukushima, the entire process was conducted through the visual feedback from the onboard cameras and teleoperated ground visual assistant. Using the risk-aware tethered visual assistant proposed in this research, the experiments utilize an autonomous robot agent to provide better viewpoint for the primary robot operator. The experiment scenario is displayed in Fig. \ref{fig::drop}.

\begin{figure}[]
\centering
	\includegraphics[width =1 \columnwidth]{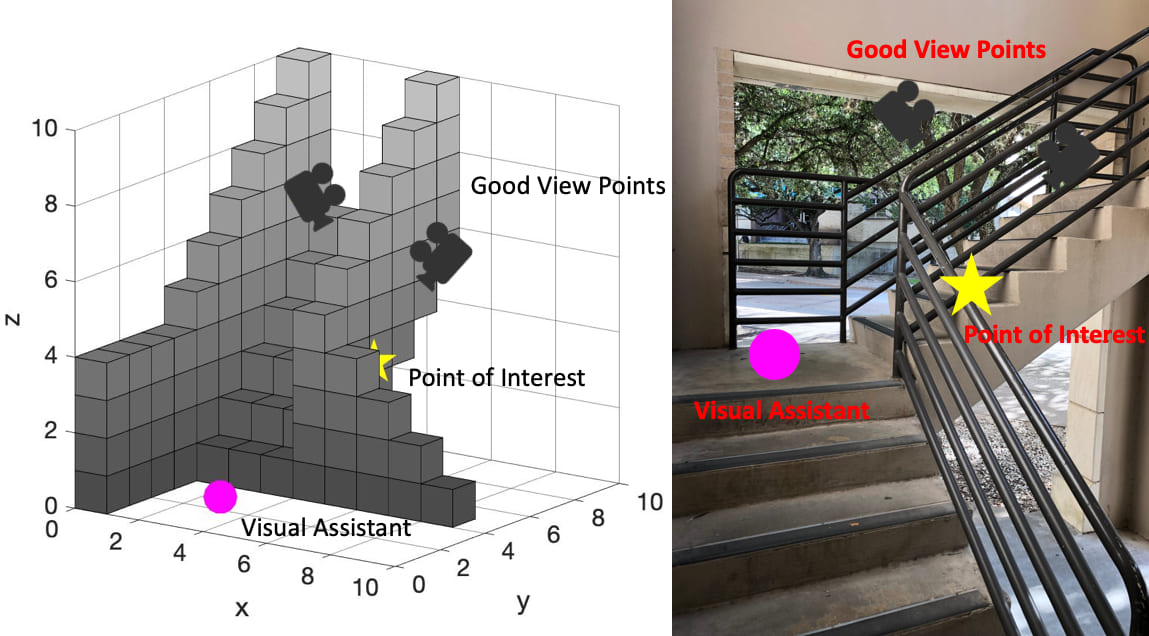}
	\caption{Map And Rewards: Greyscale voxels represent occupied spaces (obstacles) in the map, whose greyscale corresponds to height. The magenta circle represents where the tethered aerial visual assistant initially locates. The yellow star behind the railings is the Point of Interest, i.e. the insertion point of the manipulator arm between the railings. The two cameras are good view points. The one on the left is slightly better, 1.0, than the one on the right, 0.9. }
	\label{fig::map_and_rewards}
\end{figure}

Ideally, a map of the staircase is built by the ground robot's onboard LiDAR. But due to equipment problems, the portion of the staircase (scanning from the second level), where the marsupial robot team locates, is pre-mapped and manually refined (Fig. \ref{fig::map_and_rewards}). For the scope of this dissertation, the map is assumed to be complete and handling incomplete map remains the topic for future research. Due to the low tolerance of the tethered UAV platform, the obstacles are inflated for safety. Each voxel is 0.4m in dimension, which is the default onboard flight accuracy of the UAV platform (tested by experiments \cite{xiao2018motion}). The UAV initially locates on the landing platform of the primary ground robot, shown as the magenta circle in Fig. \ref{fig::map_and_rewards}. The insertion point of the manipulator arm through the staircase railings are denoted by the yellow star behind the railings, as the visual assistance Point of Interest. This is assumed to be pre-defined and static, so no tracking of the PoI is going on. Viewpoint quality scores are out of the scope of this research as mentioned in Chapter \ref{chapter::high_level}. but for the purpose of this demonstration, two good viewpoints are assumed to be in the map, shown as the two cameras. The one on the left is given a slightly better viewpoint quality (1 \emph{vs.} 0.9, the scale of the score could be arbitrary at this point). The UAV is deployed to the desired viewpoint before the teleoperation of the ground robot happens. 

\section{Planning and Risk Representation Results and Discussions}
In order to represent risk of locomoting in this unstructured or confined environment, six (out of 16) different risk elements are considered. They are distance to closest obstacle and visibility (locale-dependent), action length and turn (action-dependent), tether length and number of tether contacts (traverse-dependent). The choice of these six risk elements are due to considerations of their relevance to this particular robot platform in this particular unstructured or confined scenario, the practicality or availability of necessary risk information, and the representativeness of the three major risk categories. As discussed in detail in Chapter \ref{chapter::risk_representation}, risk is the probability of the robot not being able to finish the path. 

\begin{figure}[]
\centering
	\includegraphics[width =0.7 \columnwidth]{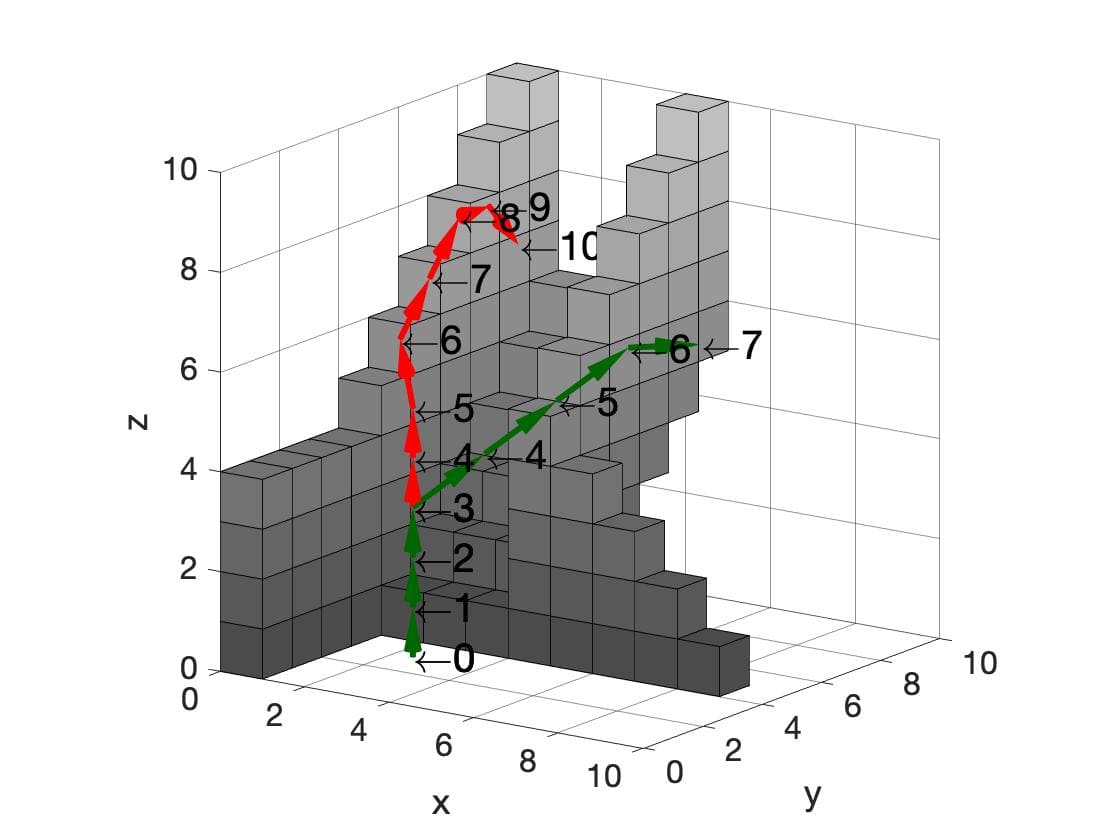}
	\caption{Minimum-risk Paths of UAV Leading to the Two Good Viewpoints: Red path has higher risk (0.714) while green path has lower risk (0.575). The numbers indicate the indices of the states on the paths. }
	\label{fig::green_red_paths}
\end{figure}

Considering the only two rewarding states in the entire state space, only paths containing these two states can yield a positive (non-zero) utility value. To maximize utility, only paths leading to them need to be considered. Using the high-level risk-aware planner described in Chapter \ref{chapter::high_level}, the minimum-risk path to these two rewarding states are planned, shown in Fig. \ref{fig::green_red_paths}. As mentioned earlier, the choice of two good viewpoints and therefore two paths with two different risk values is sufficient to demonstrate if the pass/failure rate of the physical path executions could be reflected by the risk representation. 

The red path aims at the best viewpoint between the two staircase railings. Since a direct path from the initial location to the best viewpoint needs to go through spaces confined by staircase railings and walls, the UAV maneuvers through those types of spaces to maintain relatively large clearance, i.e. remain far away from obstacles and high visibility, but at the cost of a longer path and more turns. Reasoning with the formal risk definition and explicit representation proposed in Chapter \ref{chapter::risk_representation}, the planner decides the extra length and turns worth the extra effort to go to the left viewpoint, when being compared with closer distance and low visibility. The risk associated with the red path is evaluated to be 0.714. The detailed risk representation for each state and each individual risk element on the red path is shown in Tab. \ref{tab::red_path_details}. The last column shows the state risk the robot faces at each state.

\begin{table}[]
\centering
\caption{Detailed Risk Representation for Red Path}
\label{tab::red_path_details}
\begin{tabular}{|c|c|c|c|c|c|c|c|}
\hline
\textbf{Index} & \textbf{Dist.} & \textbf{Vis.} & \textbf{Act. Len.} & \textbf{Turn} & \textbf{Tet. Len.} & \textbf{Cont. \#} & \textbf{State Risk}\\ \hline
\textbf{0}     & 0.01              & 0.02                & 0                    & 0             & 0.01                 & 0     & 0.04               \\ \hline
\textbf{1}     & 0.01              & 0.02                & 0.04                 & 0             & 0.01                 & 0        & 0.08            \\ \hline
\textbf{2}     & 0.01              & 0.02                & 0.04                 & 0             & 0.01                 & 0       & 0.08             \\ \hline
\textbf{3}     & 0.01              & 0.02                & 0.04                 & 0             & 0.01                 & 0         & 0.08           \\ \hline
\textbf{4}     & 0.01              & 0.01                & 0.04                 & 0             & 0.02                 & 0          & 0.08          \\ \hline
\textbf{5}     & 0.01              & 0.01                & 0.04                 & 0             & 0.02                 & 0      &0.08              \\ \hline
\textbf{6}     & 0.01              & 0.01                & 0.06                 & 0.05          & 0.02                 & 0        & 0.14            \\ \hline
\textbf{7}     & 0.01              & 0.01                & 0.05                 & 0.05          & 0.03                 & 0      & 0.14              \\ \hline
\textbf{8}     & 0.01              & 0.01                & 0.05                 & 0             & 0.04                 & 0      & 0.11              \\ \hline
\textbf{9}     & 0.02              & 0.02                & 0.04                 & 0.05          & 0.04                 & 0         & 0.16           \\ \hline
\textbf{10}    & 0.03              & 0.04                & 0.05                 & 0.05          & 0.04                 & 0     & 0.19               \\ \hline
\end{tabular}
\end{table}

\begin{table}[]
\centering
\caption{Detailed Risk Representation for Green Path}
\label{tab::green_path_details}
\begin{tabular}{|c|c|c|c|c|c|c|c|}
\hline
\textbf{Index} & \textbf{Dist.} & \textbf{Vis.} & \textbf{Act. Len.} & \textbf{Turn} & \textbf{Tet. Len.} & \textbf{Cont. \#} & \textbf{State Risk}\\ \hline
\textbf{0}     & 0.01              & 0.02                & 0                      & 0             & 0.01                   & 0        & 0.04               \\ \hline
\textbf{1}     & 0.01              & 0.02                & 0.04                   & 0             & 0.01                   & 0       & 0.08               \\ \hline
\textbf{2}     & 0.01              & 0.02                & 0.04                   & 0             & 0.01                   & 0    & 0.08                   \\ \hline
\textbf{3}     & 0.01              & 0.02                & 0.04                   & 0             & 0.01                   & 0    & 0.08                   \\ \hline
\textbf{4}     & 0.03              & 0.01                & 0.06                   & 0.05          & 0.02                   & 0  & 0.16                     \\ \hline
\textbf{5}     & 0.04              & 0.01                & 0.06                   & 0             & 0.02                   & 0  & 0.12                     \\ \hline
\textbf{6}     & 0.02              & 0.01                & 0.06                   & 0             & 0.03                   & 0   & 0.12                    \\ \hline
\textbf{7}     & 0.01              & 0                   & 0.05                   & 0.05          & 0.03                   & 0   & 0.13                    \\ \hline
\end{tabular}
\end{table}

The green path aims at the second best viewpoint in the wide open space in the middle of the staircase. Going there straight from the initial location needs to closely pass by the top of the railings. The planner chooses to make a slight detour to enlarge the clearance. However, maximizing distance and visibility has longer path and more turns as cost, so the planner chooses a compromise in between, shown as the 45\degree~middle segment on the green path: the UAV does not fully sacrifice path length and twistiness for clearance, so it cuts through the free space with a straighter path and slightly (not completely) avoids the obstacles. The risk associated with the green path is evaluated to be 0.575. The detailed risk representation for each state and each individual risk element on the green path is shown in Tab. \ref{tab::green_path_details}. The last column shows the state risk the robot faces at each state. No contact points are formed in either cases. 

It is worth to note that using the traditional state-dependent only risk representation, the red path has a lower additive risk, because it maintains a relatively low state-dependent risk at most of the states on the path. The green path, however, would have higher risk, due to the compromise of locale-dependent risk elements (distance and visibility) for action-dependent risk elements (action length and turns). Although overall the compromise reduces the path risk, it cannot be reflected by the traditional state-dependent risk representation. 

In terms of reward and utility, the simple assumption of viewpoint reward only at the last state can yield utility values of 1.401 and 1.565 for the red and green path, respectively. Tab. \ref{tab::path_rewards_risk_utility} shows the comparison of path rewards, risk, and utility. The lower stage reward maximizer in Chapter \ref{chapter::high_level} picks the green path as the optimal visual assistance path due to its higher utility value. 

\begin{table}[]
\centering
\caption{Path Rewards, Risk, and Utility Comparison}
\label{tab::path_rewards_risk_utility}
\begin{tabular}{|c|c|c|c|}
\hline
                    & \textbf{Rewards} & \textbf{Risk} & \textbf{Utility} \\ \hline
\textbf{Red Path}   & 1                & 0.714         & 1.401            \\ \hline
\textbf{Green Path} & 0.9              & 0.575         & \textbf{1.565}   \\ \hline
\end{tabular}
\end{table}

\section{Physical Experiments Results and Discussions}
The two planned paths (red and green path in Fig. \ref{fig::green_red_paths}) are implemented autonomously on the physical tethered UAV using the low level motion suite (chapter \ref{chapter::low_level}). It is worth to note that the evaluation of the improvement in primary ground robot teleoperation performance is dependent on the viewpoint quality study and is therefore out of the scope of this work. This research only focuses on risk-aware planning and tethered flight in unstructured or confined environments, therefore the experiments are only used to validate the risk definition and representation and implement the components in the low level tethered motion suite. Only a simple comparison between the onboard camera view and visual assistant view is given to showcase the expected improved viewpoint in the summary. 

\begin{figure}[]
\centering
	\includegraphics[width =1 \columnwidth]{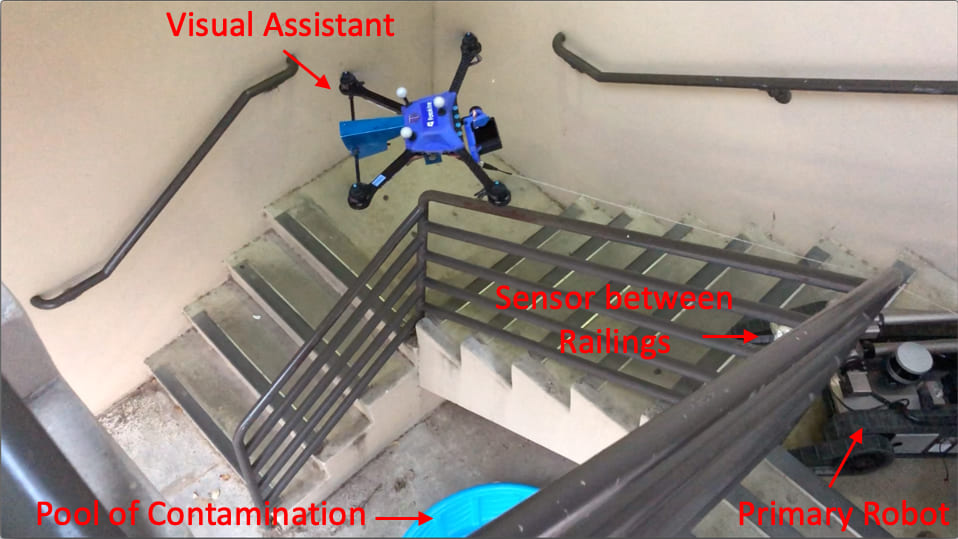}
	\caption{Tethered Aerial Visual Assistant Executing Green Path}
	\label{fig::fk_pb_pool}
\end{figure}

Ten experimental trials each are conducted for the red and green path. For all twenty trials, the components in the low level motion suite are used. To avoid singularity, position control is chosen as the motion primitives in the experiments. The path execution is manually terminated when the tethered UAV is about to collide with the obstacles (either the staircase wall or staircase railings) or the UAV starts to oscillate, loses localization, and therefore is unable to reach the next target waypoint. Fig. \ref{fig::fk_pb_pool} shows the aerial visual assistant executing the green path. The pool of contamination locates at the bottom of the staircase. The sensor on the gripper of the primary robot's manipulator arm needs to be inserted between the railings. The target viewpoint locates in the open space in the middle of the staircase. Additionally, the straighter and shorter path with fewer turns makes the green path safer than the red one. 

\begin{figure}[]
\centering
\subfloat[Red Path Execution]{\includegraphics[width=0.7\columnwidth]{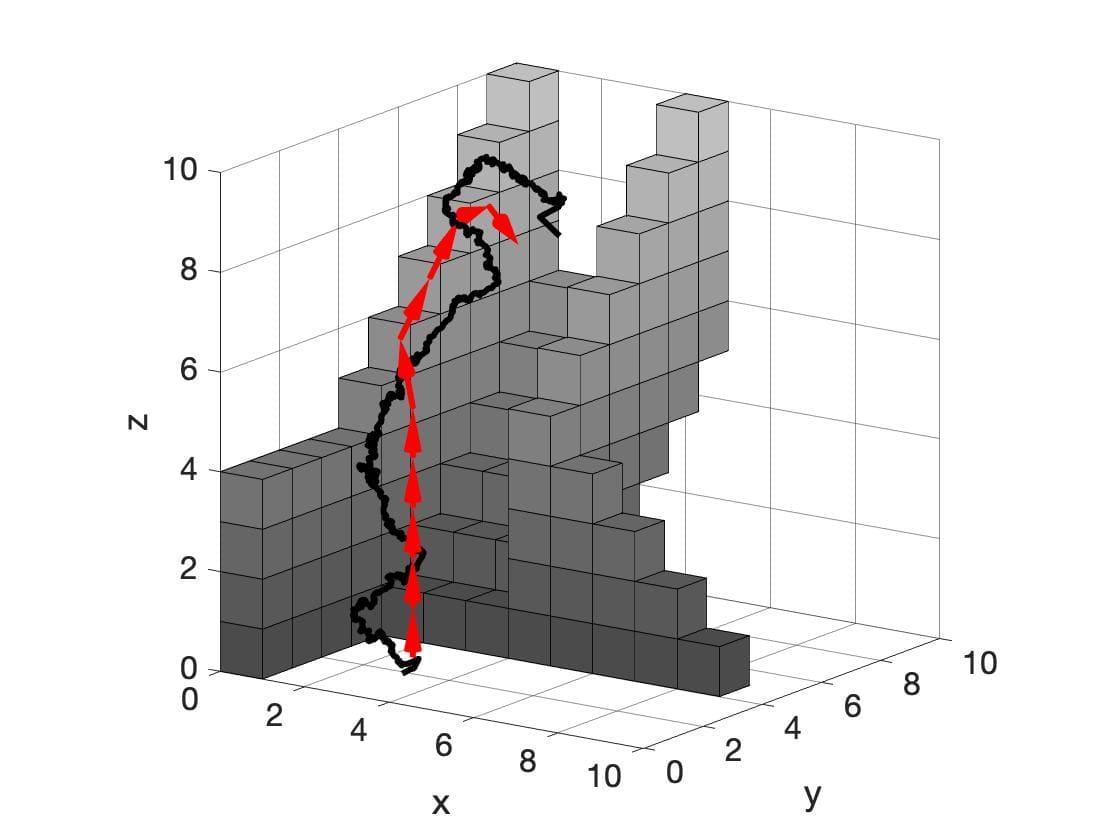}%
\label{fig::red_path_real}}\\
\subfloat[Green Path Execution]{\includegraphics[width=0.7\columnwidth]{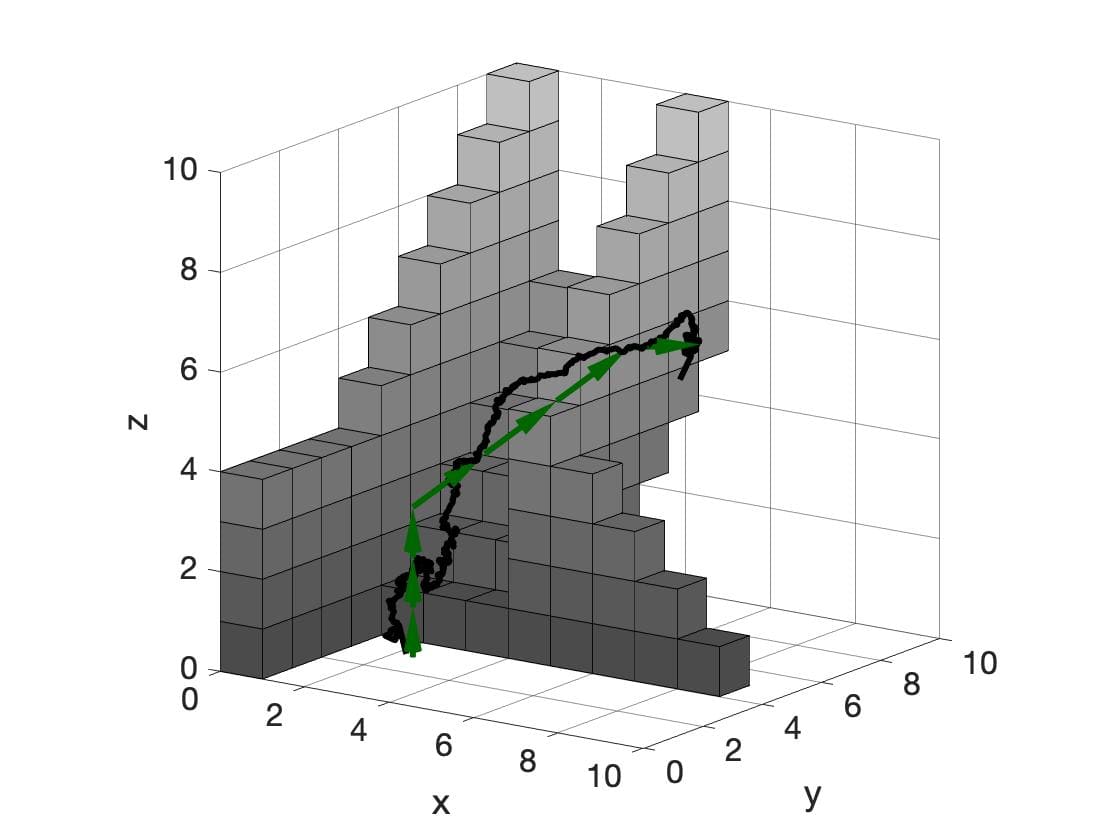}%
\label{fig::green_path_real}}\\
\caption{Example Success Trials (Black) for Red and Green Path}
\label{fig::red_green_path_real}
\end{figure}

Fig. \ref{fig::red_green_path_real} shows an example of two successful path executions for red and green path. The black trajectory is the actual trajectory sensed by the robot onboard sensor using the localization model presented in Chapter \ref{chapter::low_level}. It needs to be pointed out that the ideal data collection method would be external devices such as MoCap. But since it is not possible to set up an entire MoCap studio in real-world staircase scenario and the walls and railings may block the MoCap cameras from capturing the reflective markers, UAV onboard tether-based sensing is the only practical option for ground truth. For detailed flight accuracy analysis, readers could refer to the experiments presented in Chapter \ref{chapter::experiments}, whose experiments are conducted in well-engineered controlled lab environments with MoCap system. 

The results of all 20 experimental trials are shown in Tab. \ref{tab::trials_20}. In Tab. \ref{tab::trials_20}, the results of the 20 trials are listed, either success or failure. If the trial is a failure case, the reason of the failure/termination is specified as well. It would be ideal if each failure could be directly attributed to the responsible risk element. However, it is not apparent which risk element actually causes path execution failure. Furthermore, other risk elements which are not considered by the risk-aware planner in this particular context may contribute to the failure as well, but it is not practical to directly attribute them to the failure. Only observations could be made and the ostensible reasons for failure are listed. 

\begin{table}[]
\centering
\caption{Experimental Trials and Success/Failure}
\label{tab::trials_20}
\begin{tabular}{|c|c|c|}
\hline
Trial \#     & Red Path              & Green Path                         \\ \hline
1            & Collision w. railings & Collision w. wall                  \\ \hline
2            & Oscillation           & \textbf{Success}                   \\ \hline
3            & \textbf{Success}      & Oscillation                        \\ \hline
4            & Collision w. wall     & \textbf{Success}                   \\ \hline
5            & Oscillation      & Collision w. railings              \\ \hline
6            & Oscillation           & \textbf{Success}                   \\ \hline
7            & \textbf{Success}               & Collision w. wall                  \\ \hline
8            & Collision w. wall     & Contacts formed, localization lost \\ \hline
9            & Collision w. railings & \textbf{Success}                   \\ \hline
10           & Oscillation           & Contacts formed, localization lost \\ \hline
Success Rate & 0.2                   & 0.4                                \\ \hline
\end{tabular}
\end{table}

For the red path, only two out of the ten experimental trials are successful. The other eight trials fail due to different reasons: trial 1 and 9 fail because the UAV contacts with the staircase railings, while trial 4 and 8 contacts the wall. The most important reason for failure is oscillation. This happens primarily when the UAV is maneuvering to avoid obstacles and maintain a high clearance. The turning and long path have the potential of inducing extra turbulence in the confined staircase, therefore the rotorcraft can no longer maintain stability. Over-compensating the turbulence may cause overshoot, and the oscillation further leads to collision or not being able to reach a certain waypoint. The success rate is only 20\%, which shows that the 0.714 risk is close but actually an underestimate of risk. On the right hand side of Tab. \ref{tab::trials_20}, green path execution achieves 40\% success rate, which is close to the 0.575 risk value but a slight underestimate as well. While in trial 1, 5, and 7 the UAV collides with the obstacles and it starts oscillate in trial 3, another important failure reason comes into play for the green path: due to the closeness to the railings, contact point may be accidentally formed, deteriorating the localization accuracy. In trial 8 and 10, the contact point even causes loss of localization so that the UAV cannot reach the next waypoint. But overall speaking, the relatively open space in the center of the staircase and the straightness and shortness contribute to a less risky path. Although the risk value caused by each individual risk elements is only an empirical estimation and is therefore different from the ideal true value, six failures and eight failures out of ten trials are sufficiently close to the 0.575 and 0.714 risk value, respectively. Basically, for both cases, the proposed theoretical risk representation framework matches closely with the failure rate in physical experiments in practice. 

\begin{figure}[]
\centering
\subfloat[Failure Locations during Red Path Execution]{\includegraphics[width=0.7\columnwidth]{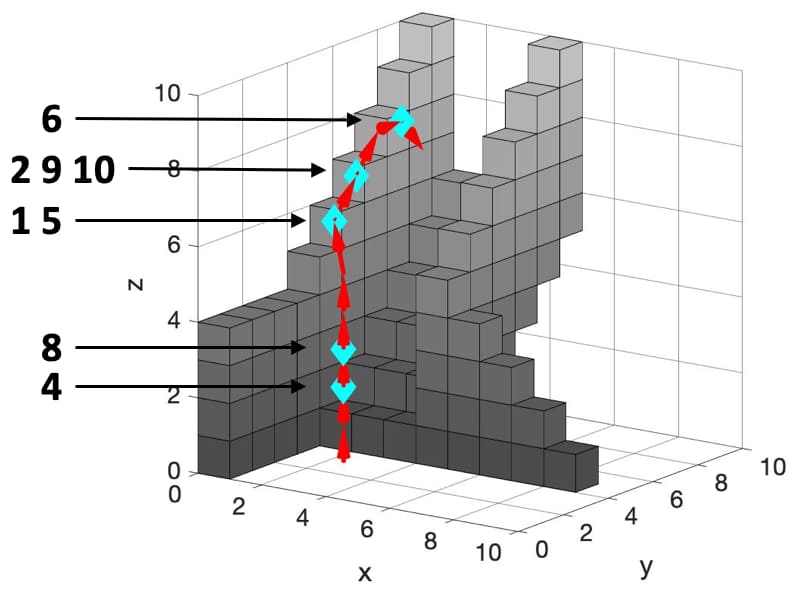}%
\label{fig::red_failures}}\\
\subfloat[Failure Locations during Green Path Execution]{\includegraphics[width=0.7\columnwidth]{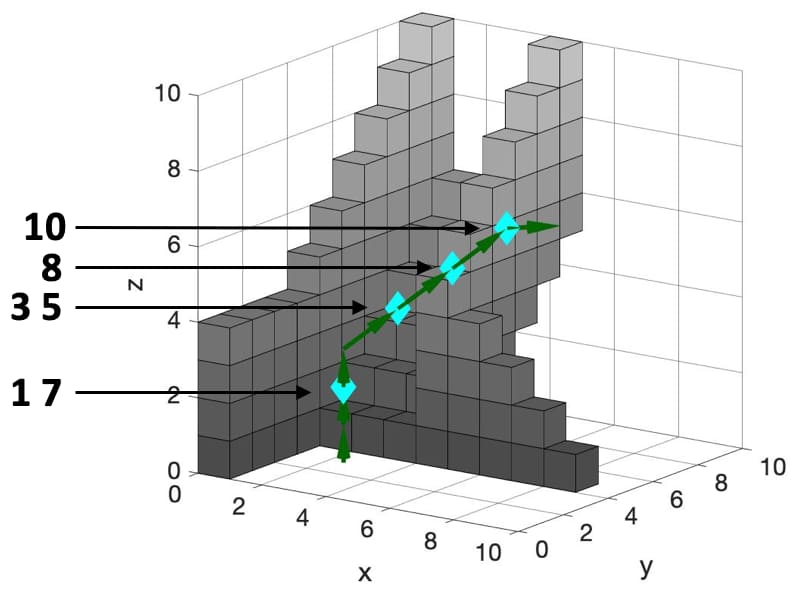}%
\label{fig::green_failures}}\\
\caption{Failure Locations on Both Paths: The numbers correspond to the trial number in Tab. \ref{tab::trials_20}, indicating this particular trial is terminated at the state denoted by cyan diamonds.  }
\label{fig::failure_locations}
\end{figure}

Fig. \ref{fig::failure_locations} shows the locations of failure (shown as cyan diamonds), i.e. the state on the path where path execution is terminated. The numbers on the left correspond to the failure trial numbers in Tab. \ref{tab::trials_20}. Some failure locations (cyan diamonds) only have one failure trial, while others may have multiple. Most failure locations for both cases are in the top part of the path, due to either complex trajectory shape (longer path and more turns for red path) or closeness to obstacles (collision or tether contact with obstacles for green path). It matches with the state risk values in the last column of Tab. \ref{tab::red_path_details} and Tab. \ref{tab::green_path_details}: the high state risk index values are correlated with more failure cases at that particular state in the physical experiments. Inspecting the failure reasons (Tab. \ref{tab::trials_20}) and failure locations (Fig. \ref{fig::failure_locations}), it could be seen that for red path most failures are caused by action-dependent risk elements while the effect of locale-dependent risk elements is minimized. But for green path, due to the sacrifice of locale-dependent risk elements for shorter path length and fewer turns (action-dependent risk elements), obstacles near states cause more possibility of failure to finish the path. 

\section{Summary of Integrated Demonstration}
The presented integrated demonstration utilizes all the contributions presented by this dissertation: the formal risk reasoning framework with definition and representation, the risk-aware planner that minimizes risk while considering reward, and the tethered motion suite to realize tethered flight in indoor unstructured or confined environments. Although quantitative analysis of tethered motion is not possible due to the lack of data collection apparatus in real-world environment, qualitative (success or failure) results and UAV onboard sensing are utilized to evaluate the experiments. 

The entire co-robots team, the teleoperated ground robot, autonomous visual assistant, and human operator, is introduced, and the system architecture used for the integrated demonstration is presented. This integrated demonstration duplicates a real-world robotic teleoperation mission in Fukushima: sensor insertion through railings for contamination level readings. All the approaches and contributions proposed by this dissertation is demonstrated in this real-world unstructured or confined environment. 

For the formal risk reasoning framework and risk-aware planning, the risk representation and planning results in the real-world unstructured or confined environments validate the proposed risk definition and representation from Chapter \ref{chapter::risk_representation}. The comparison between the red and green paths in Fig. \ref{fig::green_red_paths} favors the green path, despite the fact that red path could be regarded safer using conventional state-dependent risk representation. Physical execution of both paths further validates the claim: four out of ten green path executions are successful, while only two out of ten red path executions succeed. The practice matches with the proposed theory. Failure reasons and locations for path execution are presented, analyzed, and discussed using the results of real-world physical experiments. 

\begin{figure}[]
\centering
\subfloat[First Person View from Primary Robot Onboard Camera]{\includegraphics[width=0.5\columnwidth]{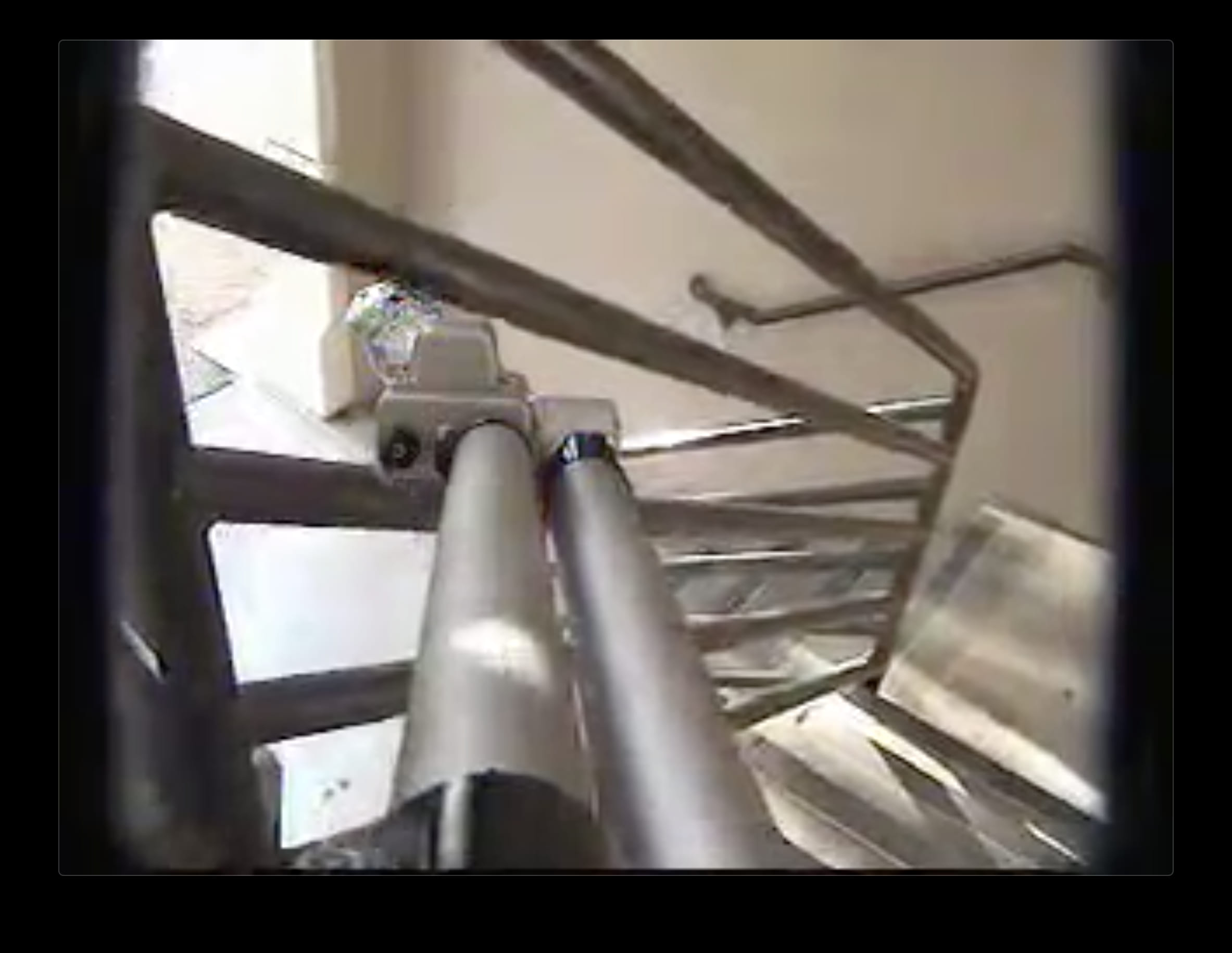}%
\label{fig::packbot_view}}\\
\subfloat[Third Person View from Aerial Visual Assistant]{\includegraphics[width=0.5\columnwidth]{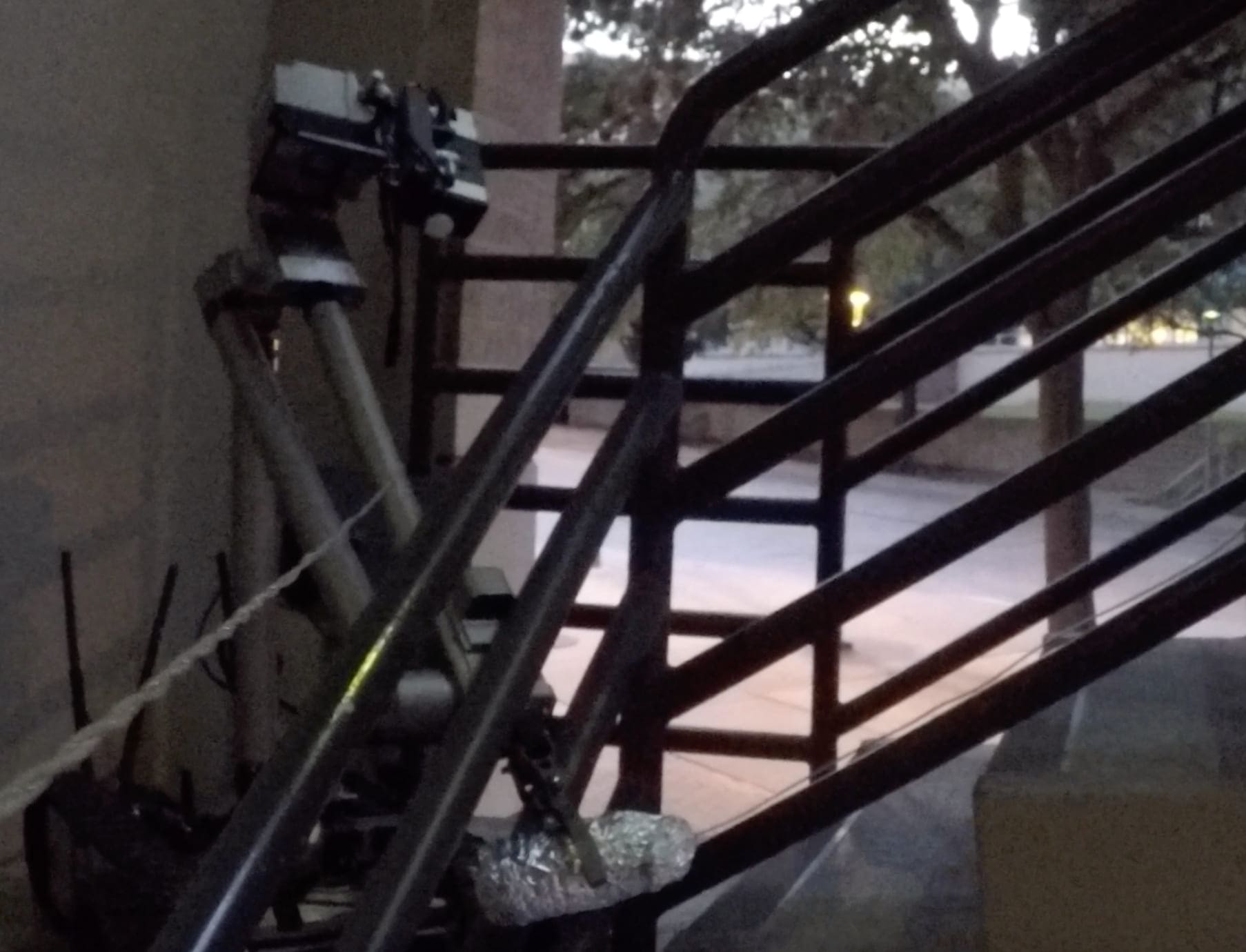}%
\label{fig::fk_view}}\\
\caption{Viewpoint Comparison from Onboard Camera and Visual Assistant: Depth perception is completely missing in the first person view from primary robot onboard camera, which is provided by the third person view from aerial visual assistant. With better viewpoint, teleoperation performance is exepcted to be improved. }
\label{fig::view_comparison}
\end{figure}

This chapter focuses on the validation of the proposed risk reasoning and risk-aware planning framework, along with the implementation of tethered motion suite for tethered flight in unstructured or confined environments. The improvement of viewpoint or teleoperation performance through aerial visual assistance is not the topic of this dissertation, but the comparison of onboard camera view and visual assistant view is presented in Fig. \ref{fig::view_comparison}, to qualitatively showcase the benefits brought by the whole visual assistance system. As shown in Fig. \ref{fig::packbot_view}, the limited first person view from the primary robot onboard camera does not have depth perception. It is difficult for the operator to decide if the sensor has gone through the gap between railings. Releasing the gripper before going through the railings cannot place the sensor at the right place (pool of contamination), while keeping the insertion may damage the manipulator arm. However, with the third person view provided by the aerial visual assistant (Fig. \ref{fig::fk_view}), depth perception becomes clear and the relative position between the gripper and railings are easy to discern. The improved teleoperation viewpoint is expected to improve teleoperation performance in unstructured or confined environments. The sensor is successfully dropped from the gripper to the pool of contamination with the help of the aerial visual assistant (Fig. \ref{fig::drop}).  

To summarize this chapter, integrated demonstration is conducted, as a combination of all the approaches proposed by this dissertation: the formal risk reasoning framework with definition and representation, the risk-aware planner that minimizes risk while considering reward, and the tethered motion suite to realize tethered flight in indoor unstructured or confined environments. After implementing all the approaches on a physical tethered UAV in a real-world unstructured or confined environment, it is found out that the proposed risk-aware planner can find minimum-risk path based on the newly proposed risk reasoning framework. Furthermore, the risk representation theory matches with practice in real application. The physical failure rate is close to the theoretically computed risk index value. Lastly, tethered flight could be enabled  in real-world obstacle-occupied spaces through the cooperation of all individual components in the low level motion suite. 

\chapter{SUMMARY AND CONCLUSIONS}
\label{chapter::summary_and_conclusions}
Motivated by the real world problem of visual assistance for robotic teleoperation with novel tasks in highly occluded environments, this dissertation develops formal theories to plan trust-worthy robot movement in unstructured or confined environments: a formal reasoning framework for robot motion risk and a risk-aware planner that can plan minimum risk path based on the newly proposed risk definition and representation. Built upon these formal theories for robots in general locomoting in unstructured or confined environments, risk-aware paths are also implemented on a tethered aerial visual assistant robot: this dissertation also presents a complete motion suite developed for tethered aerial vehicle in indoor GPS-denied obstacle-occupied spaces, opening up a new regime of indoor aerial locomotion: tethered flight. 



As part of the proposed tools to enable trust-worthy locomotion, the first contribution of this dissertation is a formal reasoning framework about robot motion risk in unstructured or confined environments. In this work, robot motion risk is formally defined as the probability of the robot not being able to finish the path. This formal and general definition unifies most adverse effects of the environment on the safety of the agent into one single numerical metric, instead of only considering risk as (chance) constraints caused by obstacles within Cartesian space. It is also applicable to any robotic agents locomoting in unstructured or confined environments. Using this metric, safety of robot locomotion could be explicitly reasoned, quantified, and compared. The use of propositional logic and probability theory provides fundamental reasoning and derivation of robot motion risk, reveals one of its important properties, history-dependency, and allows a formal approach of combining risk effects from both time domain and multiple risk sources. The discovered longitudinal dependency gives insights on a deeper understanding of what risk is for locomoting robots and the simple lateral independence assumption provides us with the leverage to combine a variety of adverse effects into one single metric of interest. This formally corrects the ill-supported temporal independencies, or when admitting dependence it alleviates the inevitable conservatism caused by ellipsoidal relaxation or Boole's bound. The proposed universe of risk elements is composed of three major risk categories, locale-dependent, action-dependent, and traverse-dependent risk elements, and is comprehensive to capture most safety-related concerns. 

The second contribution of this dissertation is a risk-aware planner which works on the newly proposed risk framework. The planner is also able to maximize mission reward simultaneously. The motivating visual assistance problem is defined into a path planning problem with reward risk tradeoff. This problem definition is an abstraction of the visual assistance problem and could be used as a general guideline to formulate path planning problems with reward and risk tradeoff. The goal state is planned simultaneously with the path, instead of being pre-defined or arbitrated. It is proved that this risk-aware reward-maximizing problem is well-defined when being converted to a graph-search query. An exact algorithm is presented which guarantees optimality at the cost of complexity. In order to make the problem tractable even in large scale, an approximate algorithm with upper stage minimizing risk while lower stage maximizing reward (utility) is proposed which sacrifices optimality for speed. As a major contribution of this dissertation, the upper stage risk-aware planner is also a stand-alone algorithm to plan minimum risk path in an absolute sense using the new risk definition and representation, instead of a feasible path within a probability bound, such as chance of constraint violation. Locale-dependent and action-dependent risk elements could be optimally addressed using the prosed planner, proved by mathematical induction, while traverse-dependent risk needs more look-back into the history dependency and is therefore more computationally intensive. Tradeoff between history dependency depth and computation is discussed. 

With a high level risk-aware path planned, the visual assistance mission is then implemented on a tethered UAV. The third contribution of this dissertation is a complete motion suite for tethered aerial vehicles, including sensing, (low level) planning, and actuation. As the key component to the motion suite, tether is maximally utilized, while its disadvantages, especially in unstructured of confined environments, are mitigated. The motion suite starts with a tether-based localizer, which allows localization in indoor GPS-denied environments with negligible computation. This creates another important benefit of using a tether for indoor light, convenient localization, in addition to the existing power-over-tether and mission-critical considerations. The two sets of motion primitives allow the tethered agent to adopt motion plans of any free-flying aerial vehicles and reside in the intuitive Cartesian space. The tether contact planning techniques address the issues brought in by the tether with existence of obstacles. By carefully reducing the reachable workspace or planning tether contact point(s) with the environments, the tethered agent can also fly in a similar way as its tetherless counterpart does. The motion executor for tether takes care of all the necessary transformation in accordance with the tether so the tethered agent can fulfill any motion possible for free flying UAVs. Lastly, as a complement to the deliberate risk-aware approach, a reactive visual servoing method is developed as an stand-alone individual member of the motion suite, providing visual assistance from a constant 6-DoF configuration with respect to a visual stimuli. Experiments are conducted for all the aforementioned system components and results are presented and discussed. 

Based on the three aforementioned contributions, the entire visual assistance system is integrated and demonstrated in a real-world unstructured or confined environment. The integrated demonstration validates the proposed risk reasoning framework, risk-aware planning, and implementation of the tethered motion suite using real scenarios similar to the motivating application of this research, Fukushima Daichi nuclear disaster decommissioning, out of well-engineered and controlled lab environments. The results are presented, analyzed, and discussed. It is shown that the proposed risk theory matches with real-world practice, and the developed tethered aerial locomoter is capable of handling real unstructured or confined environments.

In conclusion, this dissertation proposes a formal reasoning framework for robot motion risk in unstructured or confined environments, presents a risk-aware planner that conforms with the newly proposed risk definition and representation, and opens up new possibilities of autonomous, safe, and extended indoor aerial locomotion, tethered flight, with a complete tether-based motion suite. 


\let\oldbibitem\bibitem
\renewcommand{\bibitem}{\setlength{\itemsep}{0pt}\oldbibitem}
\bibliographystyle{ieeetr}

\addcontentsline{toc}{chapter}{REFERENCES}
\renewcommand{\bibname}{{\normalsize\rm REFERENCES}}

\bibliography{myReference}


\end{document}